\documentclass[num-refs, twocolumn]{wiley-article}

\usepackage{siunitx}

\usepackage{comment}
\usepackage{biblatex}
\usepackage{makecell}
\usepackage{subcaption}
\usepackage[ruled,linesnumbered]{algorithm2e}
\usepackage{graphicx}
\usepackage{setspace}
\usepackage{booktabs}
\usepackage{multicol}
\usepackage{multirow}

\usepackage{tikz}
\usepackage{pgfplots}
\usetikzlibrary{patterns,shapes.arrows, shapes.geometric}

\usepackage{textcomp}
\usepackage{siunitx}
\pgfplotsset{compat=1.12}
\usepackage[utf8]{inputenc}
\usepackage{pgfplots}
\usepackage{url}
\usepackage{pifont}
\usepackage{xcolor}

\papertype{Research Article}
\paperfield{Accepted for publication in Journal of Field Robotics, Wiley\\DOI: \url{https://doi.org/10.1002/rob.70084}}

\title{Development and Adaptation of Robotic Vision in the Real-World: the Challenge of Door Detection}

\author[1]{Michele Antonazzi}
\author[1]{Matteo Luperto}
\author[1]{N. Alberto Borghese}
\author[1]{Michele Antonazzi, Matteo Luperto, N. Alberto Borghese, Nicola Basilico}

\runningauthor{Michele Antonazzi et al.}

\begin{document}

\begin{frontmatter}
\maketitle

\begin{abstract} 
\small
Autonomous service robots are becoming increasingly common in human-centric, long-term deployments in unstructured indoor environments. \emph{Robotic vision} is a crucial capability, enabling robots to perceive and interpret high-level environmental features from visual input. While data-driven approaches based on deep learning have advanced the capabilities of vision systems, applying these techniques in real robotic scenarios still presents unique methodological challenges.
Conventional datasets often do not represent the object categories that a service robot needs to detect. More importantly, state-of-the-art models struggle to address the demanding perception constraints faced by service robots, posing the need of adaptations to the specific environments in which the robots operate. We devise a method that addresses these challenges by leveraging photorealistic simulations to create synthetic visual datasets from a robot's perspective. This approach balances data quality with acquisition costs, enabling the training of deep, general-purpose detectors tailored for service robots.
We then demonstrate the benefits of qualifying a general detector for the domain in which the robot is deployed, studying the trade-off between data-acquisition efforts and performance improvement. We evaluate our method using a representative selection of prominent deep-learning object detectors for the challenge of recognizing, in real-time, the presence and traversability of doorways. This task, which we refer to as \emph{door detection}, is fundamental to numerous significant robotic tasks, such as tracking the changing topology of dynamic environments.
We conduct an extensive experimental campaign in the field, considering different real-world setups while emulating the typical challenges encountered in long-term deployments of service robots. Our key findings demonstrate that simulation and qualification techniques can significantly reduce costs associated with domain adaptation for service robots. While simulation allows embedding the robot's perspective during the training of end-to-end robotic vision modules, qualification is essential to improve their robustness over challenging detection instances, thus reaching the performance level typically required by realistic long-term deployments of service robots.
\end{abstract}

\end{frontmatter}
\newcommand{\DDDtwo}{$\mathcal{D}_{\texttt{DD2}}$}
\newcommand{\DiG}{$\mathcal{D}_{\texttt{iG}}$}
\newcommand{\DG}{$\mathcal{D}_{\texttt{G}}$}
\newcommand{\DDDtwoG}{$\mathcal{D}_{\texttt{DD2+G}}$}
\newcommand{\Dreal}{$\mathcal{D}_{\texttt{real}}$}
\newcommand{\DrealEnv}{$\mathcal{D}_{\texttt{real,e}}$}
\newcommand{\DrealTest}{$\mathcal{D}^{\texttt{T}}_{\texttt{real,e}}$}

\newcommand{\Classrooms}{\texttt{Classrooms}}
\newcommand{\Offices}{\texttt{Offices}}
\newcommand{\Laboratories}{\texttt{Laboratories}}
\newcommand{\House}{\texttt{House}}

\newcommand{\Illuminations}{\texttt{Illuminations}}
\newcommand{\Occlusions}{\texttt{Occlusions}}

\newcommand{\RevOne}[1]{{\color{black}#1}}
\newcommand{\revI}[1]{{\color{black}#1}} 

\newcommand{\revTwo}[1]{#1}

\newcommand{\revZ}[1]{#1} 

\newcommand{\rcomment}[2][]{\noindent\textsc{\\Comment\textit{#1}:~}{\emph{#2}}}
\newcommand{\ncomment}[1]{\textcolor{gray}{\noindent\textsc{\\Comment (not answered):~}{\emph{#1}}}}
\newcommand{\response}[1]{\noindent\textsc{\\Response:~}{#1}}
\newcommand{\notes}[1]{\textcolor{teal}{\noindent\textbf{\\Notes:~}{#1}}}

\newcommand{\red}[1]{\textcolor{red}{#1}}

\definecolor{ForestGreen}{RGB}{34,139,34}
\definecolor{GoldenRod}{HTML}{FFDF42}
\definecolor{DeepBlue}{HTML}{2D2F92}
\newcommand{\doorsquare}{\ding{110}}%
\newcommand{\doorcross}{\ding{54}}%
\newcommand{\circleroom}{\ding{108}}%
\newcommand{\redcross}{\textcolor{red}{\doorcross}}%
\newcommand{\greencross}{\textcolor{ForestGreen}{\doorcross}}%
\newcommand{\yellowcross}{\textcolor{GoldenRod}{\doorcross}}%
\newcommand{\graycross}{\textcolor{gray}{\doorcross}}%
\newcommand{\blueroom}{\textcolor{DeepBlue}{\circleroom}}%
\newcommand{\greensquare}{\textcolor{ForestGreen}{\doorsquare}}%
\newcommand{\redsquare}{\textcolor{red}{\doorsquare}}%

\section{Introduction}\label{sec:intro}
Mobile service robots represent a flourishing technology increasingly employed across a range of human-centric, real-world domains such as office and domestic environments. Typically deployed for the long term, these robots find in \emph{robotic vision} - the ability to understand semantic knowledge from visual robot perceptions in real time - a cornerstone to achieving high-level autonomy and operational awareness. Indeed, one of the key capabilities is the one of detecting objects, which can significantly enhance a variety of sub-tasks across increasing levels of abstraction, including localization, navigation, scene understanding, and planning.

Recent developments in object detection, largely driven by research in deep learning, have unlocked remarkable possibilities for addressing this real-world AI challenge, facilitating the creation of highly effective vision modules for mobile robots. However, despite the abundance and diversity of available models, their practical application in field robotics continues to pose methodological and practical challenges. 
\revI{First, in real-world deployments, service robots are subject to specific perception constraints and frequently encounter challenging recognition instances, such as partially occluded or poorly positioned viewpoints. These challenges are intrinsic to the domain of service robots, as they are designed to operate in environments marked by dynamism and clutter. This condition induces critical domain shifts over the state-of-the-art object detectors, which are typically trained over datasets that largely neglect the noisy, constrained, and challenging operational conditions that a robot faces on the field.
Then, service robots often have to identify specific types of objects; however, these objects are usually not represented on conventional datasets such as COCO~\cite{coco} or Pascal VOC~\cite{pascal}, which are customarily used to train publicly-available object detectors. Thus, new task-related datasets should be collected and labeled to train the models equipped by the robot; still, the data acquisition procedure is costly, especially if real robots are used for this purpose.}

\revI{Service robot deployments, however, offer unique characteristics that can be leveraged to tackle the above challenges. In long-term human-centric deployments, robots are likely to encounter the same object multiple times and from different viewpoints. In these settings, objects of the same class often share similar visual features. Consider, for instance, chairs in a home versus an office. While office chairs may differ remarkably from those of a private residence, within each environment, it is likely to observe multiple chairs of similar style. These consistencies can be used to improve the perceptual capabilities of the robot in its target environment, thereby compensating the domain shift. Devising a comprehensive methodology to achieve this is still an open problem.}

\begin{figure*}[ht]
\centering
\includegraphics[width=\linewidth]{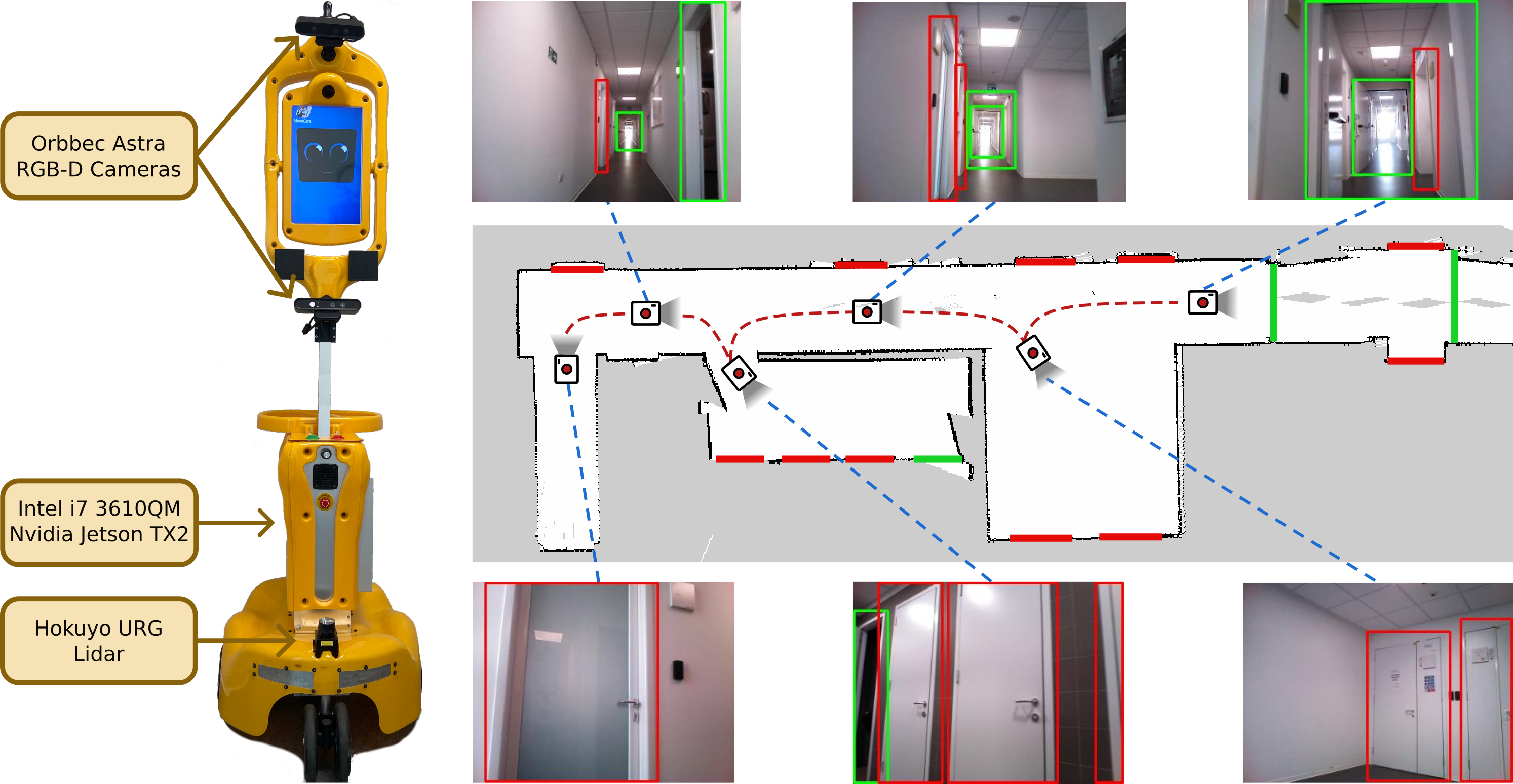}
\caption{Giraff-X~\cite{GIRAFF}, the service robot adopted in this work, performing door-status detection \revTwo{while navigating in an indoor environment. The green (red) bounding boxes represent open (closed) doors.}}
\label{fig:giraff}
\end{figure*}

\revI{We investigate the above challenge in the context of \emph{door detection}, a particularly significant detection task which can be exploited to enhance the long-term navigation capabilities of service robots~\cite{dynamicmaps, longtermnavigation}.}
Doors are key environmental features for a mobile robot: they represent the topological connections between adjacent sub-regions of the free space whose traversability, importantly, might not always be possible. 
\revI{For this reason, we address the typical operational requirements of indoor service robots by defining a door as a variable-traversability passage.
This approach is broader and hence more complex than adopting the more conventional concept of ``explicit'' door (as defined in~\cite{whiting2007topology}), which is based on physical structures with components like a leaf, a hinge, or any other related furniture.}
Door detection is the capability for a robot to recognize in real-time the location and traversability status (\texttt{open} or \texttt{closed}) of such passages. This problem is most effectively addressed as an object detection task using RGB images \revI{as illustrated in Figure~\ref{fig:giraff}: a mobile robot navigates in an environment to perform its tasks; at the same time, it acquires images through its onboard camera. For each image, it infers in real-time the bounding boxes of doors, distinguishing between open doors (depicted in green) and closed ones (depicted in red), also highlighted on the map.
Performing door detection leveraging RGB data is primarily due to the limitations of alternative technologies.} For example, laser range scanners, while robust and precise for distance measurements, are typically constrained by a 2D field of view, \revI{making it difficult to disambiguate a planar surface (such as a wall) from a closed door, especially in settings like a corridor where the doors are perpendicular to the motion of the robot.} In the same line, RGB–D cameras often exhibit a limited depth sensing range, making them
less reliable over longer distances or in larger indoor spaces
\revI{See, for example, the doors depicted in Figure~\ref{fig:giraff}: depth and LiDAR data alone are not adequate to detect the presence and the status of such challenging door instances when observed from constrained viewpoints (e.g., doors in images of the first row of Figure~\ref{fig:giraff} are difficult to detect with LiDAR data). Furthermore, both types of sensors struggle with transparent or highly reflective surfaces, that are common in doors. An example is depicted in the first robot's perception (first image second row) of Figure~\ref{fig:giraff}, where depth sensors are not able to see the closed door with a glass panel on it.} \revTwo{These limitations are also relevant in sensor fusion approaches (combining RGB with 3D data) as they inherit the challenges associated with the depth sensing. As previously mentioned, our target task often involves the detection of transparent or reflective surfaces, situations in which the depth data are missing or noisy. See for example the third image of the first row of Figure~\ref{fig:giraff}: the doors are perpendicular to the robot point-of-view and some of them are a glass panel. In these cases, 3D data can provide little knowledge to detect the status of doors, and the robot can rely only on vision. Integrating such unreliable information with knowledge acquired from RGB images is not straightforward, as it may represent an additional source of error. Consequently, our work focuses on RGB-only perception, which provides more consistent and robust visual cues for identifying the status of doors in the real-world.}

In this work, we demonstrate how the integration of simulation frameworks and domain adaptation techniques can create an effective pipeline for constructing door detectors specifically tailored for mobile service robots. 
\revTwo{We provide an extensive evaluation of the domain shift affecting the deep learning-based object detectors used for robotic perception. Our goal is to delineate an effective pipeline for the development and deployment of end-to-end architectures compliant with the requirements of service robots. At first, we demonstrate how simulation can be leveraged for training generic models able to obtain acceptable performance in novel environments. Specifically, we show that matching the photorealism encountered by robots in the real-world is not enough. To be consistent with the robotics domain, being compliant with the robot's perception model is an essential requirement to ensure the robustness of end-to-end modules for robotic vision. Secondly, we show the relevance of qualification to the operational environment of the robot and how it enables the detection of the target objects in challenging instances. We prove that fine-tuning a general detector with a small batch of high-quality annotated data strongly enhances the perception capabilities of mobile robots: this makes qualification not only essential, but also affordable. Our contributions can be summarized as follows.}

\begin{itemize}
\item We analyze the trade-offs involved in using simulations to train a door detector capable of generalizing effectively across various environments. We delineate the desiderata of this process and propose a framework based on simulation performed in 3D real-world models, Gibson~\cite{gibson}, that achieves a balance between data photorealism and acquisition costs. \revTwo{Leveraging this procedure, we collect and release a visual dataset for door-status detection acquired in $10$ photorealistic environments from the robot point of view.}
\item We explore fine-tuning to qualify detectors to a robot's target environment, demonstrating how leveraging typical operational settings of service robots can enhance performance in challenging instances.
\item We argue that performance metrics used to evaluate computer-vision models are not well suited to be used in a robotic context, and we propose new evaluation metrics specifically designed to better reflect our setting.
\item We conduct an extensive experimental campaign with three \revI{state-of-the-art and widely adopted} deep-learning models for object detection, providing insights into how domain shifts in our scenario are influenced by the nature of the training datasets. \revI{We further validate our findings by assessing the robustness of our general and environment-specific door detectors to typical domain shifts occurring in long-term deployments inside the same environment. \revTwo{To achieve this, we collected a dataset for door-status detection in four real-world environments using our robotic platform~\cite{GIRAFF}, and we have made it publicly available.}}
\item We evaluate the impact of different \revI{door detection methods} on a downstream task, that is \revI{evaluating} the current traversability status of the whole environment.
\end{itemize}

The contributions of this paper build upon and extend our earlier work presented in~\cite{antonazzi2023enhancing}, where we have illustrated a preliminary version of the findings presented here. In this work, we significantly extend the experimental evaluation also by assessing multiple models for object detection. 

In the next section, we briefly survey the related works relevant to this study. Section~\ref{sec:method} motivates and outlines our methods while the remainder of the paper is devoted to experimental analysis. Section~\ref{sec:evaluation} extensively analyses the performance of our general and qualified detectors, \revI{also testing how their detection abilities influence the downstream task of topology mapping (Section~\ref{sec:onTheFieldExp}).}
\revTwo{A discussion on the lessons learned is provided in Section~\ref{sec:lessons_learned} while the concluding remarks and future directions are reported in Section~\ref{sec:conclusions}.}

\section{State of the Art}\label{sec:sota}

Mobile service robots are a cutting-edge technology that is increasingly being adopted in a variety of real-world scenarios. These robots are typically employed to assist humans in various tasks, often unfolding in indoor industrial, or domestic environments~\cite{lee2021serviceSurvey}. Among recent and representative application domains are healthcare, where robots assist patients and caregivers~\cite{holland2021service}; logistics, where they carry out repetitive tasks like item deliveries or environmental monitoring~\cite{fragapane2021planning};  at-home caregiving, where they aid in day-by-day activities like cleaning or providing additional services such as personal assistance, wellbeing monitoring, and social entertainment~\cite{do2018rish, GIRAFF}.

Although the adoption of mobile service robots is increasing, their deployment in indoor human-centric environments such as houses, offices, hospitals, and schools, continues to present substantial research challenges. Unlike industrial settings, where there is a higher degree of control and predictability, the lived-in setup is typically unstructured and dynamic. This complexity arises both from the physical layout of such environments, that is how rooms, walls, and furniture are disposed, and by their visual aspect, which is complex and semantic-rich; consequently, two different environments of the same type can have very different features. As examples of recent research works show, in human-centric environments, tasks such as \revZ{user understanding}~\cite{Ishikawa_2021}, activity recognition~\cite{SoHAM_2022}, and even path planning~\cite{kumaar2023mobile} face additional challenges.

To properly work inside these complex settings, a robot should be able to understand relevant properties through its vision. In this domain, one of the fundamental problems is the one of real-time object detection. 
Indeed, such functionality is key for service robots, enhancing their capabilities and enabling autonomous behavior in various situations~\cite{alatise2020review}. Object detection is typically performed from images acquired with RGB cameras, and the recent advancements in deep learning applied to computer vision~\cite{chai2021deep} have led to the development of high-performance pre-trained models, which can be exploited to obtain \emph{robotic vision} modules running locally on mobile robots. These models, among which prominent examples are YOLO~\cite{yolov5}, DETR~\cite{detr}, and Faster R-CNN~\cite{fasterrcnn}, offer significant advantages in object detection tasks for service robots, but their field deployment poses a set of engineering and methodological challenges, especially in unstructured environments. Our work focuses on tackling common problems of object detection in human-centric environments by assessing a specific and widely relevant case study, door detection.

The ability of mobile robots to detect doors is key for indoor operations. The traversability status of a door (\texttt{open} or \texttt{closed}) enables the availability of passageways between sub-areas of an environment. In turn, traversability directly determines the environment's topology, ultimately affecting the robot's navigational routes and accessibility of the areas therein.  The location of doors is important for tasks such as room segmentation~\cite{segmentationsurvey}, which entails partitioning the map of an environment into semantically meaningful areas or rooms. This knowledge is also beneficial in predicting the layout of rooms not yet observed during exploration~\cite{ECMR21} or in identifying temporary unreachable locations during mapping. Furthermore, it plays a key role in place categorization, a process where rooms on the occupancy map are assigned semantic labels (like ``corridor'' or ``office'') based on their visual appearance~\cite{scenerecognitiononjectdetection, placecategorization}. Recent studies have shown that a robot's ability to recognize doors can significantly enhance its navigation capabilities in long-term scenarios. For instance, the work presented in~\cite{dynamicmaps} models the periodic changes in dynamic environments over extended periods. Similarly, the study in~\cite{longtermnavigation} introduces a navigation system designed for robots functioning in indoor environments for long periods, particularly where the traversability of the area varies over time.

The use of object detection methods is the mainstream approach to tackle door detection with mobile robots. Initial seminal methods in this domain relied on the extraction of handcrafted features~\cite{humanoid, sonarandivisualdoordetection, edgeandcornerdoorsdetector}, such as edges~\cite{canny} and corners~\cite{cornerdetector}, to describe the characteristic rectangular shape of doors. However, the requirement to explicitly define and combine these features is a significant limitation of these approaches. This constraint hampers their robustness and adaptability, especially when dealing with the highly variable images encountered in real dynamic environments.

Deep learning end-to-end methods have brought significant improvements in the field. Their ability to automatically learn features that characterize an object class, and robustly handling variations in scale, position, rotation, and lighting, is a major advantage that has led to their widespread use in mobile robotics. A pioneering method for door recognition in mobile robot navigation was introduced in~\cite{detectdoorsfeature}. This method utilizes color and shape as key features to detect doors in office environments, employing two neural classifiers to identify these elements in images. These features are then integrated using a heuristic algorithm to determine if they form a typical door structure. The study in~\cite{doorsandnavigation} presents a method for door detection aimed at enhancing the autonomous navigation of mobile robots. A convolutional neural network is trained to identify doors in indoor settings, demonstrating its utility in aiding a robot's efficiency in traversing passages. Additionally, recent research has explored the integration of RGB vision with other sensors commonly used in robot navigation~\cite{kim2022improvement} and the identification of doors and their handles to enable interactions like grasping~\cite{doorcabinet, cupec2023teaching, JangRAL2023_Manipulator}. For example, the research in~\cite{doorcabinet} utilizes a YOLO-based deep learning framework~\cite{yolo} for the detection of door Regions Of Interest (ROI). This approach specifically targets the handles by focusing on the area encapsulated within the door's ROI, effectively locating the handles for interaction purposes.

The studies previously mentioned offer approaches to the door detection task in scenarios similar to the one we address, yet they do so only to a limited extent. A notable limitation in these studies is the absence of training from a mobile robot's perspective. Additionally, these methods often do not exploit the typical operational conditions of a robot. Our work introduces a strategy specifically tailored to address these shortcomings. Viewing this from a broader angle, our method addresses the domain adaptation challenge (a well-recognized issue in deep learning at large~\cite{yosinski2014transferable}) within the realm of door detection using mobile robots in unstructured indoor environments. Similar findings are provided by the work of \cite{zhao2022opend}, where a dataset of door handles for training robotic manipulation methods to open/close doors is presented. Our proposed technique exploits the technique of fine-tuning of pre-trained deep neural models, a practice extensively employed in autonomous mobile robotics both relying on manually labeled data~\cite{zimmermanlocalizationobjectdetection, zimmerman_longterm_objectdetection_localization} and self-supervised methods~\cite{zurbrugg2022semanticadaptation, Lajoie2023}. 

\section{General and Qualified Door Detection for Service Robots}\label{sec:method}


We focus on a service robot designed to autonomously operate in human-centric indoor environments. We assume a widely-used hardware setup, where the robot is equipped with one or more RGB cameras for vision. The primary objective is developing real-time door detection, namely the task of processing an RGB image acquired by the robot to determine the bounding box and binary traversability status (\texttt{open} or \texttt{closed}) for each visible door. 

\begin{figure*}[!h]
	\centering
 \includegraphics[width=0.9\linewidth]{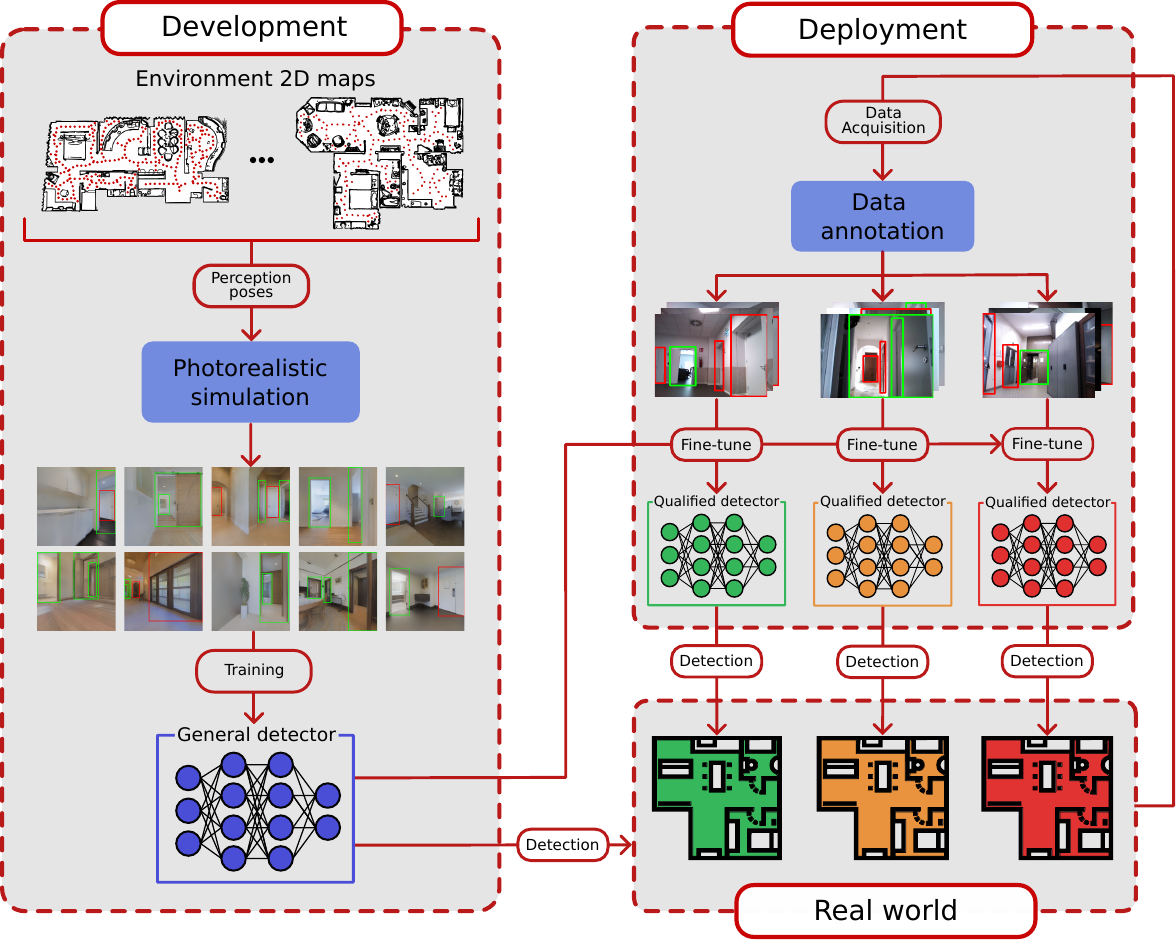}
 \caption{\revTwo{A general overview of our pipeline for the development and deployment of deep learning-based object detectors for robotic vision. At first, we build the General Detector (GD), a module that exhibits acceptable performance operating in novel environments, not included during the training phase. For doing this, we introduce a novel simulation framework to reduce the effort for acquiring training datasets from the robot's perspective. After the robot deployment, we study the domain shift experienced by the GD and we perform a fine-tuning, obtaining a Qualified Detector (QD), enhancing the detection performance in the operational environment of the robot.}}
 \label{fig:methodology}
 \end{figure*}

The method we propose, graphically summarized in Figure~\ref{fig:methodology}, is structured around the two principal phases that define the lifecycle of a mobile service robot: the robot's \emph{development} phase and the subsequent \emph{deployment} phase. 
With this method, we aim to identify, experimentally evaluate, and solve some of the challenges that are intertwined with the usage of vision-based object detection methods on mobile robots.

The development phase for a service robot involves preparing and configuring the platform, including the installation and setup of hardware and software components. The objective here is to setup a robot that is ready to meet the challenges of real-world environments. This 
\revZ{phase}'s focus lies in the domain of visual perception capabilities, aiming to equip the robot with vision skills that perform satisfactorily across various environments, thereby ensuring a high level of generalizability. The development phase is the starting ground for addressing our door-detection task. Our approach involves creating a \emph{General Detector} (GD), designed to recognize doors while adhering to the perception constraints of service robots and maintaining consistent performance in various environments. A significant part of our method involves utilizing simulation to develop a photorealistic visual dataset, representing typical visual perceptions of a robot. This dataset is then used to train a GD, ensuring it achieves baseline performance in the real world.
 

During the deployment phase, the service robot is introduced for autonomous operation in a target environment, usually for an extended period. This phase often involves a domain shift, presenting challenges to the performance of the previously developed GD. This is because the pre-built computer vision methods, focusing on the model's development, present difficulties introduced by our environmental setup that prevent their straightforward use on autonomous mobile robots. Given the long-term nature of this phase, there is an opportunity to incrementally fine-tune the GD with data collected in the target environment, aligning it more closely with the specific visual features at hand, thus obtaining a \emph{Qualified Detector} (QD). This detector can exploit the fact that usually multiple instances of the same object within the same environment present similar features, that are stable in time. Doors and windows are good examples of this fact: within a target environment, most of them are usually produced by the same manufacturer and are of the same type. The process of adapting the detection model to the target environment may require the collection and annotation of data for which we propose a method demonstrating a trade-off between the effort required and the resulting performance improvements.


\subsection{Training a General Detector}\label{sec:gd}

The recent trends in object detection suggest that a straightforward way to address Robotic Vision is to plug and play a deep detector in a robotic platform~\cite{objectdetectionsurvey}. Despite the availability of a large number of effective models, when faced with reality this simple approach presents several engineering challenges and an established method to provide a GD for service robots still needs a comprehensive investigation. 


\revTwo{State-of-the-art object detectors are typically trained on prominent datasets (such as Pascal VOC~\cite{pascal}, ImageNet~\cite{imagenetchallenge}, or, most commonly, MS COCO~\cite{coco}) composed of thousands of images acquired from application-agnostic viewpoints in diverse contexts, including both indoor and outdoor settings. However, when these models are applied to robotics, two primary challenges arise. First, the dataset could not extensively represent the object of interest, compromising the detector's ability to recognize it. Our survey of existing datasets reveals that doors frequently suffer from this lack of representation. This is primarily due to our broad definition of a door as a variable-traversability passage, introduced in Section~\ref{sec:intro}. Secondly, even when the object of interest is well-represented, distribution shifts between the training data and real-world scenarios can significantly affect performance. This widely recognized yet largely unresolved issue is particularly problematic in our context, as the shift can affect multiple aspects: the input data, the feature space, and the data collection process itself. (Here we rely on the discussion about domain shift types of~\cite{lee2022surgical}).}

\begin{figure*}[!htbp]
    \centering
    \subfloat[\label{fig:deep_doors2_pov}]{\includegraphics[width=0.21\linewidth]{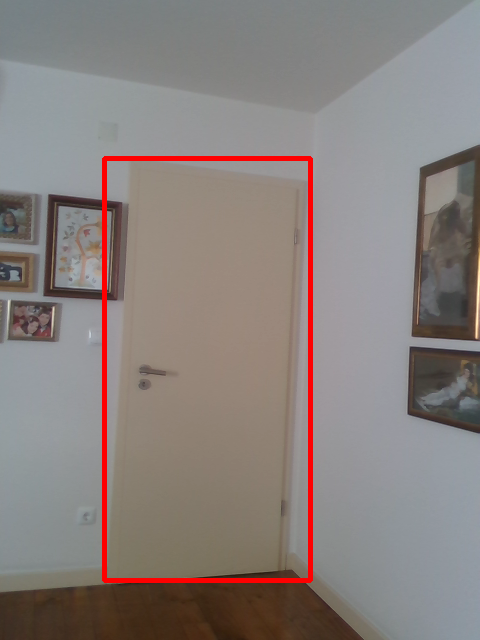}}
    \hfil
    \subfloat[\label{fig:deep_doors2_outdoor}]{\includegraphics[width=0.21\linewidth]{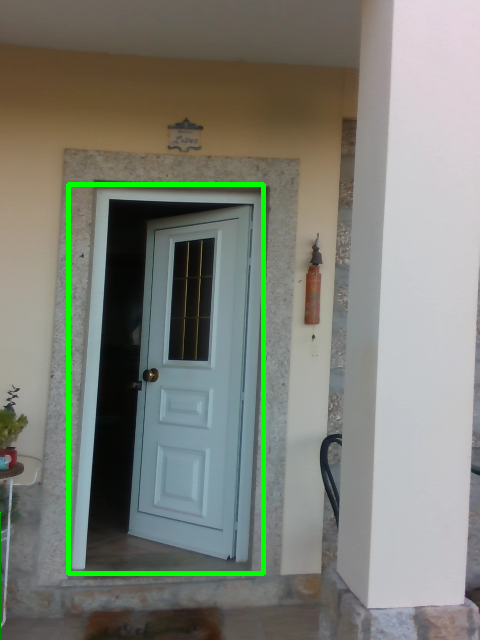}}
    \hfil
    \subfloat[\label{fig:deep_doors2_partial}]{\includegraphics[width=0.21\linewidth]{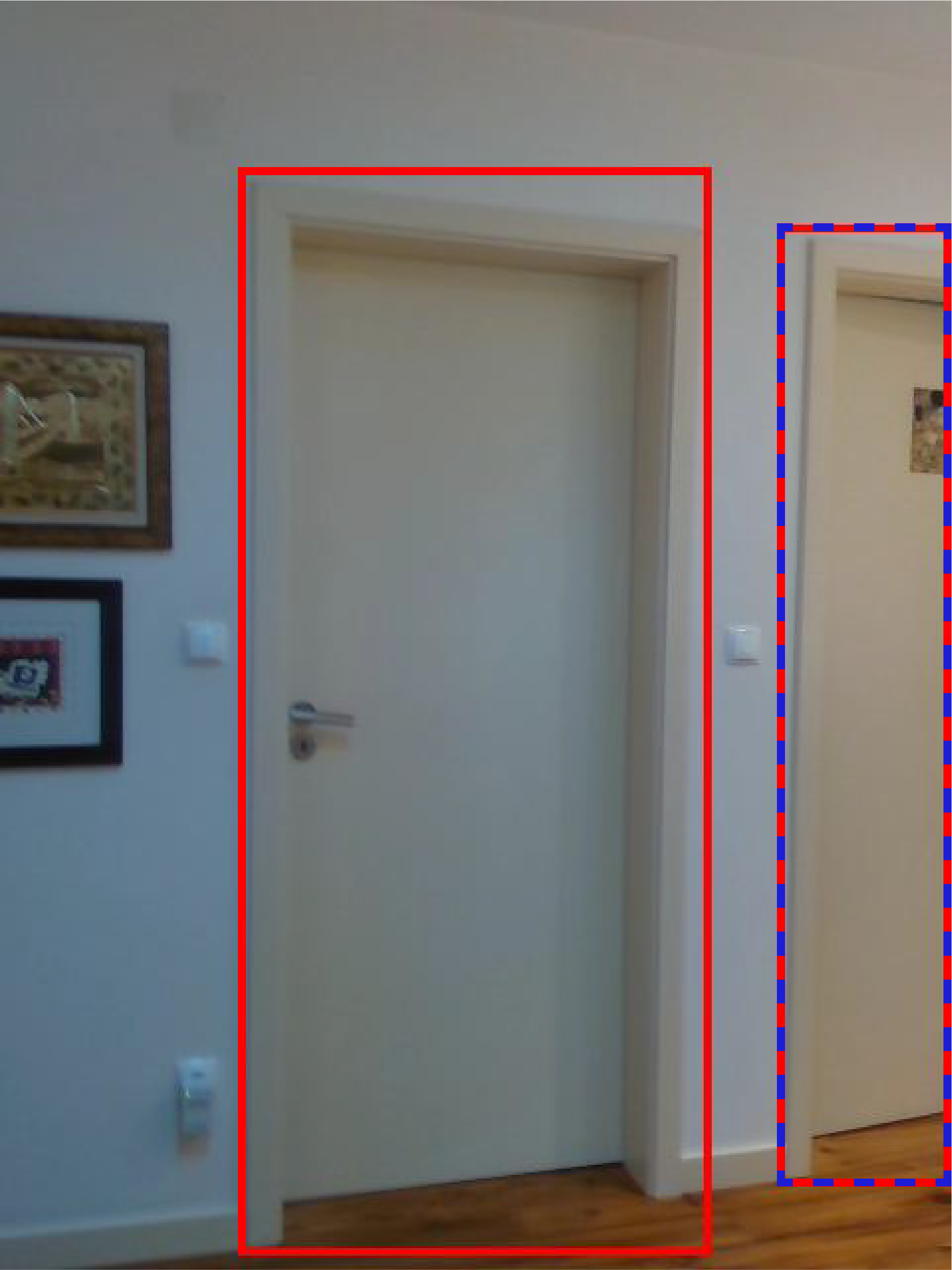}}
    \hfil
    \subfloat[\label{fig:deep_doors2_nested}]{\includegraphics[width=0.21\linewidth]{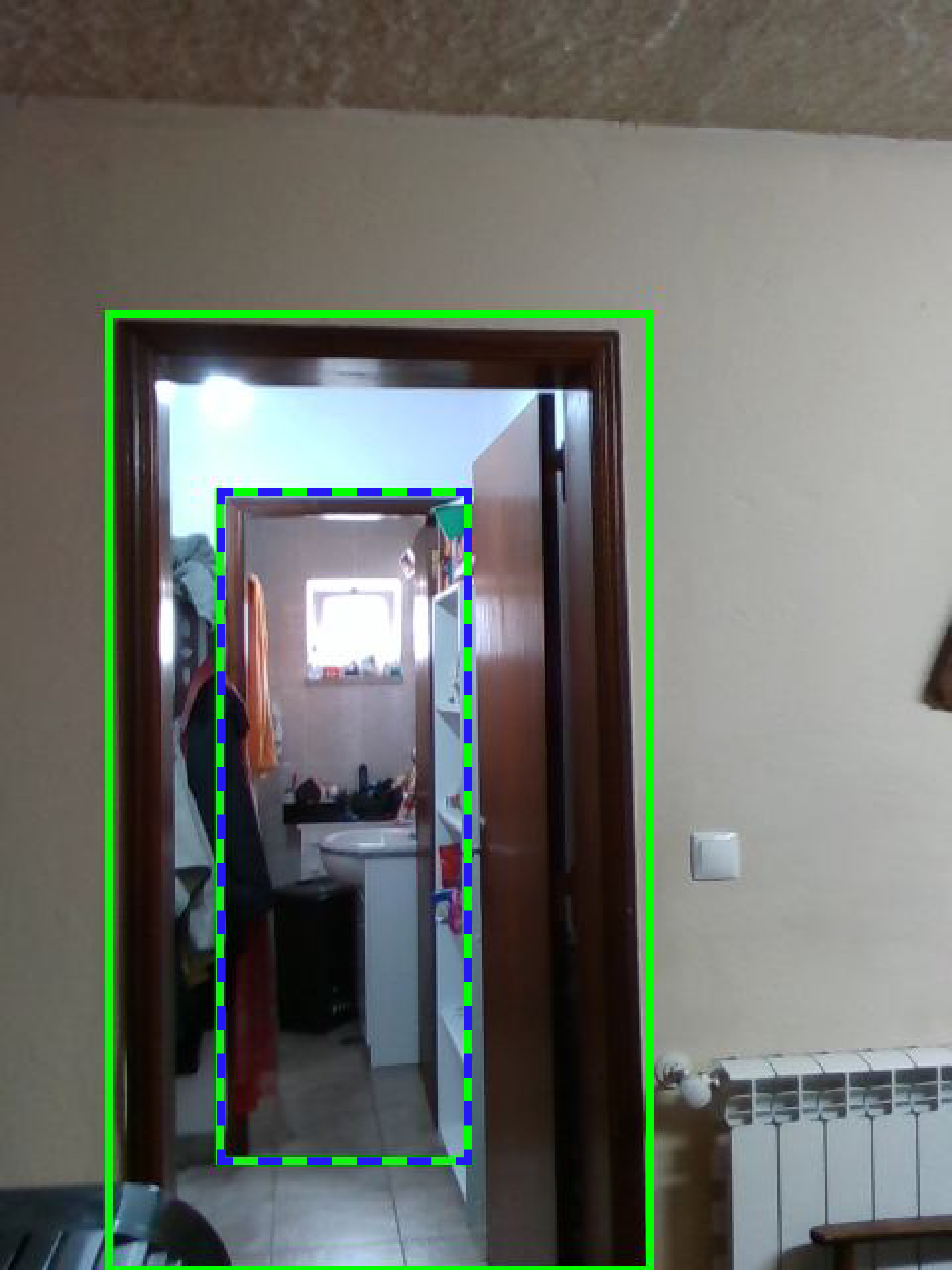}}
	\caption{Examples from the DD2 dataset~\cite{deepdoors2} of open and closed doors (in green and red, respectively). The dashed bounding boxes represent missing annotations.}
	\label{fig:deepdoors2_examples}
\end{figure*}

Perhaps the dataset for door detection that is most relevant to our work is \emph{DeepDoors2} (DD2)~\cite{deepdoors2}, which contains around $3000$ images of doors, each annotated with a bounding box and traversability status\footnote{Note how a dataset of this size is customary in several object detection tasks: as an example, in MS COCO, the average number of examples per class, is $3200$; the only exception is the category \texttt{person}, which has more examples, over $10$K. }. However, in our scenario, DD2 is susceptible to performance degradation due to distribution shifts, a fact that becomes expected already upon examining some of its examples. The images in DD2 are captured from human-like perspectives, often showing the door fully visible and centrally located, as depicted in the indoor and outdoor examples of Figure~\ref{fig:deep_doors2_pov} and Figure~\ref{fig:deep_doors2_outdoor}, respectively. This dataset overlooks instances such as partially visible or nested doors, which are common in robots' perceptions. Labels are provided only for doors that are completely within the frame and distinct enough for clear identification, as shown in the dashed bounding boxes of Figure~\ref{fig:deep_doors2_partial} (partially visible door) and Figure~\ref{fig:deep_doors2_nested} (nested doors). These shortcomings are, to varying degrees, present in nearly all conventional computer vision datasets~\cite{imagenetchallenge, pascal, coco, deepdoors2}, reflecting their inherent limitations in capturing a robot's visual perception model~\cite{surveydeeplimits}. As our experimental campaign will demonstrate concretely, these limitations significantly affect performance.


To address them, one common method is fine-tuning a large-scale pre-trained model (such as one trained on MS COCO~\cite{coco}) with new examples that better represent the target object distribution. This approach is prevalent, especially in robotics~\cite{zimmerman_longterm_objectdetection_localization, chiattisurgicalfinetuning, doorcabinet}, and the strategy we evaluate in this work is based on it. Ideally, creating an effective door detector through fine-tuning requires a dataset that:
\begin{itemize}
\item demonstrates a high level of photorealism (to withstand distribution shifts at the input level);
\item encompasses a variety of indoor environments with diverse features (to withstand distribution shifts at the feature level);
\item accurately reflects the robot's perspective and perception model (to withstand distribution shifts in the data acquisition process).
\end{itemize}
 Currently, no dataset fulfilling these criteria exists in the literature, as efficiently collecting it is still an open problem.
 An alternative to this issue is to use labeled sequences of images obtained by a robot or by a mobile platform, such as in ScanNet \cite{dai2017scannet} or SUN3D \cite{xiao2013sun3d}. However, these sequences are usually collected within single rooms and, as they are based on fixed trajectories, do not allow the sampling of new viewpoints from different perspectives that may be encountered by the robot while navigating. 
 The most straightforward approach would involve an extensive data collection campaign using robots in real-world environments, gathering image samples and manually labeling them. However, the logistics and costs associated with this method are prohibitively high and well-recognized among robotics professionals. In the following, we tackle this problem by exploiting simulation~\cite{collins2021review}, an approach frequently employed in robotics to mitigate the large costs of on-the-field experimentation. The empirical results we present later will demonstrate how, with appropriate design measures, simulations can provide a dataset from which an effective door detector can be trained.

\subsubsection{The Proposed Simulation Framework}\label{sec:simulationframework}

Common 3D physics simulators like Gazebo~\cite{koenig2004design, kenta2016gazebosim} or CoppeliaSim~\cite{lei2017coppeliadrl, jian2017coppeliavisualtrajectory} are widely adopted for prototyping control software in robotics before real-world deployment~\cite{collins2021review}. However, their lack of a sophisticated rendering pipeline for realistic visual perceptions makes them unsuitable for our setting. Early efforts to address this limitation have involved the use of 3D game engines, such as Unity3D~\cite{yoshiaki2017unitysim} or Unreal~\cite{USARSIM}, to recreate complex robotics scenarios, including unique environments or specialized physical laws, as seen in autonomous vehicles~\cite{dosovitskiy2017carla, shah2018airsim}, UAVs~\cite{pushkal2019ursim}, or surgical robotics~\cite{tagliabue2020surgicalsimulationunity}. Despite their adaptability, customizing these game engines for a particular robotic application can be challenging~\cite{melo2019simulatorscomparison}. Specifically, for the task considered in this paper, this would involve manually creating synthetic scenes that accurately reflect the structural features of real indoor environments, a task that is more aligned with environment design than robotics engineering.



Recently, the introduction of interactive realistic simulations for embodied AIs, as iGibson~\cite{igibson}, has helped mitigate the limitations of traditional simulators for indoor robotic tasks. iGibson comes with 15 artificially constructed home-sized scenes, which are developed by populating layouts of actual environments with 3D controllable objects whose configuration, shape, material, and texture can be automatically changed. Unlike the simulators mentioned earlier, iGibson seamlessly integrates with the Robot Operating System (ROS), facilitating the collection of extensive, high-quality annotated image datasets from a robot's perspective. However, despite these significant advantages, simulators based on synthetic scenes still fall short in achieving the crucial aspect of photorealism. This limitation is something we empirically assess in our experimental campaign. \revI{Similar findings are identified in the work of \cite{VLNAV}, which shows how, for the task of visual navigation for an autonomous mobile robot, the higher the performance in simulation, the higher the gap in performance with a real robot, as the robot models often overfit on the synthetic features of the simulated environment, that are different to those of real ones.}


In addressing these challenges, we adopted an approach that balances the photorealism of real-world data acquisition with the automation benefits of synthetic simulations. Our solution relies on Gibson~\cite{gibson}, a simulator designed for embodied agents with an emphasis on enhancing the photorealism of visual perceptions. \revTwo{Gibson employs scene datasets scanned directly from real environments (such as Matterport3D~\cite{matterport} and Stanford-2D-3Ds~\cite{stanford2d3d}) that accurately capture and replicate the challenges typical of the real world.} Additionally, it incorporates a neural rendering pipeline to further bridge the sim-to-real gap. 
\revI{These enhancements aid in the effective transfer of models trained within the simulator to real-world environments.} \revTwo{Leveraging these features, we developed a simulation framework based on Gibson in conjunction with Matterport3D. It is a comprehensive RGB-D dataset comprising 90 digitized real scenes also including semantic tagging for both instance and category-level segmentation.} This combination allows us to achieve a balance between photorealism and the controlled conditions necessary for effective simulation.

\begin{figure*}[h!]
    \centering
    \subfloat[\label{fig:matterport_issues_meshes_furniture}]{\includegraphics[width=0.24\linewidth]{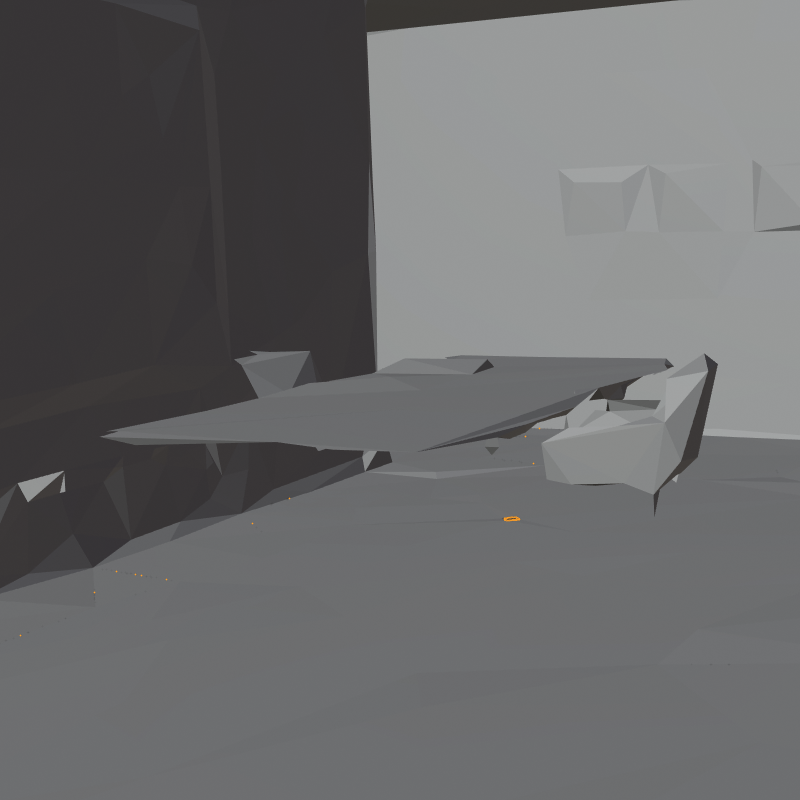}}
    \hfill
    \subfloat[\label{fig:matterport_issues_hole}]{\includegraphics[width=0.24\linewidth]{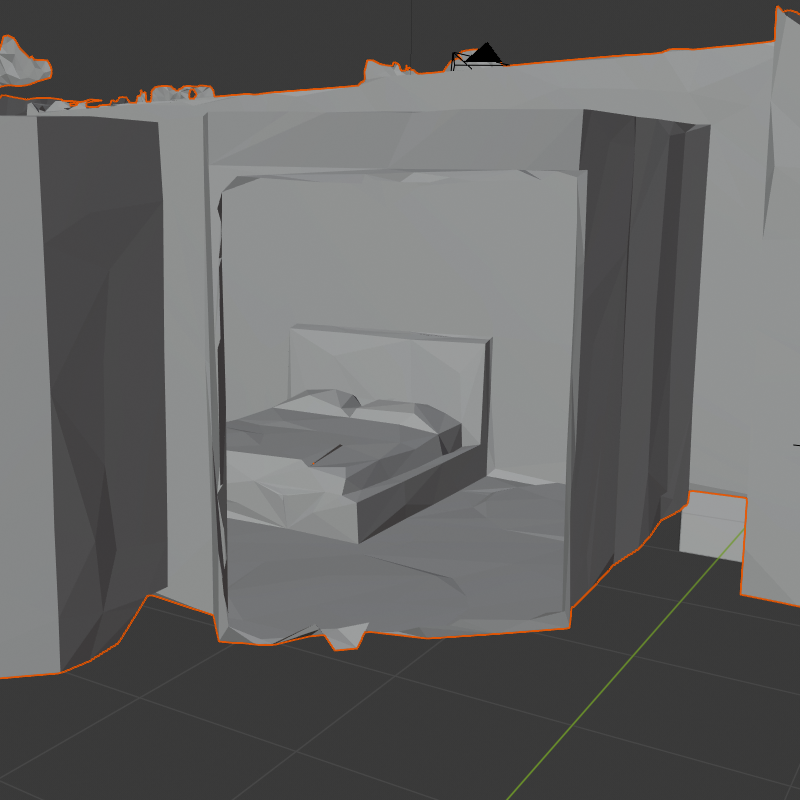}}
    \hfill
    \subfloat[\label{fig:matterport_issues_floor_plan}]{\includegraphics[width=0.24\linewidth]{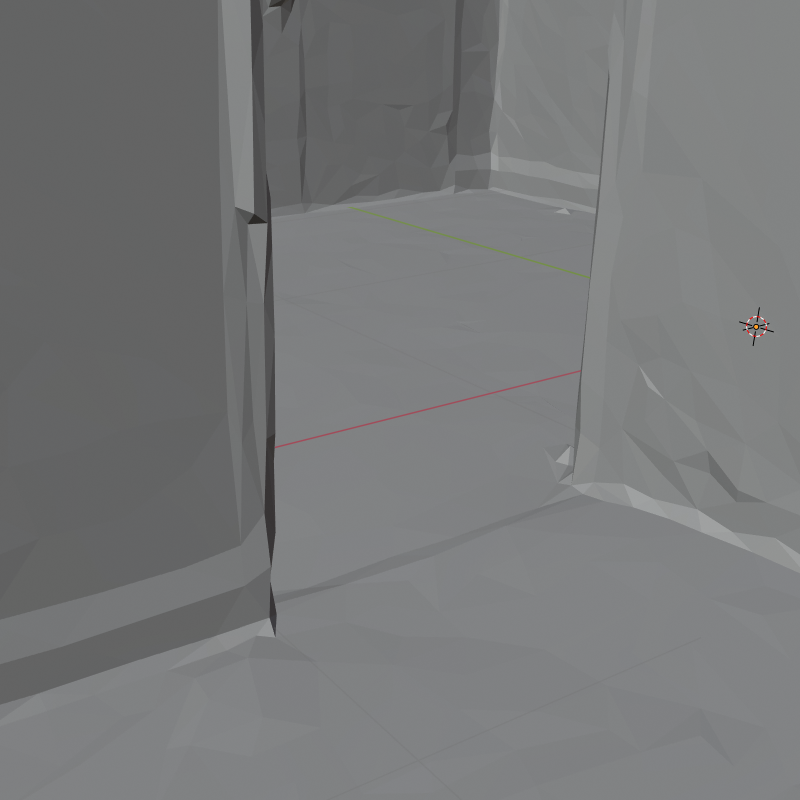}}
    \hfill
    \subfloat[\label{fig:gibson_example}]{\includegraphics[width=0.24\linewidth]{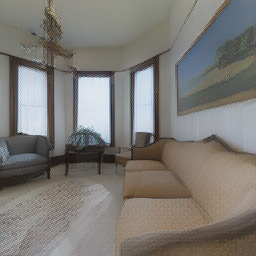}}

	\caption{Matterport3D mesh malformations. (a) The table and chairs have no legs. (b) A wall in the bedroom, in front of the bed, is missing. (c) Floor surface irregularities. (d) An example of perception acquired from the Gibson-based simulation framework.}
    \label{fig:matterport_inaccuraciess}
\end{figure*}

Gibson provides a middleware for controlling a ROS-based virtual robot. While camera perceptions can be easily simulated (setting resolution and FOV), navigation encounters several technical limitations. First, conducting a real-time acquisition campaign, even in a simulated environment, can be time-intensive. Moreover, this approach does not offer complete control over the data acquisition process, as much of it depends on the navigation stack of the simulated robot. Additionally, the 3D polygonal meshes of the Matterport3D environments, which are digitized from real-world settings, are often cluttered and noisy. This results in various issues: furniture models appear malformed and incomplete (as shown in Figure~\ref{fig:matterport_issues_meshes_furniture}, where the legs of the table are not modeled), walls frequently have holes near windows or mirrors (see Figure~\ref{fig:matterport_issues_hole}, where the bed should be behind a mirror and a wall, which are missing), and the surfaces of floors are irregular (see Figure~\ref{fig:matterport_issues_floor_plan}). 
\revI{These artifacts mostly concern the 3D meshes and not the images; as a result of this, there is a mismatch between the features of the images and the depth features perceived by the robot (e.g., with a LiDAR). As an example, in Figure~\ref{fig:matterport_issues_meshes_furniture} the robot sees (image) a table, which is not perceived in the corresponding depth sensor reading; in Figure~\ref{fig:matterport_issues_floor_plan} the robot sees a flat, smooth, pavement surface, where the 3D meshes are bumpy and are inaccurate. As a result, the robot's autonomous navigation is prone to failures and not robust. At the same time, image data are not much affected by these errors thanks to the Gibson's rendering pipeline that, using a neural network, corrects possible visual artifacts (as in Figure~\ref{fig:gibson_example}).} 


To address these shortcomings, we have developed an enhanced version of Gibson, introducing a highly controllable simulation mechanism. This upgraded simulation framework\footnote{The pre-compiled python package is available at \url{https://pypi.org/project/gibson/}, the source code can be found at \url{https://github.com/micheleantonazzi/GibsonEnv}.} allows to script robot teleporting actions to any location relaxing some constraints from the physics engine such as gravity or collisions. Such a capability can enable large-scale batch data acquisition without the risk of operational failures. With this system, the robot can effectively operate over uneven floor surfaces and across different floors without encountering issues related to architectural barriers, like stairs or elevators. This approach significantly streamlines the data-gathering process, ensuring efficient and uninterrupted data collection in simulated environments. Figure~\ref{fig:gibson_example} shows an example of an acquisition obtained with this simulation framework.


\subsubsection{Pose Extraction}\label{sec:poseextraction}

To effectively exploit the simulation framework described above it is crucial to ensure that the data acquired aligns with the perception model of a service robot. To model how a robot perceives human-centric environments, we rely on the experience obtained in a long-term deployment of service robots, described in \cite{GIRAFF}. To achieve this, we propose a method for principled selection of \emph{perception poses}. First, data acquisition should occur from locations within the free space that also maintain a minimum clearance from the nearest obstacles. Additionally, these locations ought to be strategically positioned along the shortest paths connecting key areas of the environment. This positioning is key as these paths are the most likely to be covered by a robot during its service time. Third, it is important to distribute the locations uniformly throughout the environment to minimize redundancy and to ensure comprehensive coverage of the environment's visual features. Our method is composed of three distinct phases.

\begin{figure*}[!htb]
    \centering
    \begin{subfigure}[t]{0.3\textwidth}
        \centering
        \includegraphics[width=\linewidth]{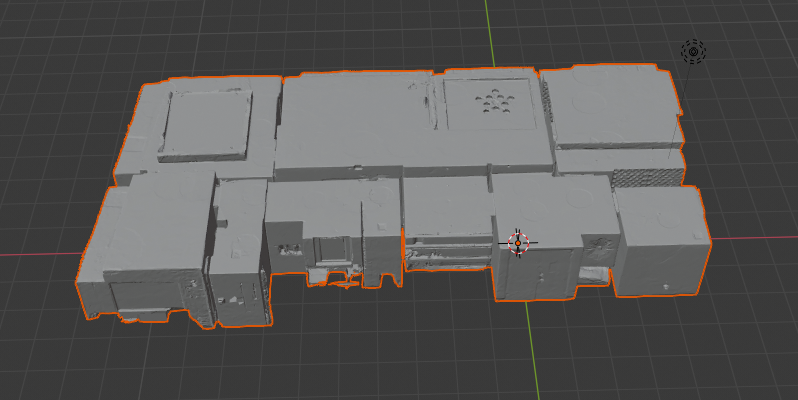}
        \caption{A 3D mesh of an environment obtained from the simulator}
        \label{fig:pose_estimator_3dmesh}
    \end{subfigure}
    \hfill
    \begin{subfigure}[t]{0.3\textwidth}
        \centering
        \includegraphics[width=\linewidth]{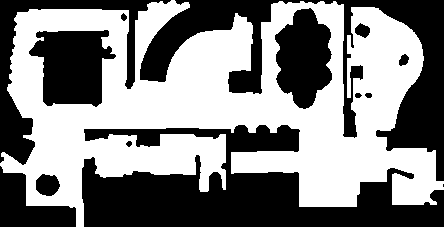}
        \caption{Map of free ($F$, white) and obstacle ($O$, black) space}
        \label{fig:pose_estimator_navigable_area}
    \end{subfigure}
    \hfill
    \begin{subfigure}[t]{0.3\textwidth}
        \centering
        \includegraphics[width=\linewidth]{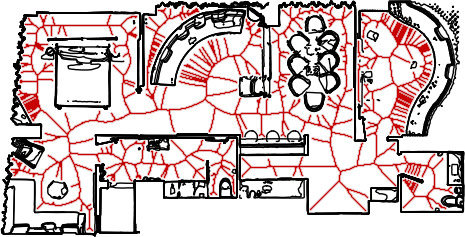}
        \caption{Grid cells ($\mathcal{G}_0$, in red) lying on the Voronoi boundary}
        \label{fig:pose_estimator_voronoi_bitmap}
    \end{subfigure}

    \vspace{1em}

    \begin{subfigure}[t]{0.45\textwidth}
        \centering
        \includegraphics[width=\linewidth]{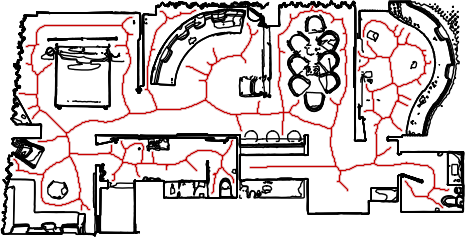}
        \caption{Navigation graph ($\mathcal{G}$, in red)}
        \label{fig:pose_estimator_navigation_graph}
    \end{subfigure}
    \hfill
    \begin{subfigure}[t]{0.45\textwidth}
        \centering
        \includegraphics[width=\linewidth]{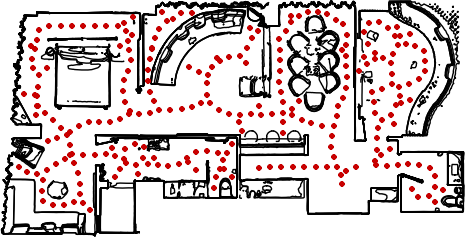}
        \caption{2D locations (in red) of the perception poses ($P$)}
        \label{fig:pose_estimator_subsampled}
    \end{subfigure}

    \caption{Different phases of our pose extraction method. \revTwo{Starting from (a) the 3D mesh of the environment, our pipeline extracts (b) the 2D map of the traversable area. Then, it generates (c) the boundary of the Voronoi graph using obstacle contours pixels as centroids, that is pruned and cleaned obtaining (d) the navigation graph, which emulates a path compliant with those delineated by the navigation stack of a real robot. Finally, our procedure samples from the navigation graph (e) the 2D poses from which to acquire the robot's perceptions.}}
    \label{fig:pose_estimator}
\end{figure*}

The initial phase focuses on generating a 2D map of the environment from the 3D mesh provided by the simulation framework. This process involves aggregating obstacles identified through multiple cross-sections of the 3D mesh, which are created using parallel planes starting from a few centimeters over the floor level. The resulting map undergoes erosion and dilation to eliminate small gaps between obstacles so as to exclude areas that are unreachable or too close to obstacles. 
Figure~\ref{fig:pose_estimator} presents some key outcomes of these steps. Specifically to this first phase, Figure~\ref{fig:pose_estimator_3dmesh} illustrates the 3D mesh of a simulated environment, while Figure~\ref{fig:pose_estimator_navigable_area} displays the corresponding 2D map.

In the second phase, the extracted map is used to compute a \emph{navigation graph}, a data structure that represents the principal routes likely to be traversed by a robot. This process is detailed in Algorithm~\ref{alg:navGraph}. The map, denoted as $\mathcal{M}=(F,O)$, comprises free and obstacle points sets denoted as $F, O \subseteq \mathbb{R}^2$, respectively. Initially, the algorithm identifies the contours of the obstacle shapes in $O$, resulting in a set of vertices $O_v \subseteq \mathbb{R}^2$ (line~\ref{alg:navGraph:Contours}). This set contains the minimum number of vertices to represent the obstacle shapes without information loss. These vertices are used as basis points for calculating the Voronoi boundary within the free space $F$ (line~\ref{alg:navGraph:Voronoi}). This boundary, separating Voronoi cells that cover $F$, is structured as an undirected graph with vertices $V_0 \subseteq \mathbb{R}^2$ and edges $E_0 \subseteq V_0 \times V_0$. The algorithm then overlays the Voronoi boundary onto a grid map that discretizes the free space $F$ at a resolution $\epsilon$. Each grid cell $c_i$ has an area of $\epsilon^2$ and is centered at coordinates $(c_i^x, c_i^y)$. A partial grid $\mathcal{G}_0$ is formed by selecting free space grid cells that contain at least one point from $O_v$ (line~\ref{alg:navGraph:Grid}, also illustrated in Figure~\ref{fig:pose_estimator_voronoi_bitmap}). Subsequently, $\mathcal{G}_0$ undergoes a heuristic filtration to eliminate spurious cells, specifically those with a degree (number of adjacent cells, assuming 8-connectivity) of $1$ or less (lines~\ref{alg:navGraph:SpuriousBegin}-\ref{alg:navGraph:SpuriousEnd}), targeting isolated or excessively narrow grid branches. The final step involves a skeletonization process~\cite{zhang1984fast} to further simplify the grid structure (line~\ref{alg:navGraph:Skeleton}). This involves converting $\mathcal{G}_0$ into a bitmap, applying the skeletonization algorithm, and then reconstructing a final grid $\mathcal{G}$, which effectively represents the navigation graph. An example of the obtained result is shown in Figure~\ref{fig:pose_estimator_navigation_graph}.

\SetKwComment{Comment}{/* }{ */}
\SetKwRepeat{Do}{do}{while}
\SetKwInput{KwInput}{Input}
\SetKwInput{KwOutput}{Output}
\begin{algorithm}[htb]
\caption{Compute navigation graph}\label{alg:navGraph}
\KwInput{$\mathcal{M}=(F,O)$, the 2D map of the environment}
\KwOutput{$\mathcal{G}$, the navigation graph}
$O_v \gets findContours(O)$\;\label{alg:navGraph:Contours}
$(V_0, E_0) \gets VoronoiBoundary(F, O_v)$\;\label{alg:navGraph:Voronoi}
$\mathcal{G}_0 \gets Grid_{\epsilon}(F,V_0, E_0)$\;\label{alg:navGraph:Grid}
\Do{filter}{\label{alg:navGraph:SpuriousBegin}
	$filter \gets false$\;
	\For{$c \in \mathcal{G}_0$}{
		\If{$degree(c) \le 1$}{
			$\mathcal{G}_0 \gets \mathcal{G}_0 \setminus c$ \Comment*[r]{Filter spurious cell}
			$filter \gets true$\;
		}
	}
}\label{alg:navGraph:SpuriousEnd}
$\mathcal{G} \gets Skeletonize(\mathcal{G}_0)$\label{alg:navGraph:Skeleton} \Comment*[r]{Apply skeletonization}
\end{algorithm}

In the third phase, the navigation graph $\mathcal{G}$ is utilized to determine the poses for data acquisition, a process detailed in Algorithm~\ref{alg:poseExtract}. A perception pose is defined by the tuple $(x, y, h, \theta)$, where $x,y$ are the 2D coordinates on the map, corresponding to the center of a cell in $\mathcal{G}$. From this location, the robot acquires an image at height $h$ and orientation $\theta$. Essentially, the algorithm performs a depth-first search on $\mathcal{G}$, generating a cluster of poses each time a distance $D$ is covered on the grid. This is achieved using a stack $S$ and a set of explored cells, denoted as $EXP$. The functions $d(\cdot, \cdot)$ and $\mathcal{N}(\cdot)$, applied over $\mathcal{G}$, compute the distance between cell pairs and identify the set of adjacent cells for a given cell, respectively (assuming again 8-connectivity). The exploration initiates from a randomly selected cell (line~\ref{alg:poseExtract:randIinit}), and whenever a distance of at least $D$ is covered (line~\ref{alg:poseExtract:coveredD}), $16$ poses centered on the current cell $c$ are added to the set $P$. These poses are generated by iterating over two height values ($h_{high}$ and $h_{low}$) and $8$ different orientations ranging from $0$ to $2\pi$ in $\frac{\pi}{4}$ increments (lines~\ref{alg:poseExtract:addPoseBegin}-\ref{alg:poseExtract:addPoseEnd}). An example of the set of 2D locations obtained over the navigation graph is depicted in Figure~\ref{fig:pose_estimator_subsampled}.

\begin{algorithm}[htb]
\caption{Pose extraction}\label{alg:poseExtract}
\KwInput{$\mathcal{G}$, the navigation graph; $D$, a distance threshold}
\KwOutput{$P$, the set of poses}
$S$, $EXP$, $P$ $\leftarrow \emptyset$, $d \gets 0$ \Comment*[r]{Initialization}
$cur \gets randomCell(\mathcal{G})$\;\label{alg:poseExtract:randIinit}
$S.push(cur)$\;
\While{$S$ not empty}{
	$c \gets S.pop()$\;
	$EXP \gets EXP \cup \{c\}$\;
	$d \gets d + dist(cur, c)$\;
	$cur \gets c$\;
	\If{$d \ge D$}{\label{alg:poseExtract:coveredD}
		\For{$h \in \{h_{high}, h_{low}\}$}{\label{alg:poseExtract:addPoseBegin}
			\For{$i \in \{0, 1, \ldots, 7\}$}{
				$P \gets P \cup (c^x, c^y, h, \frac{\pi i}{4})$\label{alg:poseExtract:addPose} \Comment*[r]{Add perception pose}
			}
		}\label{alg:poseExtract:addPoseEnd}
	$d \gets 0$\;
	}
	\For{$c' \in \mathcal{N}(c) \setminus EXP$}{
		$S.push(c')$\;
	}
}
\end{algorithm}

\subsection{Training a Qualified Detector for a Target Environment}\label{sec:qualification}

\begin{figure*}[!htbp]
	\centering
 \includegraphics[width=\linewidth]{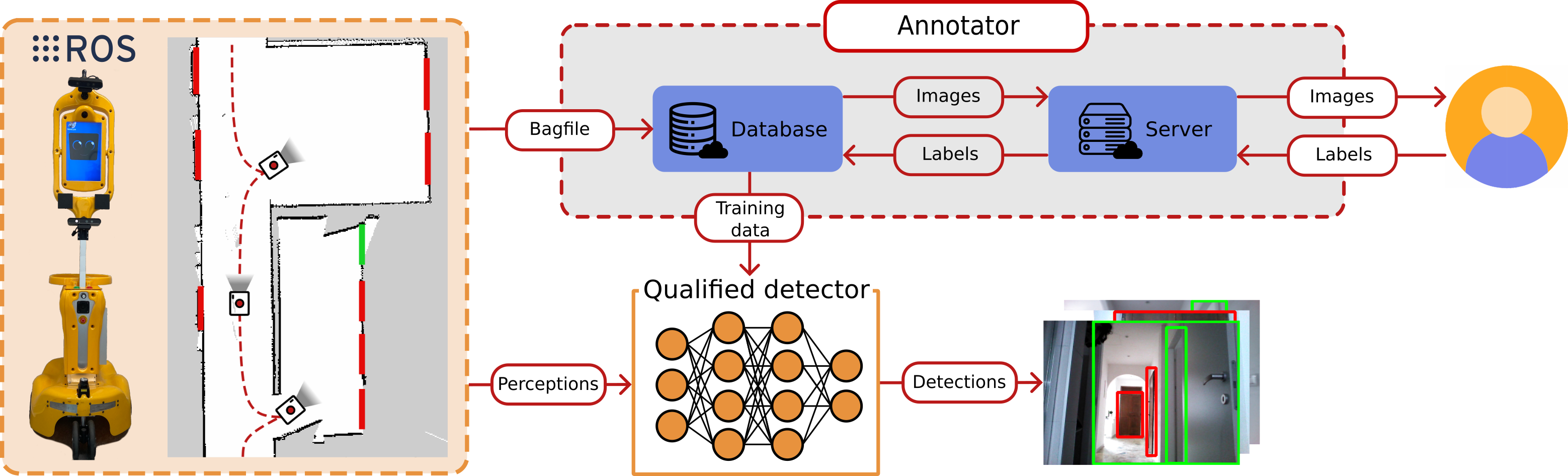}
 \caption{A general overview of the qualification procedure. \revTwo{During the initial phase of the robot's deployment, where it is tasked to acquire a map of the new environment, it collects new perceptions inside a ROS bag file. This file is then uploaded to our web-based annotation tool that extracts the RGB images and provides an interface for manual annotation. The new labels are finally exported and used to fine-tune the general detector in a qualified version with enhanced performance when used in the specific environment in which the robot will operate.}}
 \label{fig:qualification}
 \end{figure*}

During the deployment phase, the robot is set up for long-term operation in a specific target environment, denoted as $e$. A critical requirement for autonomous navigation is obtaining an on-site map. This process often involves a technician who either directly operates the robot or assists it in exploring the environment to acquire a map for later use. (We experienced this setup during an extensive experimental campaign conducted in the scope of an assistive robotics study where service robots have been installed in several private apartments~\cite{movecare, GIRAFF}. Beyond this, we deem that the setup is common and highly representative to a very large number of on-the-field installations.) In this exploration phase, the robot has the opportunity to collect additional data, particularly images of the environment captured with its onboard RGB camera. A selected portion of these images can be labeled with doors and utilized to fine-tune the general detector developed in Section~\ref{sec:gd}, tailoring it specifically to environment $e$. We refer to the adapted version of this detector as the \emph{qualified detector} for environment $e$ and we denote it as $QD_e$. A general overview of the proposed methodology is illustrated in Figure~\ref{fig:qualification}.

An intriguing approach would be to automatically label the additional acquired data in what essentially would be an instance of an unsupervised domain adaptation problem. One method is to use \emph{pseudo-labels}, generated by applying our general detector to the new samples, a technique common in semi-supervised learning~\cite{semisupervised_learning}. However, our preliminary experiments showed a significant performance drop of about $20\%$ with this method compared to results with the general detector. This decline can be attributed to the inherent inaccuracy of pseudo-labels, as also observed in recent studies~\cite{pseudolabelsRAL}. While pseudo-labels may improve performance in tasks where precise labels are less critical (such as semantic segmentation~\cite{zurbrugg2022semanticadaptation}), their lack of accuracy makes them unsuitable for object detection tasks, such as the one we consider. Indeed, re-training with missed or hallucinated bounding boxes produces a drift in the model in which errors keep getting reinforced. Exploring more advanced techniques for unsupervised domain adaptation (as discussed in~\cite{oza2023unsupervised}) is beyond the scope of this paper, where our aim is to empirically assess the trade-offs in enhancing a general detector. Consequently, we opt for manual labeling, which can be conveniently done during the robot's installation phase. This approach is widely accepted in robotics, e.g., see the work presented in~\cite{zimmermanlocalizationobjectdetection}, where manual annotations have been used to fine-tune an object detector for long-term localization tasks.

To facilitate this process, we have developed and released a ROS-integrated data annotation tool\footnotemark{}. 
This tool allows transferring robot perceptions from a ROS bag into a database. It then samples these perceptions at a given frequency and presents them to a technician, providing an interface for easy bounding box annotation. To enhance efficiency, bounding boxes from one image are retained in subsequent images, leveraging the robot’s slow movement to reduce labeling workload and reuse prior annotations.

With our experimental campaign, we prove the benefits that the qualification procedure brings to the robot's performance, studying also the trade-off between the effort between labeling costs and the model performance gain. We empirically show that a relatively limited effort is sufficient to obtain remarkably better results in object detection. In addition, we show that this procedure is more effective when applied to a GD trained with data from the robot's point of view. 

\section{Evaluation}\label{sec:evaluation}

\footnotetext{\revI{The datasets, models, scripts, and tools used for our experiments are available at \url{https://aislab.di.unimi.it/research/doordetection}}}

In this section\footnotemark[\value{footnote}], we evaluate the performance of our door detectors by presenting an extensive experimental campaign %
\revZ{of the full workflow of our method (Figure~\ref{fig:methodology}).} In Section~\ref{sec:ExperimentalSetting} we describe our experimental setting by detailing our model selection (Section~\ref{sec:ExperimentalSetting:ModelSel}), the datasets used for the trials and the details of their preparation (Section~\ref{sec:ExperimentalSetting:Datasets}), \revI{the procedures and the hyperparameters used for training and testing the detectors (Section~\ref{sec:ExperimentalSetting:Training}),} and the evaluation metrics we propose to adopt (Section~\ref{sec:ExperimentalSetting:Metrics}). We present and discuss the obtained results both with our general detectors (Section.~\ref{sec:Results:GD}) and with the qualified ones (Section~\ref{sec:Results:QD}). \RevOne{We then assess the effectiveness of the qualification procedure in long--term robot deployments by testing the robustness of the qualified detectors on data with feature shifts and focusing on challenging door instances (results and discussion are in Section~\ref{sec:Results:LongTerm}). After that,} we show how the increase in performance due to our pipeline is general regardless of the object detection method used. To do so, we compare the results obtained with three popular object detection architectures and we select the configuration that better suits our target problem~(Section~\ref{sec:Results:Comparison}). \revI{Finally, we study the impact of different performing door detectors on topology mapping, a downstream task useful to improve the long--term navigation capabilities of service robots that requires door detection~\cite{dynamicmaps, longtermnavigation}. This last evaluation is reported in  Section~\ref{sec:onTheFieldExp}.}

\subsection{Experimental Setting}\label{sec:ExperimentalSetting}

\subsubsection{Model Selection}\label{sec:ExperimentalSetting:ModelSel}

Research in deep learning for object detection primarily explored three types of deep learning architectures. Initially, the focus was on two--stage detectors, which were then followed by the development of one--stage models. More recently, considerable interest has been devoted to Transformers.

Two--stage detectors (such as~\cite{fastrcnn, fasterrcnn}) employ an architecture featuring two parts. The initial part generates \emph{proposals}, namely regions likely containing objects of interest. The second part classifies and refines these proposals in a coarse--to--fine fashion. Following a more end--to--end approach, one--stage detectors (such as~\cite{yolov3, ssd}), perform object recognition in a single step. They simultaneously predict both the locations and the labels of objects using predefined bounding boxes, known as \emph{anchors}, which are distributed uniformly across the image. Recently, Transformer--based models (such as ~\cite{detr, dynamicdetr}) have gained importance as a novel paradigm in object detection. These models first create a spatial feature map from the input image, which is then processed by a Transformer~\cite{transformer}. This process allows for the parallel prediction of multiple objects’ labels and locations, with the added advantage of considering inter--object relationships through the use of attention. (See~\cite{objectdetectionsurvey} for a more comprehensive survey of these techniques.)

\revTwo{In our experimental campaign, we selected a representative model for each architecture type, based on availability and deployment feasibility on a robotic platform.}
\revI{These models are chosen as they are widely used and stable release, to mimic a choice of a robot practitioner that is selecting such methods for a long--term deployment. However, other (more recent) models of the same families of object detectors can be used instead.}

For the two--stage architecture, we selected Faster R--CNN~\cite{fasterrcnn} as implemented in the PyTorch Hub framework. This model includes a Feature Pyramid Network (FPN) backbone based on ResNet--50~\cite{resnet}, coupled with a Region Proposal Network (RPN)~\cite{fasterrcnn} and a classifier for bounding box regression~\cite{fastrcnn}, totaling around 41 million parameters. For one--stage detectors, we opted for the medium--sized variant of YOLOv5~\cite{yolov5}, which has approximately 20 million parameters. Both these two architectures apply a non--maximum suppression procedure to discard bounding boxes with a high overlap (for any pair of bounding boxes with an overlap of $50\%$ or more, the one with the lower confidence is removed).
As for the Transformer--based model, we selected DETR~\cite{detr}, which integrates a ResNet--50 backbone~\cite{resnet} with a Transformer module~\cite{transformer} and a four--layer MLP, summing up to 41 million weights in total. DETR requires setting a critical hyperparameter, $N$, which defines the fixed number of bounding boxes predicted per image. We set $N$ to $10$, a value slightly higher than the maximum number of doors observed in any single image in our datasets, to ensure comprehensive detection without excessive computational burden.

\revI{Recently, zero--shot architectures have been proposed as a promising solution also for the task of object detection, showing remarkable results. This family of methods can be particularly interesting as it does not require additional datasets to be adapted to new tasks. Thus, we performed some preliminary examples on our task of door detection, in order to add these families of models to the three investigated here. We tested two models: Language--SAM, which combines Grounding Dino~\cite{dino} and Segment Anything~\cite{sam}, and the Transformer--based OWL--ViT~\cite{owl-vit}, two state--of--the--art zero--shot object detectors in which object categories are specified as textual queries. We prompted both models with ``open door'' and ``closed door''. 
However, we observed that the performance of models, while being able to detect doors, was significantly lower than those of one--stage, two--stages, and Transformer--based ones. In particular, most of the doors are detected multiple times, both closed and open, with similar confidence, making it difficult to disambiguate such detection to a single category. Consequently, we deem that these models are not mature enough yet to be used on challenging tasks such as robotic vision and we have not used zero--shot architectures for further evaluation. }

\subsubsection{Datasets}\label{sec:ExperimentalSetting:Datasets}

In our experiments, we considered a total of four datasets composed of images and their relative door--status annotations. 

The first dataset, which we refer to as \DDDtwo, is derived from the DD2 dataset~\cite{deepdoors2} discussed in Section~\ref{sec:gd}. This dataset includes 3000 real--world images taken from a human perspective, as provided in DD2. In these images, doors are marked as open, semi--closed, or closed. For the purposes of our experiments, we re--labeled the dataset to include ground truth data for complex examples that were not initially annotated (similar to those shown in Figure~\ref{fig:deepdoors2_examples}). Additionally, considering the operational constraints of a robot, which may not be able to navigate through partially opened doors, we categorized the doors marked as semi--closed as closed.

\begin{figure*}[!htb]
	\centering
\begin{tabular}{@{}c@{ }c@{ }c@{ }c@{ }c@{ }}
    \rotatebox[origin=c]{90}{{\small \DiG}}&
    \raisebox{-0.47\height}{\includegraphics[width=0.23\linewidth]{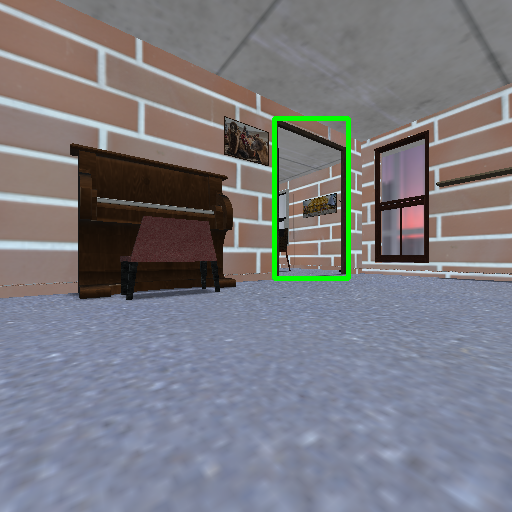}}&
    \raisebox{-0.47\height}{\includegraphics[width=0.23\linewidth]{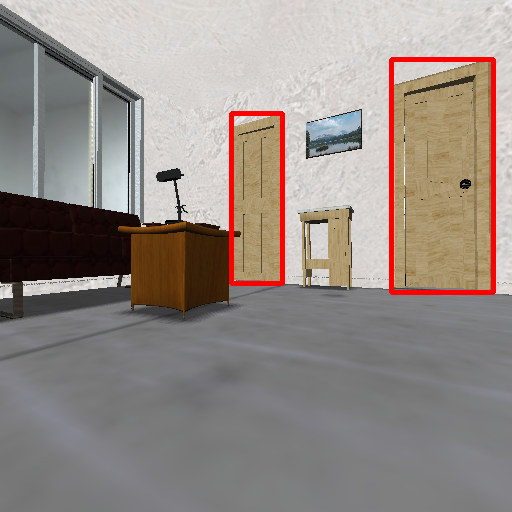}}&
    \raisebox{-0.47\height}{\includegraphics[width=0.23\linewidth]{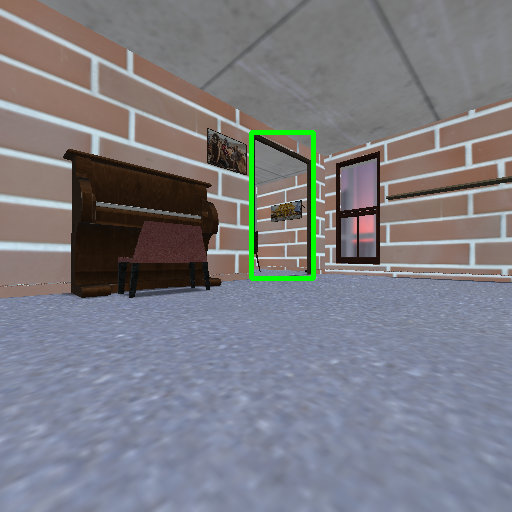}}&
    \raisebox{-0.47\height}{\includegraphics[width=0.23\linewidth]{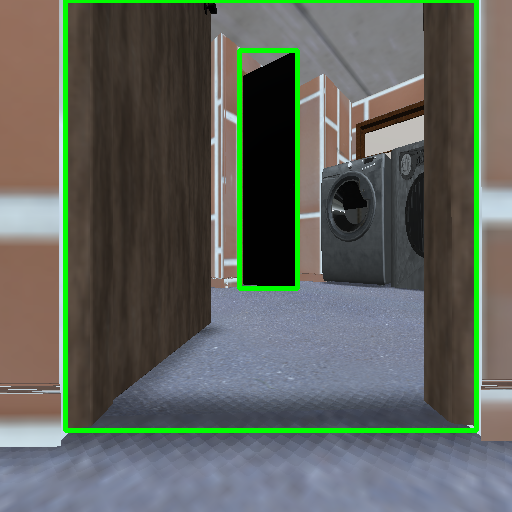}}
    \\\addlinespace[0.13cm]
    \rotatebox[origin=c]{90}{{\small \DG{}}}&
    \raisebox{-0.47\height}{\includegraphics[width=0.23\linewidth]{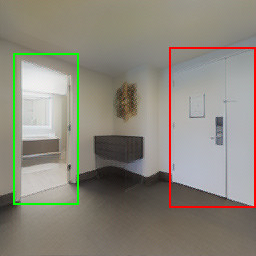}}&
    \raisebox{-0.47\height}{\includegraphics[width=0.23\linewidth]{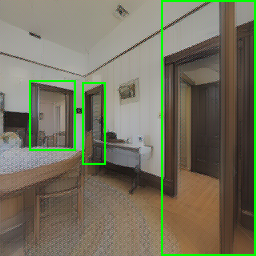}}&
    \raisebox{-0.47\height}{\includegraphics[width=0.23\linewidth]{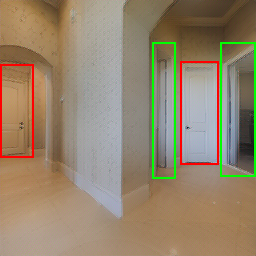}}&
    \raisebox{-0.47\height}{\includegraphics[width=0.23\linewidth]{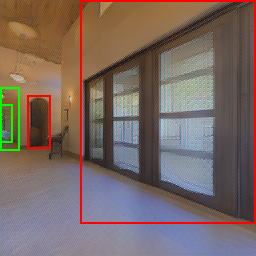}}
\end{tabular}
	\caption{Example of annotated images obtained from simulations.}
	\label{fig:igibson_and_gibson}
\end{figure*}

The second dataset, which we refer to as \DiG, was generated using the iGibson simulator~\cite{igibson}. iGibson provides 15 artificial environments, designed to mirror the structural features of real indoor scenes. To capture data from the perspectives of robots, we implemented a pose extraction mechanism akin to the one outlined in Section~\ref{sec:poseextraction}. (The details of this method are not elaborated here, as our later results will show its limited performance.) By integrating this pose extraction process with the ability to control door configurations within the simulation, we successfully generated a large batch of around 35000 instances that were automatically annotated using the semantic data provided by the simulator. (Some examples are reported in Figure~\ref{fig:igibson_and_gibson}.)

Our third dataset, referred to as \DG, was created using the Gibson--based simulation framework described in Section~\ref{sec:gd}. This dataset comprises images generated from perception poses derived using Algorithms~\ref{alg:navGraph} and~\ref{alg:poseExtract}. For this dataset, we set the distance parameter $D$ to $\SI{1}{\meter}$, and used two different robot embodiments with heights of $h_{low} = \SI{0.1}{\meter}$ and $h_{high} = \SI{0.7}{\meter}$ across 10 diverse Matterport3D environments, including small apartments and large villas with multiple floors and varied furniture styles. In processing these images, we utilized the semantic frames provided by Matterport3D, where each pixel is classified into an object category. We filtered out images without doors (i.e., where pixels labeled as ``door'' constituted less than $2.5\%$ of the total image). Subsequently, we automatically generated bounding box proposals around door instances. This pre--processing step significantly simplified the final phase of manually verifying and completing the annotations, which was carried out by human operators. The resulting \DG{} dataset comprises 5457 images all captured from the perspective of a mobile robot. \revTwo{The dataset contains approximately $6000$ door instances labeled as \texttt{open} and around $3000$ labeled as \texttt{closed}.} See some examples in Figure~\ref{fig:igibson_and_gibson}, where also the enhanced photorealism with respect to \DiG{} can be appreciated.

\begin{figure*}[!htb]
\renewcommand\thesubfigure{$e_{\arabic{subfigure}}$}
    \centering
    \subfloat[\Classrooms\label{fig:classrooms}]{\includegraphics[width=0.24\linewidth]{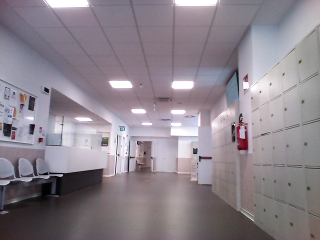}}
    \hfill
    \subfloat[\Offices\label{fig:offices}]{\includegraphics[width=0.24\linewidth]{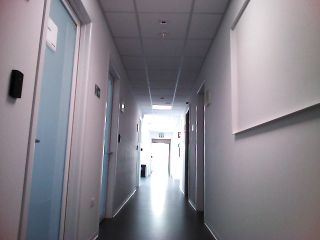}}
    \hfill
    \subfloat[\Laboratories\label{fig:laboratories}]{\includegraphics[width=0.24\linewidth]{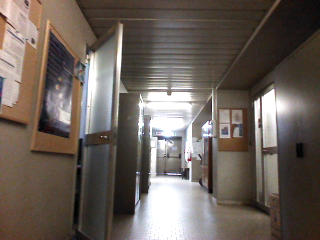}}
    \hfill
    \subfloat[\House\label{fig:house}]{\includegraphics[width=0.24\linewidth]{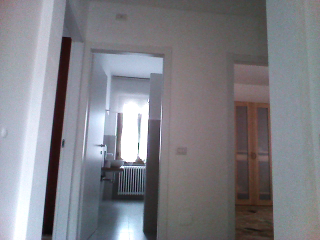}}

	\caption{Real environments considered in this work.}
    \label{fig:environments_examples}
\end{figure*}

The final dataset in our study, named \Dreal{}, is collected from a real deployment scenario of a service robot. This dataset consists of images acquired by a Giraff--X platform~\cite{movecare, GIRAFF}, as depicted in Figure~\ref{fig:giraff}, during the exploration in 4 distinct indoor settings. These environments, as depicted in Figure~\ref{fig:environments_examples}, include a variety of settings. There is a university facility characterized by open spaces and classrooms (referred to as \Classrooms), the floors of a department consisting of narrow corridors and regularly arranged offices (denoted as \Offices), a research facility with laboratories (labeled as \Laboratories), and a private apartment (identified as \House). (In Figure~\ref{fig:P1}--\ref{fig:P4} the floor plans of \Classrooms{} and \Offices{} are shown.)
Data collection was performed using an Orbbec Astra RGB--D camera (the lower camera attached to the robot in Figure~\ref{fig:giraff}), capturing 320x240 RGB images at a rate of 1 fps.  \revTwo{The dataset is composed of $3669$ images in which open and closed doors are equally distributed (approximately $4000$ instances per label)}. The images were then manually annotated \revTwo{with a particular attention on challenging door instances that are particularly relevant for our experimental campaign. }

\definecolor{darkred}{RGB}{139, 0, 0}
\definecolor{darkorange}{RGB}{204, 102, 0}
\definecolor{darkgreen}{RGB}{0, 100, 0}

\newcommand{\highR}{\textcolor{darkred}{\textbf{High}}{~}}
\newcommand{\highG}{\textcolor{darkgreen}{\textbf{High}}{~}}
\newcommand{\medium}{\textcolor{darkorange}{\textbf{Medium}}{~}}
\newcommand{\lowG}{\textcolor{darkgreen}{\textbf{Low}}{~}}
\newcommand{\lowR}{\textcolor{darkred}{\textbf{Low}}{~}}

\newcommand{\DDphoto}{\highG -- Real--world images}
\newcommand{\DDlabel}{\highR  -- Manual labeling is required}
\newcommand{\DDacquire}{\medium -- Acquisitions taken by an operator}
\newcommand{\DDnumber}{$\approx3000$ images, from several environments}

\newcommand{\iGphoto}{\lowR -- The simulator uses synthetic graphics}
\newcommand{\iGlabel}{\lowG  -- Labels provided by the simulator}
\newcommand{\iGacquire}{\lowG -- Automatized batch acquisition}
\newcommand{\iGnumber}{$\approx35000$ images, from $15$ different environments}

\newcommand{\Gphoto}{\medium -- Real--world scans with sim--2--real rendering}
\newcommand{\Glabel}{\medium  -- Manual labeling aided by simulator}
\newcommand{\Gacquire}{\lowG -- Automatized batch acquisition}
\newcommand{\Gnumber}{$\approx5500$ images, from $10$ different environments}

\newcommand{\REphoto}{\highG -- Real--world images}
\newcommand{\RElabel}{\highR  -- Manual labeling is required}
\newcommand{\REacquire}{\highR -- Real--robot deployment required}
\newcommand{\REnumber}{\revTwo{$\approx3700$} images, from $4$ different environments}

\newcommand{\Yes}{\textcolor{darkgreen}{\textbf{Yes}}}
\newcommand{\No}{\textcolor{darkred}{\textbf{No}}}

\begin{table*}[!h]
\centering
\begin{scriptsize}
\renewcommand{\arraystretch}{1.8}
\begin{tabularx}{\textwidth}{lXXXcX}
\toprule
 & \emph{Acquisition Effort} & \emph{Labeling Effort} & \emph{Photorealism} & \emph{Robot POV} & \emph{Num. of Examples} \\
\midrule
$\mathcal{D}_{\texttt{DD2}}$ & \DDacquire & \DDlabel & \DDphoto & \No & \DDnumber \\
$\mathcal{D}_{\texttt{iG}}$ & \iGacquire & \iGlabel & \iGphoto & \Yes & \iGnumber \\
$\mathcal{D}_\texttt{G}$ & \Gacquire & \Glabel & \Gphoto & \Yes & \Gnumber \\
$\mathcal{D}_{\texttt{real}}$ & \REacquire & \RElabel & \REphoto & \Yes & \REnumber \\
\bottomrule
\end{tabularx}
\end{scriptsize}
\caption{Overview of the main features of the datasets we built in this work.}
\label{tab:datasets_comparison}
\end{table*}

These datasets collectively offer a comprehensive overview of the trade--offs involved in training a door detector. \DDDtwo{} showcases what is typically available in literature but comes with significant drawbacks: the extensive effort needed for labeling and the lack in representing a robot's perception model. \DiG{} and \DG{}, on the other hand, are products of efforts to address this limitation through the use of simulation frameworks. \DiG{} maximizes the advantages of simulated data collection: images are acquired and annotated in large batches, automatically, and from a robot--centric perspective. However, this comes with a critical downside given by the lack of photorealism. Our results will demonstrate that \DG{} achieves a more favorable compromise, allowing for batch data collection from the robot’s viewpoint with reasonable effort, while ensuring a decent degree of photorealism and easing the manual annotation process. \Dreal{}, representing the ideal data set, offers the most authentic data but its high acquisition costs make it impractical for large--scale training. Table~\ref{tab:datasets_comparison} summarizes these points, giving a broad comparison of the key characteristics of each dataset together with the number of samples exploited in this work.

\revI{\subsubsection{Training and Testing}}\label{sec:ExperimentalSetting:Training}




\revI{The general detectors are obtained by re--training the pre--trained versions of DETR, YOLOv5, and Faster R--CNN on COCO 2017~\cite{coco} using the following datasets: \DiG, \DDDtwo, \DG, and \DDDtwoG. The last dataset, \DDDtwoG, is obtained by joining the examples of \DDDtwo{} and \DG.
We reduced the output layers of the three models to match the number of predicted object categories from 80 to 2. Then, we set our training parameters after a preliminary
experimental campaign that explored various batch sizes (\{1, 2, 4, 16, 32\}) and epoch numbers (\{20, 40, 60\}). 
Training is performed keeping the first layers of the models' backbones frozen, as reported in Table~\ref{tab:hyperparam}.
For Faster R--CNN and YOLOv5, we trained for 60 epochs with a batch size of 4, while DETR was trained for 60 epochs with a batch size of 1. We kept the other training hyperparameters (e.g., optimizer, learning rate, ...) as in~\cite{detr, yolov5, fasterrcnn} and we report the main ones in Table~\ref{tab:hyperparam}.
We test the general detectors in each one of the 4 real environments ${e_1, e_2, e_3, e_4}$ of \Dreal{}, depicted in Figure~\ref{fig:environments_examples}. For each environment $e$, we retain the randomly chosen $25\%$  of the images as test set, called \DrealTest{}.}

\revI{Then, we proceed with the qualification of the general detectors trained with \DDDtwo, \DG, and \DDDtwoG{} on the environments of \Dreal. The GD based on \DiG{} is not used, due to its unsatisfactory performance in the real world (see Section~\ref{sec:Results:GD}). To ease presentation, we say that a QD is \emph{based on} a dataset $\mathcal{D}_\texttt{x}$ when it is obtained from a GD trained on such a dataset. Considering each real environment $e$, we performed a series of fine--tuning rounds of each general detector using increasing amounts of data from $e$ (without considering the examples in \DrealTest). Doing this, we obtained a set of qualified detectors denoted as $QD^{15}_{e}$, $QD^{25}_{e}$, $QD^{50}_{e}$, and $QD^{75}_{e}$, where the superscripts denote the percentage of examples randomly chosen from \DrealEnv{} (the real data acquired in environment $e$) used for fine--tuning and can be interpreted as an indicator of the cost to acquire and label the additional samples. The fine--tuning is conducted using the same training parameters reported in Table~\ref{tab:hyperparam}, reducing the epochs to 40. Each qualified detector $QD^{x}_{e}$ is tested in the corresponding environment $e$ using the previously defined test set \DrealTest{} (random $25\%$ of images from \DrealEnv{} not used in any qualification round).
\begin{table*}[h]
\centering
\begin{small}
\revI{
\begin{tabular}{c|c|c|c}
\toprule

Hyperparameter &  DETR & YOLOv5& Faster R--CNN \\
\midrule

Epochs (GD/QD) & 60/40  & 60/40 & 60/40\\
Fixed layers & 11 & 27 & 11\\
Batch size & 1  & 4 & 4 \\
Optimizer & AdamW~\cite{adamw} & SGD & SGD   \\
Learning rate & $10^{-5}$  & $10^{-2}$ & $10^{-3}$ \\
Weight decay & $10^{-4}$ & $5 \times 10^{-4}$ & $5 \times 10^{-4}$ \\
Momentum  & -- & $9.37\times 10^{-1}$ & $9 \times 10 ^{-1}$ \\
Scheduler & --  & LambdaLR & StepLR \\
Step size  & --   & $1$ & $3$ \\


\bottomrule
\end{tabular}
}
\end{small}
\caption{\revI{Hyperparameters used for training the general and the qualified detectors based on DETR~\cite{detr}, YOLOv5~\cite{yolov5}, and Faster R--CNN~\cite{fasterrcnn}. ``Frozen layers'' refers to the CNN backbone layers keeping fixed during training (starting from the first). The learning rate of the CNN backbone of DETR is further decreased to $10^{-6}$ as in the original implementation~\cite{detr}. The learning rate scheduler LambdaLR linearly reduces the learning rate by subtracting $\lambda = 1.65\times 10^{-4}$ every epoch while StepLR multiplies the learning rate by a factor $\gamma = 10^{-1}$ every 3 epochs.}}
\label{tab:hyperparam}
\end{table*}}

\subsubsection{Performance Metrics}\label{sec:ExperimentalSetting:Metrics}

Our first performance metric is the mean Average Precision score (mAP), which averages the AP across all object categories (in our case, \texttt{open} and \texttt{closed} doors). The AP, as defined in~\cite{pascal}, is the area under the precision/recall curve that is interpolated at 11 evenly spaced recall levels. In our evaluation, we refine this approach by introducing additional interpolation points at each recall level where the precision reaches a local maximum. This enhancement provides a more detailed approximation of the precision/recall curve, resulting in a more accurate assessment. \revTwo{To better align the AP to our robotics context, where object detection is used for the robot's decision--making, we set the threshold of the Intersection over Union (IoU) area for positive predictions $\rho_a = 50\%$. This tailors the AP to measure the correctness of door states instead of penalizing marginal localization errors that do not prevent the bounding boxes from being used to carry out robotics downstream tasks.  Furthermore, we consider in the AP calculation only those bounding boxes with a confidence value $\geq 75\%$, thus reflecting the operational need of mobile robots in considering only high--confident predictions to avoid wrong decisions and prevent failures.}

\begin{figure*}[!htb]
\centering
\subfloat[]{
  \includegraphics[width=0.4\linewidth]{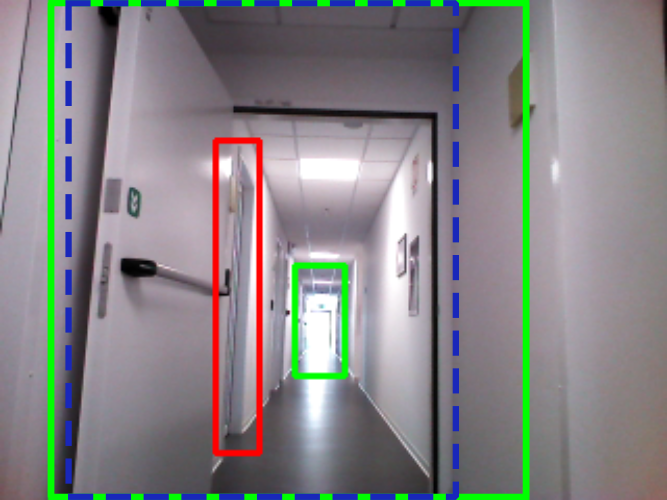}
  \label{fig:ex_metric_bbox_localization}
}
\hfill
\subfloat[]{
  \includegraphics[width=0.4\linewidth]{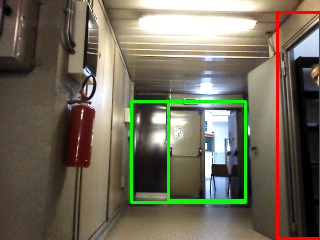}
  \label{fig:ex_wrong_label}
}
\caption{Errors made by a detector on \mbox{Giraff--X} (Figure~\ref{fig:giraff}). In~(a) the foreground green bounding box is only slightly misaligned compared to its ground truth (in dashed blue). The error affects the AP but not the robot's typical task. Similarly, in~(b) the two large green bounding boxes at the corridor's end correctly refer to the same open door; on the right, the closed door is a false positive. While the two errors affect the mAP similarly, the former is of little interest in the robotic domain, but the latter is critical for a navigating robot.}
\label{fig:metric_examples}
\end{figure*}

While the mAP is a widely accepted metric for object detection tasks, it has notable limitations in our robotic context. On one hand, certain errors disproportionately affect the AP relative to their actual impact on the robot's functionality. For instance, as illustrated in Figure~\ref{fig:ex_metric_bbox_localization}, minor inaccuracies in bounding box localization may have minimal effect on a service robot that is often primarily concerned with recognizing a door's traversability status rather than its precise localization. Furthermore, the AP treats multiple bounding boxes for the same door, as seen in Figure~\ref{fig:ex_wrong_label}, as false positives. However, a robot can resolve such ambiguities using additional information like its estimated pose and the map of the environment. On the other hand, the AP may not adequately reflect the severity of errors in identifying a door's traversability status if the bounding box is otherwise accurate. Once again, these errors are treated as false positives but, in our scenario, incorrectly classifying a \texttt{closed} door as \texttt{open} (or \emph{vice versa}) can significantly impact the robot's efficiency, especially when these classifications inform the robot’s decisions. An example of this type of error is depicted in Figure~\ref{fig:ex_wrong_label}.

Given these shortcomings, we suggest incorporating additional metrics better suited to the specific needs of the robotic application domain where door detection is crucial. These metrics are based on the premise that a service robot will invariably employ a method to sift through and select the most reliable predictions from a door detector. This process typically involves prioritizing high--confidence predictions and aggregating multiple bounding boxes that are localized in the same image region. The following definitions aim to encapsulate this approach, as well as enable the assessment of the asymmetrical nature of detection errors as previously discussed.

\revTwo{The overall procedure for the calculation of the additional metrics is detailed in Algorithm~\ref{alg:opi}.} Consider the $i$--th image $x^i \in X$ and call $Y^i$ and $\hat{Y}^i$ the set of doors present in that image and the set of predictions computed by the detector, respectively \revTwo{(line~\ref{alg:opi:line:for})}. Given a predicted bounding box $\hat{y}$, we denote as $c(\hat{y})$ the confidence associated to it by the detector and we select those predictions whose confidence is above a threshold $\rho_c$, that is $\hat{Y}_c^i=\{\hat{y} \in \hat{Y}^i \mid c(\hat{y}) \ge \rho_c\}$ \revTwo{(line~\ref{alg:opi:line:confidence})}. Given two bounding boxes $y_1$ and $y_2$, we denote as $a_{I}(y_1,y_2)$ and $a_{U}(y_1,y_2)$ the area of their intersection and union, respectively. We compute the set of \emph{Background False Detections} ($BFD$) as the confident predictions that cannot be assigned to any real door based on a threshold $\rho_a$ on their maximum Intersection Over Union area (IOU) \revTwo{(line~\ref{alg:opi:line:bfd})}. Formally, 
$$BFD^i=\Bigg\{\hat{y} \in \hat{Y}_c^i \Big| \max_{y \in Y^i}\frac{a_{I}(\hat{y},y)}{a_{U}(\hat{y},y)} < \rho_a\Bigg\}.$$
BFDs occur when a robot mistakenly identifies a door in locations where none exists, such as on a wall or a closet. As previously discussed, this type of error relates to the mislocalization of doors. In principle, a robot might correct such errors using information from its navigation stack. For example, the robot could infer from its map that a door cannot exist in a place designated as a wall. Therefore, provided these errors are not excessively frequent, they are generally deemed acceptable within typical robotic scenarios.

Confident predictions that, instead, are well localized and have an above--threshold IOU for at least one door in the image are contained in a set called $\hat{Y}_{c,a}^i = \hat{Y}_c^i \setminus BFD^i$. This allows us to define, for each ground truth door $y$, the set of predictions that are confident and whose area is maximally matched with it \revTwo{(line~\ref{alg:opi:line:b})}, formally
$$B(y) = \Bigg\{\hat{y} \in \hat{Y}_{c,a}^i \Big| \arg\max_{y \in Y^i}\frac{a_{I}(\hat{y},y)}{a_{U}(\hat{y},y)}=y\Bigg\}.$$
(Notice that, provided that ties are broken, the same prediction can never be matched to more than one door.)

Finally, we define $\hat{y}^*=\arg\max_{\hat{l} \in B(y)}c(\hat{y})$ as the most confident prediction for door $y$ \revTwo{(line~\ref{alg:opi:line:ystar})}, and it is this prediction we focus on, discarding any other predictions for the same door. \revTwo{We denote as $l(\hat{y})$ the label assigned to the prediction $\hat{y}$ by the object detector.} If $\hat{y}^*$ correctly predicts the traversability of door $y$, it is included in the set of true positives ($TP^i$) \revTwo{(line~\ref{alg:opi:line:tpadd})}. Conversely, if $\hat{y}^*$ incorrectly predicts the traversability status, it is assigned to the set of false positives ($FP^i$) \revTwo{(line~\ref{alg:opi:line:fpadd})}. A false positive substantially differs from a BFD, as an FP is potentially more consequential. An FP can lead the robot to incorrectly assess a critical aspect of the environment's topology, such as mistaking a closed door for an open passage, which could significantly impact its decisions (notice how, in this example, the environment's map cannot be exploited to fix the error).
In our evaluation, we apply the aforementioned method across all images, defining $$TP_{\%} = \frac{\sum_i |TP^i|}{\overline{Y}},~ FP_{\%} = \frac{\sum_i |FP^i|}{\overline{Y}}, ~\text{and}~ BFD_{\%} = \frac{\sum_i |BFD^i|}{\overline{Y}},$$
where $\overline{Y} = \sum_i |Y^i|$. We call these \emph{Operational Performance Indicators} (OPI), they represent the rates of true positives, false positives, and BFDs, respectively. In our experiments, the confidence threshold $\rho_{c}$ is set to $75\%$, and the IOU threshold $\rho_{a}$ is set to $50\%$.

\revTwo{
\begin{algorithm}[htb]

\caption{Calculation of the Operational Performance Indicators}\label{alg:opi}
\KwInput{$Y = \{Y^i\}$, $\hat{Y} = \{\hat{Y}^i\}$: the sets of ground truth and predicted doors divided for each image $i$
}
\KwOutput{$TP_\%$, $FP_\%$, $BFD_\%$, the Operational Performance Indicators}
$TP_\%$, $FP_\%$, $BFD_\%, \overline{Y} \leftarrow 0$ 
\For{$Y^i, \hat{Y}^i \in Y, \hat{Y}$} {\label{alg:opi:line:for}
$\overline{Y} \leftarrow \overline{Y} + |Y^i|$\;
$\hat{Y}_c^i\leftarrow\{\hat{y} \in \hat{Y}^i \mid c(\hat{y}) \ge \rho_c\}$\Comment*[r]{Select the most confident prediction}\label{alg:opi:line:confidence}
$BFD^i \leftarrow \big\{\hat{y} \in \hat{Y}_c^i \big| \max_{y \in Y^i}\frac{a_{I}(\hat{y},y)}{a_{U}(\hat{y},y)} < \rho_a\big\} $\label{alg:opi:line:bfd}\;
$TP^i, FP^i \leftarrow \emptyset$\;
\For{$y \in Y^i$}{
    $B(y) \leftarrow \big\{\hat{y} \in \hat{Y}_{c}^i \setminus BFD^i \big| \arg\max_{y \in Y^i}\frac{a_{I}(\hat{y},y)}{a_{U}(\hat{y},y)}=y\big\}$\label{alg:opi:line:b}\;
    $\hat{y}^*=\arg\max_{\hat{l} \in B(y)}c(\hat{y})$\label{alg:opi:line:ystar}\;
    \uIf{$l(\hat{y}^*) = l(y)$}{
        $TP^i \leftarrow TP^i \cup \{\hat{y}^*\}$\label{alg:opi:line:tpadd}\;
    }
    \Else{
    $FP^i \leftarrow FP^i \cup \{\hat{y}^*\}$\label{alg:opi:line:fpadd}\;
    }
}
$TP_\%, \leftarrow TP_\% + |TP^i|$\; 
$FP_\% \leftarrow FP_\% + |FP^i|$\; 
$BFD_\% \leftarrow BFD_\% + |BFD^i|$\; 
}
$TP_\%, FP_\%, BFD_\% \leftarrow \frac{TP_\%}{\overline{Y}}, \frac{FP_\%}{\overline{Y}},\frac{BFD_\%}{\overline{Y}}$\;

\end{algorithm}
}

\subsection{Evaluation of General Detectors}\label{sec:Results:GD}




\begin{table*}[!h]
\setlength\tabcolsep{3pt}
\setlength\extrarowheight{2pt}
\centering

\begin{tabular}{cc|cccc|cccc|cccc}
\toprule
     &  & \multicolumn{4}{c|}{DETR~\cite{detr}} & \multicolumn{4}{c|}{YOLOv5~\cite{yolov5}} & \multicolumn{4}{c}{Faster~R-CNN~\cite{fasterrcnn}} \\[2pt]
        \textbf{Env.} & \textbf{Dataset} &\textbf{mAP}$\uparrow$& $\mathbf{TP_\%}$$\uparrow$ &  $\mathbf{FP_\%}$$\downarrow$ & $\mathbf{BFD_\%}$$\downarrow$ & \textbf{mAP}$\uparrow$ & $\mathbf{TP_\%}$$\uparrow$ &  $\mathbf{FP_\%}$$\downarrow$ & $\mathbf{BFD_\%}$$\downarrow$ & \textbf{mAP}$\uparrow$ & $\mathbf{TP_\%}$$\uparrow$ &  $\mathbf{FP_\%}$$\downarrow$ & $\mathbf{BFD_\%}$$\downarrow$ \\
	\midrule
\multicolumn{1}{c}{\multirow{4}{*}{$e_1$}} &\DiG& 0 & 1 & \textbf{0} & 26 & 0 & 0 & \textbf{0} & \textbf{0} & 2 & 2 & \textbf{0} & \textbf{2}\\
\multicolumn{1}{c}{} &\DDDtwo& 13 & 18 & 9 & \textbf{13} & 2 & 3 & 2 & \underline{1} & 18 & 25 & 14 & \underline{9}\\
\multicolumn{1}{c}{} &\DG& \underline{26} & \underline{30} & 7 & 22 & \underline{30} & \underline{31} & \underline{1} & 8 & \underline{20} & \underline{25} & \underline{6} & 11\\
\multicolumn{1}{c}{} &\DDDtwoG& \textbf{32} & \textbf{37} & \underline{6} & \underline{17} & \textbf{32} & \textbf{34} & 2 & 3 & \textbf{34} & \textbf{43} & 10 & 14\\
[2pt]\hline
\multicolumn{1}{c}{\multirow{4}{*}{$e_2$}} &\DiG& 0 & 1 & \textbf{1} & 22 & 0 & 0 & \textbf{0} & \textbf{0} & 0 & 1 & \textbf{0} & \textbf{3}\\
\multicolumn{1}{c}{} &\DDDtwo& 14 & 19 & 8 & \textbf{17} & 3 & 5 & \underline{1} & \underline{3} & \underline{22} & \underline{27} & \underline{4} & 18\\
\multicolumn{1}{c}{} &\DG& \textbf{28} & \textbf{36} & \underline{6} & 21 & \underline{14} & \underline{21} & 9 & 9 & 14 & 17 & 4 & \underline{10}\\
\multicolumn{1}{c}{} &\DDDtwoG& \underline{24} & \underline{31} & 10 & \underline{19} & \textbf{16} & \textbf{24} & 10 & 9 & \textbf{27} & \textbf{34} & 5 & 20\\
[2pt]\hline
\multicolumn{1}{c}{\multirow{4}{*}{$e_3$}} &\DiG& 0 & 2 & \textbf{0} & 35 & 0 & 0 & \textbf{0} & \textbf{1} & 0 & 1 & \textbf{1} & \underline{11}\\
\multicolumn{1}{c}{} &\DDDtwo& 9 & 15 & \underline{3} & \textbf{30} & 3 & 3 & \underline{0} & \underline{1} & \underline{10} & \underline{20} & 8 & 40\\
\multicolumn{1}{c}{} &\DG& \underline{13} & \underline{19} & 6 & \underline{33} & \underline{4} & \underline{6} & 3 & 10 & 2 & 4 & \underline{2} & \textbf{10}\\
\multicolumn{1}{c}{} &\DDDtwoG& \textbf{16} & \textbf{24} & 4 & 44 & \textbf{6} & \textbf{10} & 2 & 12 & \textbf{14} & \textbf{24} & 8 & 34\\
[2pt]\hline
\multicolumn{1}{c}{\multirow{4}{*}{$e_4$}} &\DiG& 1 & 5 & \textbf{3} & 25 & 0 & 0 & \textbf{1} & \underline{4} & 1 & 3 & \underline{7} & \underline{7}\\
\multicolumn{1}{c}{} &\DDDtwo& 22 & 20 & 14 & \textbf{9} & 14 & 12 & 3 & \textbf{1} & \underline{31} & \underline{35} & 9 & 14\\
\multicolumn{1}{c}{} &\DG& \underline{31} & \textbf{40} & \underline{9} & \underline{11} & \underline{16} & \underline{22} & \underline{2} & 4 & 12 & 18 & \textbf{4} & \textbf{6}\\
\multicolumn{1}{c}{} &\DDDtwoG& \textbf{32} & \underline{35} & 10 & 13 & \textbf{30} & \textbf{34} & 7 & 7 & \textbf{48} & \textbf{49} & 7 & 16\\
\bottomrule
\end{tabular}

\caption{Real-World performance of general detectors. The best and second-best results among the training datasets are highlighted in bold and underlined, respectively.}
\label{tab:gd_results}
\end{table*}

\begin{figure*}[!htb]
    \centering
    \begin{subfigure}[t]{\textwidth}
    \centering
    \begin{tikzpicture}[scale=0.8]

\definecolor{crimson2143940}{RGB}{214,39,40}
\definecolor{darkgray176}{RGB}{176,176,176}
\definecolor{darkorange25512714}{RGB}{255,127,14}
\definecolor{forestgreen4416044}{RGB}{44,160,44}
\definecolor{lightgray204}{RGB}{204,204,204}
\definecolor{steelblue31119180}{RGB}{31,119,180}

\begin{axis}[
hide axis,
legend columns=6,
legend cell align={left},
legend style={
/tikz/every even column/.append style={column sep=0.3cm},
  fill opacity=0.8,
  draw opacity=1,
  text opacity=1,
  at={(0.5,0.95)},
  anchor=north,
  draw=lightgray204
},
xmin=-0.315, xmax=4.855,
y grid style={darkgray176},
ymin=0, ymax=130
]
\addlegendimage{ybar,area legend,draw=steelblue31119180,fill=steelblue31119180}
\addlegendentry{\DiG}
\addlegendimage{ybar,area legend,draw=forestgreen4416044,fill=forestgreen4416044,opacity=0.9,thick}
\addlegendentry{\DDDtwo}
\addlegendimage{ybar,area legend,draw=darkorange25512714,fill=darkorange25512714,opacity=0.9,thick}
\addlegendentry{\DG}
\addlegendimage{ybar,area legend,draw=crimson2143940,fill=crimson2143940,opacity=0.9,thick}
\addlegendentry{\DDDtwoG}
\addlegendimage{ybar,area legend,draw=black,fill=none,opacity=0.9,thick,postaction={pattern=north east lines, fill opacity=0.9}}
\addlegendentry{Closed door}
\addlegendimage{ybar,area legend,draw=black,fill=none,opacity=0.9,thick}
\addlegendentry{Open door}
\end{axis}
\end{tikzpicture}
    \end{subfigure}
    \\[-0.8cm]
    \begin{subfigure}[t]{0.48\textwidth}
        \centering
\begin{tikzpicture}[scale=0.63]

\definecolor{crimson2143940}{RGB}{214,39,40}
\definecolor{darkgray176}{RGB}{176,176,176}
\definecolor{darkorange25512714}{RGB}{255,127,14}
\definecolor{forestgreen4416044}{RGB}{44,160,44}
\definecolor{lightgray204}{RGB}{204,204,204}
\definecolor{steelblue31119180}{RGB}{31,119,180}

\begin{axis}[
width=12cm,
height=7cm,
legend columns=3,
legend cell align={left},
legend style={
/tikz/every even column/.append style={column sep=0.3cm},
  fill opacity=0.8,
  draw opacity=1,
  text opacity=1,
  at={(0.5,0.95)},
  anchor=north,
  draw=lightgray204
},
tick align=outside,
title={\(\displaystyle e_1\) -- \Classrooms},
x grid style={darkgray176},
xmajorticks=false,
xmin=-0.178, xmax=2.858,
xtick style={color=black},
y grid style={darkgray176},
ylabel={mAP},
ymin=0, ymax=50,
ytick pos=left,
ytick style={color=black}
]
\draw[draw=black,fill=steelblue31119180,opacity=0.9,thick,postaction={pattern=north east lines, fill opacity=0.9}] (axis cs:-0.04,0) rectangle (axis cs:0.12,0);
\draw[draw=black,fill=steelblue31119180,opacity=0.9,thick,postaction={pattern=north east lines, fill opacity=0.9}] (axis cs:0.96,0) rectangle (axis cs:1.12,0);
\draw[draw=black,fill=steelblue31119180,opacity=0.9,thick,postaction={pattern=north east lines, fill opacity=0.9}] (axis cs:1.96,0) rectangle (axis cs:2.12,0);
\draw[draw=black,fill=steelblue31119180,opacity=0.9,thick] (axis cs:-0.04,0) rectangle (axis cs:0.12,0);
\draw[draw=black,fill=steelblue31119180,opacity=0.9,thick] (axis cs:0.96,0) rectangle (axis cs:1.12,0.5);
\draw[draw=black,fill=steelblue31119180,opacity=0.9,thick] (axis cs:1.96,0) rectangle (axis cs:2.12,1.5);
\draw[draw=black,fill=forestgreen4416044,opacity=0.9,thick,postaction={pattern=north east lines, fill opacity=0.9}] (axis cs:0.16,0) rectangle (axis cs:0.32,4);
\draw[draw=black,fill=forestgreen4416044,opacity=0.9,thick,postaction={pattern=north east lines, fill opacity=0.9}] (axis cs:1.16,0) rectangle (axis cs:1.32,2);
\draw[draw=black,fill=forestgreen4416044,opacity=0.9,thick,postaction={pattern=north east lines, fill opacity=0.9}] (axis cs:2.16,0) rectangle (axis cs:2.32,4);
\draw[draw=black,fill=forestgreen4416044,opacity=0.9,thick] (axis cs:0.16,4) rectangle (axis cs:0.32,13);
\draw[draw=black,fill=forestgreen4416044,opacity=0.9,thick] (axis cs:1.16,2) rectangle (axis cs:1.32,2.5);
\draw[draw=black,fill=forestgreen4416044,opacity=0.9,thick] (axis cs:2.16,4) rectangle (axis cs:2.32,18.5);
\draw[draw=black,fill=darkorange25512714,opacity=0.9,thick,postaction={pattern=north east lines, fill opacity=0.9}] (axis cs:0.36,0) rectangle (axis cs:0.52,11.5);
\draw[draw=black,fill=darkorange25512714,opacity=0.9,thick,postaction={pattern=north east lines, fill opacity=0.9}] (axis cs:1.36,0) rectangle (axis cs:1.52,16);
\draw[draw=black,fill=darkorange25512714,opacity=0.9,thick,postaction={pattern=north east lines, fill opacity=0.9}] (axis cs:2.36,0) rectangle (axis cs:2.52,11);
\draw[draw=black,fill=darkorange25512714,opacity=0.9,thick] (axis cs:0.36,11.5) rectangle (axis cs:0.52,25.5);
\draw[draw=black,fill=darkorange25512714,opacity=0.9,thick] (axis cs:1.36,16) rectangle (axis cs:1.52,29.5);
\draw[draw=black,fill=darkorange25512714,opacity=0.9,thick] (axis cs:2.36,11) rectangle (axis cs:2.52,20.5);
\draw[draw=black,fill=crimson2143940,opacity=0.9,thick,postaction={pattern=north east lines, fill opacity=0.9}] (axis cs:0.56,0) rectangle (axis cs:0.72,19.5);
\draw[draw=black,fill=crimson2143940,opacity=0.9,thick,postaction={pattern=north east lines, fill opacity=0.9}] (axis cs:1.56,0) rectangle (axis cs:1.72,19);
\draw[draw=black,fill=crimson2143940,opacity=0.9,thick,postaction={pattern=north east lines, fill opacity=0.9}] (axis cs:2.56,0) rectangle (axis cs:2.72,22);
\draw[draw=black,fill=crimson2143940,opacity=0.9,thick] (axis cs:0.56,19.5) rectangle (axis cs:0.72,32.5);
\draw[draw=black,fill=crimson2143940,opacity=0.9,thick] (axis cs:1.56,19) rectangle (axis cs:1.72,32);
\draw[draw=black,fill=crimson2143940,opacity=0.9,thick] (axis cs:2.56,22) rectangle (axis cs:2.72,34.5);
\end{axis}

\end{tikzpicture}
    \end{subfigure}
    \hfill
    \begin{subfigure}[t]{0.48\textwidth}
        \centering
\begin{tikzpicture}[scale=0.63]

\definecolor{crimson2143940}{RGB}{214,39,40}
\definecolor{darkgray176}{RGB}{176,176,176}
\definecolor{darkorange25512714}{RGB}{255,127,14}
\definecolor{forestgreen4416044}{RGB}{44,160,44}
\definecolor{lightgray204}{RGB}{204,204,204}
\definecolor{steelblue31119180}{RGB}{31,119,180}

\begin{axis}[
width=12cm,
height=7cm,
legend columns=3,
legend cell align={left},
legend style={
/tikz/every even column/.append style={column sep=0.3cm},
  fill opacity=0.8,
  draw opacity=1,
  text opacity=1,
  at={(0.5,0.95)},
  anchor=north,
  draw=lightgray204
},
tick align=outside,
title={\(\displaystyle e_2\) -- \Offices},
x grid style={darkgray176},
xmajorticks=false,
xmin=-0.178, xmax=2.858,
xtick style={color=black},
y grid style={darkgray176},
ymajorticks=false,
ymin=0, ymax=50,
ytick style={color=black}
]
\draw[draw=black,fill=steelblue31119180,opacity=0.9,thick,postaction={pattern=north east lines, fill opacity=0.9}] (axis cs:-0.04,0) rectangle (axis cs:0.12,0);
\draw[draw=black,fill=steelblue31119180,opacity=0.9,thick,postaction={pattern=north east lines, fill opacity=0.9}] (axis cs:0.96,0) rectangle (axis cs:1.12,0);
\draw[draw=black,fill=steelblue31119180,opacity=0.9,thick,postaction={pattern=north east lines, fill opacity=0.9}] (axis cs:1.96,0) rectangle (axis cs:2.12,0);
\draw[draw=black,fill=steelblue31119180,opacity=0.9,thick] (axis cs:-0.04,0) rectangle (axis cs:0.12,0);
\draw[draw=black,fill=steelblue31119180,opacity=0.9,thick] (axis cs:0.96,0) rectangle (axis cs:1.12,0.5);
\draw[draw=black,fill=steelblue31119180,opacity=0.9,thick] (axis cs:1.96,0) rectangle (axis cs:2.12,0.5);
\draw[draw=black,fill=forestgreen4416044,opacity=0.9,thick,postaction={pattern=north east lines, fill opacity=0.9}] (axis cs:0.16,0) rectangle (axis cs:0.32,2);
\draw[draw=black,fill=forestgreen4416044,opacity=0.9,thick,postaction={pattern=north east lines, fill opacity=0.9}] (axis cs:1.16,0) rectangle (axis cs:1.32,1);
\draw[draw=black,fill=forestgreen4416044,opacity=0.9,thick,postaction={pattern=north east lines, fill opacity=0.9}] (axis cs:2.16,0) rectangle (axis cs:2.32,1);
\draw[draw=black,fill=forestgreen4416044,opacity=0.9,thick] (axis cs:0.16,2) rectangle (axis cs:0.32,13.5);
\draw[draw=black,fill=forestgreen4416044,opacity=0.9,thick] (axis cs:1.16,1) rectangle (axis cs:1.32,3);
\draw[draw=black,fill=forestgreen4416044,opacity=0.9,thick] (axis cs:2.16,1) rectangle (axis cs:2.32,21.5);
\draw[draw=black,fill=darkorange25512714,opacity=0.9,thick,postaction={pattern=north east lines, fill opacity=0.9}] (axis cs:0.36,0) rectangle (axis cs:0.52,6);
\draw[draw=black,fill=darkorange25512714,opacity=0.9,thick,postaction={pattern=north east lines, fill opacity=0.9}] (axis cs:1.36,0) rectangle (axis cs:1.52,4);
\draw[draw=black,fill=darkorange25512714,opacity=0.9,thick,postaction={pattern=north east lines, fill opacity=0.9}] (axis cs:2.36,0) rectangle (axis cs:2.52,2);
\draw[draw=black,fill=darkorange25512714,opacity=0.9,thick] (axis cs:0.36,6) rectangle (axis cs:0.52,27.5);
\draw[draw=black,fill=darkorange25512714,opacity=0.9,thick] (axis cs:1.36,4) rectangle (axis cs:1.52,13.5);
\draw[draw=black,fill=darkorange25512714,opacity=0.9,thick] (axis cs:2.36,2) rectangle (axis cs:2.52,13.5);
\draw[draw=black,fill=crimson2143940,opacity=0.9,thick,postaction={pattern=north east lines, fill opacity=0.9}] (axis cs:0.56,0) rectangle (axis cs:0.72,6.5);
\draw[draw=black,fill=crimson2143940,opacity=0.9,thick,postaction={pattern=north east lines, fill opacity=0.9}] (axis cs:1.56,0) rectangle (axis cs:1.72,5.5);
\draw[draw=black,fill=crimson2143940,opacity=0.9,thick,postaction={pattern=north east lines, fill opacity=0.9}] (axis cs:2.56,0) rectangle (axis cs:2.72,5);
\draw[draw=black,fill=crimson2143940,opacity=0.9,thick] (axis cs:0.56,6.5) rectangle (axis cs:0.72,24);
\draw[draw=black,fill=crimson2143940,opacity=0.9,thick] (axis cs:1.56,5.5) rectangle (axis cs:1.72,16.5);
\draw[draw=black,fill=crimson2143940,opacity=0.9,thick] (axis cs:2.56,5) rectangle (axis cs:2.72,27);
\end{axis}

\end{tikzpicture}
    \end{subfigure}
    \\[-0.8cm]
    \begin{subfigure}[t]{0.48\textwidth}
        \centering
\begin{tikzpicture}[scale=0.63]

\definecolor{crimson2143940}{RGB}{214,39,40}
\definecolor{darkgray176}{RGB}{176,176,176}
\definecolor{darkorange25512714}{RGB}{255,127,14}
\definecolor{forestgreen4416044}{RGB}{44,160,44}
\definecolor{lightgray204}{RGB}{204,204,204}
\definecolor{steelblue31119180}{RGB}{31,119,180}

\begin{axis}[
width=12cm,
height=7cm,
legend columns=3,
legend cell align={left},
legend style={
/tikz/every even column/.append style={column sep=0.3cm},
  fill opacity=0.8,
  draw opacity=1,
  text opacity=1,
  at={(0.5,0.95)},
  anchor=north,
  draw=lightgray204
},
tick align=outside,
tick pos=left,
title={\(\displaystyle e_3\) -- \Laboratories},
x grid style={darkgray176},
xlabel={Model},
xmin=-0.178, xmax=2.858,
xtick style={color=black},
xtick={0.34,1.34,2.34},
xticklabels={DETR~\cite{detr},YOLOv5~\cite{yolov5},Faster~R--CNN~\cite{fasterrcnn}},
y grid style={darkgray176},
ylabel={mAP},
ymin=0, ymax=50,
ytick style={color=black}
]
\draw[draw=black,fill=steelblue31119180,opacity=0.9,thick,postaction={pattern=north east lines, fill opacity=0.9}] (axis cs:-0.04,0) rectangle (axis cs:0.12,0);
\draw[draw=black,fill=steelblue31119180,opacity=0.9,thick,postaction={pattern=north east lines, fill opacity=0.9}] (axis cs:0.96,0) rectangle (axis cs:1.12,0);
\draw[draw=black,fill=steelblue31119180,opacity=0.9,thick,postaction={pattern=north east lines, fill opacity=0.9}] (axis cs:1.96,0) rectangle (axis cs:2.12,0);
\draw[draw=black,fill=steelblue31119180,opacity=0.9,thick] (axis cs:-0.04,0) rectangle (axis cs:0.12,0.5);
\draw[draw=black,fill=steelblue31119180,opacity=0.9,thick] (axis cs:0.96,0) rectangle (axis cs:1.12,0);
\draw[draw=black,fill=steelblue31119180,opacity=0.9,thick] (axis cs:1.96,0) rectangle (axis cs:2.12,0);
\draw[draw=black,fill=forestgreen4416044,opacity=0.9,thick,postaction={pattern=north east lines, fill opacity=0.9}] (axis cs:0.16,0) rectangle (axis cs:0.32,1.5);
\draw[draw=black,fill=forestgreen4416044,opacity=0.9,thick,postaction={pattern=north east lines, fill opacity=0.9}] (axis cs:1.16,0) rectangle (axis cs:1.32,2);
\draw[draw=black,fill=forestgreen4416044,opacity=0.9,thick,postaction={pattern=north east lines, fill opacity=0.9}] (axis cs:2.16,0) rectangle (axis cs:2.32,3.5);
\draw[draw=black,fill=forestgreen4416044,opacity=0.9,thick] (axis cs:0.16,1.5) rectangle (axis cs:0.32,9);
\draw[draw=black,fill=forestgreen4416044,opacity=0.9,thick] (axis cs:1.16,2) rectangle (axis cs:1.32,3);
\draw[draw=black,fill=forestgreen4416044,opacity=0.9,thick] (axis cs:2.16,3.5) rectangle (axis cs:2.32,10);
\draw[draw=black,fill=darkorange25512714,opacity=0.9,thick,postaction={pattern=north east lines, fill opacity=0.9}] (axis cs:0.36,0) rectangle (axis cs:0.52,2);
\draw[draw=black,fill=darkorange25512714,opacity=0.9,thick,postaction={pattern=north east lines, fill opacity=0.9}] (axis cs:1.36,0) rectangle (axis cs:1.52,0.5);
\draw[draw=black,fill=darkorange25512714,opacity=0.9,thick,postaction={pattern=north east lines, fill opacity=0.9}] (axis cs:2.36,0) rectangle (axis cs:2.52,1);
\draw[draw=black,fill=darkorange25512714,opacity=0.9,thick] (axis cs:0.36,2) rectangle (axis cs:0.52,13);
\draw[draw=black,fill=darkorange25512714,opacity=0.9,thick] (axis cs:1.36,0.5) rectangle (axis cs:1.52,3.5);
\draw[draw=black,fill=darkorange25512714,opacity=0.9,thick] (axis cs:2.36,1) rectangle (axis cs:2.52,2);
\draw[draw=black,fill=crimson2143940,opacity=0.9,thick,postaction={pattern=north east lines, fill opacity=0.9}] (axis cs:0.56,0) rectangle (axis cs:0.72,5.5);
\draw[draw=black,fill=crimson2143940,opacity=0.9,thick,postaction={pattern=north east lines, fill opacity=0.9}] (axis cs:1.56,0) rectangle (axis cs:1.72,3);
\draw[draw=black,fill=crimson2143940,opacity=0.9,thick,postaction={pattern=north east lines, fill opacity=0.9}] (axis cs:2.56,0) rectangle (axis cs:2.72,9);
\draw[draw=black,fill=crimson2143940,opacity=0.9,thick] (axis cs:0.56,5.5) rectangle (axis cs:0.72,15.5);
\draw[draw=black,fill=crimson2143940,opacity=0.9,thick] (axis cs:1.56,3) rectangle (axis cs:1.72,5.5);
\draw[draw=black,fill=crimson2143940,opacity=0.9,thick] (axis cs:2.56,9) rectangle (axis cs:2.72,14);
\end{axis}

\end{tikzpicture}
    \end{subfigure}
    \hfill
    \begin{subfigure}[t]{0.48\textwidth}
        \centering
\begin{tikzpicture}[scale=0.63]

\definecolor{crimson2143940}{RGB}{214,39,40}
\definecolor{darkgray176}{RGB}{176,176,176}
\definecolor{darkorange25512714}{RGB}{255,127,14}
\definecolor{forestgreen4416044}{RGB}{44,160,44}
\definecolor{lightgray204}{RGB}{204,204,204}
\definecolor{steelblue31119180}{RGB}{31,119,180}

\begin{axis}[
width=12cm,
height=7cm,
legend columns=3,
legend cell align={left},
legend style={
/tikz/every even column/.append style={column sep=0.3cm},
  fill opacity=0.8,
  draw opacity=1,
  text opacity=1,
  at={(0.5,0.95)},
  anchor=north,
  draw=lightgray204
},
tick align=outside,
title={\(\displaystyle e_4\) -- \House},
x grid style={darkgray176},
xlabel={Model},
xmin=-0.178, xmax=2.858,
xtick pos=left,
xtick style={color=black},
xtick={0.34,1.34,2.34},
xticklabels={DETR~\cite{detr},YOLOv5~\cite{yolov5},Faster~R--CNN~\cite{fasterrcnn}},
y grid style={darkgray176},
ymajorticks=false,
ymin=0, ymax=50,
ytick style={color=black}
]
\draw[draw=black,fill=steelblue31119180,opacity=0.9,thick,postaction={pattern=north east lines, fill opacity=0.9}] (axis cs:-0.04,0) rectangle (axis cs:0.12,0);
\draw[draw=black,fill=steelblue31119180,opacity=0.9,thick,postaction={pattern=north east lines, fill opacity=0.9}] (axis cs:0.96,0) rectangle (axis cs:1.12,0);
\draw[draw=black,fill=steelblue31119180,opacity=0.9,thick,postaction={pattern=north east lines, fill opacity=0.9}] (axis cs:1.96,0) rectangle (axis cs:2.12,0.5);
\draw[draw=black,fill=steelblue31119180,opacity=0.9,thick] (axis cs:-0.04,0) rectangle (axis cs:0.12,1);
\draw[draw=black,fill=steelblue31119180,opacity=0.9,thick] (axis cs:0.96,0) rectangle (axis cs:1.12,0);
\draw[draw=black,fill=steelblue31119180,opacity=0.9,thick] (axis cs:1.96,0.5) rectangle (axis cs:2.12,1);
\draw[draw=black,fill=forestgreen4416044,opacity=0.9,thick,postaction={pattern=north east lines, fill opacity=0.9}] (axis cs:0.16,0) rectangle (axis cs:0.32,19);
\draw[draw=black,fill=forestgreen4416044,opacity=0.9,thick,postaction={pattern=north east lines, fill opacity=0.9}] (axis cs:1.16,0) rectangle (axis cs:1.32,14.5);
\draw[draw=black,fill=forestgreen4416044,opacity=0.9,thick,postaction={pattern=north east lines, fill opacity=0.9}] (axis cs:2.16,0) rectangle (axis cs:2.32,21.5);
\draw[draw=black,fill=forestgreen4416044,opacity=0.9,thick] (axis cs:0.16,19) rectangle (axis cs:0.32,21.5);
\draw[draw=black,fill=forestgreen4416044,opacity=0.9,thick] (axis cs:1.16,14.5) rectangle (axis cs:1.32,14.5);
\draw[draw=black,fill=forestgreen4416044,opacity=0.9,thick] (axis cs:2.16,21.5) rectangle (axis cs:2.32,31);
\draw[draw=black,fill=darkorange25512714,opacity=0.9,thick,postaction={pattern=north east lines, fill opacity=0.9}] (axis cs:0.36,0) rectangle (axis cs:0.52,9);
\draw[draw=black,fill=darkorange25512714,opacity=0.9,thick,postaction={pattern=north east lines, fill opacity=0.9}] (axis cs:1.36,0) rectangle (axis cs:1.52,3);
\draw[draw=black,fill=darkorange25512714,opacity=0.9,thick,postaction={pattern=north east lines, fill opacity=0.9}] (axis cs:2.36,0) rectangle (axis cs:2.52,4);
\draw[draw=black,fill=darkorange25512714,opacity=0.9,thick] (axis cs:0.36,9) rectangle (axis cs:0.52,31);
\draw[draw=black,fill=darkorange25512714,opacity=0.9,thick] (axis cs:1.36,3) rectangle (axis cs:1.52,16.5);
\draw[draw=black,fill=darkorange25512714,opacity=0.9,thick] (axis cs:2.36,4) rectangle (axis cs:2.52,12);
\draw[draw=black,fill=crimson2143940,opacity=0.9,thick,postaction={pattern=north east lines, fill opacity=0.9}] (axis cs:0.56,0) rectangle (axis cs:0.72,18.5);
\draw[draw=black,fill=crimson2143940,opacity=0.9,thick,postaction={pattern=north east lines, fill opacity=0.9}] (axis cs:1.56,0) rectangle (axis cs:1.72,15.5);
\draw[draw=black,fill=crimson2143940,opacity=0.9,thick,postaction={pattern=north east lines, fill opacity=0.9}] (axis cs:2.56,0) rectangle (axis cs:2.72,32);
\draw[draw=black,fill=crimson2143940,opacity=0.9,thick] (axis cs:0.56,18.5) rectangle (axis cs:0.72,32.5);
\draw[draw=black,fill=crimson2143940,opacity=0.9,thick] (axis cs:1.56,15.5) rectangle (axis cs:1.72,30);
\draw[draw=black,fill=crimson2143940,opacity=0.9,thick] (axis cs:2.56,32) rectangle (axis cs:2.72,47.5);
\end{axis}

\end{tikzpicture}
    \end{subfigure}
    \\[-0.9cm]
    \caption{mAP of the general detectors trained with the 4 datasets in real environments.}
    \label{fig:ap_gd}
\end{figure*}

\begin{figure*}[!h]
	\centering
    \begin{tabular}{@{}c@{ }c@{ }c@{ }c@{ }c@{ }}
    \rotatebox[origin=c]{90}{{\small DETR~\cite{detr}}}&
    \raisebox{-0.45\height}{\includegraphics[width=0.23\linewidth]{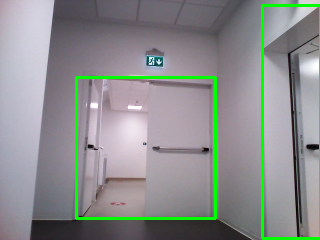}}&
    \raisebox{-0.45\height}{\includegraphics[width=0.23\linewidth]{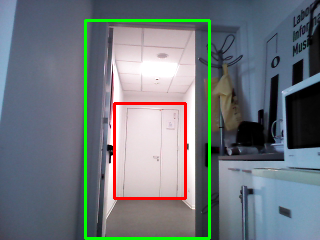}}&
    \raisebox{-0.45\height}{\includegraphics[width=0.23\linewidth]{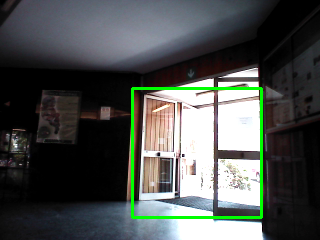}}&
    \raisebox{-0.45\height}{\includegraphics[width=0.23\linewidth]{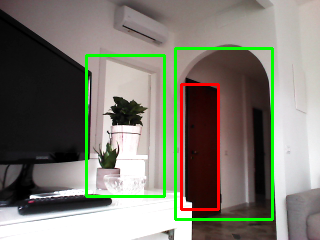}}\\\addlinespace[0.13cm]
    \rotatebox[origin=c]{90}{{\small YOLOv5~\cite{yolov5}}}&
    \raisebox{-0.45\height}{\includegraphics[width=0.23\linewidth]{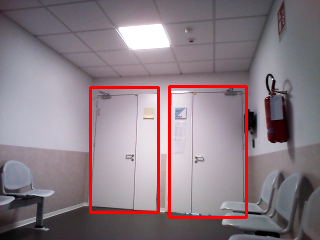}}&
    \raisebox{-0.45\height}{\includegraphics[width=0.23\linewidth]{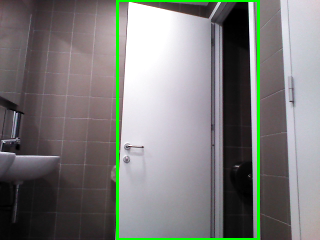}}&
    \raisebox{-0.45\height}{\includegraphics[width=0.23\linewidth]{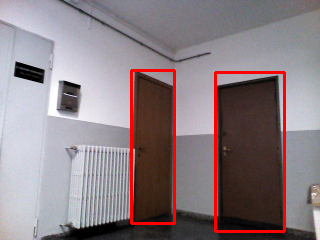}}&
    \raisebox{-0.45\height}{\includegraphics[width=0.23\linewidth]{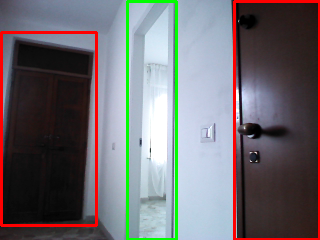}}\\\addlinespace[0.13cm]
    \rotatebox[origin=c]{90}{{\small Faster R--CNN~\cite{fasterrcnn}}}&
    \raisebox{-0.45\height}{\includegraphics[width=0.23\linewidth]{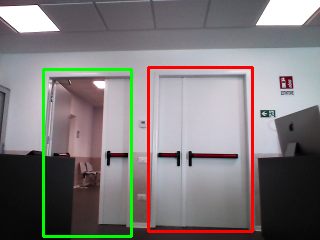}}&
    \raisebox{-0.45\height}{\includegraphics[width=0.23\linewidth]{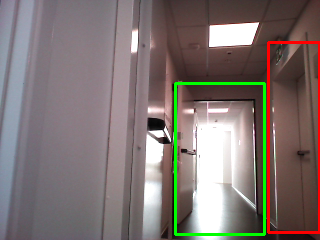}}&
    \raisebox{-0.45\height}{\includegraphics[width=0.23\linewidth]{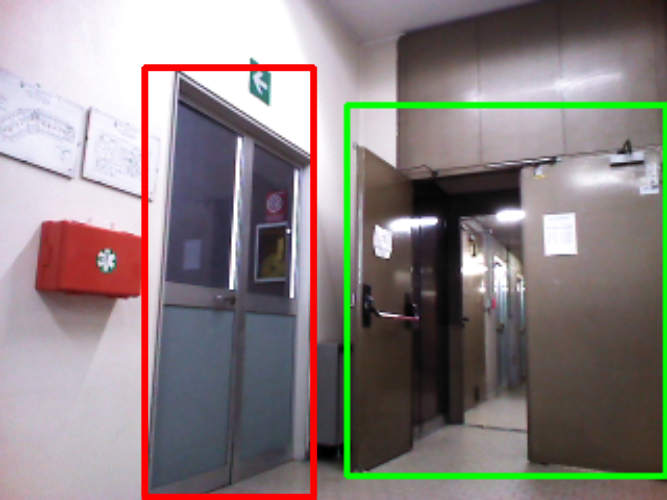}}&
    \raisebox{-0.45\height}{\includegraphics[width=0.23\linewidth]{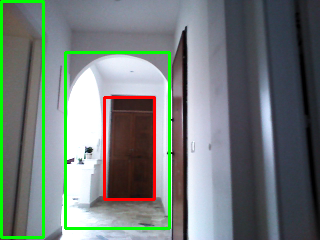}}\\\addlinespace[0.1cm]
    &$e_1$ -- \Classrooms & $e_2$ -- \Offices & $e_3$ -- \Laboratories & $e_4$ -- \House
\end{tabular}
	\caption{Real--world door instances correctly recognized by GDs trained on \DDDtwoG{}.}
	\label{fig:gd_examples}
\end{figure*}

\begin{figure*}[!htb]
    \centering
    \begin{subfigure}[t]{\textwidth}
        \centering
        \begin{tikzpicture}[scale=0.8]

\definecolor{crimson2143940}{RGB}{214,39,40}
\definecolor{darkgray176}{RGB}{176,176,176}
\definecolor{darkorange25512714}{RGB}{255,127,14}
\definecolor{forestgreen4416044}{RGB}{44,160,44}
\definecolor{lightgray204}{RGB}{204,204,204}
\definecolor{steelblue31119180}{RGB}{31,119,180}

\begin{axis}[
hide axis,
legend columns=7,
legend cell align={left},
legend style={
/tikz/every even column/.append style={column sep=0.3cm},
  fill opacity=0.8,
  draw opacity=1,
  text opacity=1,
  at={(0.5,0.95)},
  anchor=north,
  draw=lightgray204
},
xmin=-0.315, xmax=4.855,
y grid style={darkgray176},
ymin=0, ymax=130
]
\addlegendimage{ybar,area legend,draw=steelblue31119180,fill=steelblue31119180}
\addlegendentry{\Large\DiG}
\addlegendimage{ybar,area legend,draw=forestgreen4416044,fill=forestgreen4416044,opacity=0.9,thick}
\addlegendentry{\Large\DDDtwo}
\addlegendimage{ybar,area legend,draw=darkorange25512714,fill=darkorange25512714,opacity=0.9,thick}
\addlegendentry{\Large\DG}
\addlegendimage{ybar,area legend,draw=crimson2143940,fill=crimson2143940,opacity=0.9,thick}
\addlegendentry{\Large\DDDtwoG}
\addlegendimage{ybar,area legend,draw=black,fill=none,opacity=0.9,thick}
\addlegendentry{\Large $TP_{\%}$}
\addlegendimage{ybar,area legend,draw=black,fill=none,opacity=0.9,thick,postaction={pattern=north east lines, fill opacity=0.9}}
\addlegendentry{\Large $FP_{\%}$}
\addlegendimage{black,mark=*}
\addlegendentry{\Large $BFD_{\%}$}
\end{axis}
\end{tikzpicture}
    \end{subfigure}
    \\[-0.8cm]
    \begin{subfigure}[t]{0.48\textwidth}
        \centering
\begin{tikzpicture}[scale=0.63]

\definecolor{crimson2143940}{RGB}{214,39,40}
\definecolor{darkgray176}{RGB}{176,176,176}
\definecolor{darkorange25512714}{RGB}{255,127,14}
\definecolor{forestgreen4416044}{RGB}{44,160,44}
\definecolor{lightgray204}{RGB}{204,204,204}
\definecolor{steelblue31119180}{RGB}{31,119,180}

\begin{axis}[
width=12cm,
height=8cm,
legend columns=4,
legend cell align={left},
legend style={
/tikz/every even column/.append style={column sep=0.3cm},
  fill opacity=0.8,
  draw opacity=1,
  text opacity=1,
  at={(0.5,0.97)},
  anchor=north,
  draw=lightgray204
},
tick align=outside,
title={\LARGE \(\displaystyle e_1\) -- \Classrooms},
x grid style={darkgray176},
xmajorticks=false,
xmin=-0.178, xmax=2.858,
xtick style={color=black},
y grid style={darkgray176},
ylabel={\Large\%},
ymin=-50, ymax=60,
ytick pos=left,
ytick style={color=black},
ytick={-60,-40,-20,0,20,40,60},
yticklabels={60,40,20,0,20,40,60}
]
\draw[draw=black,fill=steelblue31119180,opacity=0.9,thick] (axis cs:-0.04,0) rectangle (axis cs:0.12,1);
\draw[draw=black,fill=steelblue31119180,opacity=0.9,thick] (axis cs:0.96,0) rectangle (axis cs:1.12,0);
\draw[draw=black,fill=steelblue31119180,opacity=0.9,thick] (axis cs:1.96,0) rectangle (axis cs:2.12,2);
\draw[draw=black,fill=steelblue31119180,thick,postaction={pattern=north east lines}] (axis cs:-0.04,0) rectangle (axis cs:0.12,0);
\draw[draw=black,fill=steelblue31119180,thick,postaction={pattern=north east lines}] (axis cs:0.96,0) rectangle (axis cs:1.12,0);
\draw[draw=black,fill=steelblue31119180,thick,postaction={pattern=north east lines}] (axis cs:1.96,0) rectangle (axis cs:2.12,0);
\draw[draw=black,fill=forestgreen4416044,opacity=0.9,thick] (axis cs:0.16,0) rectangle (axis cs:0.32,18);
\draw[draw=black,fill=forestgreen4416044,opacity=0.9,thick] (axis cs:1.16,0) rectangle (axis cs:1.32,3);
\draw[draw=black,fill=forestgreen4416044,opacity=0.9,thick] (axis cs:2.16,0) rectangle (axis cs:2.32,25);
\draw[draw=black,fill=forestgreen4416044,thick,postaction={pattern=north east lines}] (axis cs:0.16,0) rectangle (axis cs:0.32,-9);
\draw[draw=black,fill=forestgreen4416044,thick,postaction={pattern=north east lines}] (axis cs:1.16,0) rectangle (axis cs:1.32,-2);
\draw[draw=black,fill=forestgreen4416044,thick,postaction={pattern=north east lines}] (axis cs:2.16,0) rectangle (axis cs:2.32,-14);
\draw[draw=black,fill=darkorange25512714,opacity=0.9,thick] (axis cs:0.36,0) rectangle (axis cs:0.52,30);
\draw[draw=black,fill=darkorange25512714,opacity=0.9,thick] (axis cs:1.36,0) rectangle (axis cs:1.52,31);
\draw[draw=black,fill=darkorange25512714,opacity=0.9,thick] (axis cs:2.36,0) rectangle (axis cs:2.52,25);
\draw[draw=black,fill=darkorange25512714,thick,postaction={pattern=north east lines}] (axis cs:0.36,0) rectangle (axis cs:0.52,-7);
\draw[draw=black,fill=darkorange25512714,thick,postaction={pattern=north east lines}] (axis cs:1.36,0) rectangle (axis cs:1.52,-1);
\draw[draw=black,fill=darkorange25512714,thick,postaction={pattern=north east lines}] (axis cs:2.36,0) rectangle (axis cs:2.52,-6);
\draw[draw=black,fill=crimson2143940,opacity=0.9,thick] (axis cs:0.56,0) rectangle (axis cs:0.72,37);
\draw[draw=black,fill=crimson2143940,opacity=0.9,thick] (axis cs:1.56,0) rectangle (axis cs:1.72,34);
\draw[draw=black,fill=crimson2143940,opacity=0.9,thick] (axis cs:2.56,0) rectangle (axis cs:2.72,43);
\draw[draw=black,fill=crimson2143940,thick,postaction={pattern=north east lines}] (axis cs:0.56,0) rectangle (axis cs:0.72,-6);
\draw[draw=black,fill=crimson2143940,thick,postaction={pattern=north east lines}] (axis cs:1.56,0) rectangle (axis cs:1.72,-2);
\draw[draw=black,fill=crimson2143940,thick,postaction={pattern=north east lines}] (axis cs:2.56,0) rectangle (axis cs:2.72,-10);
\path [draw=black, ultra thick]
(axis cs:0.04,0)
--(axis cs:0.04,-26);

\path [draw=black, ultra thick]
(axis cs:1.04,0)
--(axis cs:1.04,0);

\path [draw=black, ultra thick]
(axis cs:2.04,0)
--(axis cs:2.04,-2);

\addplot [semithick, black, mark=*, mark size=2, mark options={solid}, only marks, forget plot]
table {%
0.04 -26
1.04 -0
2.04 -2
};
\path [draw=black, ultra thick]
(axis cs:0.24,0)
--(axis cs:0.24,-13);

\path [draw=black, ultra thick]
(axis cs:1.24,0)
--(axis cs:1.24,-1);

\path [draw=black, ultra thick]
(axis cs:2.24,0)
--(axis cs:2.24,-9);

\addplot [semithick, black, mark=*, mark size=2, mark options={solid}, only marks, forget plot]
table {%
0.24 -13
1.24 -1
2.24 -9
};
\path [draw=black, ultra thick]
(axis cs:0.44,0)
--(axis cs:0.44,-22);

\path [draw=black, ultra thick]
(axis cs:1.44,0)
--(axis cs:1.44,-8);

\path [draw=black, ultra thick]
(axis cs:2.44,0)
--(axis cs:2.44,-11);

\addplot [semithick, black, mark=*, mark size=2, mark options={solid}, only marks, forget plot]
table {%
0.44 -22
1.44 -8
2.44 -11
};
\path [draw=black, ultra thick]
(axis cs:0.64,0)
--(axis cs:0.64,-17);

\path [draw=black, ultra thick]
(axis cs:1.64,0)
--(axis cs:1.64,-3);

\path [draw=black, ultra thick]
(axis cs:2.64,0)
--(axis cs:2.64,-14);

\addplot [semithick, black, mark=*, mark size=2, mark options={solid}, only marks, forget plot]
table {%
0.64 -17
1.64 -3
2.64 -14
};
\addplot [black, forget plot]
table {%
-0.178 7.105427357601e-15
2.858 7.105427357601e-15
};
\end{axis}

\end{tikzpicture}
    \end{subfigure}
    \hfill
    \begin{subfigure}[t]{0.48\textwidth}
        \centering
\begin{tikzpicture}[scale=0.63]

\definecolor{crimson2143940}{RGB}{214,39,40}
\definecolor{darkgray176}{RGB}{176,176,176}
\definecolor{darkorange25512714}{RGB}{255,127,14}
\definecolor{forestgreen4416044}{RGB}{44,160,44}
\definecolor{lightgray204}{RGB}{204,204,204}
\definecolor{steelblue31119180}{RGB}{31,119,180}

\begin{axis}[
width=12cm,
height=8cm,
legend columns=4,
legend cell align={left},
legend style={
/tikz/every even column/.append style={column sep=0.3cm},
  fill opacity=0.8,
  draw opacity=1,
  text opacity=1,
  at={(0.5,0.97)},
  anchor=north,
  draw=lightgray204
},
tick align=outside,
title={\LARGE\(\displaystyle e_2\) -- \Offices},
x grid style={darkgray176},
xmajorticks=false,
xmin=-0.178, xmax=2.858,
xtick style={color=black},
y grid style={darkgray176},
ymajorticks=false,
ymin=-50, ymax=60,
ytick style={color=black},
ytick={-60,-40,-20,0,20,40,60},
yticklabels={60,40,20,0,20,40,60}
]
\draw[draw=black,fill=steelblue31119180,opacity=0.9,thick] (axis cs:-0.04,0) rectangle (axis cs:0.12,1);
\draw[draw=black,fill=steelblue31119180,opacity=0.9,thick] (axis cs:0.96,0) rectangle (axis cs:1.12,0);
\draw[draw=black,fill=steelblue31119180,opacity=0.9,thick] (axis cs:1.96,0) rectangle (axis cs:2.12,1);
\draw[draw=black,fill=steelblue31119180,thick,postaction={pattern=north east lines}] (axis cs:-0.04,0) rectangle (axis cs:0.12,-1);
\draw[draw=black,fill=steelblue31119180,thick,postaction={pattern=north east lines}] (axis cs:0.96,0) rectangle (axis cs:1.12,0);
\draw[draw=black,fill=steelblue31119180,thick,postaction={pattern=north east lines}] (axis cs:1.96,0) rectangle (axis cs:2.12,0);
\draw[draw=black,fill=forestgreen4416044,opacity=0.9,thick] (axis cs:0.16,0) rectangle (axis cs:0.32,19);
\draw[draw=black,fill=forestgreen4416044,opacity=0.9,thick] (axis cs:1.16,0) rectangle (axis cs:1.32,5);
\draw[draw=black,fill=forestgreen4416044,opacity=0.9,thick] (axis cs:2.16,0) rectangle (axis cs:2.32,27);
\draw[draw=black,fill=forestgreen4416044,thick,postaction={pattern=north east lines}] (axis cs:0.16,0) rectangle (axis cs:0.32,-8);
\draw[draw=black,fill=forestgreen4416044,thick,postaction={pattern=north east lines}] (axis cs:1.16,0) rectangle (axis cs:1.32,-1);
\draw[draw=black,fill=forestgreen4416044,thick,postaction={pattern=north east lines}] (axis cs:2.16,0) rectangle (axis cs:2.32,-4);
\draw[draw=black,fill=darkorange25512714,opacity=0.9,thick] (axis cs:0.36,0) rectangle (axis cs:0.52,36);
\draw[draw=black,fill=darkorange25512714,opacity=0.9,thick] (axis cs:1.36,0) rectangle (axis cs:1.52,21);
\draw[draw=black,fill=darkorange25512714,opacity=0.9,thick] (axis cs:2.36,0) rectangle (axis cs:2.52,17);
\draw[draw=black,fill=darkorange25512714,thick,postaction={pattern=north east lines}] (axis cs:0.36,0) rectangle (axis cs:0.52,-6);
\draw[draw=black,fill=darkorange25512714,thick,postaction={pattern=north east lines}] (axis cs:1.36,0) rectangle (axis cs:1.52,-9);
\draw[draw=black,fill=darkorange25512714,thick,postaction={pattern=north east lines}] (axis cs:2.36,0) rectangle (axis cs:2.52,-4);
\draw[draw=black,fill=crimson2143940,opacity=0.9,thick] (axis cs:0.56,0) rectangle (axis cs:0.72,31);
\draw[draw=black,fill=crimson2143940,opacity=0.9,thick] (axis cs:1.56,0) rectangle (axis cs:1.72,24);
\draw[draw=black,fill=crimson2143940,opacity=0.9,thick] (axis cs:2.56,0) rectangle (axis cs:2.72,34);
\draw[draw=black,fill=crimson2143940,thick,postaction={pattern=north east lines}] (axis cs:0.56,0) rectangle (axis cs:0.72,-10);
\draw[draw=black,fill=crimson2143940,thick,postaction={pattern=north east lines}] (axis cs:1.56,0) rectangle (axis cs:1.72,-10);
\draw[draw=black,fill=crimson2143940,thick,postaction={pattern=north east lines}] (axis cs:2.56,0) rectangle (axis cs:2.72,-5);
\path [draw=black, ultra thick]
(axis cs:0.04,0)
--(axis cs:0.04,-22);

\path [draw=black, ultra thick]
(axis cs:1.04,0)
--(axis cs:1.04,0);

\path [draw=black, ultra thick]
(axis cs:2.04,0)
--(axis cs:2.04,-3);

\addplot [semithick, black, mark=*, mark size=2, mark options={solid}, only marks, forget plot]
table {%
0.04 -22
1.04 -0
2.04 -3
};
\path [draw=black, ultra thick]
(axis cs:0.24,0)
--(axis cs:0.24,-17);

\path [draw=black, ultra thick]
(axis cs:1.24,0)
--(axis cs:1.24,-3);

\path [draw=black, ultra thick]
(axis cs:2.24,0)
--(axis cs:2.24,-18);

\addplot [semithick, black, mark=*, mark size=2, mark options={solid}, only marks, forget plot]
table {%
0.24 -17
1.24 -3
2.24 -18
};
\path [draw=black, ultra thick]
(axis cs:0.44,0)
--(axis cs:0.44,-21);

\path [draw=black, ultra thick]
(axis cs:1.44,0)
--(axis cs:1.44,-9);

\path [draw=black, ultra thick]
(axis cs:2.44,0)
--(axis cs:2.44,-10);

\addplot [semithick, black, mark=*, mark size=2, mark options={solid}, only marks, forget plot]
table {%
0.44 -21
1.44 -9
2.44 -10
};
\path [draw=black, ultra thick]
(axis cs:0.64,0)
--(axis cs:0.64,-19);

\path [draw=black, ultra thick]
(axis cs:1.64,0)
--(axis cs:1.64,-9);

\path [draw=black, ultra thick]
(axis cs:2.64,0)
--(axis cs:2.64,-20);

\addplot [semithick, black, mark=*, mark size=2, mark options={solid}, only marks, forget plot]
table {%
0.64 -19
1.64 -9
2.64 -20
};
\addplot [black, forget plot]
table {%
-0.178 7.105427357601e-15
2.858 7.105427357601e-15
};
\end{axis}

\end{tikzpicture}
    \end{subfigure}
    \\[-0.8cm]
    \begin{subfigure}[t]{0.48\textwidth}
        \centering
\begin{tikzpicture}[scale=0.63]

\definecolor{crimson2143940}{RGB}{214,39,40}
\definecolor{darkgray176}{RGB}{176,176,176}
\definecolor{darkorange25512714}{RGB}{255,127,14}
\definecolor{forestgreen4416044}{RGB}{44,160,44}
\definecolor{lightgray204}{RGB}{204,204,204}
\definecolor{steelblue31119180}{RGB}{31,119,180}

\begin{axis}[
width=12cm,
height=8cm,
legend columns=4,
legend cell align={left},
legend style={
/tikz/every even column/.append style={column sep=0.3cm},
  fill opacity=0.8,
  draw opacity=1,
  text opacity=1,
  at={(0.5,0.97)},
  anchor=north,
  draw=lightgray204
},
tick align=outside,
tick pos=left,
title={\LARGE\(\displaystyle e_3\) -- \Laboratories},
x grid style={darkgray176},
xlabel={\Large Model},
xmin=-0.178, xmax=2.858,
xtick style={color=black},
xtick={0.34,1.34,2.34},
xticklabels={\Large DETR~\cite{detr},\Large YOLOv5~\cite{yolov5},\Large Faster~R--CNN~\cite{fasterrcnn}},
y grid style={darkgray176},
ylabel={\Large\%},
ymin=-50, ymax=60,
ytick style={color=black},
ytick={-60,-40,-20,0,20,40,60},
yticklabels={60,40,20,0,20,40,60}
]
\draw[draw=black,fill=steelblue31119180,opacity=0.9,thick] (axis cs:-0.04,0) rectangle (axis cs:0.12,2);
\draw[draw=black,fill=steelblue31119180,opacity=0.9,thick] (axis cs:0.96,0) rectangle (axis cs:1.12,0);
\draw[draw=black,fill=steelblue31119180,opacity=0.9,thick] (axis cs:1.96,0) rectangle (axis cs:2.12,1);
\draw[draw=black,fill=steelblue31119180,thick,postaction={pattern=north east lines}] (axis cs:-0.04,0) rectangle (axis cs:0.12,0);
\draw[draw=black,fill=steelblue31119180,thick,postaction={pattern=north east lines}] (axis cs:0.96,0) rectangle (axis cs:1.12,0);
\draw[draw=black,fill=steelblue31119180,thick,postaction={pattern=north east lines}] (axis cs:1.96,0) rectangle (axis cs:2.12,-1);
\draw[draw=black,fill=forestgreen4416044,opacity=0.9,thick] (axis cs:0.16,0) rectangle (axis cs:0.32,15);
\draw[draw=black,fill=forestgreen4416044,opacity=0.9,thick] (axis cs:1.16,0) rectangle (axis cs:1.32,3);
\draw[draw=black,fill=forestgreen4416044,opacity=0.9,thick] (axis cs:2.16,0) rectangle (axis cs:2.32,20);
\draw[draw=black,fill=forestgreen4416044,thick,postaction={pattern=north east lines}] (axis cs:0.16,0) rectangle (axis cs:0.32,-3);
\draw[draw=black,fill=forestgreen4416044,thick,postaction={pattern=north east lines}] (axis cs:1.16,0) rectangle (axis cs:1.32,0);
\draw[draw=black,fill=forestgreen4416044,thick,postaction={pattern=north east lines}] (axis cs:2.16,0) rectangle (axis cs:2.32,-8);
\draw[draw=black,fill=darkorange25512714,opacity=0.9,thick] (axis cs:0.36,0) rectangle (axis cs:0.52,19);
\draw[draw=black,fill=darkorange25512714,opacity=0.9,thick] (axis cs:1.36,0) rectangle (axis cs:1.52,6);
\draw[draw=black,fill=darkorange25512714,opacity=0.9,thick] (axis cs:2.36,0) rectangle (axis cs:2.52,4);
\draw[draw=black,fill=darkorange25512714,thick,postaction={pattern=north east lines}] (axis cs:0.36,0) rectangle (axis cs:0.52,-6);
\draw[draw=black,fill=darkorange25512714,thick,postaction={pattern=north east lines}] (axis cs:1.36,0) rectangle (axis cs:1.52,-3);
\draw[draw=black,fill=darkorange25512714,thick,postaction={pattern=north east lines}] (axis cs:2.36,0) rectangle (axis cs:2.52,-2);
\draw[draw=black,fill=crimson2143940,opacity=0.9,thick] (axis cs:0.56,0) rectangle (axis cs:0.72,24);
\draw[draw=black,fill=crimson2143940,opacity=0.9,thick] (axis cs:1.56,0) rectangle (axis cs:1.72,10);
\draw[draw=black,fill=crimson2143940,opacity=0.9,thick] (axis cs:2.56,0) rectangle (axis cs:2.72,24);
\draw[draw=black,fill=crimson2143940,thick,postaction={pattern=north east lines}] (axis cs:0.56,0) rectangle (axis cs:0.72,-4);
\draw[draw=black,fill=crimson2143940,thick,postaction={pattern=north east lines}] (axis cs:1.56,0) rectangle (axis cs:1.72,-2);
\draw[draw=black,fill=crimson2143940,thick,postaction={pattern=north east lines}] (axis cs:2.56,0) rectangle (axis cs:2.72,-8);
\path [draw=black, ultra thick]
(axis cs:0.04,0)
--(axis cs:0.04,-35);

\path [draw=black, ultra thick]
(axis cs:1.04,0)
--(axis cs:1.04,-1);

\path [draw=black, ultra thick]
(axis cs:2.04,0)
--(axis cs:2.04,-11);

\addplot [semithick, black, mark=*, mark size=2, mark options={solid}, only marks, forget plot]
table {%
0.04 -35
1.04 -1
2.04 -11
};
\path [draw=black, ultra thick]
(axis cs:0.24,0)
--(axis cs:0.24,-30);

\path [draw=black, ultra thick]
(axis cs:1.24,0)
--(axis cs:1.24,-1);

\path [draw=black, ultra thick]
(axis cs:2.24,0)
--(axis cs:2.24,-40);

\addplot [semithick, black, mark=*, mark size=2, mark options={solid}, only marks, forget plot]
table {%
0.24 -30
1.24 -1
2.24 -40
};
\path [draw=black, ultra thick]
(axis cs:0.44,0)
--(axis cs:0.44,-33);

\path [draw=black, ultra thick]
(axis cs:1.44,0)
--(axis cs:1.44,-10);

\path [draw=black, ultra thick]
(axis cs:2.44,0)
--(axis cs:2.44,-10);

\addplot [semithick, black, mark=*, mark size=2, mark options={solid}, only marks, forget plot]
table {%
0.44 -33
1.44 -10
2.44 -10
};
\path [draw=black, ultra thick]
(axis cs:0.64,0)
--(axis cs:0.64,-44);

\path [draw=black, ultra thick]
(axis cs:1.64,0)
--(axis cs:1.64,-12);

\path [draw=black, ultra thick]
(axis cs:2.64,0)
--(axis cs:2.64,-34);

\addplot [semithick, black, mark=*, mark size=2, mark options={solid}, only marks, forget plot]
table {%
0.64 -44
1.64 -12
2.64 -34
};
\addplot [black, forget plot]
table {%
-0.178 0
2.858 0
};
\end{axis}

\end{tikzpicture}
    \end{subfigure}
    \hfill
    \begin{subfigure}[t]{0.48\textwidth}
        \centering
\begin{tikzpicture}[scale=0.63]

\definecolor{crimson2143940}{RGB}{214,39,40}
\definecolor{darkgray176}{RGB}{176,176,176}
\definecolor{darkorange25512714}{RGB}{255,127,14}
\definecolor{forestgreen4416044}{RGB}{44,160,44}
\definecolor{lightgray204}{RGB}{204,204,204}
\definecolor{steelblue31119180}{RGB}{31,119,180}

\begin{axis}[
width=12cm,
height=8cm,
legend columns=4,
legend cell align={left},
legend style={
/tikz/every even column/.append style={column sep=0.3cm},
  fill opacity=0.8,
  draw opacity=1,
  text opacity=1,
  at={(0.5,0.97)},
  anchor=north,
  draw=lightgray204
},
tick align=outside,
title={\LARGE\(\displaystyle e_4\) -- \House},
x grid style={darkgray176},
xlabel={\Large Model},
xmin=-0.178, xmax=2.858,
xtick pos=left,
xtick style={color=black},
xtick={0.34,1.34,2.34},
xticklabels={\Large DETR~\cite{detr},\Large YOLOv5~\cite{yolov5},\Large Faster~R--CNN~\cite{fasterrcnn}},
y grid style={darkgray176},
ymajorticks=false,
ymin=-50, ymax=60,
ytick style={color=black},
ytick={-60,-40,-20,0,20,40,60},
yticklabels={60,40,20,0,20,40,60}
]
\draw[draw=black,fill=steelblue31119180,opacity=0.9,thick] (axis cs:-0.04,0) rectangle (axis cs:0.12,5);
\draw[draw=black,fill=steelblue31119180,opacity=0.9,thick] (axis cs:0.96,0) rectangle (axis cs:1.12,0);
\draw[draw=black,fill=steelblue31119180,opacity=0.9,thick] (axis cs:1.96,0) rectangle (axis cs:2.12,3);
\draw[draw=black,fill=steelblue31119180,thick,postaction={pattern=north east lines}] (axis cs:-0.04,0) rectangle (axis cs:0.12,-3);
\draw[draw=black,fill=steelblue31119180,thick,postaction={pattern=north east lines}] (axis cs:0.96,0) rectangle (axis cs:1.12,-1);
\draw[draw=black,fill=steelblue31119180,thick,postaction={pattern=north east lines}] (axis cs:1.96,0) rectangle (axis cs:2.12,-7);
\draw[draw=black,fill=forestgreen4416044,opacity=0.9,thick] (axis cs:0.16,0) rectangle (axis cs:0.32,20);
\draw[draw=black,fill=forestgreen4416044,opacity=0.9,thick] (axis cs:1.16,0) rectangle (axis cs:1.32,12);
\draw[draw=black,fill=forestgreen4416044,opacity=0.9,thick] (axis cs:2.16,0) rectangle (axis cs:2.32,35);
\draw[draw=black,fill=forestgreen4416044,thick,postaction={pattern=north east lines}] (axis cs:0.16,0) rectangle (axis cs:0.32,-14);
\draw[draw=black,fill=forestgreen4416044,thick,postaction={pattern=north east lines}] (axis cs:1.16,0) rectangle (axis cs:1.32,-3);
\draw[draw=black,fill=forestgreen4416044,thick,postaction={pattern=north east lines}] (axis cs:2.16,0) rectangle (axis cs:2.32,-9);
\draw[draw=black,fill=darkorange25512714,opacity=0.9,thick] (axis cs:0.36,0) rectangle (axis cs:0.52,40);
\draw[draw=black,fill=darkorange25512714,opacity=0.9,thick] (axis cs:1.36,0) rectangle (axis cs:1.52,22);
\draw[draw=black,fill=darkorange25512714,opacity=0.9,thick] (axis cs:2.36,0) rectangle (axis cs:2.52,18);
\draw[draw=black,fill=darkorange25512714,thick,postaction={pattern=north east lines}] (axis cs:0.36,0) rectangle (axis cs:0.52,-9);
\draw[draw=black,fill=darkorange25512714,thick,postaction={pattern=north east lines}] (axis cs:1.36,0) rectangle (axis cs:1.52,-2);
\draw[draw=black,fill=darkorange25512714,thick,postaction={pattern=north east lines}] (axis cs:2.36,0) rectangle (axis cs:2.52,-4);
\draw[draw=black,fill=crimson2143940,opacity=0.9,thick] (axis cs:0.56,0) rectangle (axis cs:0.72,35);
\draw[draw=black,fill=crimson2143940,opacity=0.9,thick] (axis cs:1.56,0) rectangle (axis cs:1.72,34);
\draw[draw=black,fill=crimson2143940,opacity=0.9,thick] (axis cs:2.56,0) rectangle (axis cs:2.72,49);
\draw[draw=black,fill=crimson2143940,thick,postaction={pattern=north east lines}] (axis cs:0.56,0) rectangle (axis cs:0.72,-10);
\draw[draw=black,fill=crimson2143940,thick,postaction={pattern=north east lines}] (axis cs:1.56,0) rectangle (axis cs:1.72,-7);
\draw[draw=black,fill=crimson2143940,thick,postaction={pattern=north east lines}] (axis cs:2.56,0) rectangle (axis cs:2.72,-7);
\path [draw=black, ultra thick]
(axis cs:0.04,0)
--(axis cs:0.04,-25);

\path [draw=black, ultra thick]
(axis cs:1.04,0)
--(axis cs:1.04,-4);

\path [draw=black, ultra thick]
(axis cs:2.04,0)
--(axis cs:2.04,-7);

\addplot [semithick, black, mark=*, mark size=2, mark options={solid}, only marks, forget plot]
table {%
0.04 -25
1.04 -4
2.04 -7
};
\path [draw=black, ultra thick]
(axis cs:0.24,0)
--(axis cs:0.24,-9);

\path [draw=black, ultra thick]
(axis cs:1.24,0)
--(axis cs:1.24,-1);

\path [draw=black, ultra thick]
(axis cs:2.24,0)
--(axis cs:2.24,-14);

\addplot [semithick, black, mark=*, mark size=2, mark options={solid}, only marks, forget plot]
table {%
0.24 -9
1.24 -1
2.24 -14
};
\path [draw=black, ultra thick]
(axis cs:0.44,0)
--(axis cs:0.44,-11);

\path [draw=black, ultra thick]
(axis cs:1.44,0)
--(axis cs:1.44,-4);

\path [draw=black, ultra thick]
(axis cs:2.44,0)
--(axis cs:2.44,-6);

\addplot [semithick, black, mark=*, mark size=2, mark options={solid}, only marks, forget plot]
table {%
0.44 -11
1.44 -4
2.44 -6
};
\path [draw=black, ultra thick]
(axis cs:0.64,0)
--(axis cs:0.64,-13);

\path [draw=black, ultra thick]
(axis cs:1.64,0)
--(axis cs:1.64,-7);

\path [draw=black, ultra thick]
(axis cs:2.64,0)
--(axis cs:2.64,-16);

\addplot [semithick, black, mark=*, mark size=2, mark options={solid}, only marks, forget plot]
table {%
0.64 -13
1.64 -7
2.64 -16
};
\addplot [black, forget plot]
table {%
-0.178 1.4210854715202e-14
2.858 1.4210854715202e-14
};
\end{axis}

\end{tikzpicture}
    \end{subfigure}
    \\[-0.9cm]
    \caption{Operational performance indicators of the GDs trained with the 4 datasets.}
    \label{fig:ex_metric_gd}
\end{figure*}

\revI{In this section, we evaluate our pipeline for synthesizing a GD using the training parameters of Section~\ref{sec:ExperimentalSetting:Training}. The performance metrics are detailed in Table~\ref{tab:gd_results} and visually summarized in Figures~\ref{fig:ap_gd} and~\ref{fig:ex_metric_gd}. The mAP bars in Figure~\ref{fig:ap_gd} are composed of a dashed and an undashed part, stating the AP contributions of the two labels (\texttt{open} door and \texttt{closed} door) to the final mAP value. Ideally, we want the dashed and undashed parts of the same size, i.e. a model that is equally able to detect both categories.  The same representation is used for similar plots reporting mAP (Figures~\ref{fig:ap_qualification},~\ref{fig:ap_diff_conf},~and~\ref{fig:ap_comparison_detectors}).}
First, notice how the general detectors trained on \DiG{} exhibit very poor performance, as indicated by the blue bars in the figures. To elaborate, the YOLOv5--based GD correctly identified only one door instance (in $e_4$ -- \House). Meanwhile, its counterparts, DETR and Faster R--CNN, incur a high number of errors in terms of $FP_\%$ and $BFD_\%$ (as illustrated in Figure~\ref{fig:ex_metric_gd}), which outweighs their very limited number of correct detections. These unsurprising outcomes confirm the intuition that training with simulations, even those designed to replicate real environmental features, is ineffective if they lack photorealism. This conclusion is further supported by observing the significant performance improvements achieved when transitioning from training with \DiG{} to \DDDtwo{} (among our training datasets, the one that maximizes photorealism).


An interesting and perhaps counter--intuitive observation emerges when comparing the training results of \DDDtwo{} (real--world images) with \DG{} (our simulation framework outlined in Section~\ref{sec:gd}). Common intuition suggests that a detector trained on real--world data should outperform one trained on a simulation, even if photorealistic. However, as shown by the mAP scores in Figure~\ref{fig:ap_gd}  and the $TP_{\%}$ in Figure~\ref{fig:ex_metric_gd}, we see that two out of the three detectors, specifically those based on DETR and YOLOv5, actually have better performance when trained on \DG{} rather than \DDDtwo. This result indicates that while photorealism, a characteristic highly present in \DDDtwo, is important, it is not the unique key feature for creating effective general detectors for robots. It appears that the slightly compromised visual quality in \DG{} might be effectively balanced by a closer alignment with a robot's perception model, thereby reducing, to some extent, the sim--to--real gap. This also suggests that in real robot deployments, the shift in data distribution might be more significantly influenced by the data acquisition process rather than by the characteristics of the input space.

This trend does not hold for the detector based on Faster R--CNN, which shows better results with \DDDtwo. Upon closer examination, this can be attributed to the Region Proposal Network, which, by localizing and classifying bounding boxes based on features extracted from the Pyramid Backbone, is more sensitive to the photorealistic quality of images. To support this observation, we consider the performance of Faster R--CNN trained on \DDDtwoG, a dataset that combines \DDDtwo's high photorealism with \DG's representation of the robot's viewpoint. As indicated by the red bars in Figure~\ref{fig:ap_gd}, Faster R--CNN's performance improves, while DETR and YOLOv5 are only slightly impacted by the absence of real--world data. The $TP_{\%}$ in Figure~\ref{fig:ex_metric_gd} shows that the correct door status detections with \DDDtwoG{} slightly surpass those with \DG. In some cases, our simulated data even yield better results, as Ad by DETR's performance in environments $e_2$ and $e_3$. However, it's noteworthy that mixing training data often leads to an increase in erroneous detections, as evidenced by the $FP_{\%}$ and $BFD_{\%}$ indicators in Figure~\ref{fig:ex_metric_gd}.


These results prove the effectiveness of our simulation framework, which strikes a balance between photorealism and alignment with the robot's perception model. This approach is hence both viable and efficient for building general door detectors, reducing training costs while still achieving an acceptable performance level. The general detectors we developed are capable of accurately recognizing doors across diverse real--world environments, demonstrating a fair level of generalization. However, this strength is mainly evident in straightforward door instances, and less so in more complex ones involving occluded views, multiple nested doors, or combinations of these. Figure~\ref{fig:gd_examples} showcases some representative examples where our GDs excel. To bridge the gap in identifying such difficult cases, it is essential to qualify the general detectors for the target environment where they are set to operate.


\subsection{Evaluation of Qualified Detectors}\label{sec:Results:QD}
\revI{In this section, we assess how the process of qualifying a model to the robot's target environments enhances performances when compared with those of a general detector. The detailed results can be found in Table~\ref{tab:qualifications_results}.
Data are collected by using all three methods (DETR, YOLOv5, Faster R--CNN) and averaged. The same setting is also used in Section~\ref{sec:Results:LongTerm} and for the remainder of this work. 
}

\begin{table*}[!h]
\setlength\tabcolsep{3pt}
\setlength\extrarowheight{2pt}
\centering

\begin{tabular}{cc|cccc|cccc|cccc}
\toprule
         & & \multicolumn{4}{c|}{\DDDtwo} & \multicolumn{4}{c|}{\DG} &\multicolumn{4}{c}{\DDDtwoG} \\
        \textbf{Env.} & \textbf{Exp.} & \textbf{mAP}$\uparrow$& $\mathbf{TP}_{\%}$$\uparrow$ &  $\mathbf{FP}_{\%}$$\downarrow$ & $\mathbf{BFD}_{\%}$$\downarrow$ & \textbf{mAP}$\uparrow$& $\mathbf{TP}_{\%}$$\uparrow$ &  $\mathbf{FP}_{\%}$$\downarrow$ & $\mathbf{BFD}_{\%}$$\downarrow$ & \textbf{mAP}$\uparrow$& $\mathbf{TP}_{\%}$$\uparrow$ &  $\mathbf{FP}_{\%}$$\downarrow$ & $\mathbf{BFD}_{\%}$$\downarrow$ \\
\midrule
\multicolumn{1}{c}{\multirow{5}{*}[-1.5ex]{$e_1$}} &$GD$&$11 \pm 8 $&$15 \pm 11 $&$8 \pm 6 $&$\textbf{8} \pm 6 $&$25 \pm 5 $&$29 \pm 3 $&$\textbf{5} \pm 3 $&$14 \pm 7 $&$\textbf{33} \pm 1 $&$\textbf{38} \pm 5 $&$6 \pm 4 $&$11 \pm 7 $\\[2pt]
\multicolumn{1}{c}{} &$QD_{e}^{15}$&$49 \pm 21 $&$53 \pm 19 $&$\textbf{3} \pm 2 $&$\textbf{10} \pm 9 $&$63 \pm 12 $&$67 \pm 9 $&$3 \pm 1 $&$15 \pm 15 $&$\textbf{63} \pm 7 $&$\textbf{67} \pm 7 $&$3 \pm 2 $&$14 \pm 10 $\\[2pt]
\multicolumn{1}{c}{} &$QD_{e}^{25}$&$59 \pm 20 $&$63 \pm 18 $&$\textbf{2} \pm 1 $&$16 \pm 15 $&$72 \pm 14 $&$75 \pm 12 $&$3 \pm 2 $&$\textbf{14} \pm 13 $&$\textbf{73} \pm 12 $&$\textbf{76} \pm 9 $&$2 \pm 2 $&$15 \pm 12 $\\[2pt]
\multicolumn{1}{c}{} &$QD_{e}^{50}$&$72 \pm 19 $&$76 \pm 16 $&$\textbf{1} \pm 1 $&$13 \pm 13 $&$\textbf{81} \pm 9 $&$83 \pm 7 $&$1 \pm 1 $&$\textbf{11} \pm 9 $&$80 \pm 10 $&$\textbf{83} \pm 9 $&$1 \pm 1 $&$11 \pm 8 $\\[2pt]
\multicolumn{1}{c}{} &$QD_{e}^{75}$&$78 \pm 15 $&$81 \pm 13 $&$\textbf{1} \pm 1 $&$11 \pm 9 $&$85 \pm 9 $&$87 \pm 7 $&$1 \pm 1 $&$\textbf{9} \pm 7 $&$\textbf{85} \pm 8 $&$\textbf{87} \pm 7 $&$1 \pm 1 $&$9 \pm 6 $\\[2pt]
\hline
\multicolumn{1}{c}{\multirow{5}{*}[-1.5ex]{$e_2$}} &$GD$&$13 \pm 9 $&$17 \pm 11 $&$\textbf{4} \pm 4 $&$\textbf{13} \pm 8 $&$18 \pm 8 $&$25 \pm 10 $&$6 \pm 3 $&$13 \pm 7 $&$\textbf{22} \pm 5 $&$\textbf{30} \pm 5 $&$8 \pm 3 $&$16 \pm 6 $\\[2pt]
\multicolumn{1}{c}{} &$QD_{e}^{15}$&$48 \pm 21 $&$53 \pm 19 $&$3 \pm 1 $&$\textbf{16} \pm 16 $&$\textbf{65} \pm 12 $&$\textbf{69} \pm 9 $&$\textbf{2} \pm 1 $&$24 \pm 25 $&$62 \pm 11 $&$66 \pm 10 $&$3 \pm 1 $&$24 \pm 19 $\\[2pt]
\multicolumn{1}{c}{} &$QD_{e}^{25}$&$60 \pm 18 $&$66 \pm 17 $&$\textbf{3} \pm 2 $&$23 \pm 24 $&$71 \pm 8 $&$74 \pm 7 $&$3 \pm 1 $&$\textbf{17} \pm 16 $&$\textbf{72} \pm 10 $&$\textbf{76} \pm 8 $&$3 \pm 0 $&$20 \pm 19 $\\[2pt]
\multicolumn{1}{c}{} &$QD_{e}^{50}$&$70 \pm 17 $&$74 \pm 14 $&$\textbf{2} \pm 0 $&$18 \pm 19 $&$80 \pm 8 $&$83 \pm 6 $&$2 \pm 1 $&$\textbf{14} \pm 11 $&$\textbf{80} \pm 7 $&$\textbf{84} \pm 7 $&$2 \pm 1 $&$16 \pm 17 $\\[2pt]
\multicolumn{1}{c}{} &$QD_{e}^{75}$&$75 \pm 13 $&$80 \pm 9 $&$\textbf{2} \pm 0 $&$18 \pm 18 $&$\textbf{84} \pm 5 $&$\textbf{86} \pm 4 $&$2 \pm 1 $&$\textbf{12} \pm 11 $&$82 \pm 7 $&$85 \pm 6 $&$2 \pm 1 $&$18 \pm 15 $\\[2pt]
\hline
\multicolumn{1}{c}{\multirow{5}{*}[-1.5ex]{$e_3$}} &$GD$&$7 \pm 4 $&$13 \pm 9 $&$\textbf{4} \pm 4 $&$24 \pm 20 $&$6 \pm 6 $&$10 \pm 8 $&$4 \pm 2 $&$\textbf{18} \pm 13 $&$\textbf{12} \pm 5 $&$\textbf{19} \pm 8 $&$5 \pm 3 $&$30 \pm 16 $\\[2pt]
\multicolumn{1}{c}{} &$QD_{e}^{15}$&$41 \pm 23 $&$48 \pm 20 $&$\textbf{2} \pm 1 $&$\textbf{26} \pm 20 $&$55 \pm 16 $&$63 \pm 11 $&$5 \pm 3 $&$27 \pm 20 $&$\textbf{56} \pm 14 $&$\textbf{63} \pm 10 $&$4 \pm 2 $&$33 \pm 24 $\\[2pt]
\multicolumn{1}{c}{} &$QD_{e}^{25}$&$54 \pm 25 $&$59 \pm 21 $&$\textbf{3} \pm 2 $&$\textbf{21} \pm 18 $&$64 \pm 19 $&$70 \pm 13 $&$3 \pm 3 $&$26 \pm 28 $&$\textbf{68} \pm 13 $&$\textbf{75} \pm 9 $&$3 \pm 2 $&$21 \pm 15 $\\[2pt]
\multicolumn{1}{c}{} &$QD_{e}^{50}$&$68 \pm 21 $&$74 \pm 16 $&$\textbf{2} \pm 1 $&$19 \pm 18 $&$76 \pm 13 $&$81 \pm 10 $&$2 \pm 2 $&$\textbf{17} \pm 15 $&$\textbf{76} \pm 15 $&$\textbf{81} \pm 12 $&$3 \pm 2 $&$19 \pm 17 $\\[2pt]
\multicolumn{1}{c}{} &$QD_{e}^{75}$&$76 \pm 19 $&$81 \pm 15 $&$\textbf{1} \pm 1 $&$14 \pm 14 $&$82 \pm 12 $&$86 \pm 9 $&$1 \pm 1 $&$\textbf{12} \pm 10 $&$\textbf{82} \pm 11 $&$\textbf{86} \pm 8 $&$1 \pm 1 $&$15 \pm 16 $\\[2pt]
\hline
\multicolumn{1}{c}{\multirow{5}{*}[-1.5ex]{$e_4$}} &$GD$&$22 \pm 8 $&$22 \pm 12 $&$9 \pm 6 $&$8 \pm 7 $&$20 \pm 10 $&$27 \pm 12 $&$\textbf{5} \pm 4 $&$\textbf{7} \pm 4 $&$\textbf{37} \pm 9 $&$\textbf{39} \pm 8 $&$8 \pm 2 $&$12 \pm 5 $\\[2pt]
\multicolumn{1}{c}{} &$QD_{e}^{15}$&$60 \pm 18 $&$64 \pm 19 $&$2 \pm 1 $&$18 \pm 12 $&$76 \pm 4 $&$77 \pm 4 $&$\textbf{1} \pm 1 $&$\textbf{14} \pm 7 $&$\textbf{76} \pm 6 $&$\textbf{78} \pm 8 $&$3 \pm 2 $&$18 \pm 16 $\\[2pt]
\multicolumn{1}{c}{} &$QD_{e}^{25}$&$70 \pm 18 $&$73 \pm 17 $&$\textbf{2} \pm 1 $&$14 \pm 11 $&$79 \pm 7 $&$81 \pm 5 $&$3 \pm 1 $&$14 \pm 9 $&$\textbf{82} \pm 8 $&$\textbf{83} \pm 8 $&$4 \pm 2 $&$\textbf{12} \pm 9 $\\[2pt]
\multicolumn{1}{c}{} &$QD_{e}^{50}$&$81 \pm 10 $&$84 \pm 10 $&$2 \pm 1 $&$13 \pm 11 $&$\textbf{91} \pm 4 $&$92 \pm 3 $&$\textbf{1} \pm 1 $&$\textbf{8} \pm 6 $&$90 \pm 5 $&$\textbf{92} \pm 3 $&$1 \pm 1 $&$8 \pm 7 $\\[2pt]
\multicolumn{1}{c}{} &$QD_{e}^{75}$&$90 \pm 8 $&$91 \pm 6 $&$1 \pm 1 $&$5 \pm 4 $&$96 \pm 2 $&$96 \pm 2 $&$\textbf{0} \pm 1 $&$4 \pm 3 $&$\textbf{96} \pm 2 $&$\textbf{96} \pm 2 $&$0 \pm 0 $&$\textbf{3} \pm 2 $\\ 
\bottomrule
\end{tabular}

\caption{Results of the qualification procedure \revTwo{(averaged over detectors, together with the standard deviations)} when the $GD$ is trained with the \DDDtwo, \DG, and \DDDtwoG. \revI{Bold entries indicate the best performance on each metric across the three datasets.}}
\label{tab:qualifications_results}
\end{table*}

\begin{figure*}[!htb]
    \centering
    \begin{subfigure}[t]{\textwidth}
        \centering
    \begin{tikzpicture}[scale=0.8]

\definecolor{crimson2143940}{RGB}{214,39,40}
\definecolor{darkgray176}{RGB}{176,176,176}
\definecolor{forestgreen4416044}{RGB}{44,160,44}
\definecolor{lightgray204}{RGB}{204,204,204}
\definecolor{steelblue31119180}{RGB}{31,119,180}
\definecolor{darkorange25512714}{RGB}{255,127,14}

\begin{axis}[
hide axis,
legend columns=5,
legend cell align={left},
legend style={
/tikz/every even column/.append style={column sep=0.3cm},
  fill opacity=0.8,
  draw opacity=1,
  text opacity=1,
  at={(0.5,0.95)},
  anchor=north,
  draw=lightgray204
},
xmin=-0.315, xmax=4.855,
y grid style={darkgray176},
ymin=0, ymax=130
]
\addlegendimage{ybar,area legend,draw=forestgreen4416044,fill=forestgreen4416044}
\addlegendentry{\Large \DDDtwo}
\addlegendimage{ybar,area legend,draw=darkorange25512714,fill=darkorange25512714,opacity=0.9,thick}
\addlegendentry{\Large \DG}
\addlegendimage{ybar,area legend,draw=crimson2143940,fill=crimson2143940,opacity=0.9,thick}
\addlegendentry{\Large \DDDtwoG}
\addlegendimage{ybar,area legend,draw=black,fill=none,opacity=0.9,thick,postaction={pattern=north east lines, fill opacity=0.9}}
\addlegendentry{\large Closed door}
\addlegendimage{ybar,area legend,draw=black,fill=none,opacity=0.9,thick}
\addlegendentry{\Large Open door}
\end{axis}
\end{tikzpicture}
    \end{subfigure}
    \\[-0.8cm]
    \begin{subfigure}[t]{0.48\textwidth}
        \centering
\begin{tikzpicture}[scale=0.63]

\definecolor{crimson2143940}{RGB}{214,39,40}
\definecolor{darkgray176}{RGB}{176,176,176}
\definecolor{darkorange25512714}{RGB}{255,127,14}
\definecolor{forestgreen4416044}{RGB}{44,160,44}
\definecolor{lightgray204}{RGB}{204,204,204}

\begin{axis}[
width=12cm,
height=7cm,
legend columns=3,
legend cell align={left},
legend style={
/tikz/every even column/.append style={column sep=0.3cm},
  fill opacity=0.8,
  draw opacity=1,
  text opacity=1,
  at={(0.5,0.95)},
  anchor=north,
  draw=lightgray204
},
tick align=outside,
title={\LARGE\(\displaystyle e_1\) -- \Classrooms},
x grid style={darkgray176},
xmajorticks=false,
xmin=-0.315, xmax=4.855,
xtick style={color=black},
y grid style={darkgray176},
ylabel={\Large mAP},
ymin=0, ymax=100,
ytick pos=left,
ytick style={color=black}
]
\draw[draw=black,fill=forestgreen4416044,opacity=0.9,thick,postaction={pattern=north east lines, fill opacity=0.9}] (axis cs:-0.08,0) rectangle (axis cs:0.12,3.33333333333333);
\draw[draw=black,fill=forestgreen4416044,opacity=0.9,thick,postaction={pattern=north east lines, fill opacity=0.9}] (axis cs:0.92,0) rectangle (axis cs:1.12,23.5);
\draw[draw=black,fill=forestgreen4416044,opacity=0.9,thick,postaction={pattern=north east lines, fill opacity=0.9}] (axis cs:1.92,0) rectangle (axis cs:2.12,30.3333333333333);
\draw[draw=black,fill=forestgreen4416044,opacity=0.9,thick,postaction={pattern=north east lines, fill opacity=0.9}] (axis cs:2.92,0) rectangle (axis cs:3.12,37.5);
\draw[draw=black,fill=forestgreen4416044,opacity=0.9,thick,postaction={pattern=north east lines, fill opacity=0.9}] (axis cs:3.92,0) rectangle (axis cs:4.12,40);
\draw[draw=black,fill=forestgreen4416044,opacity=0.9,thick] (axis cs:-0.08,3.33333333333333) rectangle (axis cs:0.12,11.3333333333333);
\draw[draw=black,fill=forestgreen4416044,opacity=0.9,thick] (axis cs:0.92,23.5) rectangle (axis cs:1.12,48.8333333333333);
\draw[draw=black,fill=forestgreen4416044,opacity=0.9,thick] (axis cs:1.92,30.3333333333333) rectangle (axis cs:2.12,59.1666666666667);
\draw[draw=black,fill=forestgreen4416044,opacity=0.9,thick] (axis cs:2.92,37.5) rectangle (axis cs:3.12,72.1666666666667);
\draw[draw=black,fill=forestgreen4416044,opacity=0.9,thick] (axis cs:3.92,40) rectangle (axis cs:4.12,77.8333333333333);
\draw[draw=black,fill=darkorange25512714,opacity=0.9,thick,postaction={pattern=north east lines, fill opacity=0.9}] (axis cs:0.17,0) rectangle (axis cs:0.37,12.8333333333333);
\draw[draw=black,fill=darkorange25512714,opacity=0.9,thick,postaction={pattern=north east lines, fill opacity=0.9}] (axis cs:1.17,0) rectangle (axis cs:1.37,31.5);
\draw[draw=black,fill=darkorange25512714,opacity=0.9,thick,postaction={pattern=north east lines, fill opacity=0.9}] (axis cs:2.17,0) rectangle (axis cs:2.37,36.6666666666667);
\draw[draw=black,fill=darkorange25512714,opacity=0.9,thick,postaction={pattern=north east lines, fill opacity=0.9}] (axis cs:3.17,0) rectangle (axis cs:3.37,41.1666666666667);
\draw[draw=black,fill=darkorange25512714,opacity=0.9,thick,postaction={pattern=north east lines, fill opacity=0.9}] (axis cs:4.17,0) rectangle (axis cs:4.37,42.1666666666667);
\draw[draw=black,fill=darkorange25512714,opacity=0.9,thick] (axis cs:0.17,12.8333333333333) rectangle (axis cs:0.37,25.1666666666667);
\draw[draw=black,fill=darkorange25512714,opacity=0.9,thick] (axis cs:1.17,31.5) rectangle (axis cs:1.37,62.8333333333333);
\draw[draw=black,fill=darkorange25512714,opacity=0.9,thick] (axis cs:2.17,36.6666666666667) rectangle (axis cs:2.37,72.1666666666667);
\draw[draw=black,fill=darkorange25512714,opacity=0.9,thick] (axis cs:3.17,41.1666666666667) rectangle (axis cs:3.37,80.6666666666667);
\draw[draw=black,fill=darkorange25512714,opacity=0.9,thick] (axis cs:4.17,42.1666666666667) rectangle (axis cs:4.37,84.8333333333333);
\draw[draw=black,fill=crimson2143940,opacity=0.9,thick,postaction={pattern=north east lines, fill opacity=0.9}] (axis cs:0.42,0) rectangle (axis cs:0.62,20.1666666666667);
\draw[draw=black,fill=crimson2143940,opacity=0.9,thick,postaction={pattern=north east lines, fill opacity=0.9}] (axis cs:1.42,0) rectangle (axis cs:1.62,30.6666666666667);
\draw[draw=black,fill=crimson2143940,opacity=0.9,thick,postaction={pattern=north east lines, fill opacity=0.9}] (axis cs:2.42,0) rectangle (axis cs:2.62,39.3333333333333);
\draw[draw=black,fill=crimson2143940,opacity=0.9,thick,postaction={pattern=north east lines, fill opacity=0.9}] (axis cs:3.42,0) rectangle (axis cs:3.62,40.3333333333333);
\draw[draw=black,fill=crimson2143940,opacity=0.9,thick,postaction={pattern=north east lines, fill opacity=0.9}] (axis cs:4.42,0) rectangle (axis cs:4.62,43.5);
\draw[draw=black,fill=crimson2143940,opacity=0.9,thick] (axis cs:0.42,20.1666666666667) rectangle (axis cs:0.62,33);
\draw[draw=black,fill=crimson2143940,opacity=0.9,thick] (axis cs:1.42,30.6666666666667) rectangle (axis cs:1.62,63.1666666666667);
\draw[draw=black,fill=crimson2143940,opacity=0.9,thick] (axis cs:2.42,39.3333333333333) rectangle (axis cs:2.62,72.8333333333333);
\draw[draw=black,fill=crimson2143940,opacity=0.9,thick] (axis cs:3.42,40.3333333333333) rectangle (axis cs:3.62,80.3333333333333);
\draw[draw=black,fill=crimson2143940,opacity=0.9,thick] (axis cs:4.42,43.5) rectangle (axis cs:4.62,84.8333333333333);
\path [draw=black, semithick]
(axis cs:0.02,3.2041677344952)
--(axis cs:0.02,19.4624989321715);

\path [draw=black, semithick]
(axis cs:1.02,27.6122746101373)
--(axis cs:1.02,70.0543920565293);

\path [draw=black, semithick]
(axis cs:2.02,38.9038106652701)
--(axis cs:2.02,79.4295226680632);

\path [draw=black, semithick]
(axis cs:3.02,53.1316112863077)
--(axis cs:3.02,91.2017220470256);

\path [draw=black, semithick]
(axis cs:4.02,63.3218436326398)
--(axis cs:4.02,92.3448230340268);

\addplot [semithick, black, mark=-, mark size=3, mark options={solid}, only marks, forget plot]
table {%
0.02 3.2041677344952
1.02 27.6122746101373
2.02 38.9038106652701
3.02 53.1316112863077
4.02 63.3218436326398
};
\addplot [semithick, black, mark=-, mark size=3, mark options={solid}, only marks, forget plot]
table {%
0.02 19.4624989321715
1.02 70.0543920565293
2.02 79.4295226680632
3.02 91.2017220470256
4.02 92.3448230340268
};
\path [draw=black, semithick]
(axis cs:0.27,20.6574169138438)
--(axis cs:0.27,29.6759164194896);

\path [draw=black, semithick]
(axis cs:1.27,50.9344538342656)
--(axis cs:1.27,74.732212832401);

\path [draw=black, semithick]
(axis cs:2.27,57.9862811523961)
--(axis cs:2.27,86.3470521809372);

\path [draw=black, semithick]
(axis cs:3.27,71.2150354141614)
--(axis cs:3.27,90.1182979191719);

\path [draw=black, semithick]
(axis cs:4.27,76.313748053912)
--(axis cs:4.27,93.3529186127547);

\addplot [semithick, black, mark=-, mark size=3, mark options={solid}, only marks, forget plot]
table {%
0.27 20.6574169138438
1.27 50.9344538342656
2.27 57.9862811523961
3.27 71.2150354141614
4.27 76.313748053912
};
\addplot [semithick, black, mark=-, mark size=3, mark options={solid}, only marks, forget plot]
table {%
0.27 29.6759164194896
1.27 74.732212832401
2.27 86.3470521809372
3.27 90.1182979191719
4.27 93.3529186127547
};
\path [draw=black, semithick]
(axis cs:0.52,31.6771243444677)
--(axis cs:0.52,34.3228756555323);

\path [draw=black, semithick]
(axis cs:1.52,55.9152300273976)
--(axis cs:1.52,70.4181033059357);

\path [draw=black, semithick]
(axis cs:2.52,61.1893076835276)
--(axis cs:2.52,84.4773589831391);

\path [draw=black, semithick]
(axis cs:3.52,70.2297036225149)
--(axis cs:3.52,90.4369630441518);

\path [draw=black, semithick]
(axis cs:4.52,77.3111439357355)
--(axis cs:4.52,92.3555227309312);

\addplot [semithick, black, mark=-, mark size=3, mark options={solid}, only marks, forget plot]
table {%
0.52 31.6771243444677
1.52 55.9152300273976
2.52 61.1893076835276
3.52 70.2297036225149
4.52 77.3111439357355
};
\addplot [semithick, black, mark=-, mark size=3, mark options={solid}, only marks, forget plot]
table {%
0.52 34.3228756555323
1.52 70.4181033059357
2.52 84.4773589831391
3.52 90.4369630441518
4.52 92.3555227309312
};
\end{axis}

\end{tikzpicture}
    \end{subfigure}
    \hfill
    \begin{subfigure}[t]{0.48\textwidth}
        \centering
\begin{tikzpicture}[scale=0.63]

\definecolor{crimson2143940}{RGB}{214,39,40}
\definecolor{darkgray176}{RGB}{176,176,176}
\definecolor{darkorange25512714}{RGB}{255,127,14}
\definecolor{forestgreen4416044}{RGB}{44,160,44}
\definecolor{lightgray204}{RGB}{204,204,204}

\begin{axis}[
width=12cm,
height=7cm,
legend columns=3,
legend cell align={left},
legend style={
/tikz/every even column/.append style={column sep=0.3cm},
  fill opacity=0.8,
  draw opacity=1,
  text opacity=1,
  at={(0.5,0.95)},
  anchor=north,
  draw=lightgray204
},
tick align=outside,
title={\LARGE \(\displaystyle e_2\) -- \Offices},
x grid style={darkgray176},
xmajorticks=false,
xmin=-0.315, xmax=4.855,
xtick style={color=black},
y grid style={darkgray176},
ymajorticks=false,
ymin=0, ymax=100,
ytick style={color=black}
]
\draw[draw=black,fill=forestgreen4416044,opacity=0.9,thick,postaction={pattern=north east lines, fill opacity=0.9}] (axis cs:-0.08,0) rectangle (axis cs:0.12,1.33333333333333);
\draw[draw=black,fill=forestgreen4416044,opacity=0.9,thick,postaction={pattern=north east lines, fill opacity=0.9}] (axis cs:0.92,0) rectangle (axis cs:1.12,22.6666666666667);
\draw[draw=black,fill=forestgreen4416044,opacity=0.9,thick,postaction={pattern=north east lines, fill opacity=0.9}] (axis cs:1.92,0) rectangle (axis cs:2.12,26.6666666666667);
\draw[draw=black,fill=forestgreen4416044,opacity=0.9,thick,postaction={pattern=north east lines, fill opacity=0.9}] (axis cs:2.92,0) rectangle (axis cs:3.12,33.8333333333333);
\draw[draw=black,fill=forestgreen4416044,opacity=0.9,thick,postaction={pattern=north east lines, fill opacity=0.9}] (axis cs:3.92,0) rectangle (axis cs:4.12,35.3333333333333);
\draw[draw=black,fill=forestgreen4416044,opacity=0.9,thick] (axis cs:-0.08,1.33333333333333) rectangle (axis cs:0.12,12.6666666666667);
\draw[draw=black,fill=forestgreen4416044,opacity=0.9,thick] (axis cs:0.92,22.6666666666667) rectangle (axis cs:1.12,48.3333333333333);
\draw[draw=black,fill=forestgreen4416044,opacity=0.9,thick] (axis cs:1.92,26.6666666666667) rectangle (axis cs:2.12,60.1666666666667);
\draw[draw=black,fill=forestgreen4416044,opacity=0.9,thick] (axis cs:2.92,33.8333333333333) rectangle (axis cs:3.12,69.6666666666667);
\draw[draw=black,fill=forestgreen4416044,opacity=0.9,thick] (axis cs:3.92,35.3333333333333) rectangle (axis cs:4.12,74.8333333333333);
\draw[draw=black,fill=darkorange25512714,opacity=0.9,thick,postaction={pattern=north east lines, fill opacity=0.9}] (axis cs:0.17,0) rectangle (axis cs:0.37,4);
\draw[draw=black,fill=darkorange25512714,opacity=0.9,thick,postaction={pattern=north east lines, fill opacity=0.9}] (axis cs:1.17,0) rectangle (axis cs:1.37,30.1666666666667);
\draw[draw=black,fill=darkorange25512714,opacity=0.9,thick,postaction={pattern=north east lines, fill opacity=0.9}] (axis cs:2.17,0) rectangle (axis cs:2.37,34.1666666666667);
\draw[draw=black,fill=darkorange25512714,opacity=0.9,thick,postaction={pattern=north east lines, fill opacity=0.9}] (axis cs:3.17,0) rectangle (axis cs:3.37,39.3333333333333);
\draw[draw=black,fill=darkorange25512714,opacity=0.9,thick,postaction={pattern=north east lines, fill opacity=0.9}] (axis cs:4.17,0) rectangle (axis cs:4.37,40.5);
\draw[draw=black,fill=darkorange25512714,opacity=0.9,thick] (axis cs:0.17,4) rectangle (axis cs:0.37,18.1666666666667);
\draw[draw=black,fill=darkorange25512714,opacity=0.9,thick] (axis cs:1.17,30.1666666666667) rectangle (axis cs:1.37,64.6666666666667);
\draw[draw=black,fill=darkorange25512714,opacity=0.9,thick] (axis cs:2.17,34.1666666666667) rectangle (axis cs:2.37,70.8333333333333);
\draw[draw=black,fill=darkorange25512714,opacity=0.9,thick] (axis cs:3.17,39.3333333333333) rectangle (axis cs:3.37,79.6666666666667);
\draw[draw=black,fill=darkorange25512714,opacity=0.9,thick] (axis cs:4.17,40.5) rectangle (axis cs:4.37,84.1666666666667);
\draw[draw=black,fill=crimson2143940,opacity=0.9,thick,postaction={pattern=north east lines, fill opacity=0.9}] (axis cs:0.42,0) rectangle (axis cs:0.62,5.66666666666667);
\draw[draw=black,fill=crimson2143940,opacity=0.9,thick,postaction={pattern=north east lines, fill opacity=0.9}] (axis cs:1.42,0) rectangle (axis cs:1.62,28.3333333333333);
\draw[draw=black,fill=crimson2143940,opacity=0.9,thick,postaction={pattern=north east lines, fill opacity=0.9}] (axis cs:2.42,0) rectangle (axis cs:2.62,34.3333333333333);
\draw[draw=black,fill=crimson2143940,opacity=0.9,thick,postaction={pattern=north east lines, fill opacity=0.9}] (axis cs:3.42,0) rectangle (axis cs:3.62,39);
\draw[draw=black,fill=crimson2143940,opacity=0.9,thick,postaction={pattern=north east lines, fill opacity=0.9}] (axis cs:4.42,0) rectangle (axis cs:4.62,39.1666666666667);
\draw[draw=black,fill=crimson2143940,opacity=0.9,thick] (axis cs:0.42,5.66666666666667) rectangle (axis cs:0.62,22.5);
\draw[draw=black,fill=crimson2143940,opacity=0.9,thick] (axis cs:1.42,28.3333333333333) rectangle (axis cs:1.62,62.5);
\draw[draw=black,fill=crimson2143940,opacity=0.9,thick] (axis cs:2.42,34.3333333333333) rectangle (axis cs:2.62,71.5);
\draw[draw=black,fill=crimson2143940,opacity=0.9,thick] (axis cs:3.42,39) rectangle (axis cs:3.62,80.5);
\draw[draw=black,fill=crimson2143940,opacity=0.9,thick] (axis cs:4.42,39.1666666666667) rectangle (axis cs:4.62,82);
\path [draw=black, semithick]
(axis cs:0.02,3.38855622685548)
--(axis cs:0.02,21.9447771064779);

\path [draw=black, semithick]
(axis cs:1.02,26.8779532776612)
--(axis cs:1.02,69.7887133890055);

\path [draw=black, semithick]
(axis cs:2.02,42.0812213214911)
--(axis cs:2.02,78.2521120118423);

\path [draw=black, semithick]
(axis cs:3.02,52.4523315550995)
--(axis cs:3.02,86.8810017782338);

\path [draw=black, semithick]
(axis cs:4.02,61.4885465859427)
--(axis cs:4.02,88.178120080724);

\addplot [semithick, black, mark=-, mark size=3, mark options={solid}, only marks, forget plot]
table {%
0.02 3.38855622685548
1.02 26.8779532776612
2.02 42.0812213214911
3.02 52.4523315550995
4.02 61.4885465859427
};
\addplot [semithick, black, mark=-, mark size=3, mark options={solid}, only marks, forget plot]
table {%
0.02 21.9447771064779
1.02 69.7887133890055
2.02 78.2521120118423
3.02 86.8810017782338
4.02 88.178120080724
};
\path [draw=black, semithick]
(axis cs:0.27,10.0837628980119)
--(axis cs:0.27,26.2495704353214);

\path [draw=black, semithick]
(axis cs:1.27,52.767787167599)
--(axis cs:1.27,76.5655461657344);

\path [draw=black, semithick]
(axis cs:2.27,63.0658798681793)
--(axis cs:2.27,78.6007867984874);

\path [draw=black, semithick]
(axis cs:3.27,71.7557038091295)
--(axis cs:3.27,87.5776295242039);

\path [draw=black, semithick]
(axis cs:4.27,79.0594821846518)
--(axis cs:4.27,89.2738511486815);

\addplot [semithick, black, mark=-, mark size=3, mark options={solid}, only marks, forget plot]
table {%
0.27 10.0837628980119
1.27 52.767787167599
2.27 63.0658798681793
3.27 71.7557038091295
4.27 79.0594821846518
};
\addplot [semithick, black, mark=-, mark size=3, mark options={solid}, only marks, forget plot]
table {%
0.27 26.2495704353214
1.27 76.5655461657344
2.27 78.6007867984874
3.27 87.5776295242039
4.27 89.2738511486815
};
\path [draw=black, semithick]
(axis cs:0.52,17.091673086804)
--(axis cs:0.52,27.908326913196);

\path [draw=black, semithick]
(axis cs:1.52,51.2416697508023)
--(axis cs:1.52,73.7583302491977);

\path [draw=black, semithick]
(axis cs:2.52,61.9606079858305)
--(axis cs:2.52,81.0393920141695);

\path [draw=black, semithick]
(axis cs:3.52,73.4112765606211)
--(axis cs:3.52,87.5887234393789);

\path [draw=black, semithick]
(axis cs:4.52,74.9112765606211)
--(axis cs:4.52,89.0887234393789);

\addplot [semithick, black, mark=-, mark size=3, mark options={solid}, only marks, forget plot]
table {%
0.52 17.091673086804
1.52 51.2416697508023
2.52 61.9606079858305
3.52 73.4112765606211
4.52 74.9112765606211
};
\addplot [semithick, black, mark=-, mark size=3, mark options={solid}, only marks, forget plot]
table {%
0.52 27.908326913196
1.52 73.7583302491977
2.52 81.0393920141695
3.52 87.5887234393789
4.52 89.0887234393789
};
\end{axis}

\end{tikzpicture}
    \end{subfigure}
    \\[-0.8cm]
    \begin{subfigure}[t]{0.48\textwidth}
        \centering
\begin{tikzpicture}[scale=0.63]

\definecolor{crimson2143940}{RGB}{214,39,40}
\definecolor{darkgray176}{RGB}{176,176,176}
\definecolor{darkorange25512714}{RGB}{255,127,14}
\definecolor{forestgreen4416044}{RGB}{44,160,44}
\definecolor{lightgray204}{RGB}{204,204,204}

\begin{axis}[
width=12cm,
height=7cm,
legend columns=3,
legend cell align={left},
legend style={
/tikz/every even column/.append style={column sep=0.3cm},
  fill opacity=0.8,
  draw opacity=1,
  text opacity=1,
  at={(0.5,0.95)},
  anchor=north,
  draw=lightgray204
},
tick align=outside,
tick pos=left,
title={\LARGE \(\displaystyle e_3\) -- \Laboratories},
x grid style={darkgray176},
xlabel={\Large Qualification rounds},
xmin=-0.315, xmax=4.855,
xtick style={color=black},
xtick={0.27,1.27,2.27,3.27,4.27},
xticklabels={
 \Large \(\displaystyle GD\),
  \Large\(\displaystyle QD_{e}^{15}\),
  \Large\(\displaystyle QD_{e}^{25}\),
  \Large\(\displaystyle QD_{e}^{50}\),
  \Large\(\displaystyle QD_{e}^{75}\)
},
y grid style={darkgray176},
ylabel={\Large mAP},
ymin=0, ymax=100,
ytick style={color=black}
]
\draw[draw=black,fill=forestgreen4416044,opacity=0.9,thick,postaction={pattern=north east lines, fill opacity=0.9}] (axis cs:-0.08,0) rectangle (axis cs:0.12,2.33333333333333);
\draw[draw=black,fill=forestgreen4416044,opacity=0.9,thick,postaction={pattern=north east lines, fill opacity=0.9}] (axis cs:0.92,0) rectangle (axis cs:1.12,19.6666666666667);
\draw[draw=black,fill=forestgreen4416044,opacity=0.9,thick,postaction={pattern=north east lines, fill opacity=0.9}] (axis cs:1.92,0) rectangle (axis cs:2.12,24.6666666666667);
\draw[draw=black,fill=forestgreen4416044,opacity=0.9,thick,postaction={pattern=north east lines, fill opacity=0.9}] (axis cs:2.92,0) rectangle (axis cs:3.12,36);
\draw[draw=black,fill=forestgreen4416044,opacity=0.9,thick,postaction={pattern=north east lines, fill opacity=0.9}] (axis cs:3.92,0) rectangle (axis cs:4.12,37.1666666666667);
\draw[draw=black,fill=forestgreen4416044,opacity=0.9,thick] (axis cs:-0.08,2.33333333333333) rectangle (axis cs:0.12,7.33333333333333);
\draw[draw=black,fill=forestgreen4416044,opacity=0.9,thick] (axis cs:0.92,19.6666666666667) rectangle (axis cs:1.12,41.1666666666667);
\draw[draw=black,fill=forestgreen4416044,opacity=0.9,thick] (axis cs:1.92,24.6666666666667) rectangle (axis cs:2.12,54.3333333333333);
\draw[draw=black,fill=forestgreen4416044,opacity=0.9,thick] (axis cs:2.92,36) rectangle (axis cs:3.12,68.3333333333333);
\draw[draw=black,fill=forestgreen4416044,opacity=0.9,thick] (axis cs:3.92,37.1666666666667) rectangle (axis cs:4.12,76.1666666666667);
\draw[draw=black,fill=darkorange25512714,opacity=0.9,thick,postaction={pattern=north east lines, fill opacity=0.9}] (axis cs:0.17,0) rectangle (axis cs:0.37,1.16666666666667);
\draw[draw=black,fill=darkorange25512714,opacity=0.9,thick,postaction={pattern=north east lines, fill opacity=0.9}] (axis cs:1.17,0) rectangle (axis cs:1.37,25.6666666666667);
\draw[draw=black,fill=darkorange25512714,opacity=0.9,thick,postaction={pattern=north east lines, fill opacity=0.9}] (axis cs:2.17,0) rectangle (axis cs:2.37,32.1666666666667);
\draw[draw=black,fill=darkorange25512714,opacity=0.9,thick,postaction={pattern=north east lines, fill opacity=0.9}] (axis cs:3.17,0) rectangle (axis cs:3.37,38.6666666666667);
\draw[draw=black,fill=darkorange25512714,opacity=0.9,thick,postaction={pattern=north east lines, fill opacity=0.9}] (axis cs:4.17,0) rectangle (axis cs:4.37,40);
\draw[draw=black,fill=darkorange25512714,opacity=0.9,thick] (axis cs:0.17,1.16666666666667) rectangle (axis cs:0.37,6.16666666666667);
\draw[draw=black,fill=darkorange25512714,opacity=0.9,thick] (axis cs:1.17,25.6666666666667) rectangle (axis cs:1.37,54.6666666666667);
\draw[draw=black,fill=darkorange25512714,opacity=0.9,thick] (axis cs:2.17,32.1666666666667) rectangle (axis cs:2.37,64.5);
\draw[draw=black,fill=darkorange25512714,opacity=0.9,thick] (axis cs:3.17,38.6666666666667) rectangle (axis cs:3.37,76.5);
\draw[draw=black,fill=darkorange25512714,opacity=0.9,thick] (axis cs:4.17,40) rectangle (axis cs:4.37,82.1666666666667);
\draw[draw=black,fill=crimson2143940,opacity=0.9,thick,postaction={pattern=north east lines, fill opacity=0.9}] (axis cs:0.42,0) rectangle (axis cs:0.62,5.83333333333333);
\draw[draw=black,fill=crimson2143940,opacity=0.9,thick,postaction={pattern=north east lines, fill opacity=0.9}] (axis cs:1.42,0) rectangle (axis cs:1.62,26.8333333333333);
\draw[draw=black,fill=crimson2143940,opacity=0.9,thick,postaction={pattern=north east lines, fill opacity=0.9}] (axis cs:2.42,0) rectangle (axis cs:2.62,33.1666666666667);
\draw[draw=black,fill=crimson2143940,opacity=0.9,thick,postaction={pattern=north east lines, fill opacity=0.9}] (axis cs:3.42,0) rectangle (axis cs:3.62,39);
\draw[draw=black,fill=crimson2143940,opacity=0.9,thick,postaction={pattern=north east lines, fill opacity=0.9}] (axis cs:4.42,0) rectangle (axis cs:4.62,40.8333333333333);
\draw[draw=black,fill=crimson2143940,opacity=0.9,thick] (axis cs:0.42,5.83333333333333) rectangle (axis cs:0.62,11.6666666666667);
\draw[draw=black,fill=crimson2143940,opacity=0.9,thick] (axis cs:1.42,26.8333333333333) rectangle (axis cs:1.62,55.5);
\draw[draw=black,fill=crimson2143940,opacity=0.9,thick] (axis cs:2.42,33.1666666666667) rectangle (axis cs:2.62,67.8333333333333);
\draw[draw=black,fill=crimson2143940,opacity=0.9,thick] (axis cs:3.42,39) rectangle (axis cs:3.62,76);
\draw[draw=black,fill=crimson2143940,opacity=0.9,thick] (axis cs:4.42,40.8333333333333) rectangle (axis cs:4.62,82.1666666666667);
\path [draw=black, semithick]
(axis cs:0.02,3.54739443613315)
--(axis cs:0.02,11.1192722305335);

\path [draw=black, semithick]
(axis cs:1.02,18.625962743911)
--(axis cs:1.02,63.7073705894224);

\path [draw=black, semithick]
(axis cs:2.02,29.5022372372418)
--(axis cs:2.02,79.1644294294249);

\path [draw=black, semithick]
(axis cs:3.02,47.5948827322558)
--(axis cs:3.02,89.0717839344109);

\path [draw=black, semithick]
(axis cs:4.02,57.3495645164063)
--(axis cs:4.02,94.983768816927);

\addplot [semithick, black, mark=-, mark size=3, mark options={solid}, only marks, forget plot]
table {%
0.02 3.54739443613315
1.02 18.625962743911
2.02 29.5022372372418
3.02 47.5948827322558
4.02 57.3495645164063
};
\addplot [semithick, black, mark=-, mark size=3, mark options={solid}, only marks, forget plot]
table {%
0.02 11.1192722305335
1.02 63.7073705894224
2.02 79.1644294294249
3.02 89.0717839344109
4.02 94.983768816927
};
\path [draw=black, semithick]
(axis cs:0.27,0.20148994394224)
--(axis cs:0.27,12.1318433893911);

\path [draw=black, semithick]
(axis cs:1.27,38.4931286080558)
--(axis cs:1.27,70.8402047252775);

\path [draw=black, semithick]
(axis cs:2.27,45.3884851463836)
--(axis cs:2.27,83.6115148536164);

\path [draw=black, semithick]
(axis cs:3.27,63.271243444677)
--(axis cs:3.27,89.7287565553229);

\path [draw=black, semithick]
(axis cs:4.27,70.0698351255839)
--(axis cs:4.27,94.2634982077494);

\addplot [semithick, black, mark=-, mark size=3, mark options={solid}, only marks, forget plot]
table {%
0.27 0.20148994394224
1.27 38.4931286080558
2.27 45.3884851463836
3.27 63.271243444677
4.27 70.0698351255839
};
\addplot [semithick, black, mark=-, mark size=3, mark options={solid}, only marks, forget plot]
table {%
0.27 12.1318433893911
1.27 70.8402047252775
2.27 83.6115148536164
3.27 89.7287565553229
4.27 94.2634982077494
};
\path [draw=black, semithick]
(axis cs:0.52,6.27377010421219)
--(axis cs:0.52,17.0595632291211);

\path [draw=black, semithick]
(axis cs:1.52,41.103819951112)
--(axis cs:1.52,69.896180048888);

\path [draw=black, semithick]
(axis cs:2.52,54.5166770963745)
--(axis cs:2.52,81.1499895702921);

\path [draw=black, semithick]
(axis cs:3.52,61.0919484841244)
--(axis cs:3.52,90.9080515158756);

\path [draw=black, semithick]
(axis cs:4.52,70.6956887949535)
--(axis cs:4.52,93.6376445383798);

\addplot [semithick, black, mark=-, mark size=3, mark options={solid}, only marks, forget plot]
table {%
0.52 6.27377010421219
1.52 41.103819951112
2.52 54.5166770963745
3.52 61.0919484841244
4.52 70.6956887949535
};
\addplot [semithick, black, mark=-, mark size=3, mark options={solid}, only marks, forget plot]
table {%
0.52 17.0595632291211
1.52 69.896180048888
2.52 81.1499895702921
3.52 90.9080515158756
4.52 93.6376445383798
};
\end{axis}

\end{tikzpicture}
    \end{subfigure}
    \hfill
    \begin{subfigure}[t]{0.48\textwidth}
        \centering
\begin{tikzpicture}[scale=0.63]

\definecolor{crimson2143940}{RGB}{214,39,40}
\definecolor{darkgray176}{RGB}{176,176,176}
\definecolor{darkorange25512714}{RGB}{255,127,14}
\definecolor{forestgreen4416044}{RGB}{44,160,44}
\definecolor{lightgray204}{RGB}{204,204,204}

\begin{axis}[
width=12cm,
height=7cm,
legend columns=3,
legend cell align={left},
legend style={
/tikz/every even column/.append style={column sep=0.3cm},
  fill opacity=0.8,
  draw opacity=1,
  text opacity=1,
  at={(0.5,0.95)},
  anchor=north,
  draw=lightgray204
},
tick align=outside,
title={\LARGE \(\displaystyle e_4\) -- \House},
x grid style={darkgray176},
xlabel={\Large Qualification rounds},
xmin=-0.315, xmax=4.855,
xtick pos=left,
xtick style={color=black},
xtick={0.27,1.27,2.27,3.27,4.27},
xticklabels={
  \Large \(\displaystyle GD\),
  \Large\(\displaystyle QD_{e}^{15}\),
  \Large\(\displaystyle QD_{e}^{25}\),
  \Large\(\displaystyle QD_{e}^{50}\),
  \Large\(\displaystyle QD_{e}^{75}\)
},
y grid style={darkgray176},
ymajorticks=false,
ymin=0, ymax=100,
ytick style={color=black}
]
\draw[draw=black,fill=forestgreen4416044,opacity=0.9,thick,postaction={pattern=north east lines, fill opacity=0.9}] (axis cs:-0.08,0) rectangle (axis cs:0.12,18.3333333333333);
\draw[draw=black,fill=forestgreen4416044,opacity=0.9,thick,postaction={pattern=north east lines, fill opacity=0.9}] (axis cs:0.92,0) rectangle (axis cs:1.12,32.8333333333333);
\draw[draw=black,fill=forestgreen4416044,opacity=0.9,thick,postaction={pattern=north east lines, fill opacity=0.9}] (axis cs:1.92,0) rectangle (axis cs:2.12,36.6666666666667);
\draw[draw=black,fill=forestgreen4416044,opacity=0.9,thick,postaction={pattern=north east lines, fill opacity=0.9}] (axis cs:2.92,0) rectangle (axis cs:3.12,39.8333333333333);
\draw[draw=black,fill=forestgreen4416044,opacity=0.9,thick,postaction={pattern=north east lines, fill opacity=0.9}] (axis cs:3.92,0) rectangle (axis cs:4.12,44.6666666666667);
\draw[draw=black,fill=forestgreen4416044,opacity=0.9,thick] (axis cs:-0.08,18.3333333333333) rectangle (axis cs:0.12,22.3333333333333);
\draw[draw=black,fill=forestgreen4416044,opacity=0.9,thick] (axis cs:0.92,32.8333333333333) rectangle (axis cs:1.12,59.8333333333333);
\draw[draw=black,fill=forestgreen4416044,opacity=0.9,thick] (axis cs:1.92,36.6666666666667) rectangle (axis cs:2.12,70.5);
\draw[draw=black,fill=forestgreen4416044,opacity=0.9,thick] (axis cs:2.92,39.8333333333333) rectangle (axis cs:3.12,81);
\draw[draw=black,fill=forestgreen4416044,opacity=0.9,thick] (axis cs:3.92,44.6666666666667) rectangle (axis cs:4.12,89.6666666666667);
\draw[draw=black,fill=darkorange25512714,opacity=0.9,thick,postaction={pattern=north east lines, fill opacity=0.9}] (axis cs:0.17,0) rectangle (axis cs:0.37,5.33333333333333);
\draw[draw=black,fill=darkorange25512714,opacity=0.9,thick,postaction={pattern=north east lines, fill opacity=0.9}] (axis cs:1.17,0) rectangle (axis cs:1.37,40.3333333333333);
\draw[draw=black,fill=darkorange25512714,opacity=0.9,thick,postaction={pattern=north east lines, fill opacity=0.9}] (axis cs:2.17,0) rectangle (axis cs:2.37,40.1666666666667);
\draw[draw=black,fill=darkorange25512714,opacity=0.9,thick,postaction={pattern=north east lines, fill opacity=0.9}] (axis cs:3.17,0) rectangle (axis cs:3.37,46.1666666666667);
\draw[draw=black,fill=darkorange25512714,opacity=0.9,thick,postaction={pattern=north east lines, fill opacity=0.9}] (axis cs:4.17,0) rectangle (axis cs:4.37,49);
\draw[draw=black,fill=darkorange25512714,opacity=0.9,thick] (axis cs:0.17,5.33333333333333) rectangle (axis cs:0.37,19.8333333333333);
\draw[draw=black,fill=darkorange25512714,opacity=0.9,thick] (axis cs:1.17,40.3333333333333) rectangle (axis cs:1.37,75.5);
\draw[draw=black,fill=darkorange25512714,opacity=0.9,thick] (axis cs:2.17,40.1666666666667) rectangle (axis cs:2.37,79.1666666666667);
\draw[draw=black,fill=darkorange25512714,opacity=0.9,thick] (axis cs:3.17,46.1666666666667) rectangle (axis cs:3.37,91.3333333333333);
\draw[draw=black,fill=darkorange25512714,opacity=0.9,thick] (axis cs:4.17,49) rectangle (axis cs:4.37,96.3333333333333);
\draw[draw=black,fill=crimson2143940,opacity=0.9,thick,postaction={pattern=north east lines, fill opacity=0.9}] (axis cs:0.42,0) rectangle (axis cs:0.62,22);
\draw[draw=black,fill=crimson2143940,opacity=0.9,thick,postaction={pattern=north east lines, fill opacity=0.9}] (axis cs:1.42,0) rectangle (axis cs:1.62,40.6666666666667);
\draw[draw=black,fill=crimson2143940,opacity=0.9,thick,postaction={pattern=north east lines, fill opacity=0.9}] (axis cs:2.42,0) rectangle (axis cs:2.62,43.1666666666667);
\draw[draw=black,fill=crimson2143940,opacity=0.9,thick,postaction={pattern=north east lines, fill opacity=0.9}] (axis cs:3.42,0) rectangle (axis cs:3.62,45);
\draw[draw=black,fill=crimson2143940,opacity=0.9,thick,postaction={pattern=north east lines, fill opacity=0.9}] (axis cs:4.42,0) rectangle (axis cs:4.62,48.1666666666667);
\draw[draw=black,fill=crimson2143940,opacity=0.9,thick] (axis cs:0.42,22) rectangle (axis cs:0.62,36.6666666666667);
\draw[draw=black,fill=crimson2143940,opacity=0.9,thick] (axis cs:1.42,40.6666666666667) rectangle (axis cs:1.62,76);
\draw[draw=black,fill=crimson2143940,opacity=0.9,thick] (axis cs:2.42,43.1666666666667) rectangle (axis cs:2.62,82.3333333333333);
\draw[draw=black,fill=crimson2143940,opacity=0.9,thick] (axis cs:3.42,45) rectangle (axis cs:3.62,90.5);
\draw[draw=black,fill=crimson2143940,opacity=0.9,thick] (axis cs:4.42,48.1666666666667) rectangle (axis cs:4.62,95.5);
\path [draw=black, semithick]
(axis cs:0.02,14.0518278341301)
--(axis cs:0.02,30.6148388325366);

\path [draw=black, semithick]
(axis cs:1.02,42.2241501230497)
--(axis cs:1.02,77.4425165436169);

\path [draw=black, semithick]
(axis cs:2.02,52.6604372250887)
--(axis cs:2.02,88.3395627749113);

\path [draw=black, semithick]
(axis cs:3.02,70.5956739766576)
--(axis cs:3.02,91.4043260233424);

\path [draw=black, semithick]
(axis cs:4.02,81.8191617553376)
--(axis cs:4.02,97.5141715779957);

\addplot [semithick, black, mark=-, mark size=3, mark options={solid}, only marks, forget plot]
table {%
0.02 14.0518278341301
1.02 42.2241501230497
2.02 52.6604372250887
3.02 70.5956739766576
4.02 81.8191617553376
};
\addplot [semithick, black, mark=-, mark size=3, mark options={solid}, only marks, forget plot]
table {%
0.02 30.6148388325366
1.02 77.4425165436169
2.02 88.3395627749113
3.02 91.4043260233424
4.02 97.5141715779957
};
\path [draw=black, semithick]
(axis cs:0.27,9.90441932759568)
--(axis cs:0.27,29.762247339071);

\path [draw=black, semithick]
(axis cs:1.27,71.0559027913422)
--(axis cs:1.27,79.9440972086578);

\path [draw=black, semithick]
(axis cs:2.27,71.8808356685352)
--(axis cs:2.27,86.4524976647981);

\path [draw=black, semithick]
(axis cs:3.27,87.2918814490059)
--(axis cs:3.27,95.3747852176607);

\path [draw=black, semithick]
(axis cs:4.27,94.0787084569219)
--(axis cs:4.27,98.5879582097448);

\addplot [semithick, black, mark=-, mark size=3, mark options={solid}, only marks, forget plot]
table {%
0.27 9.90441932759568
1.27 71.0559027913422
2.27 71.8808356685352
3.27 87.2918814490059
4.27 94.0787084569219
};
\addplot [semithick, black, mark=-, mark size=3, mark options={solid}, only marks, forget plot]
table {%
0.27 29.762247339071
1.27 79.9440972086578
2.27 86.4524976647981
3.27 95.3747852176607
4.27 98.5879582097448
};
\path [draw=black, semithick]
(axis cs:0.52,27.2018194236662)
--(axis cs:0.52,46.1315139096671);

\path [draw=black, semithick]
(axis cs:1.52,69.7550020016016)
--(axis cs:1.52,82.2449979983984);

\path [draw=black, semithick]
(axis cs:2.52,74.6793583311966)
--(axis cs:2.52,89.98730833547);

\path [draw=black, semithick]
(axis cs:3.52,85.9174243050442)
--(axis cs:3.52,95.0825756949558);

\path [draw=black, semithick]
(axis cs:4.52,93.3205505282297)
--(axis cs:4.52,97.6794494717703);

\addplot [semithick, black, mark=-, mark size=3, mark options={solid}, only marks, forget plot]
table {%
0.52 27.2018194236662
1.52 69.7550020016016
2.52 74.6793583311966
3.52 85.9174243050442
4.52 93.3205505282297
};
\addplot [semithick, black, mark=-, mark size=3, mark options={solid}, only marks, forget plot]
table {%
0.52 46.1315139096671
1.52 82.2449979983984
2.52 89.98730833547
3.52 95.0825756949558
4.52 97.6794494717703
};
\end{axis}

\end{tikzpicture}
    \end{subfigure}
    \\[-0.9cm]
    \caption{Real--world evaluation of the qualified detectors where GDs are based on different datasets. The mAP is averaged over the three models, \revTwo{for which we report also the standard deviation}.}
    \label{fig:ap_qualification}
\end{figure*}

\begin{figure*}[!htb]
    \centering
    \begin{subfigure}[t]{\textwidth}
        \centering
    \begin{tikzpicture}[scale=0.8]

\definecolor{crimson2143940}{RGB}{214,39,40}
\definecolor{darkgray176}{RGB}{176,176,176}
\definecolor{forestgreen4416044}{RGB}{44,160,44}
\definecolor{lightgray204}{RGB}{204,204,204}
\definecolor{steelblue31119180}{RGB}{31,119,180}
\definecolor{darkorange25512714}{RGB}{255,127,14}

\begin{axis}[
hide axis,
legend columns=3,
legend cell align={left},
legend style={
/tikz/every even column/.append style={column sep=0.3cm},
  fill opacity=0.8,
  draw opacity=1,
  text opacity=1,
  at={(0.5,0.95)},
  anchor=north,
  draw=lightgray204
},
xmin=-0.315, xmax=4.855,
y grid style={darkgray176},
ymin=0, ymax=130
]
\addlegendimage{forestgreen4416044,mark=*,mark size=3}
\addlegendentry{\Large$TP_{\%}$}
\addlegendimage{darkorange25512714,mark=diamond*,mark size=3}
\addlegendentry{\Large$BFD_{\%}$}
\addlegendimage{crimson2143940,mark=triangle*,mark size=3}
\addlegendentry{\Large$FP_{\%}$}
\end{axis}
\end{tikzpicture}
    \end{subfigure}
    \\[-0.8cm]
    \begin{subfigure}[t]{0.48\textwidth}
        \centering
\begin{tikzpicture}[scale=0.63]

\definecolor{crimson2143940}{RGB}{214,39,40}
\definecolor{darkgray176}{RGB}{176,176,176}
\definecolor{darkorange25512714}{RGB}{255,127,14}
\definecolor{forestgreen4416044}{RGB}{44,160,44}
\definecolor{gray}{RGB}{128,128,128}
\definecolor{lightgray204}{RGB}{204,204,204}

\begin{axis}[
width=12cm,
height=8cm,
legend columns=4,
legend cell align={left},
legend style={
/tikz/every even column/.append style={column sep=0.3cm},
  fill opacity=0.8,
  draw opacity=1,
  text opacity=1,
  at={(0.5,0.97)},
  anchor=north,
  draw=lightgray204
},
tick align=outside,
title={\LARGE \(\displaystyle e_1\) -- \Classrooms},
x grid style={darkgray176},
xmajorticks=false,
xmin=-0.7, xmax=14.7,
xtick style={color=black},
y grid style={darkgray176},
ylabel={\Large\%},
ymin=0, ymax=115,
ytick pos=left,
ytick style={color=black}
]
\path [draw=forestgreen4416044, fill=forestgreen4416044, opacity=0.2]
(axis cs:0,4.09352313327509)
--(axis cs:0,26.5731435333916)
--(axis cs:1,72.0659793692686)
--(axis cs:2,80.8142101184216)
--(axis cs:3,91.2337259051142)
--(axis cs:4,93.5299640861417)
--(axis cs:4,68.4700359138583)
--(axis cs:4,68.4700359138583)
--(axis cs:3,60.0996074282192)
--(axis cs:2,44.5191232149117)
--(axis cs:1,33.2673539640647)
--(axis cs:0,4.09352313327509)
--cycle;

\path [draw=crimson2143940, fill=crimson2143940, opacity=0.2]
(axis cs:0,2.30561955999163)
--(axis cs:0,14.361047106675)
--(axis cs:1,4.73205080756888)
--(axis cs:2,3.48803387171258)
--(axis cs:3,2)
--(axis cs:4,1.91068360252296)
--(axis cs:4,0.755983064143708)
--(axis cs:4,0.755983064143708)
--(axis cs:3,0)
--(axis cs:2,1.17863279495408)
--(axis cs:1,1.26794919243112)
--(axis cs:0,2.30561955999163)
--cycle;

\path [draw=darkorange25512714, fill=darkorange25512714, opacity=0.2]
(axis cs:0,1.55656574005888)
--(axis cs:0,13.7767675932745)
--(axis cs:1,18.955011437585)
--(axis cs:2,31.3444403322636)
--(axis cs:3,26.6499895702921)
--(axis cs:4,19.8382338814487)
--(axis cs:4,2.82843278521795)
--(axis cs:4,2.82843278521795)
--(axis cs:3,0.0166770963745488)
--(axis cs:2,1.32222633440306)
--(axis cs:1,1.71165522908163)
--(axis cs:0,1.55656574005888)
--cycle;

\path [draw=forestgreen4416044, fill=forestgreen4416044, opacity=0.2]
(axis cs:5,25.4521164130024)
--(axis cs:5,31.881216920331)
--(axis cs:6,76.4071050592108)
--(axis cs:7,87.2636867787822)
--(axis cs:8,90.3571025019018)
--(axis cs:9,93.557438524302)
--(axis cs:9,80.442561475698)
--(axis cs:9,80.442561475698)
--(axis cs:8,76.3095641647648)
--(axis cs:7,63.4029798878845)
--(axis cs:6,58.2595616074559)
--(axis cs:5,25.4521164130024)
--cycle;

\path [draw=crimson2143940, fill=crimson2143940, opacity=0.2]
(axis cs:5,1.45211641300235)
--(axis cs:5,7.88121692033098)
--(axis cs:6,3.91068360252296)
--(axis cs:7,4.19419189831861)
--(axis cs:8,2.48803387171258)
--(axis cs:9,2)
--(axis cs:9,0)
--(axis cs:9,0)
--(axis cs:8,0.178632794954082)
--(axis cs:7,1.13914143501472)
--(axis cs:6,2.75598306414371)
--(axis cs:5,1.45211641300235)
--cycle;

\path [draw=darkorange25512714, fill=darkorange25512714, opacity=0.2]
(axis cs:5,6.29555187083467)
--(axis cs:5,21.0377814624987)
--(axis cs:6,30.0996688705415)
--(axis cs:7,26.5299640861417)
--(axis cs:8,19.7404383925441)
--(axis cs:9,15.1730737653144)
--(axis cs:9,2.16025956801895)
--(axis cs:9,2.16025956801895)
--(axis cs:8,1.5928949407892)
--(axis cs:7,1.47003591385833)
--(axis cs:6,-0.0996688705414996)
--(axis cs:5,6.29555187083467)
--cycle;

\path [draw=forestgreen4416044, fill=forestgreen4416044, opacity=0.2]
(axis cs:10,33.4174243050442)
--(axis cs:10,42.5825756949558)
--(axis cs:11,74)
--(axis cs:12,85.6249065765109)
--(axis cs:13,91.1715672147821)
--(axis cs:14,93.6904358352352)
--(axis cs:14,79.6428974980982)
--(axis cs:14,79.6428974980982)
--(axis cs:13,74.1617661185513)
--(axis cs:12,67.0417600901558)
--(axis cs:11,60)
--(axis cs:10,33.4174243050442)
--cycle;

\path [draw=crimson2143940, fill=crimson2143940, opacity=0.2]
(axis cs:10,2)
--(axis cs:10,10)
--(axis cs:11,4.86085856498528)
--(axis cs:12,3.73205080756888)
--(axis cs:13,1.24401693585629)
--(axis cs:14,1.91068360252296)
--(axis cs:14,0.755983064143708)
--(axis cs:14,0.755983064143708)
--(axis cs:13,0.0893163974770408)
--(axis cs:12,0.267949192431123)
--(axis cs:11,1.80580810168139)
--(axis cs:10,2)
--cycle;

\path [draw=darkorange25512714, fill=darkorange25512714, opacity=0.2]
(axis cs:10,3.96221853750134)
--(axis cs:10,18.7044481291653)
--(axis cs:11,23.733112580361)
--(axis cs:12,26.6805475272934)
--(axis cs:13,19.6599973311979)
--(axis cs:14,14.1742372139528)
--(axis cs:14,3.15909611938056)
--(axis cs:14,3.15909611938056)
--(axis cs:13,3.0066693354688)
--(axis cs:12,2.65278580603993)
--(axis cs:11,3.60022075297233)
--(axis cs:10,3.96221853750134)
--cycle;

\addplot [semithick, forestgreen4416044, mark=*, mark size=2.5, mark options={solid}, forget plot]
table {%
0 15.3333333333333
1 52.6666666666667
2 62.6666666666667
3 75.6666666666667
4 81
};
\addplot [semithick, crimson2143940, mark=triangle*, mark size=2.5, mark options={solid}, forget plot]
table {%
0 8.33333333333333
1 3
2 2.33333333333333
3 1
4 1.33333333333333
};
\addplot [semithick, darkorange25512714, mark=diamond*, mark size=2.5, mark options={solid}, forget plot]
table {%
0 7.66666666666667
1 10.3333333333333
2 16.3333333333333
3 13.3333333333333
4 11.3333333333333
};
\addplot [semithick, forestgreen4416044, mark=*, mark size=2.5, mark options={solid}, forget plot]
table {%
5 28.6666666666667
6 67.3333333333333
7 75.3333333333333
8 83.3333333333333
9 87
};
\addplot [semithick, crimson2143940, mark=triangle*, mark size=2.5, mark options={solid}, forget plot]
table {%
5 4.66666666666667
6 3.33333333333333
7 2.66666666666667
8 1.33333333333333
9 1
};
\addplot [semithick, darkorange25512714, mark=diamond*, mark size=2.5, mark options={solid}, forget plot]
table {%
5 13.6666666666667
6 15
7 14
8 10.6666666666667
9 8.66666666666667
};
\addplot [semithick, forestgreen4416044, mark=*, mark size=2.5, mark options={solid}, forget plot]
table {%
10 38
11 67
12 76.3333333333333
13 82.6666666666667
14 86.6666666666667
};
\addplot [semithick, crimson2143940, mark=triangle*, mark size=2.5, mark options={solid}, forget plot]
table {%
10 6
11 3.33333333333333
12 2
13 0.666666666666667
14 1.33333333333333
};
\addplot [semithick, darkorange25512714, mark=diamond*, mark size=2.5, mark options={solid}, forget plot]
table {%
10 11.3333333333333
11 13.6666666666667
12 14.6666666666667
13 11.3333333333333
14 8.66666666666667
};
\addplot [black, forget plot]
table {%
-0.7 -1.77635683940025e-15
14.7 -1.77635683940025e-15
};
\path [draw=gray, draw opacity=0.7, semithick, dash pattern=on 5.55pt off 2.4pt]
(axis cs:4.5,0)
--(axis cs:4.5,115);

\path [draw=gray, draw opacity=0.7, semithick, dash pattern=on 5.55pt off 2.4pt]
(axis cs:9.5,0)
--(axis cs:9.5,115);

\draw (axis cs:2,107) node[
  text=black,
  rotate=0.0
]{\Large\DDDtwo};
\draw (axis cs:7,107) node[
  text=black,
  rotate=0.0
]{\Large\DG};
\draw (axis cs:12,107) node[
  text=black,
  rotate=0.0
]{\Large\DDDtwoG};
\end{axis}

\end{tikzpicture}
    \end{subfigure}
    \hfill
    \begin{subfigure}[t]{0.48\textwidth}
        \centering
\begin{tikzpicture}[scale=0.63]

\definecolor{crimson2143940}{RGB}{214,39,40}
\definecolor{darkgray176}{RGB}{176,176,176}
\definecolor{darkorange25512714}{RGB}{255,127,14}
\definecolor{forestgreen4416044}{RGB}{44,160,44}
\definecolor{gray}{RGB}{128,128,128}
\definecolor{lightgray204}{RGB}{204,204,204}

\begin{axis}[
width=12cm,
height=8cm,
legend columns=4,
legend cell align={left},
legend style={
/tikz/every even column/.append style={column sep=0.3cm},
  fill opacity=0.8,
  draw opacity=1,
  text opacity=1,
  at={(0.5,0.97)},
  anchor=north,
  draw=lightgray204
},
tick align=outside,
title={\LARGE\(\displaystyle e_2\) -- \Offices},
x grid style={darkgray176},
xmajorticks=false,
xmin=-0.7, xmax=14.7,
xtick style={color=black},
y grid style={darkgray176},
ymajorticks=false,
ymin=0, ymax=115,
ytick style={color=black}
]
\path [draw=forestgreen4416044, fill=forestgreen4416044, opacity=0.2]
(axis cs:0,5.86447127433996)
--(axis cs:0,28.13552872566)
--(axis cs:1,72.2873015219859)
--(axis cs:2,82.1691917259821)
--(axis cs:3,88)
--(axis cs:4,88.6295531064992)
--(axis cs:4,70.7037802268342)
--(axis cs:4,70.7037802268342)
--(axis cs:3,60)
--(axis cs:2,49.1641416073513)
--(axis cs:1,33.7126984780141)
--(axis cs:0,5.86447127433996)
--cycle;

\path [draw=crimson2143940, fill=crimson2143940, opacity=0.2]
(axis cs:0,0.821448749049086)
--(axis cs:0,7.84521791761758)
--(axis cs:1,3.82136720504592)
--(axis cs:2,4.19419189831861)
--(axis cs:3,2)
--(axis cs:4,2)
--(axis cs:4,2)
--(axis cs:4,2)
--(axis cs:3,2)
--(axis cs:2,1.13914143501472)
--(axis cs:1,1.51196612828742)
--(axis cs:0,0.821448749049086)
--cycle;

\path [draw=darkorange25512714, fill=darkorange25512714, opacity=0.2]
(axis cs:0,4.28016958306058)
--(axis cs:0,21.0531637502727)
--(axis cs:1,31.52417469626)
--(axis cs:2,46.920052760446)
--(axis cs:3,36.8355854485039)
--(axis cs:4,35.2260895880879)
--(axis cs:4,0.107243745245434)
--(axis cs:4,0.107243745245434)
--(axis cs:3,-0.168918781837224)
--(axis cs:2,-0.253386093779316)
--(axis cs:1,0.475825303739976)
--(axis cs:0,4.28016958306058)
--cycle;

\path [draw=forestgreen4416044, fill=forestgreen4416044, opacity=0.2]
(axis cs:5,14.6500138657889)
--(axis cs:5,34.6833194675445)
--(axis cs:6,78.6249065765109)
--(axis cs:7,80.9916614518127)
--(axis cs:8,88.1742372139528)
--(axis cs:9,90.4966653322656)
--(axis cs:9,82.1700013344011)
--(axis cs:9,82.1700013344011)
--(axis cs:8,77.1590961193806)
--(axis cs:7,67.6750052148539)
--(axis cs:6,60.0417600901558)
--(axis cs:5,14.6500138657889)
--cycle;

\path [draw=crimson2143940, fill=crimson2143940, opacity=0.2]
(axis cs:5,3.81672185490975)
--(axis cs:5,8.84994481175692)
--(axis cs:6,2.91068360252296)
--(axis cs:7,3.91068360252296)
--(axis cs:8,3)
--(axis cs:9,2.82136720504592)
--(axis cs:9,0.511966128287415)
--(axis cs:9,0.511966128287415)
--(axis cs:8,1)
--(axis cs:7,2.75598306414371)
--(axis cs:6,1.75598306414371)
--(axis cs:5,3.81672185490975)
--cycle;

\path [draw=darkorange25512714, fill=darkorange25512714, opacity=0.2]
(axis cs:5,6.67500521485394)
--(axis cs:5,19.9916614518127)
--(axis cs:6,48.6373699895098)
--(axis cs:7,32.6770799446971)
--(axis cs:8,25.13552872566)
--(axis cs:9,23.5731435333916)
--(axis cs:9,1.09352313327509)
--(axis cs:9,1.09352313327509)
--(axis cs:8,2.86447127433996)
--(axis cs:7,0.65625338863623)
--(axis cs:6,-0.637369989509839)
--(axis cs:5,6.67500521485394)
--cycle;

\path [draw=forestgreen4416044, fill=forestgreen4416044, opacity=0.2]
(axis cs:10,24.5350652272198)
--(axis cs:10,34.7982681061136)
--(axis cs:11,76.741663330664)
--(axis cs:12,83.8388868327985)
--(axis cs:13,90.557438524302)
--(axis cs:14,91.4434342599411)
--(axis cs:14,79.2232324067255)
--(axis cs:14,79.2232324067255)
--(axis cs:13,77.442561475698)
--(axis cs:12,68.8277798338682)
--(axis cs:11,55.9250033360027)
--(axis cs:10,24.5350652272198)
--cycle;

\path [draw=crimson2143940, fill=crimson2143940, opacity=0.2]
(axis cs:10,5.4465819873852)
--(axis cs:10,11.2200846792815)
--(axis cs:11,4)
--(axis cs:12,3)
--(axis cs:13,2.91068360252296)
--(axis cs:14,2.82136720504592)
--(axis cs:14,0.511966128287415)
--(axis cs:14,0.511966128287415)
--(axis cs:13,1.75598306414371)
--(axis cs:12,3)
--(axis cs:11,2)
--(axis cs:10,5.4465819873852)
--cycle;

\path [draw=darkorange25512714, fill=darkorange25512714, opacity=0.2]
(axis cs:10,9.91723746970178)
--(axis cs:10,22.0827625302982)
--(axis cs:11,42.5963611526676)
--(axis cs:12,38.8355854485039)
--(axis cs:13,32.8358583926487)
--(axis cs:14,32.9331845230681)
--(axis cs:14,3.06681547693192)
--(axis cs:14,3.06681547693192)
--(axis cs:13,-0.169191725982085)
--(axis cs:12,1.83108121816278)
--(axis cs:11,4.73697218066576)
--(axis cs:10,9.91723746970178)
--cycle;

\addplot [semithick, forestgreen4416044, mark=*, mark size=2.5, mark options={solid}, forget plot]
table {%
0 17
1 53
2 65.6666666666667
3 74
4 79.6666666666667
};
\addplot [semithick, crimson2143940, mark=triangle*, mark size=2.5, mark options={solid}, forget plot]
table {%
0 4.33333333333333
1 2.66666666666667
2 2.66666666666667
3 2
4 2
};
\addplot [semithick, darkorange25512714, mark=diamond*, mark size=2.5, mark options={solid}, forget plot]
table {%
0 12.6666666666667
1 16
2 23.3333333333333
3 18.3333333333333
4 17.6666666666667
};
\addplot [semithick, forestgreen4416044, mark=*, mark size=2.5, mark options={solid}, forget plot]
table {%
5 24.6666666666667
6 69.3333333333333
7 74.3333333333333
8 82.6666666666667
9 86.3333333333333
};
\addplot [semithick, crimson2143940, mark=triangle*, mark size=2.5, mark options={solid}, forget plot]
table {%
5 6.33333333333333
6 2.33333333333333
7 3.33333333333333
8 2
9 1.66666666666667
};
\addplot [semithick, darkorange25512714, mark=diamond*, mark size=2.5, mark options={solid}, forget plot]
table {%
5 13.3333333333333
6 24
7 16.6666666666667
8 14
9 12.3333333333333
};
\addplot [semithick, forestgreen4416044, mark=*, mark size=2.5, mark options={solid}, forget plot]
table {%
10 29.6666666666667
11 66.3333333333333
12 76.3333333333333
13 84
14 85.3333333333333
};
\addplot [semithick, crimson2143940, mark=triangle*, mark size=2.5, mark options={solid}, forget plot]
table {%
10 8.33333333333333
11 3
12 3
13 2.33333333333333
14 1.66666666666667
};
\addplot [semithick, darkorange25512714, mark=diamond*, mark size=2.5, mark options={solid}, forget plot]
table {%
10 16
11 23.6666666666667
12 20.3333333333333
13 16.3333333333333
14 18
};
\addplot [black, forget plot]
table {%
-0.7 -1.77635683940025e-15
14.7 -1.77635683940025e-15
};
\path [draw=gray, draw opacity=0.7, semithick, dash pattern=on 5.55pt off 2.4pt]
(axis cs:4.5,0)
--(axis cs:4.5,115);

\path [draw=gray, draw opacity=0.7, semithick, dash pattern=on 5.55pt off 2.4pt]
(axis cs:9.5,0)
--(axis cs:9.5,115);

\draw (axis cs:2,107) node[
  text=black,
  rotate=0.0
]{\Large\DDDtwo};
\draw (axis cs:7,107) node[
  text=black,
  rotate=0.0
]{\Large\DG};
\draw (axis cs:12,107) node[
  text=black,
  rotate=0.0
]{\Large\DDDtwoG};
\end{axis}

\end{tikzpicture}
    \end{subfigure}
    \\[-0.8cm]
    \begin{subfigure}[t]{0.48\textwidth}
        \centering
\begin{tikzpicture}[scale=0.63]

\definecolor{crimson2143940}{RGB}{214,39,40}
\definecolor{darkgray176}{RGB}{176,176,176}
\definecolor{darkorange25512714}{RGB}{255,127,14}
\definecolor{forestgreen4416044}{RGB}{44,160,44}
\definecolor{gray}{RGB}{128,128,128}
\definecolor{lightgray204}{RGB}{204,204,204}

\begin{axis}[
width=12cm,
height=8cm,
legend columns=4,
legend cell align={left},
legend style={
/tikz/every even column/.append style={column sep=0.3cm},
  fill opacity=0.8,
  draw opacity=1,
  text opacity=1,
  at={(0.5,0.97)},
  anchor=north,
  draw=lightgray204
},
tick align=outside,
tick pos=left,
title={\LARGE\(\displaystyle e_3\) -- \Laboratories},
x grid style={darkgray176},
xlabel={\Large\% of qualification data},
xmin=-0.7, xmax=14.7,
xtick style={color=black},
xtick={0,1,2,3,4,5,6,7,8,9,10,11,12,13,14},
xticklabels={0,15,25,50,75,0,15,25,50,75,0,15,25,50,75},
y grid style={darkgray176},
ylabel={\Large\%},
ymin=0, ymax=115,
ytick style={color=black}
]
\path [draw=forestgreen4416044, fill=forestgreen4416044, opacity=0.2]
(axis cs:0,3.92977171861256)
--(axis cs:0,21.4035616147208)
--(axis cs:1,68.0748598998847)
--(axis cs:2,80.3412683421183)
--(axis cs:3,90)
--(axis cs:4,95.3067941706652)
--(axis cs:4,66.0265391626682)
--(axis cs:4,66.0265391626682)
--(axis cs:3,58)
--(axis cs:2,38.3253983245484)
--(axis cs:1,27.9251401001153)
--(axis cs:0,3.92977171861256)
--cycle;

\path [draw=crimson2143940, fill=crimson2143940, opacity=0.2]
(axis cs:0,-0.374785217660714)
--(axis cs:0,7.70811855099405)
--(axis cs:1,2.91068360252296)
--(axis cs:2,4.19419189831861)
--(axis cs:3,3)
--(axis cs:4,2)
--(axis cs:4,0)
--(axis cs:4,0)
--(axis cs:3,1)
--(axis cs:2,1.13914143501472)
--(axis cs:1,1.75598306414371)
--(axis cs:0,-0.374785217660714)
--cycle;

\path [draw=darkorange25512714, fill=darkorange25512714, opacity=0.2]
(axis cs:0,3.40998052768201)
--(axis cs:0,43.9233528056513)
--(axis cs:1,45.6999722684223)
--(axis cs:2,38.2260895880879)
--(axis cs:3,37.114792136029)
--(axis cs:4,27.5350960418098)
--(axis cs:4,-0.201762708476481)
--(axis cs:4,-0.201762708476481)
--(axis cs:3,0.218541197304383)
--(axis cs:2,3.10724374524543)
--(axis cs:1,5.63336106491104)
--(axis cs:0,3.40998052768201)
--cycle;

\path [draw=forestgreen4416044, fill=forestgreen4416044, opacity=0.2]
(axis cs:5,1.52213885141959)
--(axis cs:5,17.8111944819137)
--(axis cs:6,73.3593432882303)
--(axis cs:7,82.9956132754817)
--(axis cs:8,91.1488915650922)
--(axis cs:9,94.7404383925441)
--(axis cs:9,76.5928949407892)
--(axis cs:9,76.5928949407892)
--(axis cs:8,70.8511084349078)
--(axis cs:7,57.6710533911849)
--(axis cs:6,51.973990045103)
--(axis cs:5,1.52213885141959)
--cycle;

\path [draw=crimson2143940, fill=crimson2143940, opacity=0.2]
(axis cs:5,1.58500066720053)
--(axis cs:5,5.7483326661328)
--(axis cs:6,8)
--(axis cs:7,6.54788358699765)
--(axis cs:8,4)
--(axis cs:9,2)
--(axis cs:9,0)
--(axis cs:9,0)
--(axis cs:8,0)
--(axis cs:7,0.118783079669015)
--(axis cs:6,2)
--(axis cs:5,1.58500066720053)
--cycle;

\path [draw=darkorange25512714, fill=darkorange25512714, opacity=0.2]
(axis cs:5,4.38761047530528)
--(axis cs:5,30.9457228580281)
--(axis cs:6,46.9886973170741)
--(axis cs:7,54.5172530985706)
--(axis cs:8,31.1695399452047)
--(axis cs:9,22.1488915650922)
--(axis cs:9,1.85110843490778)
--(axis cs:9,1.85110843490778)
--(axis cs:8,2.16379338812861)
--(axis cs:7,-1.85058643190396)
--(axis cs:6,7.67796934959257)
--(axis cs:5,4.38761047530528)
--cycle;

\path [draw=forestgreen4416044, fill=forestgreen4416044, opacity=0.2]
(axis cs:10,11.2504295646786)
--(axis cs:10,27.4162371019881)
--(axis cs:11,72.733112580361)
--(axis cs:12,84.2962197731658)
--(axis cs:13,92.5325625946708)
--(axis cs:14,93.9372539331938)
--(axis cs:14,78.0627460668062)
--(axis cs:14,78.0627460668062)
--(axis cs:13,69.4674374053292)
--(axis cs:12,66.3704468935008)
--(axis cs:11,52.6002207529723)
--(axis cs:10,11.2504295646786)
--cycle;

\path [draw=crimson2143940, fill=crimson2143940, opacity=0.2]
(axis cs:10,1.61161620336277)
--(axis cs:10,7.72171712997056)
--(axis cs:11,6)
--(axis cs:12,4.86085856498528)
--(axis cs:13,4.7483326661328)
--(axis cs:14,1.24401693585629)
--(axis cs:14,0.0893163974770408)
--(axis cs:14,0.0893163974770408)
--(axis cs:13,0.585000667200533)
--(axis cs:12,1.80580810168139)
--(axis cs:11,2)
--(axis cs:10,1.61161620336277)
--cycle;

\path [draw=darkorange25512714, fill=darkorange25512714, opacity=0.2]
(axis cs:10,13.6292944562551)
--(axis cs:10,46.3707055437449)
--(axis cs:11,56.5273735575644)
--(axis cs:12,36)
--(axis cs:13,36.0880074906351)
--(axis cs:14,30.8745078663875)
--(axis cs:14,-0.874507866387544)
--(axis cs:14,-0.874507866387544)
--(axis cs:13,1.91199250936494)
--(axis cs:12,6)
--(axis cs:11,8.80595977576895)
--(axis cs:10,13.6292944562551)
--cycle;

\addplot [semithick, forestgreen4416044, mark=*, mark size=2.5, mark options={solid}, forget plot]
table {%
0 12.6666666666667
1 48
2 59.3333333333333
3 74
4 80.6666666666667
};
\addplot [semithick, crimson2143940, mark=triangle*, mark size=2.5, mark options={solid}, forget plot]
table {%
0 3.66666666666667
1 2.33333333333333
2 2.66666666666667
3 2
4 1
};
\addplot [semithick, darkorange25512714, mark=diamond*, mark size=2.5, mark options={solid}, forget plot]
table {%
0 23.6666666666667
1 25.6666666666667
2 20.6666666666667
3 18.6666666666667
4 13.6666666666667
};
\addplot [semithick, forestgreen4416044, mark=*, mark size=2.5, mark options={solid}, forget plot]
table {%
5 9.66666666666667
6 62.6666666666667
7 70.3333333333333
8 81
9 85.6666666666667
};
\addplot [semithick, crimson2143940, mark=triangle*, mark size=2.5, mark options={solid}, forget plot]
table {%
5 3.66666666666667
6 5
7 3.33333333333333
8 2
9 1
};
\addplot [semithick, darkorange25512714, mark=diamond*, mark size=2.5, mark options={solid}, forget plot]
table {%
5 17.6666666666667
6 27.3333333333333
7 26.3333333333333
8 16.6666666666667
9 12
};
\addplot [semithick, forestgreen4416044, mark=*, mark size=2.5, mark options={solid}, forget plot]
table {%
10 19.3333333333333
11 62.6666666666667
12 75.3333333333333
13 81
14 86
};
\addplot [semithick, crimson2143940, mark=triangle*, mark size=2.5, mark options={solid}, forget plot]
table {%
10 4.66666666666667
11 4
12 3.33333333333333
13 2.66666666666667
14 0.666666666666667
};
\addplot [semithick, darkorange25512714, mark=diamond*, mark size=2.5, mark options={solid}, forget plot]
table {%
10 30
11 32.6666666666667
12 21
13 19
14 15
};
\addplot [black, forget plot]
table {%
-0.7 3.5527136788005e-15
14.7 3.5527136788005e-15
};
\path [draw=gray, draw opacity=0.7, semithick, dash pattern=on 5.55pt off 2.4pt]
(axis cs:4.5,0)
--(axis cs:4.5,115);

\path [draw=gray, draw opacity=0.7, semithick, dash pattern=on 5.55pt off 2.4pt]
(axis cs:9.5,0)
--(axis cs:9.5,115);

\draw (axis cs:2,107) node[
  text=black,
  rotate=0.0
]{\Large\DDDtwo};
\draw (axis cs:7,107) node[
  text=black,
  rotate=0.0
]{\Large\DG};
\draw (axis cs:12,107) node[
  text=black,
  rotate=0.0
]{\Large\DDDtwoG};
\end{axis}

\end{tikzpicture}
    \end{subfigure}
    \hfill
    \begin{subfigure}[t]{0.48\textwidth}
        \centering
\begin{tikzpicture}[scale=0.63]

\definecolor{crimson2143940}{RGB}{214,39,40}
\definecolor{darkgray176}{RGB}{176,176,176}
\definecolor{darkorange25512714}{RGB}{255,127,14}
\definecolor{forestgreen4416044}{RGB}{44,160,44}
\definecolor{gray}{RGB}{128,128,128}
\definecolor{lightgray204}{RGB}{204,204,204}

\begin{axis}[
width=12cm,
height=8cm,
legend columns=4,
legend cell align={left},
legend style={
/tikz/every even column/.append style={column sep=0.3cm},
  fill opacity=0.8,
  draw opacity=1,
  text opacity=1,
  at={(0.5,0.97)},
  anchor=north,
  draw=lightgray204
},
tick align=outside,
title={\LARGE \(\displaystyle e_4\) -- \House},
x grid style={darkgray176},
xlabel={\Large\% of qualification data},
xmin=-0.7, xmax=14.7,
xtick pos=left,
xtick style={color=black},
xtick={0,1,2,3,4,5,6,7,8,9,10,11,12,13,14},
xticklabels={0,15,25,50,75,0,15,25,50,75,0,15,25,50,75},
y grid style={darkgray176},
ymajorticks=false,
ymin=0, ymax=115,
ytick style={color=black}
]
\path [draw=forestgreen4416044, fill=forestgreen4416044, opacity=0.2]
(axis cs:0,10.657146741242)
--(axis cs:0,34.0095199254247)
--(axis cs:1,82.4372109670681)
--(axis cs:2,89.9866613290624)
--(axis cs:3,93.5393920141695)
--(axis cs:4,97.0957671739953)
--(axis cs:4,84.237566159338)
--(axis cs:4,84.237566159338)
--(axis cs:3,74.4606079858305)
--(axis cs:2,56.6800053376043)
--(axis cs:1,44.8961223662652)
--(axis cs:0,10.657146741242)
--cycle;

\path [draw=crimson2143940, fill=crimson2143940, opacity=0.2]
(axis cs:0,3.15909611938056)
--(axis cs:0,14.1742372139528)
--(axis cs:1,3)
--(axis cs:2,2.91068360252296)
--(axis cs:3,3)
--(axis cs:4,1.82136720504592)
--(axis cs:4,-0.488033871712585)
--(axis cs:4,-0.488033871712585)
--(axis cs:3,1)
--(axis cs:2,1.75598306414371)
--(axis cs:1,1)
--(axis cs:0,3.15909611938056)
--cycle;

\path [draw=darkorange25512714, fill=darkorange25512714, opacity=0.2]
(axis cs:0,1.442561475698)
--(axis cs:0,14.557438524302)
--(axis cs:1,29.9235591004758)
--(axis cs:2,24.3593432882303)
--(axis cs:3,23.1706341710592)
--(axis cs:4,9)
--(axis cs:4,1)
--(axis cs:4,1)
--(axis cs:3,2.16269916227418)
--(axis cs:2,2.97399004510304)
--(axis cs:1,6.74310756619086)
--(axis cs:0,1.442561475698)
--cycle;

\path [draw=forestgreen4416044, fill=forestgreen4416044, opacity=0.2]
(axis cs:5,14.947736112502)
--(axis cs:5,38.3855972208313)
--(axis cs:6,80.8452179176176)
--(axis cs:7,86.4649347727802)
--(axis cs:8,95.3883837966372)
--(axis cs:9,97.8608585649853)
--(axis cs:9,94.8058081016814)
--(axis cs:9,94.8058081016814)
--(axis cs:8,89.2782828700294)
--(axis cs:7,76.2017318938864)
--(axis cs:6,73.8214487490491)
--(axis cs:5,14.947736112502)
--cycle;

\path [draw=crimson2143940, fill=crimson2143940, opacity=0.2]
(axis cs:5,1.39444872453601)
--(axis cs:5,8.60555127546399)
--(axis cs:6,2)
--(axis cs:7,3.24401693585629)
--(axis cs:8,1.91068360252296)
--(axis cs:9,0.910683602522959)
--(axis cs:9,-0.244016935856293)
--(axis cs:9,-0.244016935856293)
--(axis cs:8,0.755983064143708)
--(axis cs:7,2.08931639747704)
--(axis cs:6,0)
--(axis cs:5,1.39444872453601)
--cycle;

\path [draw=darkorange25512714, fill=darkorange25512714, opacity=0.2]
(axis cs:5,3.39444872453601)
--(axis cs:5,10.605551275464)
--(axis cs:6,21.211102550928)
--(axis cs:7,23.1182979191719)
--(axis cs:8,14.361047106675)
--(axis cs:9,7)
--(axis cs:9,1)
--(axis cs:9,1)
--(axis cs:8,2.30561955999163)
--(axis cs:7,4.21503541416145)
--(axis cs:6,6.78889744907202)
--(axis cs:5,3.39444872453601)
--cycle;

\path [draw=forestgreen4416044, fill=forestgreen4416044, opacity=0.2]
(axis cs:10,30.9468362497273)
--(axis cs:10,47.7198304169394)
--(axis cs:11,86)
--(axis cs:12,90.8388868327985)
--(axis cs:13,95.4641016151377)
--(axis cs:14,97.7483326661328)
--(axis cs:14,93.5850006672005)
--(axis cs:14,93.5850006672005)
--(axis cs:13,88.5358983848623)
--(axis cs:12,75.8277798338682)
--(axis cs:11,70)
--(axis cs:10,30.9468362497273)
--cycle;

\path [draw=crimson2143940, fill=crimson2143940, opacity=0.2]
(axis cs:10,6.26794919243112)
--(axis cs:10,9.73205080756888)
--(axis cs:11,5.64273441009184)
--(axis cs:12,5.73205080756888)
--(axis cs:13,2.48803387171258)
--(axis cs:14,0)
--(axis cs:14,0)
--(axis cs:14,0)
--(axis cs:13,0.178632794954082)
--(axis cs:12,2.26794919243112)
--(axis cs:11,1.02393225657483)
--(axis cs:10,6.26794919243112)
--cycle;

\path [draw=darkorange25512714, fill=darkorange25512714, opacity=0.2]
(axis cs:10,7.41742430504416)
--(axis cs:10,16.5825756949558)
--(axis cs:11,34.0934769394311)
--(axis cs:12,20.5440037453175)
--(axis cs:13,15.3571025019018)
--(axis cs:14,5.41499933279947)
--(axis cs:14,1.2516673338672)
--(axis cs:14,1.2516673338672)
--(axis cs:13,1.30956416476484)
--(axis cs:12,3.45599625468247)
--(axis cs:11,1.90652306056892)
--(axis cs:10,7.41742430504416)
--cycle;

\addplot [semithick, forestgreen4416044, mark=*, mark size=2.5, mark options={solid}, forget plot]
table {%
0 22.3333333333333
1 63.6666666666667
2 73.3333333333333
3 84
4 90.6666666666667
};
\addplot [semithick, crimson2143940, mark=triangle*, mark size=2.5, mark options={solid}, forget plot]
table {%
0 8.66666666666667
1 2
2 2.33333333333333
3 2
4 0.666666666666667
};
\addplot [semithick, darkorange25512714, mark=diamond*, mark size=2.5, mark options={solid}, forget plot]
table {%
0 8
1 18.3333333333333
2 13.6666666666667
3 12.6666666666667
4 5
};
\addplot [semithick, forestgreen4416044, mark=*, mark size=2.5, mark options={solid}, forget plot]
table {%
5 26.6666666666667
6 77.3333333333333
7 81.3333333333333
8 92.3333333333333
9 96.3333333333333
};
\addplot [semithick, crimson2143940, mark=triangle*, mark size=2.5, mark options={solid}, forget plot]
table {%
5 5
6 1
7 2.66666666666667
8 1.33333333333333
9 0.333333333333333
};
\addplot [semithick, darkorange25512714, mark=diamond*, mark size=2.5, mark options={solid}, forget plot]
table {%
5 7
6 14
7 13.6666666666667
8 8.33333333333333
9 4
};
\addplot [semithick, forestgreen4416044, mark=*, mark size=2.5, mark options={solid}, forget plot]
table {%
10 39.3333333333333
11 78
12 83.3333333333333
13 92
14 95.6666666666667
};
\addplot [semithick, crimson2143940, mark=triangle*, mark size=2.5, mark options={solid}, forget plot]
table {%
10 8
11 3.33333333333333
12 4
13 1.33333333333333
14 0
};
\addplot [semithick, darkorange25512714, mark=diamond*, mark size=2.5, mark options={solid}, forget plot]
table {%
10 12
11 18
12 12
13 8.33333333333333
14 3.33333333333333
};
\addplot [black, forget plot]
table {%
-0.7 3.5527136788005e-15
14.7 3.5527136788005e-15
};
\path [draw=gray, draw opacity=0.7, semithick, dash pattern=on 5.55pt off 2.4pt]
(axis cs:4.5,0)
--(axis cs:4.5,115);

\path [draw=gray, draw opacity=0.7, semithick, dash pattern=on 5.55pt off 2.4pt]
(axis cs:9.5,0)
--(axis cs:9.5,115);

\draw (axis cs:2,107) node[
  text=black,
  rotate=0.0
]{\Large\DDDtwo};
\draw (axis cs:7,107) node[
  text=black,
  rotate=0.0
]{\Large\DG};
\draw (axis cs:12,107) node[
  text=black,
  rotate=0.0
]{\Large\DDDtwoG};
\end{axis}

\end{tikzpicture}
    \end{subfigure}
    \\[-0.9cm]
    \caption{Operational performance indicators \revTwo{(averages over the three models with the standard deviation)} with GDs trained on different datasets.}
    \label{fig:extended_metric_qualification_mean}
\end{figure*}

A first evident observation is that the qualification procedure boosts the performance of the general detectors for the target environment and, unsurprisingly, the performance (together with the data preparation costs) increases as more samples are included, from $QD_e^{15}$ to $QD_e^{75}$.  This can be appreciated in the mAP and $TP_{\%}$ improvements visually depicted in Figures~\ref{fig:ap_qualification}~and~\ref{fig:extended_metric_qualification_mean} and by the decreasing trend (after the first qualification round) of $FP_{\%}$ and $BFD_{\%}$ in Figure~\ref{fig:extended_metric_qualification_mean}.

\begin{figure*}[!htb]
	\centering
    \begin{tabular}{@{}c@{ }c@{ }c@{ }c@{ }c@{ }}
    \rotatebox[origin=c]{90}{{\small DETR~\cite{detr}}}&
    \raisebox{-0.45\height}{\includegraphics[width=0.23\linewidth]{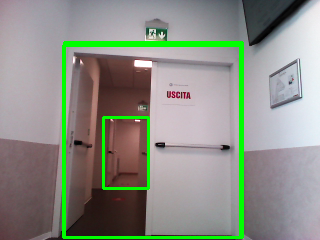}}&
    \raisebox{-0.45\height}{\includegraphics[width=0.23\linewidth]{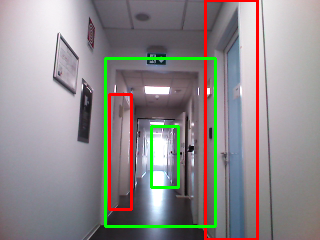}}&
    \raisebox{-0.45\height}{\includegraphics[width=0.23\linewidth]{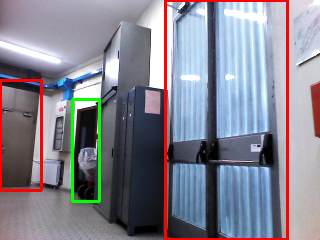}}&
    \raisebox{-0.45\height}{\includegraphics[width=0.23\linewidth]{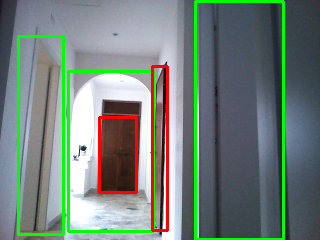}}\\\addlinespace[0.13cm]
    \rotatebox[origin=c]{90}{{\small YOLOv5~\cite{yolov5}}}&
    \raisebox{-0.45\height}{\includegraphics[width=0.23\linewidth]{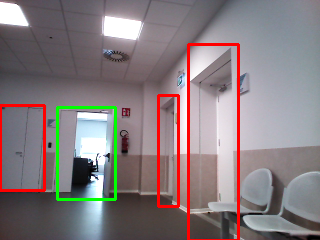}}&
    \raisebox{-0.45\height}{\includegraphics[width=0.23\linewidth]{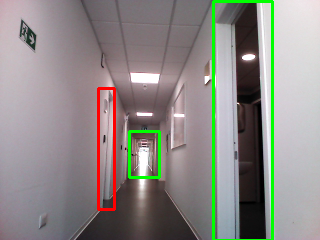}}&
    \raisebox{-0.45\height}{\includegraphics[width=0.23\linewidth]{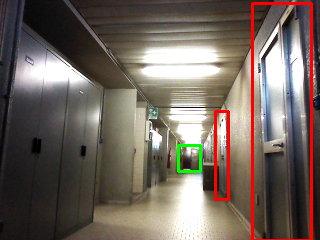}}&
    \raisebox{-0.45\height}{\includegraphics[width=0.23\linewidth]{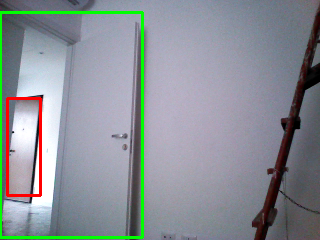}}\\\addlinespace[0.13cm]
    \rotatebox[origin=c]{90}{{\small Faster R--CNN~\cite{fasterrcnn}}}&
    \raisebox{-0.45\height}{\includegraphics[width=0.23\linewidth]{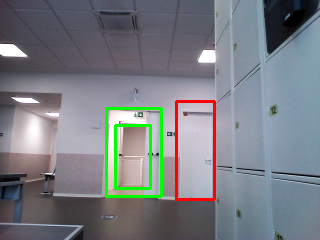}}&
    \raisebox{-0.45\height}{\includegraphics[width=0.23\linewidth]{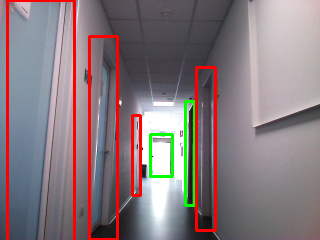}}&
    \raisebox{-0.45\height}{\includegraphics[width=0.23\linewidth]{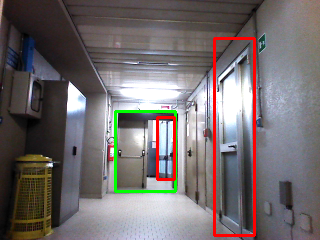}}&
    \raisebox{-0.45\height}{\includegraphics[width=0.23\linewidth]{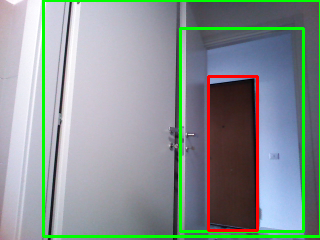}}\\\addlinespace[0.1cm]
    &$e_1$ -- \Classrooms & $e_2$ -- \Offices & $e_3$ -- \Laboratories& $e_4$ -- \House
\end{tabular}
	\caption{Challenging doors correctly detected by $QD^{15}_e$ (GDs divided by model and trained on \DG).}
	\label{fig:qd15_examples}
\end{figure*}

However, the increments follow a diminishing--returns trend, with large gains in the first qualification rounds and marginal ones as more data are used. Focusing on the average mAP and $TP_{\%}$ it can be seen how the qualified detector that scores the highest performance improvements is $QD^{15}_e$, despite requiring a relatively affordable effort for data preparation. From a practical perspective, this observation suggests how just a coarse visual inspection of the target environment might be enough to obtain an environment--specific detector whose performance is significantly better than the corresponding general one. In such a case the robot's deployment time is only marginally affected. To give a concrete idea, annotating the $15\%$ of the data collected by the robot's first exploration of a new environment (approximately 80 images) required a human operator using our tool (Section~\ref{sec:qualification}) about half an hour. Another key finding is how the improvements through qualification are distributed across various types of instances encountered by the detector. Upon direct inspection, we observed how these were particularly notable in challenging instances. Figure~\ref{fig:qd15_examples} showcases significant examples of this, illustrating how the $QD^{15}_e$ model, based on our dataset \DG{}, successfully detects doors in highly challenging instances. These include scenarios with nested or partially occluded doors and even situations where the door is hidden in the background.

It is important to notice that the dataset chosen to train the general detector does affect the benefits of the qualification. The trends observed in Figure~\ref{fig:ap_qualification} indicate that QDs based on \DDDtwo{} generally demonstrate lower performance compared to those based on \DG{} and \DDDtwoG. This observation is further supported by the data presented in Figure~\ref{fig:extended_metric_qualification_mean}. Although the error rates ($FP_{\%}$ and $BFD_{\%}$), are substantially similar, there is a noticeable difference in the $TP_{\%}$. Specifically, detectors based on \DG{} or \DDDtwoG{} show better $TP_{\%}$ performance than those based on \DDDtwo. Confirming the findings from the previous section, this again suggests that training on images not representing the robot's point of view, although taken from the real world, hits a performance limit. A simulated dataset from the robot perspective with an adequate level of photorealism (as \DG), when included in the training phase, enables the detectors to reach better performance when qualified for a target environment.

To further support the effectiveness of the method of Section~\ref{sec:gd}, we can notice that \DDDtwoG, which integrates the realism in \DDDtwo{} and the robot perception model of \DG, does not introduce significant variations in the performance of the qualified detectors when compared with those solely based on \DG{}. This can be easily seen by comparing the (substantially similar) orange and red bars of Figure~\ref{fig:ap_qualification} that refer to \DG{} and \DDDtwoG{}, respectively. In addition, while $TP_{\%}$ reaches comparable performance, Table~\ref{tab:qualifications_results} shows that \DG{} enables the qualified detectors to reduce the rate of $BFD$ with respect to \DDDtwoG.

\revI{It is important to remark that the qualification procedure is effective if the detector is qualified and then used inside the same environment, a condition that perfectly matches the practical deployments of service robots. To assess this claim, we conduct additional experiments to assess the performance of qualified detectors on data from a different distribution (i.e., acquired from a different environment). To do this, we fine--tune the general detectors using data from one or more environments and we test using images from a new one (e.g., fine--tuning QD on $e_1$, $e_2$, $e_3$, testing on $e_4$). 
We observed that when a few examples are used for fine--tuning, this procedure results in minor performance improvements when compared with a GD; still, performances are below those of $QD_e^{15}$ (trained with the data of the target environment). Moreover, using more data for qualification results in a performance drop as the qualified detectors overfit the training data that come from different environments to the one used for testing. We omit these results for the sake of brevity.}

\RevOne{\subsection{Evaluation in Challenging Settings}\label{sec:Results:LongTerm}}

\RevOne{
As discussed in Section~\ref{sec:qualification}, the advantage represented by a qualified detector is enabled by the long--term deployment of the robot in the same target environment, where the same object instances get repeatedly observed. However, while the observed doors are the same, transient changes in the environment's appearance might still take place resulting in unpredictable domain shifts.


We deem that one of the most significant shifts might occur at the \emph{feature level} of the robot's perceptions~\cite{lee2022surgical}. For a long--term deployment in a human--centric environment, we considered two possible factors of such a feature shift: the variations in illumination between day and night and dynamic camera occlusions caused by people walking around.
In the first case, changes in lighting can significantly alter the appearance of doors and these variations in illumination can be widespread throughout the entire environment (e.g., day/night), or being localized (e.g., light reflections). In the second case, dynamic actors walking freely within the environment can obstruct the robot's view, especially in confined areas such as narrow corridors or passageways.
To test the robustness of our approach with light variations, we included in our real--world dataset (following the same procedure of Section~\ref{sec:ExperimentalSetting:Datasets}) additional data from $e_1$ and $e_2$ during nighttime, when only artificial light is present and some rooms are entirely dark. For camera occlusions, we acquired new data in $e_2$ while having people intentionally walking by the robot or loitering in its vicinity.
We used these data to test our qualified detectors (Section~\ref{sec:Results:QD}), which were trained during daytime hours when the environment was sparsely populated (as is typical during a deployment phase). 
The metrics' average performance obtained by DETR~\cite{detr}, YOLOv5~\cite{yolov5}, and Faster R--CNN~\cite{fasterrcnn} are detailed in Table~\ref{tab:ap_extended_metric_results_diff_cond} and visually shown in Figure~\ref{fig:ap_diff_conf} (mAP) and Figure~\ref{fig:extended_metric_diff_cond} ($TP_{\%}$, $FP_{\%}$, and $BFD_{\%}$).
}

\begin{table*}[!htb]

\setlength\tabcolsep{2.2pt}
\setlength\extrarowheight{2pt}
\centering
\begin{tabular}{c|cc|cccc|cccc|cccc}
\toprule
        & & & \multicolumn{4}{c|}{\DDDtwo} & \multicolumn{4}{c|}{\DG} &\multicolumn{4}{c}{\DDDtwoG} \\
        &\textbf{Env.} & \textbf{Exp.} & \textbf{mAP}$\uparrow$& $\mathbf{TP}_{\%}$$\uparrow$ &  $\mathbf{FP}_{\%}$$\downarrow$ & $\mathbf{BFD}_{\%}$$\downarrow$ & \textbf{mAP}$\uparrow$& $\mathbf{TP}_{\%}$$\uparrow$ &  $\mathbf{FP}_{\%}$$\downarrow$ & $\mathbf{BFD}_{\%}$$\downarrow$ & \textbf{mAP}$\uparrow$& $\mathbf{TP}_{\%}$$\uparrow$ &  $\mathbf{FP}_{\%}$$\downarrow$ & $\mathbf{BFD}_{\%}$$\downarrow$ \\
\midrule
\multirow{10}{*}[-2ex]{\rotatebox[origin=c]{90}{Nighttime  - light variations}}&\multicolumn{1}{c}{\multirow{5}{*}[-1.5ex]{$e_1$}} &$GD$&$10 \pm 6 $&$15 \pm 8 $&$5 \pm 2 $&$\textbf{9} \pm 6 $&$26 \pm 2 $&$31 \pm 2 $&$\textbf{4} \pm 1 $&$10 \pm 4 $&$\textbf{31} \pm 5 $&$\textbf{37} \pm 6 $&$6 \pm 2 $&$13 \pm 6 $\\[2pt]
&\multicolumn{1}{c}{} &$QD_{e}^{15}$&$27 \pm 10 $&$32 \pm 10 $&$\textbf{4} \pm 2 $&$\textbf{14} \pm 11 $&$41 \pm 5 $&$47 \pm 2 $&$4 \pm 2 $&$17 \pm 13 $&$\textbf{41} \pm 3 $&$\textbf{47} \pm 1 $&$4 \pm 2 $&$17 \pm 12 $\\[2pt]
&\multicolumn{1}{c}{} &$QD_{e}^{25}$&$39 \pm 10 $&$44 \pm 9 $&$\textbf{4} \pm 2 $&$\textbf{16} \pm 12 $&$\textbf{48} \pm 5 $&$53 \pm 2 $&$4 \pm 3 $&$18 \pm 14 $&$47 \pm 5 $&$\textbf{53} \pm 3 $&$5 \pm 2 $&$18 \pm 13 $\\[2pt]
&\multicolumn{1}{c}{} &$QD_{e}^{50}$&$44 \pm 12 $&$49 \pm 11 $&$\textbf{4} \pm 2 $&$15 \pm 13 $&$52 \pm 4 $&$57 \pm 4 $&$4 \pm 1 $&$\textbf{14} \pm 11 $&$\textbf{54} \pm 7 $&$\textbf{58} \pm 6 $&$4 \pm 3 $&$17 \pm 13 $\\[2pt]
&\multicolumn{1}{c}{} &$QD_{e}^{75}$&$50 \pm 10 $&$54 \pm 9 $&$5 \pm 2 $&$13 \pm 11 $&$\textbf{56} \pm 5 $&$\textbf{60} \pm 3 $&$5 \pm 2 $&$13 \pm 11 $&$52 \pm 6 $&$57 \pm 5 $&$\textbf{4} \pm 3 $&$\textbf{12} \pm 9 $\\[2pt]
\cline{2-15}
&\multicolumn{1}{c}{\multirow{5}{*}[-1.5ex]{$e_2$}} &$GD$&$12 \pm 9 $&$18 \pm 14 $&$\textbf{2} \pm 2 $&$\textbf{9} \pm 8 $&$16 \pm 3 $&$27 \pm 4 $&$6 \pm 2 $&$17 \pm 10 $&$\textbf{24} \pm 5 $&$\textbf{34} \pm 7 $&$5 \pm 0 $&$17 \pm 8 $\\[2pt]
&\multicolumn{1}{c}{} &$QD_{e}^{15}$&$30 \pm 10 $&$39 \pm 17 $&$\textbf{3} \pm 2 $&$\textbf{19} \pm 16 $&$39 \pm 7 $&$49 \pm 11 $&$3 \pm 2 $&$27 \pm 21 $&$\textbf{42} \pm 9 $&$\textbf{51} \pm 13 $&$3 \pm 1 $&$25 \pm 19 $\\[2pt]
&\multicolumn{1}{c}{} &$QD_{e}^{25}$&$39 \pm 13 $&$47 \pm 15 $&$4 \pm 1 $&$24 \pm 19 $&$48 \pm 4 $&$56 \pm 5 $&$4 \pm 2 $&$\textbf{22} \pm 16 $&$\textbf{50} \pm 4 $&$\textbf{59} \pm 6 $&$\textbf{3} \pm 1 $&$22 \pm 16 $\\[2pt]
&\multicolumn{1}{c}{} &$QD_{e}^{50}$&$48 \pm 16 $&$54 \pm 13 $&$\textbf{3} \pm 1 $&$18 \pm 15 $&$57 \pm 8 $&$\textbf{64} \pm 5 $&$3 \pm 1 $&$\textbf{15} \pm 12 $&$\textbf{58} \pm 6 $&$63 \pm 6 $&$3 \pm 0 $&$17 \pm 10 $\\[2pt]
&\multicolumn{1}{c}{} &$QD_{e}^{75}$&$54 \pm 13 $&$60 \pm 9 $&$\textbf{3} \pm 1 $&$15 \pm 12 $&$61 \pm 7 $&$67 \pm 5 $&$3 \pm 1 $&$\textbf{15} \pm 10 $&$\textbf{62} \pm 7 $&$\textbf{68} \pm 6 $&$4 \pm 2 $&$16 \pm 12 $\\[2pt]
\hline
\hline
\multirow{5}{*}[-2ex]{\rotatebox[origin=c]{90}{Occlusions}}&\multicolumn{1}{c}{\multirow{5}{*}[-1.5ex]{$e_2$}} &$GD$&$10 \pm 7 $&$14 \pm 9 $&$\textbf{3} \pm 2 $&$14 \pm 10 $&$13 \pm 7 $&$18 \pm 8 $&$7 \pm 1 $&$\textbf{14} \pm 9 $&$\textbf{17} \pm 2 $&$\textbf{23} \pm 3 $&$7 \pm 1 $&$16 \pm 9 $\\[2pt]
&\multicolumn{1}{c}{} &$QD_{e}^{15}$&$26 \pm 11 $&$34 \pm 15 $&$\textbf{4} \pm 2 $&$\textbf{22} \pm 19 $&$38 \pm 4 $&$46 \pm 6 $&$5 \pm 2 $&$27 \pm 25 $&$\textbf{38} \pm 7 $&$\textbf{47} \pm 9 $&$5 \pm 1 $&$28 \pm 21 $\\[2pt]
&\multicolumn{1}{c}{} &$QD_{e}^{25}$&$35 \pm 14 $&$43 \pm 13 $&$6 \pm 3 $&$27 \pm 22 $&$43 \pm 6 $&$51 \pm 6 $&$5 \pm 1 $&$\textbf{23} \pm 17 $&$\textbf{44} \pm 6 $&$\textbf{52} \pm 6 $&$\textbf{5} \pm 2 $&$24 \pm 17 $\\[2pt]
&\multicolumn{1}{c}{} &$QD_{e}^{50}$&$43 \pm 17 $&$50 \pm 14 $&$5 \pm 1 $&$21 \pm 17 $&$52 \pm 7 $&$58 \pm 6 $&$\textbf{4} \pm 2 $&$\textbf{20} \pm 14 $&$\textbf{53} \pm 5 $&$\textbf{58} \pm 4 $&$4 \pm 1 $&$20 \pm 14 $\\[2pt]
&\multicolumn{1}{c}{} &$QD_{e}^{75}$&$47 \pm 11 $&$54 \pm 8 $&$\textbf{5} \pm 2 $&$22 \pm 19 $&$54 \pm 7 $&$60 \pm 8 $&$6 \pm 3 $&$\textbf{20} \pm 14 $&$\textbf{55} \pm 8 $&$\textbf{62} \pm 7 $&$5 \pm 3 $&$24 \pm 18 $\\[2pt]

\bottomrule
\end{tabular}
\caption{General and qualified detector performance  \revTwo{(averaged over detectors, together with the standard deviations)} tested in nighttime and with camera occlusions. \revI{Bold entries indicate the best performance on each metric across the three datasets.}}
\label{tab:ap_extended_metric_results_diff_cond}

\end{table*}

\begin{figure*}[!htb]
    \centering
    \begin{subfigure}[t]{\textwidth}
        \centering
    \begin{tikzpicture}[scale=0.8]

\definecolor{crimson2143940}{RGB}{214,39,40}
\definecolor{darkgray176}{RGB}{176,176,176}
\definecolor{forestgreen4416044}{RGB}{44,160,44}
\definecolor{lightgray204}{RGB}{204,204,204}
\definecolor{steelblue31119180}{RGB}{31,119,180}
\definecolor{darkorange25512714}{RGB}{255,127,14}

\begin{axis}[
hide axis,
legend columns=5,
legend cell align={left},
legend style={
/tikz/every even column/.append style={column sep=0.3cm},
  fill opacity=0.8,
  draw opacity=1,
  text opacity=1,
  at={(0.5,0.95)},
  anchor=north,
  draw=lightgray204
},
xmin=-0.315, xmax=4.855,
y grid style={darkgray176},
ymin=0, ymax=130
]
\addlegendimage{ybar,area legend,draw=forestgreen4416044,fill=forestgreen4416044}
\addlegendentry{\Large \DDDtwo}
\addlegendimage{ybar,area legend,draw=darkorange25512714,fill=darkorange25512714,opacity=0.9,thick}
\addlegendentry{\Large \DG}
\addlegendimage{ybar,area legend,draw=crimson2143940,fill=crimson2143940,opacity=0.9,thick}
\addlegendentry{\Large \DDDtwoG}
\addlegendimage{ybar,area legend,draw=black,fill=none,opacity=0.9,thick,postaction={pattern=north east lines, fill opacity=0.9}}
\addlegendentry{\large Closed door}
\addlegendimage{ybar,area legend,draw=black,fill=none,opacity=0.9,thick}
\addlegendentry{\Large Open door}
\end{axis}
\end{tikzpicture}
    \end{subfigure}
    \\[-0.8cm]
    \begin{subfigure}[t]{0.32\textwidth}
        \centering
\begin{tikzpicture}[scale=0.41]

\definecolor{crimson2143940}{RGB}{214,39,40}
\definecolor{darkgray176}{RGB}{176,176,176}
\definecolor{darkorange25512714}{RGB}{255,127,14}
\definecolor{forestgreen4416044}{RGB}{44,160,44}
\definecolor{lightgray204}{RGB}{204,204,204}

\begin{axis}[
width=12cm,
height=7cm,
legend columns=3,
legend cell align={left},
legend style={
/tikz/every even column/.append style={column sep=0.3cm},
  fill opacity=0.8,
  draw opacity=1,
  text opacity=1,
  at={(0.5,0.95)},
  anchor=north,
  draw=lightgray204
},
tick align=outside,
tick pos=left,
title={\huge \(\displaystyle e_1\) -- \Classrooms{} under light variations},
x grid style={darkgray176},
xlabel={\huge Model},
xmin=-0.315, xmax=4.855,
xtick style={color=black},
xtick={0.27,1.27,2.27,3.27,4.27},
xticklabels={
  \huge\(\displaystyle GD\),
  \huge\(\displaystyle QD_{e}^{15}\),
  \huge\(\displaystyle QD_{e}^{25}\),
  \huge\(\displaystyle QD_{e}^{50}\),
\huge  \(\displaystyle QD_{e}^{75}\)
},
y grid style={darkgray176},
ylabel={\huge mAP},
ymin=0, ymax=80,
ytick style={color=black}
]
\draw[draw=black,fill=forestgreen4416044,opacity=0.9,thick,postaction={pattern=north east lines, fill opacity=0.9}] (axis cs:-0.08,0) rectangle (axis cs:0.12,3);
\draw[draw=black,fill=forestgreen4416044,opacity=0.9,thick,postaction={pattern=north east lines, fill opacity=0.9}] (axis cs:0.92,0) rectangle (axis cs:1.12,16);
\draw[draw=black,fill=forestgreen4416044,opacity=0.9,thick,postaction={pattern=north east lines, fill opacity=0.9}] (axis cs:1.92,0) rectangle (axis cs:2.12,24.1666666666667);
\draw[draw=black,fill=forestgreen4416044,opacity=0.9,thick,postaction={pattern=north east lines, fill opacity=0.9}] (axis cs:2.92,0) rectangle (axis cs:3.12,26.3333333333333);
\draw[draw=black,fill=forestgreen4416044,opacity=0.9,thick,postaction={pattern=north east lines, fill opacity=0.9}] (axis cs:3.92,0) rectangle (axis cs:4.12,28.3333333333333);
\draw[draw=black,fill=forestgreen4416044,opacity=0.9,thick] (axis cs:-0.08,3) rectangle (axis cs:0.12,10.3333333333333);
\draw[draw=black,fill=forestgreen4416044,opacity=0.9,thick] (axis cs:0.92,16) rectangle (axis cs:1.12,27.1666666666667);
\draw[draw=black,fill=forestgreen4416044,opacity=0.9,thick] (axis cs:1.92,24.1666666666667) rectangle (axis cs:2.12,38.8333333333333);
\draw[draw=black,fill=forestgreen4416044,opacity=0.9,thick] (axis cs:2.92,26.3333333333333) rectangle (axis cs:3.12,43.5);
\draw[draw=black,fill=forestgreen4416044,opacity=0.9,thick] (axis cs:3.92,28.3333333333333) rectangle (axis cs:4.12,49.6666666666667);
\draw[draw=black,fill=darkorange25512714,opacity=0.9,thick,postaction={pattern=north east lines, fill opacity=0.9}] (axis cs:0.17,0) rectangle (axis cs:0.37,15.6666666666667);
\draw[draw=black,fill=darkorange25512714,opacity=0.9,thick,postaction={pattern=north east lines, fill opacity=0.9}] (axis cs:1.17,0) rectangle (axis cs:1.37,24.6666666666667);
\draw[draw=black,fill=darkorange25512714,opacity=0.9,thick,postaction={pattern=north east lines, fill opacity=0.9}] (axis cs:2.17,0) rectangle (axis cs:2.37,27.8333333333333);
\draw[draw=black,fill=darkorange25512714,opacity=0.9,thick,postaction={pattern=north east lines, fill opacity=0.9}] (axis cs:3.17,0) rectangle (axis cs:3.37,30.3333333333333);
\draw[draw=black,fill=darkorange25512714,opacity=0.9,thick,postaction={pattern=north east lines, fill opacity=0.9}] (axis cs:4.17,0) rectangle (axis cs:4.37,31.8333333333333);
\draw[draw=black,fill=darkorange25512714,opacity=0.9,thick] (axis cs:0.17,15.6666666666667) rectangle (axis cs:0.37,26.3333333333333);
\draw[draw=black,fill=darkorange25512714,opacity=0.9,thick] (axis cs:1.17,24.6666666666667) rectangle (axis cs:1.37,40.8333333333333);
\draw[draw=black,fill=darkorange25512714,opacity=0.9,thick] (axis cs:2.17,27.8333333333333) rectangle (axis cs:2.37,48);
\draw[draw=black,fill=darkorange25512714,opacity=0.9,thick] (axis cs:3.17,30.3333333333333) rectangle (axis cs:3.37,52.5);
\draw[draw=black,fill=darkorange25512714,opacity=0.9,thick] (axis cs:4.17,31.8333333333333) rectangle (axis cs:4.37,56.1666666666667);
\draw[draw=black,fill=crimson2143940,opacity=0.9,thick,postaction={pattern=north east lines, fill opacity=0.9}] (axis cs:0.42,0) rectangle (axis cs:0.62,19.8333333333333);
\draw[draw=black,fill=crimson2143940,opacity=0.9,thick,postaction={pattern=north east lines, fill opacity=0.9}] (axis cs:1.42,0) rectangle (axis cs:1.62,24.5);
\draw[draw=black,fill=crimson2143940,opacity=0.9,thick,postaction={pattern=north east lines, fill opacity=0.9}] (axis cs:2.42,0) rectangle (axis cs:2.62,29);
\draw[draw=black,fill=crimson2143940,opacity=0.9,thick,postaction={pattern=north east lines, fill opacity=0.9}] (axis cs:3.42,0) rectangle (axis cs:3.62,30.1666666666667);
\draw[draw=black,fill=crimson2143940,opacity=0.9,thick,postaction={pattern=north east lines, fill opacity=0.9}] (axis cs:4.42,0) rectangle (axis cs:4.62,31);
\draw[draw=black,fill=crimson2143940,opacity=0.9,thick] (axis cs:0.42,19.8333333333333) rectangle (axis cs:0.62,30.8333333333333);
\draw[draw=black,fill=crimson2143940,opacity=0.9,thick] (axis cs:1.42,24.5) rectangle (axis cs:1.62,41.1666666666667);
\draw[draw=black,fill=crimson2143940,opacity=0.9,thick] (axis cs:2.42,29) rectangle (axis cs:2.62,47.3333333333333);
\draw[draw=black,fill=crimson2143940,opacity=0.9,thick] (axis cs:3.42,30.1666666666667) rectangle (axis cs:3.62,53.8333333333333);
\draw[draw=black,fill=crimson2143940,opacity=0.9,thick] (axis cs:4.42,31) rectangle (axis cs:4.62,52.5);
\path [draw=black, semithick]
(axis cs:0.02,4.80311332565431)
--(axis cs:0.02,15.8635533410124);

\path [draw=black, semithick]
(axis cs:1.02,17.6229077985197)
--(axis cs:1.02,36.7104255348136);

\path [draw=black, semithick]
(axis cs:2.02,28.8166805324555)
--(axis cs:2.02,48.8499861342111);

\path [draw=black, semithick]
(axis cs:3.02,31.7420239836952)
--(axis cs:3.02,55.2579760163048);

\path [draw=black, semithick]
(axis cs:4.02,39.3669902963464)
--(axis cs:4.02,59.9663430369869);

\addplot [semithick, black, mark=-, mark size=3, mark options={solid}, only marks, forget plot]
table {%
0.02 4.80311332565431
1.02 17.6229077985197
2.02 28.8166805324555
3.02 31.7420239836952
4.02 39.3669902963464
};
\addplot [semithick, black, mark=-, mark size=3, mark options={solid}, only marks, forget plot]
table {%
0.02 15.8635533410124
1.02 36.7104255348136
2.02 48.8499861342111
3.02 55.2579760163048
4.02 59.9663430369869
};
\path [draw=black, semithick]
(axis cs:0.27,23.970425520207)
--(axis cs:0.27,28.6962411464596);

\path [draw=black, semithick]
(axis cs:1.27,35.8250069328944)
--(axis cs:1.27,45.8416597337722);

\path [draw=black, semithick]
(axis cs:2.27,42.7321731235736)
--(axis cs:2.27,53.2678268764264);

\path [draw=black, semithick]
(axis cs:3.27,48.0559027913422)
--(axis cs:3.27,56.9440972086578);

\path [draw=black, semithick]
(axis cs:4.27,51.4144741902056)
--(axis cs:4.27,60.9188591431278);

\addplot [semithick, black, mark=-, mark size=3, mark options={solid}, only marks, forget plot]
table {%
0.27 23.970425520207
1.27 35.8250069328944
2.27 42.7321731235736
3.27 48.0559027913422
4.27 51.4144741902056
};
\addplot [semithick, black, mark=-, mark size=3, mark options={solid}, only marks, forget plot]
table {%
0.27 28.6962411464596
1.27 45.8416597337722
2.27 53.2678268764264
3.27 56.9440972086578
4.27 60.9188591431278
};
\path [draw=black, semithick]
(axis cs:0.52,25.7017318938864)
--(axis cs:0.52,35.9649347727802);

\path [draw=black, semithick]
(axis cs:1.52,37.7632370238896)
--(axis cs:1.52,44.5700963094437);

\path [draw=black, semithick]
(axis cs:2.52,42.2261488513185)
--(axis cs:2.52,52.4405178153482);

\path [draw=black, semithick]
(axis cs:3.52,46.5132695813341)
--(axis cs:3.52,61.1533970853326);

\path [draw=black, semithick]
(axis cs:4.52,46.5)
--(axis cs:4.52,58.5);

\addplot [semithick, black, mark=-, mark size=3, mark options={solid}, only marks, forget plot]
table {%
0.52 25.7017318938864
1.52 37.7632370238896
2.52 42.2261488513185
3.52 46.5132695813341
4.52 46.5
};
\addplot [semithick, black, mark=-, mark size=3, mark options={solid}, only marks, forget plot]
table {%
0.52 35.9649347727802
1.52 44.5700963094437
2.52 52.4405178153482
3.52 61.1533970853326
4.52 58.5
};
\end{axis}

\end{tikzpicture}
    \end{subfigure}
    \hspace{0.5em}
    \begin{subfigure}[t]{0.32\textwidth}
        \centering
\begin{tikzpicture}[scale=0.42]

\definecolor{crimson2143940}{RGB}{214,39,40}
\definecolor{darkgray176}{RGB}{176,176,176}
\definecolor{darkorange25512714}{RGB}{255,127,14}
\definecolor{forestgreen4416044}{RGB}{44,160,44}
\definecolor{lightgray204}{RGB}{204,204,204}

\begin{axis}[
width=12cm,
height=7cm,
legend columns=3,
legend cell align={left},
legend style={
/tikz/every even column/.append style={column sep=0.3cm},
  fill opacity=0.8,
  draw opacity=1,
  text opacity=1,
  at={(0.5,0.95)},
  anchor=north,
  draw=lightgray204
},
tick align=outside,
title={\huge\(\displaystyle e_2\) -- \Offices{} under light variations},
x grid style={darkgray176},
xlabel={\huge Model},
xmin=-0.315, xmax=4.855,
xtick pos=left,
xtick style={color=black},
xtick={0.27,1.27,2.27,3.27,4.27},
xticklabels={
 \huge \(\displaystyle GD\),
 \huge \(\displaystyle QD_{e}^{15}\),
 \huge \(\displaystyle QD_{e}^{25}\),
 \huge \(\displaystyle QD_{e}^{50}\),
 \huge \(\displaystyle QD_{e}^{75}\)
},
y grid style={darkgray176},
ymajorticks=false,
ymin=0, ymax=80,
ytick style={color=black}
]
\draw[draw=black,fill=forestgreen4416044,opacity=0.9,thick,postaction={pattern=north east lines, fill opacity=0.9}] (axis cs:-0.08,0) rectangle (axis cs:0.12,0.333333333333333);
\draw[draw=black,fill=forestgreen4416044,opacity=0.9,thick,postaction={pattern=north east lines, fill opacity=0.9}] (axis cs:0.92,0) rectangle (axis cs:1.12,12.3333333333333);
\draw[draw=black,fill=forestgreen4416044,opacity=0.9,thick,postaction={pattern=north east lines, fill opacity=0.9}] (axis cs:1.92,0) rectangle (axis cs:2.12,16.5);
\draw[draw=black,fill=forestgreen4416044,opacity=0.9,thick,postaction={pattern=north east lines, fill opacity=0.9}] (axis cs:2.92,0) rectangle (axis cs:3.12,22);
\draw[draw=black,fill=forestgreen4416044,opacity=0.9,thick,postaction={pattern=north east lines, fill opacity=0.9}] (axis cs:3.92,0) rectangle (axis cs:4.12,25.5);
\draw[draw=black,fill=forestgreen4416044,opacity=0.9,thick] (axis cs:-0.08,0.333333333333333) rectangle (axis cs:0.12,12.3333333333333);
\draw[draw=black,fill=forestgreen4416044,opacity=0.9,thick] (axis cs:0.92,12.3333333333333) rectangle (axis cs:1.12,30.1666666666667);
\draw[draw=black,fill=forestgreen4416044,opacity=0.9,thick] (axis cs:1.92,16.5) rectangle (axis cs:2.12,39);
\draw[draw=black,fill=forestgreen4416044,opacity=0.9,thick] (axis cs:2.92,22) rectangle (axis cs:3.12,48.3333333333333);
\draw[draw=black,fill=forestgreen4416044,opacity=0.9,thick] (axis cs:3.92,25.5) rectangle (axis cs:4.12,54);
\draw[draw=black,fill=darkorange25512714,opacity=0.9,thick,postaction={pattern=north east lines, fill opacity=0.9}] (axis cs:0.17,0) rectangle (axis cs:0.37,0.666666666666667);
\draw[draw=black,fill=darkorange25512714,opacity=0.9,thick,postaction={pattern=north east lines, fill opacity=0.9}] (axis cs:1.17,0) rectangle (axis cs:1.37,16.3333333333333);
\draw[draw=black,fill=darkorange25512714,opacity=0.9,thick,postaction={pattern=north east lines, fill opacity=0.9}] (axis cs:2.17,0) rectangle (axis cs:2.37,20.5);
\draw[draw=black,fill=darkorange25512714,opacity=0.9,thick,postaction={pattern=north east lines, fill opacity=0.9}] (axis cs:3.17,0) rectangle (axis cs:3.37,25.6666666666667);
\draw[draw=black,fill=darkorange25512714,opacity=0.9,thick,postaction={pattern=north east lines, fill opacity=0.9}] (axis cs:4.17,0) rectangle (axis cs:4.37,28.6666666666667);
\draw[draw=black,fill=darkorange25512714,opacity=0.9,thick] (axis cs:0.17,0.666666666666667) rectangle (axis cs:0.37,16);
\draw[draw=black,fill=darkorange25512714,opacity=0.9,thick] (axis cs:1.17,16.3333333333333) rectangle (axis cs:1.37,39);
\draw[draw=black,fill=darkorange25512714,opacity=0.9,thick] (axis cs:2.17,20.5) rectangle (axis cs:2.37,47.6666666666667);
\draw[draw=black,fill=darkorange25512714,opacity=0.9,thick] (axis cs:3.17,25.6666666666667) rectangle (axis cs:3.37,56.6666666666667);
\draw[draw=black,fill=darkorange25512714,opacity=0.9,thick] (axis cs:4.17,28.6666666666667) rectangle (axis cs:4.37,61.3333333333333);
\draw[draw=black,fill=crimson2143940,opacity=0.9,thick,postaction={pattern=north east lines, fill opacity=0.9}] (axis cs:0.42,0) rectangle (axis cs:0.62,2.16666666666667);
\draw[draw=black,fill=crimson2143940,opacity=0.9,thick,postaction={pattern=north east lines, fill opacity=0.9}] (axis cs:1.42,0) rectangle (axis cs:1.62,17.1666666666667);
\draw[draw=black,fill=crimson2143940,opacity=0.9,thick,postaction={pattern=north east lines, fill opacity=0.9}] (axis cs:2.42,0) rectangle (axis cs:2.62,22.3333333333333);
\draw[draw=black,fill=crimson2143940,opacity=0.9,thick,postaction={pattern=north east lines, fill opacity=0.9}] (axis cs:3.42,0) rectangle (axis cs:3.62,26.8333333333333);
\draw[draw=black,fill=crimson2143940,opacity=0.9,thick,postaction={pattern=north east lines, fill opacity=0.9}] (axis cs:4.42,0) rectangle (axis cs:4.62,28.8333333333333);
\draw[draw=black,fill=crimson2143940,opacity=0.9,thick] (axis cs:0.42,2.16666666666667) rectangle (axis cs:0.62,23.6666666666667);
\draw[draw=black,fill=crimson2143940,opacity=0.9,thick] (axis cs:1.42,17.1666666666667) rectangle (axis cs:1.62,42);
\draw[draw=black,fill=crimson2143940,opacity=0.9,thick] (axis cs:2.42,22.3333333333333) rectangle (axis cs:2.62,50.3333333333333);
\draw[draw=black,fill=crimson2143940,opacity=0.9,thick] (axis cs:3.42,26.8333333333333) rectangle (axis cs:3.62,57.6666666666667);
\draw[draw=black,fill=crimson2143940,opacity=0.9,thick] (axis cs:4.42,28.8333333333333) rectangle (axis cs:4.62,62);
\path [draw=black, semithick]
(axis cs:0.02,3.21832739074979)
--(axis cs:0.02,21.4483392759169);

\path [draw=black, semithick]
(axis cs:1.02,19.6746061763086)
--(axis cs:1.02,40.6587271570248);

\path [draw=black, semithick]
(axis cs:2.02,25.5185312372872)
--(axis cs:2.02,52.4814687627128);

\path [draw=black, semithick]
(axis cs:3.02,32.8306453943554)
--(axis cs:3.02,63.8360212723113);

\path [draw=black, semithick]
(axis cs:4.02,41.2328546651963)
--(axis cs:4.02,66.7671453348037);

\addplot [semithick, black, mark=-, mark size=3, mark options={solid}, only marks, forget plot]
table {%
0.02 3.21832739074979
1.02 19.6746061763086
2.02 25.5185312372872
3.02 32.8306453943554
4.02 41.2328546651963
};
\addplot [semithick, black, mark=-, mark size=3, mark options={solid}, only marks, forget plot]
table {%
0.02 21.4483392759169
1.02 40.6587271570248
2.02 52.4814687627128
3.02 63.8360212723113
4.02 66.7671453348037
};
\path [draw=black, semithick]
(axis cs:0.27,12.8775010008008)
--(axis cs:0.27,19.1224989991992);

\path [draw=black, semithick]
(axis cs:1.27,31.9466320101671)
--(axis cs:1.27,46.0533679898329);

\path [draw=black, semithick]
(axis cs:2.27,43.6252147823393)
--(axis cs:2.27,51.7081185509941);

\path [draw=black, semithick]
(axis cs:3.27,49.1444772690688)
--(axis cs:3.27,64.1888560642645);

\path [draw=black, semithick]
(axis cs:4.27,54.2211372355494)
--(axis cs:4.27,68.4455294311173);

\addplot [semithick, black, mark=-, mark size=3, mark options={solid}, only marks, forget plot]
table {%
0.27 12.8775010008008
1.27 31.9466320101671
2.27 43.6252147823393
3.27 49.1444772690688
4.27 54.2211372355494
};
\addplot [semithick, black, mark=-, mark size=3, mark options={solid}, only marks, forget plot]
table {%
0.27 19.1224989991992
1.27 46.0533679898329
2.27 51.7081185509941
3.27 64.1888560642645
4.27 68.4455294311173
};
\path [draw=black, semithick]
(axis cs:0.52,18.3203283558849)
--(axis cs:0.52,29.0130049774485);

\path [draw=black, semithick]
(axis cs:1.52,32.98612181134)
--(axis cs:1.52,51.01387818866);

\path [draw=black, semithick]
(axis cs:2.52,46.1700013344011)
--(axis cs:2.52,54.4966653322656);

\path [draw=black, semithick]
(axis cs:3.52,51.7435481186944)
--(axis cs:3.52,63.589785214639);

\path [draw=black, semithick]
(axis cs:4.52,54.6345400686719)
--(axis cs:4.52,69.3654599313281);

\addplot [semithick, black, mark=-, mark size=3, mark options={solid}, only marks, forget plot]
table {%
0.52 18.3203283558849
1.52 32.98612181134
2.52 46.1700013344011
3.52 51.7435481186944
4.52 54.6345400686719
};
\addplot [semithick, black, mark=-, mark size=3, mark options={solid}, only marks, forget plot]
table {%
0.52 29.0130049774485
1.52 51.01387818866
2.52 54.4966653322656
3.52 63.589785214639
4.52 69.3654599313281
};
\end{axis}

\end{tikzpicture}
    \end{subfigure}
    \hspace{-0.8em}
    \begin{subfigure}[t]{0.32\textwidth}
        \centering
\begin{tikzpicture}[scale=0.42]

\definecolor{crimson2143940}{RGB}{214,39,40}
\definecolor{darkgray176}{RGB}{176,176,176}
\definecolor{darkorange25512714}{RGB}{255,127,14}
\definecolor{forestgreen4416044}{RGB}{44,160,44}
\definecolor{lightgray204}{RGB}{204,204,204}

\begin{axis}[
width=12cm,
height=7cm,
legend columns=3,
legend cell align={left},
legend style={
/tikz/every even column/.append style={column sep=0.3cm},
  fill opacity=0.8,
  draw opacity=1,
  text opacity=1,
  at={(0.5,0.95)},
  anchor=north,
  draw=lightgray204
},
tick align=outside,
title={\huge\(\displaystyle e_2\) -- \Offices{} with camera occlusions},
x grid style={darkgray176},
xlabel={\huge Model},
xmin=-0.315, xmax=4.855,
xtick pos=left,
xtick style={color=black},
xtick={0.27,1.27,2.27,3.27,4.27},
xticklabels={
\huge \(\displaystyle GD\),
  \huge\(\displaystyle QD_{e}^{15}\),
  \huge\(\displaystyle QD_{e}^{25}\),
 \huge \(\displaystyle QD_{e}^{50}\),
  \huge\(\displaystyle QD_{e}^{75}\)
},
y grid style={darkgray176},
ymajorticks=false,
ymin=0, ymax=80,
ytick style={color=black}
]
\draw[draw=black,fill=forestgreen4416044,opacity=0.9,thick,postaction={pattern=north east lines, fill opacity=0.9}] (axis cs:-0.08,0) rectangle (axis cs:0.12,1.33333333333333);
\draw[draw=black,fill=forestgreen4416044,opacity=0.9,thick,postaction={pattern=north east lines, fill opacity=0.9}] (axis cs:0.92,0) rectangle (axis cs:1.12,11.3333333333333);
\draw[draw=black,fill=forestgreen4416044,opacity=0.9,thick,postaction={pattern=north east lines, fill opacity=0.9}] (axis cs:1.92,0) rectangle (axis cs:2.12,15.5);
\draw[draw=black,fill=forestgreen4416044,opacity=0.9,thick,postaction={pattern=north east lines, fill opacity=0.9}] (axis cs:2.92,0) rectangle (axis cs:3.12,19.5);
\draw[draw=black,fill=forestgreen4416044,opacity=0.9,thick,postaction={pattern=north east lines, fill opacity=0.9}] (axis cs:3.92,0) rectangle (axis cs:4.12,22.8333333333333);
\draw[draw=black,fill=forestgreen4416044,opacity=0.9,thick] (axis cs:-0.08,1.33333333333333) rectangle (axis cs:0.12,9.66666666666667);
\draw[draw=black,fill=forestgreen4416044,opacity=0.9,thick] (axis cs:0.92,11.3333333333333) rectangle (axis cs:1.12,26);
\draw[draw=black,fill=forestgreen4416044,opacity=0.9,thick] (axis cs:1.92,15.5) rectangle (axis cs:2.12,35.1666666666667);
\draw[draw=black,fill=forestgreen4416044,opacity=0.9,thick] (axis cs:2.92,19.5) rectangle (axis cs:3.12,43.1666666666667);
\draw[draw=black,fill=forestgreen4416044,opacity=0.9,thick] (axis cs:3.92,22.8333333333333) rectangle (axis cs:4.12,46.8333333333333);
\draw[draw=black,fill=darkorange25512714,opacity=0.9,thick,postaction={pattern=north east lines, fill opacity=0.9}] (axis cs:0.17,0) rectangle (axis cs:0.37,2.83333333333333);
\draw[draw=black,fill=darkorange25512714,opacity=0.9,thick,postaction={pattern=north east lines, fill opacity=0.9}] (axis cs:1.17,0) rectangle (axis cs:1.37,16.5);
\draw[draw=black,fill=darkorange25512714,opacity=0.9,thick,postaction={pattern=north east lines, fill opacity=0.9}] (axis cs:2.17,0) rectangle (axis cs:2.37,19.8333333333333);
\draw[draw=black,fill=darkorange25512714,opacity=0.9,thick,postaction={pattern=north east lines, fill opacity=0.9}] (axis cs:3.17,0) rectangle (axis cs:3.37,24.5);
\draw[draw=black,fill=darkorange25512714,opacity=0.9,thick,postaction={pattern=north east lines, fill opacity=0.9}] (axis cs:4.17,0) rectangle (axis cs:4.37,25.8333333333333);
\draw[draw=black,fill=darkorange25512714,opacity=0.9,thick] (axis cs:0.17,2.83333333333333) rectangle (axis cs:0.37,13.1666666666667);
\draw[draw=black,fill=darkorange25512714,opacity=0.9,thick] (axis cs:1.17,16.5) rectangle (axis cs:1.37,37.6666666666667);
\draw[draw=black,fill=darkorange25512714,opacity=0.9,thick] (axis cs:2.17,19.8333333333333) rectangle (axis cs:2.37,43.3333333333333);
\draw[draw=black,fill=darkorange25512714,opacity=0.9,thick] (axis cs:3.17,24.5) rectangle (axis cs:3.37,52.1666666666667);
\draw[draw=black,fill=darkorange25512714,opacity=0.9,thick] (axis cs:4.17,25.8333333333333) rectangle (axis cs:4.37,54.3333333333333);
\draw[draw=black,fill=crimson2143940,opacity=0.9,thick,postaction={pattern=north east lines, fill opacity=0.9}] (axis cs:0.42,0) rectangle (axis cs:0.62,5);
\draw[draw=black,fill=crimson2143940,opacity=0.9,thick,postaction={pattern=north east lines, fill opacity=0.9}] (axis cs:1.42,0) rectangle (axis cs:1.62,17.1666666666667);
\draw[draw=black,fill=crimson2143940,opacity=0.9,thick,postaction={pattern=north east lines, fill opacity=0.9}] (axis cs:2.42,0) rectangle (axis cs:2.62,19.5);
\draw[draw=black,fill=crimson2143940,opacity=0.9,thick,postaction={pattern=north east lines, fill opacity=0.9}] (axis cs:3.42,0) rectangle (axis cs:3.62,25.3333333333333);
\draw[draw=black,fill=crimson2143940,opacity=0.9,thick,postaction={pattern=north east lines, fill opacity=0.9}] (axis cs:4.42,0) rectangle (axis cs:4.62,26.1666666666667);
\draw[draw=black,fill=crimson2143940,opacity=0.9,thick] (axis cs:0.42,5) rectangle (axis cs:0.62,17.3333333333333);
\draw[draw=black,fill=crimson2143940,opacity=0.9,thick] (axis cs:1.42,17.1666666666667) rectangle (axis cs:1.62,37.8333333333333);
\draw[draw=black,fill=crimson2143940,opacity=0.9,thick] (axis cs:2.42,19.5) rectangle (axis cs:2.62,44.1666666666667);
\draw[draw=black,fill=crimson2143940,opacity=0.9,thick] (axis cs:3.42,25.3333333333333) rectangle (axis cs:3.62,52.6666666666667);
\draw[draw=black,fill=crimson2143940,opacity=0.9,thick] (axis cs:4.42,26.1666666666667) rectangle (axis cs:4.62,55.3333333333333);
\path [draw=black, semithick]
(axis cs:0.02,2.66071681433877)
--(axis cs:0.02,16.6726165189946);

\path [draw=black, semithick]
(axis cs:1.02,14.7305723304154)
--(axis cs:1.02,37.2694276695846);

\path [draw=black, semithick]
(axis cs:2.02,21.6543266275271)
--(axis cs:2.02,48.6790067058062);

\path [draw=black, semithick]
(axis cs:3.02,26.3638891184953)
--(axis cs:3.02,59.9694442148381);

\path [draw=black, semithick]
(axis cs:4.02,35.5602089512761)
--(axis cs:4.02,58.1064577153906);

\addplot [semithick, black, mark=-, mark size=3, mark options={solid}, only marks, forget plot]
table {%
0.02 2.66071681433877
1.02 14.7305723304154
2.02 21.6543266275271
3.02 26.3638891184953
4.02 35.5602089512761
};
\addplot [semithick, black, mark=-, mark size=3, mark options={solid}, only marks, forget plot]
table {%
0.02 16.6726165189946
1.02 37.2694276695846
2.02 48.6790067058062
3.02 59.9694442148381
4.02 58.1064577153906
};
\path [draw=black, semithick]
(axis cs:0.27,6.34146825685225)
--(axis cs:0.27,19.9918650764811);

\path [draw=black, semithick]
(axis cs:1.27,33.8807277694665)
--(axis cs:1.27,41.4526055638669);

\path [draw=black, semithick]
(axis cs:2.27,37.0021933622591)
--(axis cs:2.27,49.6644733044075);

\path [draw=black, semithick]
(axis cs:3.27,45.1607168143388)
--(axis cs:3.27,59.1726165189946);

\path [draw=black, semithick]
(axis cs:4.27,46.9115186405959)
--(axis cs:4.27,61.7551480260708);

\addplot [semithick, black, mark=-, mark size=3, mark options={solid}, only marks, forget plot]
table {%
0.27 6.34146825685225
1.27 33.8807277694665
2.27 37.0021933622591
3.27 45.1607168143388
4.27 46.9115186405959
};
\addplot [semithick, black, mark=-, mark size=3, mark options={solid}, only marks, forget plot]
table {%
0.27 19.9918650764811
1.27 41.4526055638669
2.27 49.6644733044075
3.27 59.1726165189946
4.27 61.7551480260708
};
\path [draw=black, semithick]
(axis cs:0.52,15.3126073911696)
--(axis cs:0.52,19.354059275497);

\path [draw=black, semithick]
(axis cs:1.52,30.7211372355494)
--(axis cs:1.52,44.9455294311173);

\path [draw=black, semithick]
(axis cs:2.52,38.4148553577542)
--(axis cs:2.52,49.9184779755791);

\path [draw=black, semithick]
(axis cs:3.52,47.3672956348051)
--(axis cs:3.52,57.9660376985282);

\path [draw=black, semithick]
(axis cs:4.52,47.3125270563227)
--(axis cs:4.52,63.354139610344);

\addplot [semithick, black, mark=-, mark size=3, mark options={solid}, only marks, forget plot]
table {%
0.52 15.3126073911696
1.52 30.7211372355494
2.52 38.4148553577542
3.52 47.3672956348051
4.52 47.3125270563227
};
\addplot [semithick, black, mark=-, mark size=3, mark options={solid}, only marks, forget plot]
table {%
0.52 19.354059275497
1.52 44.9455294311173
2.52 49.9184779755791
3.52 57.9660376985282
4.52 63.354139610344
};
\end{axis}

\end{tikzpicture}
    \end{subfigure}
    \\[-0.9cm]
    \caption{\RevOne{Real--world evaluation of the detectors under light variations conditions (left, middle) and with occlusions (right). GDs are based on different datasets and the mAP is averaged over the three models \revTwo{with the standard deviation}.}}
    \label{fig:ap_diff_conf}
\end{figure*}

\begin{figure*}[!htb]
    \centering
    \begin{subfigure}[t]{\textwidth}
        \centering
    \begin{tikzpicture}[scale=0.8]

\definecolor{crimson2143940}{RGB}{214,39,40}
\definecolor{darkgray176}{RGB}{176,176,176}
\definecolor{forestgreen4416044}{RGB}{44,160,44}
\definecolor{lightgray204}{RGB}{204,204,204}
\definecolor{steelblue31119180}{RGB}{31,119,180}
\definecolor{darkorange25512714}{RGB}{255,127,14}

\begin{axis}[
hide axis,
legend columns=3,
legend cell align={left},
legend style={
/tikz/every even column/.append style={column sep=0.3cm},
  fill opacity=0.8,
  draw opacity=1,
  text opacity=1,
  at={(0.5,0.95)},
  anchor=north,
  draw=lightgray204
},
xmin=-0.315, xmax=4.855,
y grid style={darkgray176},
ymin=0, ymax=130
]
\addlegendimage{forestgreen4416044,mark=*,mark size=3}
\addlegendentry{\Large$TP_{\%}$}
\addlegendimage{darkorange25512714,mark=diamond*,mark size=3}
\addlegendentry{\Large$BFD_{\%}$}
\addlegendimage{crimson2143940,mark=triangle*,mark size=3}
\addlegendentry{\Large$FP_{\%}$}
\end{axis}
\end{tikzpicture}
    \end{subfigure}
    \\[-0.8cm]
    \begin{subfigure}[t]{0.32\textwidth}
        \centering
        
\begin{tikzpicture}[scale=0.415]

\definecolor{crimson2143940}{RGB}{214,39,40}
\definecolor{darkgray176}{RGB}{176,176,176}
\definecolor{darkorange25512714}{RGB}{255,127,14}
\definecolor{forestgreen4416044}{RGB}{44,160,44}
\definecolor{gray}{RGB}{128,128,128}
\definecolor{lightgray204}{RGB}{204,204,204}

\begin{axis}[
width=12cm,
height=8cm,
legend columns=4,
legend cell align={left},
legend style={
/tikz/every even column/.append style={column sep=0.3cm},
  fill opacity=0.8,
  draw opacity=1,
  text opacity=1,
  at={(0.5,0.97)},
  anchor=north,
  draw=lightgray204
},
tick align=outside,
tick pos=left,
title={\huge\(\displaystyle e_1\) -- \Classrooms{} under light variations},
x grid style={darkgray176},
xlabel={\huge\% of qualification data},
xmin=-0.7, xmax=14.7,
xtick style={color=black},
xtick={0,1,2,3,4,5,6,7,8,9,10,11,12,13,14},
xticklabels={0,15,25,50,75,0,15,25,50,75,0,15,25,50,75},
y grid style={darkgray176},
ylabel={\huge\%},
ymin=0, ymax=95,
ytick style={color=black}
]
\path [draw=forestgreen4416044, fill=forestgreen4416044, opacity=0.2]
(axis cs:0,7.06274606680623)
--(axis cs:0,22.9372539331938)
--(axis cs:1,42.547702297363)
--(axis cs:2,52.8382338814487)
--(axis cs:3,59.6363217812696)
--(axis cs:4,62.7404383925441)
--(axis cs:4,44.5928949407892)
--(axis cs:4,44.5928949407892)
--(axis cs:3,37.6970115520638)
--(axis cs:2,35.828432785218)
--(axis cs:1,22.1189643693036)
--(axis cs:0,7.06274606680623)
--cycle;

\path [draw=crimson2143940, fill=crimson2143940, opacity=0.2]
(axis cs:0,3.2516673338672)
--(axis cs:0,7.41499933279947)
--(axis cs:1,5.97606774342517)
--(axis cs:2,6.41499933279947)
--(axis cs:3,5.73205080756888)
--(axis cs:4,6.97606774342517)
--(axis cs:4,2.35726558990816)
--(axis cs:4,2.35726558990816)
--(axis cs:3,2.26794919243112)
--(axis cs:2,2.2516673338672)
--(axis cs:1,1.35726558990816)
--(axis cs:0,3.2516673338672)
--cycle;

\path [draw=darkorange25512714, fill=darkorange25512714, opacity=0.2]
(axis cs:0,2.98248037224745)
--(axis cs:0,15.6841862944192)
--(axis cs:1,24.8171524557852)
--(axis cs:2,28.0090057210491)
--(axis cs:3,27.9163907254513)
--(axis cs:4,23.5356537528527)
--(axis cs:4,2.46434624714726)
--(axis cs:4,2.46434624714726)
--(axis cs:3,2.75027594121542)
--(axis cs:2,3.32432761228426)
--(axis cs:1,2.51618087754818)
--(axis cs:0,2.98248037224745)
--cycle;

\path [draw=forestgreen4416044, fill=forestgreen4416044, opacity=0.2]
(axis cs:5,29)
--(axis cs:5,33)
--(axis cs:6,48.9760677434252)
--(axis cs:7,54.7320508075689)
--(axis cs:8,60.605551275464)
--(axis cs:9,63.4641016151378)
--(axis cs:9,56.5358983848622)
--(axis cs:9,56.5358983848622)
--(axis cs:8,53.394448724536)
--(axis cs:7,51.2679491924311)
--(axis cs:6,44.3572655899082)
--(axis cs:5,29)
--cycle;

\path [draw=crimson2143940, fill=crimson2143940, opacity=0.2]
(axis cs:5,2.51196612828742)
--(axis cs:5,4.82136720504592)
--(axis cs:6,6.41499933279947)
--(axis cs:7,7.22008467928146)
--(axis cs:8,5.48803387171258)
--(axis cs:9,7.41499933279947)
--(axis cs:9,3.2516673338672)
--(axis cs:9,3.2516673338672)
--(axis cs:8,3.17863279495408)
--(axis cs:7,1.4465819873852)
--(axis cs:6,2.2516673338672)
--(axis cs:5,2.51196612828742)
--cycle;

\path [draw=darkorange25512714, fill=darkorange25512714, opacity=0.2]
(axis cs:5,5.88072776946648)
--(axis cs:5,13.4526055638668)
--(axis cs:6,29.5299640861417)
--(axis cs:7,31.3170634862955)
--(axis cs:8,25.3484744279055)
--(axis cs:9,23.5356537528527)
--(axis cs:9,2.46434624714726)
--(axis cs:9,2.46434624714726)
--(axis cs:8,3.31819223876113)
--(axis cs:7,4.01626984703782)
--(axis cs:6,4.47003591385833)
--(axis cs:5,5.88072776946648)
--cycle;

\path [draw=forestgreen4416044, fill=forestgreen4416044, opacity=0.2]
(axis cs:10,31.1590961193806)
--(axis cs:10,42.1742372139528)
--(axis cs:11,48.4880338717126)
--(axis cs:12,55.8499448117569)
--(axis cs:13,64.2449979983984)
--(axis cs:14,62.1961524227066)
--(axis cs:14,51.8038475772934)
--(axis cs:14,51.8038475772934)
--(axis cs:13,51.7550020016016)
--(axis cs:12,50.8167218549098)
--(axis cs:11,46.1786327949541)
--(axis cs:10,31.1590961193806)
--cycle;

\path [draw=crimson2143940, fill=crimson2143940, opacity=0.2]
(axis cs:10,3.35726558990816)
--(axis cs:10,7.97606774342517)
--(axis cs:11,6)
--(axis cs:12,6.97606774342517)
--(axis cs:13,6.64575131106459)
--(axis cs:14,6.64575131106459)
--(axis cs:14,1.35424868893541)
--(axis cs:14,1.35424868893541)
--(axis cs:13,1.35424868893541)
--(axis cs:12,2.35726558990816)
--(axis cs:11,2)
--(axis cs:10,3.35726558990816)
--cycle;

\path [draw=darkorange25512714, fill=darkorange25512714, opacity=0.2]
(axis cs:10,6.80720138958435)
--(axis cs:10,18.526131943749)
--(axis cs:11,28.5325625946708)
--(axis cs:12,30.7671453348037)
--(axis cs:13,30.114877048604)
--(axis cs:14,20.5440037453175)
--(axis cs:14,3.45599625468247)
--(axis cs:14,3.45599625468247)
--(axis cs:13,3.885122951396)
--(axis cs:12,5.2328546651963)
--(axis cs:11,5.4674374053292)
--(axis cs:10,6.80720138958435)
--cycle;

\addplot [semithick, forestgreen4416044, mark=*, mark size=2.5, mark options={solid}, forget plot]
table {%
0 15
1 32.3333333333333
2 44.3333333333333
3 48.6666666666667
4 53.6666666666667
};
\addplot [semithick, crimson2143940, mark=triangle*, mark size=2.5, mark options={solid}, forget plot]
table {%
0 5.33333333333333
1 3.66666666666667
2 4.33333333333333
3 4
4 4.66666666666667
};
\addplot [semithick, darkorange25512714, mark=diamond*, mark size=2.5, mark options={solid}, forget plot]
table {%
0 9.33333333333333
1 13.6666666666667
2 15.6666666666667
3 15.3333333333333
4 13
};
\addplot [semithick, forestgreen4416044, mark=*, mark size=2.5, mark options={solid}, forget plot]
table {%
5 31
6 46.6666666666667
7 53
8 57
9 60
};
\addplot [semithick, crimson2143940, mark=triangle*, mark size=2.5, mark options={solid}, forget plot]
table {%
5 3.66666666666667
6 4.33333333333333
7 4.33333333333333
8 4.33333333333333
9 5.33333333333333
};
\addplot [semithick, darkorange25512714, mark=diamond*, mark size=2.5, mark options={solid}, forget plot]
table {%
5 9.66666666666667
6 17
7 17.6666666666667
8 14.3333333333333
9 13
};
\addplot [semithick, forestgreen4416044, mark=*, mark size=2.5, mark options={solid}, forget plot]
table {%
10 36.6666666666667
11 47.3333333333333
12 53.3333333333333
13 58
14 57
};
\addplot [semithick, crimson2143940, mark=triangle*, mark size=2.5, mark options={solid}, forget plot]
table {%
10 5.66666666666667
11 4
12 4.66666666666667
13 4
14 4
};
\addplot [semithick, darkorange25512714, mark=diamond*, mark size=2.5, mark options={solid}, forget plot]
table {%
10 12.6666666666667
11 17
12 18
13 17
14 12
};
\addplot [black, forget plot]
table {%
-0.7 1.77635683940025e-15
14.7 1.77635683940025e-15
};
\path [draw=gray, draw opacity=0.7, semithick, dash pattern=on 5.55pt off 2.4pt]
(axis cs:4.5,0)
--(axis cs:4.5,115);

\path [draw=gray, draw opacity=0.7, semithick, dash pattern=on 5.55pt off 2.4pt]
(axis cs:9.5,0)
--(axis cs:9.5,115);

\draw (axis cs:2,87) node[
  text=black,
  rotate=0.0
]{\huge\DDDtwo};
\draw (axis cs:7,87) node[
  text=black,
  rotate=0.0
]{\huge\DG};
\draw (axis cs:12,87) node[
  text=black,
  rotate=0.0
]{\huge\DDDtwoG};
\end{axis}

\end{tikzpicture}
    \end{subfigure}
    \hspace{0.6em}
    \begin{subfigure}[t]{0.32\linewidth}
        \centering
       
\begin{tikzpicture}[scale=0.42]

\definecolor{crimson2143940}{RGB}{214,39,40}
\definecolor{darkgray176}{RGB}{176,176,176}
\definecolor{darkorange25512714}{RGB}{255,127,14}
\definecolor{forestgreen4416044}{RGB}{44,160,44}
\definecolor{gray}{RGB}{128,128,128}
\definecolor{lightgray204}{RGB}{204,204,204}

\begin{axis}[
width=12cm,
height=8cm,
legend columns=4,
legend cell align={left},
legend style={
/tikz/every even column/.append style={column sep=0.3cm},
  fill opacity=0.8,
  draw opacity=1,
  text opacity=1,
  at={(0.5,0.97)},
  anchor=north,
  draw=lightgray204
},
tick align=outside,
title={\huge\(\displaystyle e_2\) -- \Offices{} under light variations},
x grid style={darkgray176},
xlabel={\huge\% of qualification data},
xmin=-0.7, xmax=14.7,
xtick pos=left,
xtick style={color=black},
xtick={0,1,2,3,4,5,6,7,8,9,10,11,12,13,14},
xticklabels={0,15,25,50,75,0,15,25,50,75,0,15,25,50,75},
y grid style={darkgray176},
ymajorticks=false,
ymin=0, ymax=95,
ytick style={color=black}
]
\path [draw=forestgreen4416044, fill=forestgreen4416044, opacity=0.2]
(axis cs:0,4.01626984703782)
--(axis cs:0,31.3170634862955)
--(axis cs:1,55.1691917259821)
--(axis cs:2,62.377712128529)
--(axis cs:3,67.1915343479906)
--(axis cs:4,69.7849645858386)
--(axis cs:4,50.8817020808281)
--(axis cs:4,50.8817020808281)
--(axis cs:3,41.4751323186761)
--(axis cs:2,32.2889545381377)
--(axis cs:1,22.1641416073512)
--(axis cs:0,4.01626984703782)
--cycle;

\path [draw=crimson2143940, fill=crimson2143940, opacity=0.2]
(axis cs:0,0.805808101681387)
--(axis cs:0,3.86085856498528)
--(axis cs:1,4.19419189831861)
--(axis cs:2,5)
--(axis cs:3,4)
--(axis cs:4,4)
--(axis cs:4,2)
--(axis cs:4,2)
--(axis cs:3,2)
--(axis cs:2,3)
--(axis cs:1,1.13914143501472)
--(axis cs:0,0.805808101681387)
--cycle;

\path [draw=darkorange25512714, fill=darkorange25512714, opacity=0.2]
(axis cs:0,1.25042956467857)
--(axis cs:0,17.4162371019881)
--(axis cs:1,34.8633741510085)
--(axis cs:2,43.0787840283389)
--(axis cs:3,33.1769627188082)
--(axis cs:4,27.3472141939601)
--(axis cs:4,3.3194524727066)
--(axis cs:4,3.3194524727066)
--(axis cs:3,3.48970394785845)
--(axis cs:2,4.92121597166109)
--(axis cs:1,2.46995918232488)
--(axis cs:0,1.25042956467857)
--cycle;

\path [draw=forestgreen4416044, fill=forestgreen4416044, opacity=0.2]
(axis cs:5,22.6411010564593)
--(axis cs:5,31.3588989435407)
--(axis cs:6,60.3578166916005)
--(axis cs:7,61)
--(axis cs:8,68.5825756949558)
--(axis cs:9,71.6998896235138)
--(axis cs:9,61.6334437098195)
--(axis cs:9,61.6334437098195)
--(axis cs:8,59.4174243050442)
--(axis cs:7,51)
--(axis cs:6,37.6421833083995)
--(axis cs:5,22.6411010564593)
--cycle;

\path [draw=crimson2143940, fill=crimson2143940, opacity=0.2]
(axis cs:5,4)
--(axis cs:5,8)
--(axis cs:6,4.86085856498528)
--(axis cs:7,6)
--(axis cs:8,4)
--(axis cs:9,3.91068360252296)
--(axis cs:9,2.75598306414371)
--(axis cs:9,2.75598306414371)
--(axis cs:8,2)
--(axis cs:7,2)
--(axis cs:6,1.80580810168139)
--(axis cs:5,4)
--cycle;

\path [draw=darkorange25512714, fill=darkorange25512714, opacity=0.2]
(axis cs:5,6.80090094203417)
--(axis cs:5,26.5324323912992)
--(axis cs:6,48.2837966537928)
--(axis cs:7,37.6144982852076)
--(axis cs:8,26.8803387171259)
--(axis cs:9,25.0749966639973)
--(axis cs:9,4.258336669336)
--(axis cs:9,4.258336669336)
--(axis cs:8,3.78632794954082)
--(axis cs:7,5.71883504812575)
--(axis cs:6,5.71620334620724)
--(axis cs:5,6.80090094203417)
--cycle;

\path [draw=forestgreen4416044, fill=forestgreen4416044, opacity=0.2]
(axis cs:10,27.6750052148539)
--(axis cs:10,40.9916614518127)
--(axis cs:11,64.3461475306288)
--(axis cs:12,65)
--(axis cs:13,69.0195740364107)
--(axis cs:14,74)
--(axis cs:14,62)
--(axis cs:14,62)
--(axis cs:13,57.647092630256)
--(axis cs:12,53)
--(axis cs:11,38.3205191360379)
--(axis cs:10,27.6750052148539)
--cycle;

\path [draw=crimson2143940, fill=crimson2143940, opacity=0.2]
(axis cs:10,5)
--(axis cs:10,5)
--(axis cs:11,4)
--(axis cs:12,3.91068360252296)
--(axis cs:13,3)
--(axis cs:14,5.19419189831861)
--(axis cs:14,2.13914143501472)
--(axis cs:14,2.13914143501472)
--(axis cs:13,3)
--(axis cs:12,2.75598306414371)
--(axis cs:11,2)
--(axis cs:10,5)
--cycle;

\path [draw=darkorange25512714, fill=darkorange25512714, opacity=0.2]
(axis cs:10,9.45016556472925)
--(axis cs:10,24.5498344352707)
--(axis cs:11,44.2873015219859)
--(axis cs:12,37.5884572681199)
--(axis cs:13,27.0749966639973)
--(axis cs:14,27.8803387171258)
--(axis cs:14,4.78632794954082)
--(axis cs:14,4.78632794954082)
--(axis cs:13,6.258336669336)
--(axis cs:12,6.4115427318801)
--(axis cs:11,5.71269847801409)
--(axis cs:10,9.45016556472925)
--cycle;

\addplot [semithick, forestgreen4416044, mark=*, mark size=2.5, mark options={solid}, forget plot]
table {%
0 17.6666666666667
1 38.6666666666667
2 47.3333333333333
3 54.3333333333333
4 60.3333333333333
};
\addplot [semithick, crimson2143940, mark=triangle*, mark size=2.5, mark options={solid}, forget plot]
table {%
0 2.33333333333333
1 2.66666666666667
2 4
3 3
4 3
};
\addplot [semithick, darkorange25512714, mark=diamond*, mark size=2.5, mark options={solid}, forget plot]
table {%
0 9.33333333333333
1 18.6666666666667
2 24
3 18.3333333333333
4 15.3333333333333
};
\addplot [semithick, forestgreen4416044, mark=*, mark size=2.5, mark options={solid}, forget plot]
table {%
5 27
6 49
7 56
8 64
9 66.6666666666667
};
\addplot [semithick, crimson2143940, mark=triangle*, mark size=2.5, mark options={solid}, forget plot]
table {%
5 6
6 3.33333333333333
7 4
8 3
9 3.33333333333333
};
\addplot [semithick, darkorange25512714, mark=diamond*, mark size=2.5, mark options={solid}, forget plot]
table {%
5 16.6666666666667
6 27
7 21.6666666666667
8 15.3333333333333
9 14.6666666666667
};
\addplot [semithick, forestgreen4416044, mark=*, mark size=2.5, mark options={solid}, forget plot]
table {%
10 34.3333333333333
11 51.3333333333333
12 59
13 63.3333333333333
14 68
};
\addplot [semithick, crimson2143940, mark=triangle*, mark size=2.5, mark options={solid}, forget plot]
table {%
10 5
11 3
12 3.33333333333333
13 3
14 3.66666666666667
};
\addplot [semithick, darkorange25512714, mark=diamond*, mark size=2.5, mark options={solid}, forget plot]
table {%
10 17
11 25
12 22
13 16.6666666666667
14 16.3333333333333
};
\addplot [black, forget plot]
table {%
-0.7 1.77635683940025e-15
14.7 1.77635683940025e-15
};
\path [draw=gray, draw opacity=0.7, semithick, dash pattern=on 5.55pt off 2.4pt]
(axis cs:4.5,0)
--(axis cs:4.5,115);

\path [draw=gray, draw opacity=0.7, semithick, dash pattern=on 5.55pt off 2.4pt]
(axis cs:9.5,0)
--(axis cs:9.5,115);

\draw (axis cs:2,87) node[
  text=black,
  rotate=0.0
]{\huge\DDDtwo};
\draw (axis cs:7,87) node[
  text=black,
  rotate=0.0
]{\huge\DG};
\draw (axis cs:12,87) node[
  text=black,
  rotate=0.0
]{\huge\DDDtwoG};
\end{axis}

\end{tikzpicture}
    \end{subfigure}
    \hspace{-0.8em}
    \begin{subfigure}[t]{0.32\linewidth}
        \centering
       
\begin{tikzpicture}[scale=0.42]

\definecolor{crimson2143940}{RGB}{214,39,40}
\definecolor{darkgray176}{RGB}{176,176,176}
\definecolor{darkorange25512714}{RGB}{255,127,14}
\definecolor{forestgreen4416044}{RGB}{44,160,44}
\definecolor{gray}{RGB}{128,128,128}
\definecolor{lightgray204}{RGB}{204,204,204}

\begin{axis}[
width=12cm,
height=8cm,
legend columns=4,
legend cell align={left},
legend style={
/tikz/every even column/.append style={column sep=0.3cm},
  fill opacity=0.8,
  draw opacity=1,
  text opacity=1,
  at={(0.5,0.97)},
  anchor=north,
  draw=lightgray204
},
tick align=outside,
title={\huge\(\displaystyle e_2\) -- \Offices{} with camera occlusions},
x grid style={darkgray176},
xlabel={\huge\% of qualification data},
xmin=-0.7, xmax=14.7,
xtick pos=left,
xtick style={color=black},
xtick={0,1,2,3,4,5,6,7,8,9,10,11,12,13,14},
xticklabels={0,15,25,50,75,0,15,25,50,75,0,15,25,50,75},
y grid style={darkgray176},
ymajorticks=false,
ymin=0, ymax=95,
ytick style={color=black}
]
\path [draw=forestgreen4416044, fill=forestgreen4416044, opacity=0.2]
(axis cs:0,5.45599625468247)
--(axis cs:0,22.5440037453175)
--(axis cs:1,48.5102960521415)
--(axis cs:2,56.0954913856558)
--(axis cs:3,64.767090063074)
--(axis cs:4,62.7198304169394)
--(axis cs:4,45.9468362497273)
--(axis cs:4,45.9468362497273)
--(axis cs:3,35.8995766035927)
--(axis cs:2,29.2378419476775)
--(axis cs:1,18.8230372811918)
--(axis cs:0,5.45599625468247)
--cycle;

\path [draw=crimson2143940, fill=crimson2143940, opacity=0.2]
(axis cs:0,1.13914143501472)
--(axis cs:0,4.19419189831861)
--(axis cs:1,6.41499933279947)
--(axis cs:2,8.72171712997056)
--(axis cs:3,5.82136720504592)
--(axis cs:4,7)
--(axis cs:4,3)
--(axis cs:4,3)
--(axis cs:3,3.51196612828742)
--(axis cs:2,2.61161620336277)
--(axis cs:1,2.2516673338672)
--(axis cs:0,1.13914143501472)
--cycle;

\path [draw=darkorange25512714, fill=darkorange25512714, opacity=0.2]
(axis cs:0,3.55969349108945)
--(axis cs:0,24.4403065089106)
--(axis cs:1,41.313207915828)
--(axis cs:2,49.3636155224777)
--(axis cs:3,38.0880074906351)
--(axis cs:4,40.5202591774521)
--(axis cs:4,3.47974082254786)
--(axis cs:4,3.47974082254786)
--(axis cs:3,3.91199250936494)
--(axis cs:2,5.30305114418892)
--(axis cs:1,2.68679208417203)
--(axis cs:0,3.55969349108945)
--cycle;

\path [draw=forestgreen4416044, fill=forestgreen4416044, opacity=0.2]
(axis cs:5,10.5658798681793)
--(axis cs:5,26.1007867984874)
--(axis cs:6,51.526131943749)
--(axis cs:7,56.6943804400084)
--(axis cs:8,64.0827625302982)
--(axis cs:9,67.9052111277337)
--(axis cs:9,52.761455538933)
--(axis cs:9,52.761455538933)
--(axis cs:8,51.9172374697018)
--(axis cs:7,44.638952893325)
--(axis cs:6,39.8072013895843)
--(axis cs:5,10.5658798681793)
--cycle;

\path [draw=crimson2143940, fill=crimson2143940, opacity=0.2]
(axis cs:5,6.08931639747704)
--(axis cs:5,7.24401693585629)
--(axis cs:6,6.86085856498528)
--(axis cs:7,6.48803387171258)
--(axis cs:8,6.41499933279947)
--(axis cs:9,8.18327814509025)
--(axis cs:9,3.15005518824308)
--(axis cs:9,3.15005518824308)
--(axis cs:8,2.2516673338672)
--(axis cs:7,4.17863279495408)
--(axis cs:6,3.80580810168139)
--(axis cs:5,6.08931639747704)
--cycle;

\path [draw=darkorange25512714, fill=darkorange25512714, opacity=0.2]
(axis cs:5,4.3750934234891)
--(axis cs:5,22.9582399098442)
--(axis cs:6,51.9165254233231)
--(axis cs:7,40.3493515728975)
--(axis cs:8,34.422205101856)
--(axis cs:9,34.5577255289012)
--(axis cs:9,6.10894113776542)
--(axis cs:9,6.10894113776542)
--(axis cs:8,5.57779489814404)
--(axis cs:7,5.65064842710253)
--(axis cs:6,2.75014124334361)
--(axis cs:5,4.3750934234891)
--cycle;

\path [draw=forestgreen4416044, fill=forestgreen4416044, opacity=0.2]
(axis cs:10,20.1500551882431)
--(axis cs:10,25.1832781450903)
--(axis cs:11,56)
--(axis cs:12,57.352907369744)
--(axis cs:13,62.3747852176607)
--(axis cs:14,68.9916614518127)
--(axis cs:14,55.6750052148539)
--(axis cs:14,55.6750052148539)
--(axis cs:13,54.291881449006)
--(axis cs:12,45.9804259635893)
--(axis cs:11,38)
--(axis cs:10,20.1500551882431)
--cycle;

\path [draw=crimson2143940, fill=crimson2143940, opacity=0.2]
(axis cs:10,6.17863279495408)
--(axis cs:10,8.48803387171258)
--(axis cs:11,5.82136720504592)
--(axis cs:12,6.73205080756888)
--(axis cs:13,5.48803387171258)
--(axis cs:14,7.64575131106459)
--(axis cs:14,2.35424868893541)
--(axis cs:14,2.35424868893541)
--(axis cs:13,3.17863279495408)
--(axis cs:12,3.26794919243112)
--(axis cs:11,3.51196612828742)
--(axis cs:10,6.17863279495408)
--cycle;

\path [draw=darkorange25512714, fill=darkorange25512714, opacity=0.2]
(axis cs:10,7.04176009015576)
--(axis cs:10,25.6249065765109)
--(axis cs:11,49.6952932933495)
--(axis cs:12,40.3199946623957)
--(axis cs:13,33.9102745696109)
--(axis cs:14,41.6759235456535)
--(axis cs:14,5.65740978767987)
--(axis cs:14,5.65740978767987)
--(axis cs:13,6.7563920970558)
--(axis cs:12,7.0133386709376)
--(axis cs:11,6.97137337331718)
--(axis cs:10,7.04176009015576)
--cycle;

\addplot [semithick, forestgreen4416044, mark=*, mark size=2.5, mark options={solid}, forget plot]
table {%
0 14
1 33.6666666666667
2 42.6666666666667
3 50.3333333333333
4 54.3333333333333
};
\addplot [semithick, crimson2143940, mark=triangle*, mark size=2.5, mark options={solid}, forget plot]
table {%
0 2.66666666666667
1 4.33333333333333
2 5.66666666666667
3 4.66666666666667
4 5
};
\addplot [semithick, darkorange25512714, mark=diamond*, mark size=2.5, mark options={solid}, forget plot]
table {%
0 14
1 22
2 27.3333333333333
3 21
4 22
};
\addplot [semithick, forestgreen4416044, mark=*, mark size=2.5, mark options={solid}, forget plot]
table {%
5 18.3333333333333
6 45.6666666666667
7 50.6666666666667
8 58
9 60.3333333333333
};
\addplot [semithick, crimson2143940, mark=triangle*, mark size=2.5, mark options={solid}, forget plot]
table {%
5 6.66666666666667
6 5.33333333333333
7 5.33333333333333
8 4.33333333333333
9 5.66666666666667
};
\addplot [semithick, darkorange25512714, mark=diamond*, mark size=2.5, mark options={solid}, forget plot]
table {%
5 13.6666666666667
6 27.3333333333333
7 23
8 20
9 20.3333333333333
};
\addplot [semithick, forestgreen4416044, mark=*, mark size=2.5, mark options={solid}, forget plot]
table {%
10 22.6666666666667
11 47
12 51.6666666666667
13 58.3333333333333
14 62.3333333333333
};
\addplot [semithick, crimson2143940, mark=triangle*, mark size=2.5, mark options={solid}, forget plot]
table {%
10 7.33333333333333
11 4.66666666666667
12 5
13 4.33333333333333
14 5
};
\addplot [semithick, darkorange25512714, mark=diamond*, mark size=2.5, mark options={solid}, forget plot]
table {%
10 16.3333333333333
11 28.3333333333333
12 23.6666666666667
13 20.3333333333333
14 23.6666666666667
};
\addplot [black, forget plot]
table {%
-0.7 1.77635683940025e-15
14.7 1.77635683940025e-15
};
\path [draw=gray, draw opacity=0.7, semithick, dash pattern=on 5.55pt off 2.4pt]
(axis cs:4.5,0)
--(axis cs:4.5,115);

\path [draw=gray, draw opacity=0.7, semithick, dash pattern=on 5.55pt off 2.4pt]
(axis cs:9.5,0)
--(axis cs:9.5,115);

\draw (axis cs:2,87) node[
  text=black,
  rotate=0.0
]{\huge\DDDtwo};
\draw (axis cs:7,87) node[
  text=black,
  rotate=0.0
]{\huge\DG};
\draw (axis cs:12,87) node[
  text=black,
  rotate=0.0
]{\huge\DDDtwoG};
\end{axis}

\end{tikzpicture}
    \end{subfigure}
    \\[-0.9cm]
    \caption{\RevOne{Operational performance indicators under light variations (left, middle) and with camera occlusions (right) averaged over the three models \revTwo{(with standard deviations)} with GDs trained on different
datasets.}}
    \label{fig:extended_metric_diff_cond}
\end{figure*}

\RevOne{
\begin{figure*}[!htb]
	\centering
    \begin{tabular}{@{}c@{ }c@{ }c@{ }c@{ }}
    & \multicolumn{2}{c}{\small Day/night light differences}&\small Camera occlusions \\
    & $e_1$ -- \Classrooms & $e_2$ -- \Offices& $e_2$ -- \Offices\\
    \rotatebox[origin=c]{90}{\small DETR~\cite{detr}}&
    \raisebox{-0.45\height}{\includegraphics[width=0.23\linewidth]{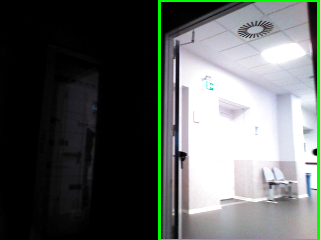}}&
    \raisebox{-0.45\height}{\includegraphics[width=0.23\linewidth]{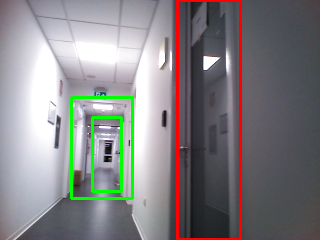}}&
    \raisebox{-0.45\height}{\includegraphics[width=0.23\linewidth]{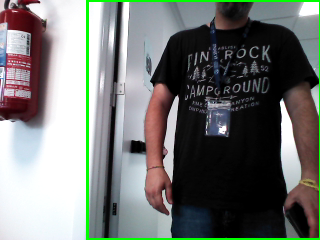}}
    \\\addlinespace[0.13cm]
    \rotatebox[origin=c]{90}{\small YOLOv5~\cite{yolov5}}&
    \raisebox{-0.45\height}{\includegraphics[width=0.23\linewidth]{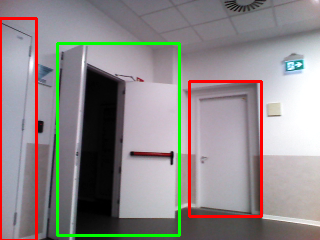}}&
     \raisebox{-0.45\height}{\includegraphics[width=0.23\linewidth]{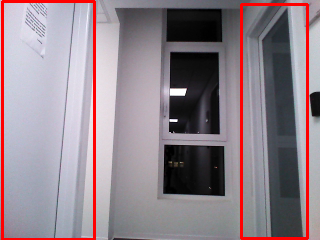}}&
    \raisebox{-0.45\height}{\includegraphics[width=0.23\linewidth]{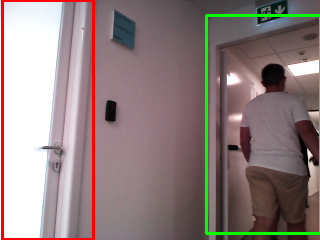}}
    \\\addlinespace[0.11cm]
    \rotatebox[origin=c]{90}{\small Faster R--CNN~\cite{fasterrcnn}}&
    \raisebox{-0.45\height}{\includegraphics[width=0.23\linewidth]{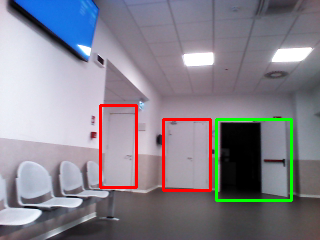}}&
     \raisebox{-0.45\height}{\includegraphics[width=0.23\linewidth]{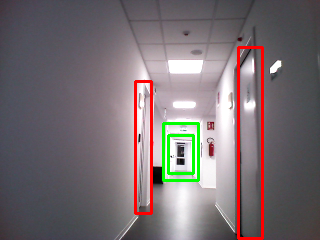}}&
    \raisebox{-0.45\height}{\includegraphics[width=0.23\linewidth]{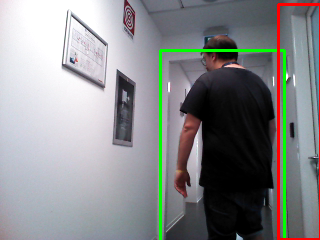}}
    \\\addlinespace[0.11cm]
\end{tabular}
	\caption{\RevOne{Challenging doors detected by $QD^{15}_e$ (based on our dataset \DG) under light changes (first and second column) and camera occlusions (third column).}}
	\label{fig:qd15_examples_diff_cond}
\end{figure*}}

\RevOne{Figure~\ref{fig:ap_diff_conf} shows how the mAP performance of the GDs are similar to those of Figure~\ref{fig:ap_qualification}, indicating that the GDs are robust to illumination changes and camera occlusions. As reported in Section~\ref{sec:Results:GD}, the general detectors based on \DDDtwo{} exhibit the worst performance while those trained with our simulated dataset \DG{} perform remarkably better, especially in $e_1$ during nighttime (see also the $TP_{\%}$ in Figure~\ref{fig:extended_metric_diff_cond}). The \DDDtwoG--based GDs, despite improving the average mAP as shown in Figure~\ref{fig:ap_gd}, increase also the performance gap between the models (see the standard deviation of the orange and red bars in Figure~\ref{fig:ap_diff_conf}). More interestingly, it can be seen in Table~\ref{tab:ap_extended_metric_results_diff_cond} how the improvement provided by the qualified detectors is maintained also in the (challenging) long--term deployment conditions of light variations and camera occlusions (Figure~\ref{fig:qd15_examples_diff_cond} reports some representative detections of $QD_e^{15}$). 
Despite this, the qualification procedure we propose enables $QD^{15}_e$ to perform door detection also in (very) challenging situations where doors are almost entirely occluded (see the first and third examples in the last column of Figure~\ref{fig:qd15_examples_diff_cond}). The performance decrease observable comparing Tables~\ref{tab:ap_extended_metric_results_diff_cond}~and~\ref{tab:qualifications_results} is a direct consequence of the fine--tune, which produces QDs that slightly overfit the conditions (different light and no limited occlusions) seen during the robot's deployment. Despite this, our method ensures a performance improvement to the GDs when used in long--term scenarios with illumination changes and dynamic obstacles hiding doors' portions, enabling the QDs to still solve challenging examples, as shown in Figure~\ref{fig:qd15_examples_diff_cond}. Once again, $QD^{15}_{e}$, albeit using a few examples for fine--tuning, ensures the best performance improvement also in challenging long--term deployment conditions.

As mentioned before, the qualified detector $QD$ can detect doors from challenging points of view, thus improving its performance in a target environment; the same does not hold for $GD$. 
To further prove the qualification's benefits, we test the performance of the detectors of Section~\ref{sec:Results:QD}, qualified with the data from the robot's initial deployment, on a new run of the robot obtained a year later and containing challenging images of doors. More precisely, we performed an additional acquisition campaign targeted at capturing only door instances from difficult viewpoints, which the robot will encounter in long--term deployments: data are acquired when the robot is navigating through the main corridor of $e_2$ -- \Offices{} (see Figure~\ref{fig:P4} for the floorplan). In such a corridor, detecting doors is particularly challenging: there are multiple doors, often far away from the robot, and perpendicular to the robot's motion. 
Note that, in this data, the doors are perceived by the robot in the same environmental settings (daylight and no dynamic obstacles) as those encountered during the initial deployment (whose data are used to train the $QD$); still, the status of some doors is different (doors that were open/closed may be closed/open). In this way, we can observe if the qualification procedure overfits to the initial training data (e.g., if the model is biased to detect a door as open/closed because in the dataset used to train the $QD$ such a door is open/closed). Examples of these changes can be seen in the first two columns of Figure~\ref{fig:examples_run2}.

The results of this experiment are in Table~\ref{tab:ap_extended_metric_run2}. It can be seen how the GDs work fairly well also on this challenging run, with performance close to that of the GD on the less challenging dataset of Table~\ref{tab:qualifications_results}. Also for this experiment, the use of our dataset \DG{} to train the GD improves its performance (mAP and $TP_{\%}$), when compared with the GD trained on \DDDtwo. Again, \DDDtwoG{} increases the detection accuracy in terms of mAP and $TP_{\%}$ of the GDs also reducing the discrepancy between the tested models (low $\sigma$ for all metrics).
More interestingly, the qualified detectors, fine--tuned with the data acquired during the robot's initial deployment (those tested in Section~\ref{sec:Results:QD}), have remarkably higher performance when tested on new, challenging, examples (see the first part of Table~\ref{tab:ap_extended_metric_run2}). (Note that the performance decrease, when compared against Table~\ref{tab:qualifications_results}, is because door images are taken from challenging points of view, while in the initial dataset many images of doors have a clear, frontal view of the target.) 
The fact that the performance of Table~\ref{tab:ap_extended_metric_results_diff_cond} is close to that of Table~\ref{tab:ap_extended_metric_run2} shows how the qualification procedure is robust against overfitting on data used for the qualification procedure; the fact that doors are observed in a given status (open/closed) during the initial deployment does not cause a drop in performance when the same door is observed, later on, with a different status (closed/open).
Once again, the $QD_{e}^{15}$ ensures the best performance improvements when compared to the subsequent qualification rounds. This is corroborated by the challenging detections reported in Figure~\ref{fig:examples_run2} (third column), where YOLOv5, based on \DG{} and qualified with the 15\% of the data collected during the robot deployment, successfully identifies challenging door instances when viewed from narrow side angles, at large distance from the camera, and with different statuses.

A robot deployed in the long term is constantly acquiring new data from its environment; some of them can be potentially used, after a labeling step, to perform further qualification runs on the QD. We have thus performed preliminary tests to evaluate the impact of this procedure. 
To do so, we compared a QD as trained in Section~\ref{sec:ExperimentalSetting:Training}, using \emph{deployment} data, with a QD that has been trained with data acquired during the initial deployment and with additional data acquired in different environmental condition (i.e., during nighttime). The former is indicated as \emph{Deployment}, the latter as \emph{Deployment + nighttime} in  Table~\ref{tab:ap_extended_metric_run2}. Note that the Deployment and the Deployment + nighttime datasets have different sizes (the latter includes the former).
The results reported in Table~\ref{tab:ap_extended_metric_run2} show that using more data for qualification enables the QDs to better identify doors from challenging perspectives and, importantly, this happens even when mixing images with a feature shift (i.e., different light conditions). Even in this case, the QD trained on our dataset \DG{} have better performances than those \DDDtwo{} (\DDDtwoG) in terms of mAP, $TP_{\%}$ and $BFD_{\%}$. In particular, $QD_{e}^{15}$ benefits more from using more data for the qualification, reaching performance close to the one reported in Table~\ref{tab:qualifications_results}. Some improved detections can be seen in the last column of Figure~\ref{fig:examples_run2}, where YOLOv5 based on \DG{} and qualified with more data manages in detecting two very challenging closed doors (the second one with changed status) on the left (first row) and the right (second row) of the corridor.

\begin{table*}[h]

\setlength\tabcolsep{1.9pt}
\setlength\extrarowheight{2pt}
\centering

\begin{tabular}{cc|cccc|cccc|cccc}
\toprule
        &  & \multicolumn{4}{c|}{\DDDtwo} & \multicolumn{4}{c|}{\DG} &\multicolumn{4}{c}{\DDDtwoG} \\
        \textbf{Exp.} & \textbf{Qual.} & \textbf{mAP}$\uparrow$& $\mathbf{TP}_{\%}$$\uparrow$ &  $\mathbf{FP}_{\%}$$\downarrow$ & $\mathbf{BFD}_{\%}$$\downarrow$ & \textbf{mAP}$\uparrow$& $\mathbf{TP}_{\%}$$\uparrow$ &  $\mathbf{FP}_{\%}$$\downarrow$ & $\mathbf{BFD}_{\%}$$\downarrow$ & \textbf{mAP}$\uparrow$& $\mathbf{TP}_{\%}$$\uparrow$ &  $\mathbf{FP}_{\%}$$\downarrow$ & $\mathbf{BFD}_{\%}$$\downarrow$ \\
\midrule

$GD$ &--&$14 \pm 11 $&$15 \pm 11 $&$\textbf{1} \pm 1 $&$\textbf{14} \pm 10 $&$16 \pm 7 $&$19 \pm 8 $&$6 \pm 1 $&$14 \pm 11 $&$\textbf{20} \pm 5 $&$\textbf{24} \pm 5 $&$5 \pm 1 $&$20 \pm 8 $\\[2pt]\hline
$QD_{e}^{15}$& \multirow{4}{*}[-1.3ex]{\rotatebox[origin=c]{90}{Deployment}} &$37 \pm 16 $&$44 \pm 18 $&$\textbf{3} \pm 2 $&$\textbf{26} \pm 24 $&$\textbf{48} \pm 10 $&$\textbf{56} \pm 10 $&$5 \pm 1 $&$31 \pm 28 $&$46 \pm 10 $&$55 \pm 13 $&$5 \pm 1 $&$33 \pm 24 $\\[2pt]
$QD_{e}^{25}$&&$45 \pm 14 $&$53 \pm 12 $&$\textbf{5} \pm 3 $&$33 \pm 30 $&$53 \pm 8 $&$60 \pm 8 $&$5 \pm 2 $&$\textbf{25} \pm 20 $&$\textbf{54} \pm 9 $&$\textbf{63} \pm 8 $&$6 \pm 1 $&$30 \pm 22 $\\[2pt]
$QD_{e}^{50}$&&$55 \pm 17 $&$62 \pm 14 $&$\textbf{4} \pm 2 $&$26 \pm 25 $&$\textbf{63} \pm 10 $&$\textbf{69} \pm 9 $&$4 \pm 2 $&$\textbf{21} \pm 18 $&$62 \pm 9 $&$69 \pm 8 $&$5 \pm 1 $&$26 \pm 19 $\\[2pt]
$QD_{e}^{75}$&&$59 \pm 15 $&$67 \pm 13 $&$\textbf{4} \pm 1 $&$25 \pm 24 $&$65 \pm 11 $&$71 \pm 11 $&$5 \pm 1 $&$\textbf{23} \pm 18 $&$\textbf{65} \pm 11 $&$\textbf{72} \pm 9 $&$6 \pm 1 $&$26 \pm 19 $\\[2pt]
\hline

$QD_{e}^{15}$&\multirow{4}{*}[-1.3ex]{\rotatebox[origin=c]{90}{\makecell{Deployment\\+\\nighttime}}}&$51 \pm 16 $&$59 \pm 14 $&$\textbf{3} \pm 1 $&$35 \pm 35 $&$\textbf{60} \pm 11 $&$\textbf{68} \pm 9 $&$4 \pm 1 $&$\textbf{29} \pm 21 $&$60 \pm 11 $&$67 \pm 9 $&$4 \pm 1 $&$30 \pm 23 $\\[2pt]
$QD_{e}^{25}$&&$54 \pm 19 $&$63 \pm 17 $&$\textbf{3} \pm 2 $&$37 \pm 31 $&$\textbf{62} \pm 10 $&$\textbf{70} \pm 7 $&$5 \pm 1 $&$\textbf{30} \pm 24 $&$61 \pm 13 $&$69 \pm 10 $&$5 \pm 1 $&$31 \pm 22 $\\[2pt]
$QD_{e}^{50}$&&$64 \pm 13 $&$71 \pm 12 $&$\textbf{4} \pm 3 $&$30 \pm 24 $&$\textbf{69} \pm 9 $&$\textbf{74} \pm 8 $&$6 \pm 1 $&$\textbf{24} \pm 16 $&$66 \pm 13 $&$73 \pm 10 $&$5 \pm 1 $&$25 \pm 20 $\\[2pt]
$QD_{e}^{75}$&&$65 \pm 13 $&$72 \pm 10 $&$\textbf{4} \pm 2 $&$32 \pm 26 $&$\textbf{70} \pm 9 $&$\textbf{76} \pm 8 $&$5 \pm 1 $&$\textbf{24} \pm 19 $&$68 \pm 12 $&$74 \pm 9 $&$5 \pm 1 $&$25 \pm 20 $\\[2pt]

\bottomrule
\end{tabular}

\caption{\RevOne{Performance of the QD's in $e_2$ -- \Offices{} focusing on challenging examples when the qualification is performed (top) using the data acquired during the first robot's deployment and (bottom) adding the images collected during nighttime. \revTwo{Results are averages and standard deviations computed over the three models.} Bold values indicate the best performance on each metric across the three datasets. }}
\label{tab:ap_extended_metric_run2}

\end{table*}
\begin{figure*}[!h]
	\centering
    \begin{tabular}{@{}c@{ }c@{ }c@{ }c@{ }}
     \scriptsize\makecell{Doors status during\\robot deployment}&\scriptsize \makecell{Door status in the\\challenging acquisition} &  \scriptsize\makecell{Qualification on the\\robot's deployment} &  \scriptsize\makecell{Qualification on the\\robot's deployment + nighttime}\\\addlinespace[0.1cm] 
    \raisebox{-0.45\height}{\includegraphics[width=0.23\linewidth]{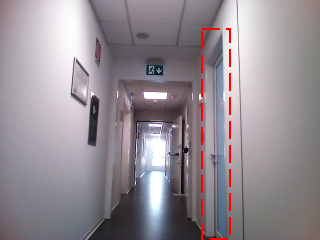}}&
    \raisebox{-0.45\height}{\includegraphics[width=0.23\linewidth]{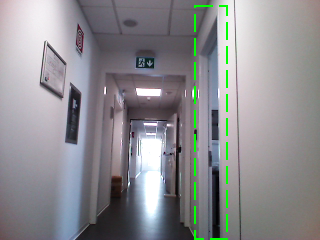}}&
    \raisebox{-0.45\height}{\includegraphics[width=0.23\linewidth]{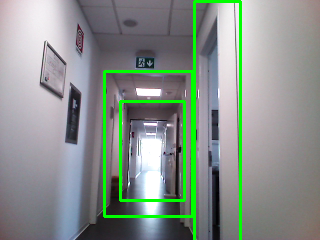}}&
    \raisebox{-0.45\height}{\includegraphics[width=0.23\linewidth]{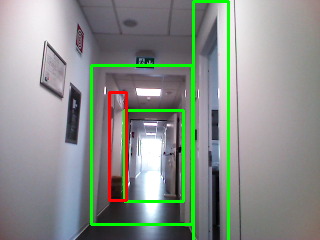}}\\\addlinespace[0.13cm]
    \raisebox{-0.45\height}{\includegraphics[width=0.23\linewidth]{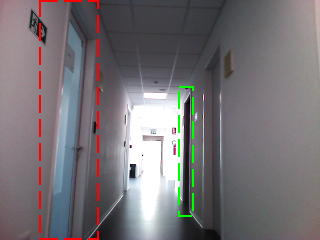}}&
    \raisebox{-0.45\height}{\includegraphics[width=0.23\linewidth]{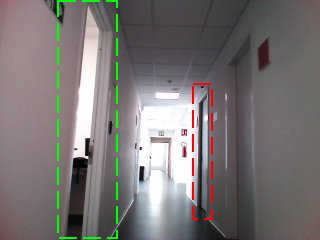}}&
    \raisebox{-0.45\height}{\includegraphics[width=0.23\linewidth]{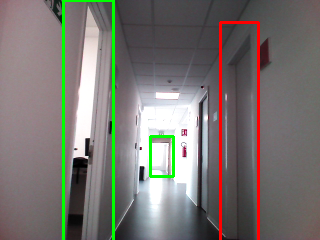}}&
    \raisebox{-0.45\height}{\includegraphics[width=0.23\linewidth]{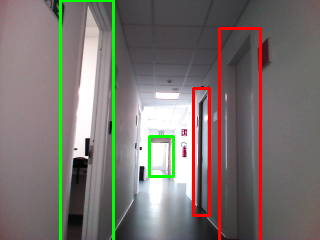}}\\\addlinespace[0.13cm]
\end{tabular}
	\caption{\RevOne{Door--status detections in the challenging run in the corridor of $e_2$ -- \Offices{} performed by the $QD_e^{15}$ based on YOLOv5 and our dataset \DG{}. The first and second columns highlight doors with a different status (in dashed green and red) between the robot deployment and the challenging run while the third and fourth columns report the detections when the qualification is performed using only the data from the robot deployment and adding the nighttime images.}}
	\label{fig:examples_run2}
\end{figure*}
}

\subsection{Model Comparison}\label{sec:Results:Comparison}

In this section, we analyze the three selected models (DETR~\cite{detr}, YOLOv5~\cite{yolov5}, and Faster R--CNN~\cite{fasterrcnn}) highlighting their strengths and weaknesses to provide insights for helping technicians working in Robotic Vision scenarios to choose the best one according to their requirements.

From our experience in setting up the three models for the specific task of door detection, DETR turned out to be the easiest to adapt. Instead of learning how to activate a set of predefined anchor boxes according to the image features, DETR directly regresses the coordinates of the bounding boxes by construction. Moreover, it does not require a non--maximum suppression step to discard multiple detections of the same object. This is achieved by its loss function that matches the (limited) inferred bounding boxes to a single target. On the contrary, the detection performance of our detectors based on YOLOv5 and Faster R--CNN are strongly influenced by the hyperparameters setting: the anchor dimension and scale should be compliant with the object shape while the non--maximum suppression procedure can delete correct bounding boxes (such as those of nested doors). In other words, while the competitors need to encode task--specific prior knowledge in the model, DETR offers the possibility to share the same configuration among different applications (such as~\cite{zimmermanlocalizationobjectdetection}). 

After these considerations, we compare the performance of the detectors (based on our dataset \DG) to study how they work, on average, in the four real environments of \Dreal. Table~\ref{tab:model_comparison} reports in detail the metrics results, depicted also in Figure~\ref{fig:ap_comparison_detectors} (mAP) and Figure~\ref{fig:extended_comparison_detectors} ($TP_{\%}$, $FP_{\%}$, and $BFD_{\%}$).

\begin{table*}[tb]
\setlength\tabcolsep{4pt}
\setlength\extrarowheight{2pt}
\centering

\begin{tabular}{c|cccc|cccc|cccc}
\toprule
     &  \multicolumn{4}{c|}{DETR~\cite{detr}} & \multicolumn{4}{c|}{YOLOv5~\cite{yolov5}} & \multicolumn{4}{c}{Faster~R--CNN~\cite{fasterrcnn}} \\[2pt]
        \textbf{Exp.} &\textbf{mAP}$\uparrow$& $\mathbf{TP_\%}$$\uparrow$ &  $\mathbf{FP_\%}$$\downarrow$ & $\mathbf{BFD_\%}$$\downarrow$ & \textbf{mAP}$\uparrow$ & $\mathbf{TP_\%}$$\uparrow$ &  $\mathbf{FP_\%}$$\downarrow$ & $\mathbf{BFD_\%}$$\downarrow$ & \textbf{mAP}$\uparrow$ & $\mathbf{TP_\%}$$\uparrow$ &  $\mathbf{FP_\%}$$\downarrow$ & $\mathbf{BFD_\%}$$\downarrow$ \\
	\midrule
 $GD$&$\textbf{24} \pm 8 $&$\textbf{31} \pm 9 $&$7 \pm 1 $&$22 \pm 9 $&$16 \pm 11 $&$20 \pm 10 $&$\textbf{4} \pm 4 $&$\textbf{8} \pm 3 $&$12 \pm 8 $&$16 \pm 9 $&$4 \pm 2 $&$9 \pm 2 $\\[2pt]
$QD_{e}^{15}$&$53 \pm 15 $&$63 \pm 11 $&$4 \pm 3 $&$35 \pm 15 $&$66 \pm 5 $&$67 \pm 5 $&$\textbf{2} \pm 1 $&$\textbf{3} \pm 3 $&$\textbf{74} \pm 7 $&$\textbf{78} \pm 4 $&$3 \pm 2 $&$22 \pm 9 $\\[2pt]
$QD_{e}^{25}$&$59 \pm 12 $&$66 \pm 9 $&$4 \pm 2 $&$35 \pm 16 $&$75 \pm 5 $&$76 \pm 5 $&$\textbf{2} \pm 1 $&$\textbf{2} \pm 1 $&$\textbf{81} \pm 4 $&$\textbf{84} \pm 3 $&$3 \pm 0 $&$16 \pm 2 $\\[2pt]
$QD_{e}^{50}$&$72 \pm 11 $&$78 \pm 8 $&$2 \pm 1 $&$21 \pm 9 $&$85 \pm 5 $&$86 \pm 5 $&$\textbf{0} \pm 1 $&$\textbf{2} \pm 0 $&$\textbf{89} \pm 4 $&$\textbf{91} \pm 3 $&$2 \pm 1 $&$15 \pm 2 $\\[2pt]
$QD_{e}^{75}$&$80 \pm 11 $&$84 \pm 8 $&$2 \pm 1 $&$17 \pm 7 $&$88 \pm 6 $&$88 \pm 6 $&$\textbf{0} \pm 0 $&$\textbf{2} \pm 1 $&$\textbf{93} \pm 4 $&$\textbf{94} \pm 3 $&$1 \pm 0 $&$10 \pm 5 $\\[2pt]
\bottomrule
\end{tabular}

\caption{\revTwo{Real--world performance obtained by the three selected models ($GDs$ are based on \DG). We report the averages and standard deviations over the four real environments in \Dreal.} \revI{Bold entries indicate the best performance on each metric across the three models.}}
\label{tab:model_comparison}
\end{table*}

\begin{figure*}[!ht]
    \centering
    \begin{subfigure}[t]{\textwidth}
        \centering
        \def\scale{0.8}
\begin{tikzpicture}[scale=\scale]

\definecolor{crimson2143940}{RGB}{214,39,40}
\definecolor{darkgray176}{RGB}{176,176,176}
\definecolor{darkorange25512714}{RGB}{255,127,14}
\definecolor{forestgreen4416044}{RGB}{44,160,44}
\definecolor{lightgray204}{RGB}{204,204,204}

\begin{axis}[
width=1.03\textwidth,
height=0.5\textwidth,
legend columns=5,
legend cell align={left},
legend style={
/tikz/every even column/.append style={column sep=0.3cm},
  fill opacity=0.8,
  draw opacity=1,
  text opacity=1,
  at={(0.5,0.98)},
  anchor=north,
  draw=lightgray204
},
tick align=outside,
tick pos=left,
title={mAP of the three detectors in real worlds},
x grid style={darkgray176},
xlabel={Environment},
xmin=-0.315, xmax=4.855,
xtick style={color=black},
xtick={0.27,1.27,2.27,3.27,4.27},
xticklabels={
  \(\displaystyle GD\),
  \(\displaystyle QD_{e}^{15}\),
  \(\displaystyle QD_{e}^{25}\),
  \(\displaystyle QD_{e}^{50}\),
  \(\displaystyle QD_{e}^{75}\)
},
y grid style={darkgray176},
ylabel={mAP},
ymin=0, ymax=115,
ytick style={color=black}
]
\draw[draw=black,fill=forestgreen4416044,opacity=0.9,thick,postaction={pattern=north east lines, fill opacity=0.9}] (axis cs:-0.08,0) rectangle (axis cs:0.12,7.125);
\draw[draw=black,fill=forestgreen4416044,opacity=0.9,thick,postaction={pattern=north east lines, fill opacity=0.9}] (axis cs:0.92,0) rectangle (axis cs:1.12,24.625);
\draw[draw=black,fill=forestgreen4416044,opacity=0.9,thick,postaction={pattern=north east lines, fill opacity=0.9}] (axis cs:1.92,0) rectangle (axis cs:2.12,28.875);
\draw[draw=black,fill=forestgreen4416044,opacity=0.9,thick,postaction={pattern=north east lines, fill opacity=0.9}] (axis cs:2.92,0) rectangle (axis cs:3.12,35.375);
\draw[draw=black,fill=forestgreen4416044,opacity=0.9,thick,postaction={pattern=north east lines, fill opacity=0.9}] (axis cs:3.92,0) rectangle (axis cs:4.12,38.75);
\draw[draw=black,fill=forestgreen4416044,opacity=0.9,thick] (axis cs:-0.08,7.125) rectangle (axis cs:0.12,24.25);
\draw[draw=black,fill=forestgreen4416044,opacity=0.9,thick] (axis cs:0.92,24.625) rectangle (axis cs:1.12,53);
\draw[draw=black,fill=forestgreen4416044,opacity=0.9,thick] (axis cs:1.92,28.875) rectangle (axis cs:2.12,58.625);
\draw[draw=black,fill=forestgreen4416044,opacity=0.9,thick] (axis cs:2.92,35.375) rectangle (axis cs:3.12,72.375);
\draw[draw=black,fill=forestgreen4416044,opacity=0.9,thick] (axis cs:3.92,38.75) rectangle (axis cs:4.12,79.75);
\draw[draw=black,fill=darkorange25512714,opacity=0.9,thick,postaction={pattern=north east lines, fill opacity=0.9}] (axis cs:0.17,0) rectangle (axis cs:0.37,5.875);
\draw[draw=black,fill=darkorange25512714,opacity=0.9,thick,postaction={pattern=north east lines, fill opacity=0.9}] (axis cs:1.17,0) rectangle (axis cs:1.37,34.5);
\draw[draw=black,fill=darkorange25512714,opacity=0.9,thick,postaction={pattern=north east lines, fill opacity=0.9}] (axis cs:2.17,0) rectangle (axis cs:2.37,38.125);
\draw[draw=black,fill=darkorange25512714,opacity=0.9,thick,postaction={pattern=north east lines, fill opacity=0.9}] (axis cs:3.17,0) rectangle (axis cs:3.37,42.625);
\draw[draw=black,fill=darkorange25512714,opacity=0.9,thick,postaction={pattern=north east lines, fill opacity=0.9}] (axis cs:4.17,0) rectangle (axis cs:4.37,42.875);
\draw[draw=black,fill=darkorange25512714,opacity=0.9,thick] (axis cs:0.17,5.875) rectangle (axis cs:0.37,15.75);
\draw[draw=black,fill=darkorange25512714,opacity=0.9,thick] (axis cs:1.17,34.5) rectangle (axis cs:1.37,66.25);
\draw[draw=black,fill=darkorange25512714,opacity=0.9,thick] (axis cs:2.17,38.125) rectangle (axis cs:2.37,75.125);
\draw[draw=black,fill=darkorange25512714,opacity=0.9,thick] (axis cs:3.17,42.625) rectangle (axis cs:3.37,84.75);
\draw[draw=black,fill=darkorange25512714,opacity=0.9,thick] (axis cs:4.17,42.875) rectangle (axis cs:4.37,87.625);
\draw[draw=black,fill=crimson2143940,opacity=0.9,thick,postaction={pattern=north east lines, fill opacity=0.9}] (axis cs:0.42,0) rectangle (axis cs:0.62,4.5);
\draw[draw=black,fill=crimson2143940,opacity=0.9,thick,postaction={pattern=north east lines, fill opacity=0.9}] (axis cs:1.42,0) rectangle (axis cs:1.62,36.625);
\draw[draw=black,fill=crimson2143940,opacity=0.9,thick,postaction={pattern=north east lines, fill opacity=0.9}] (axis cs:2.42,0) rectangle (axis cs:2.62,40.375);
\draw[draw=black,fill=crimson2143940,opacity=0.9,thick,postaction={pattern=north east lines, fill opacity=0.9}] (axis cs:3.42,0) rectangle (axis cs:3.62,46);
\draw[draw=black,fill=crimson2143940,opacity=0.9,thick,postaction={pattern=north east lines, fill opacity=0.9}] (axis cs:4.42,0) rectangle (axis cs:4.62,47.125);
\draw[draw=black,fill=crimson2143940,opacity=0.9,thick] (axis cs:0.42,4.5) rectangle (axis cs:0.62,12);
\draw[draw=black,fill=crimson2143940,opacity=0.9,thick] (axis cs:1.42,36.625) rectangle (axis cs:1.62,74);
\draw[draw=black,fill=crimson2143940,opacity=0.9,thick] (axis cs:2.42,40.375) rectangle (axis cs:2.62,81.25);
\draw[draw=black,fill=crimson2143940,opacity=0.9,thick] (axis cs:3.42,46) rectangle (axis cs:3.62,89);
\draw[draw=black,fill=crimson2143940,opacity=0.9,thick] (axis cs:4.42,47.125) rectangle (axis cs:4.62,93.25);
\path [draw=black, semithick]
(axis cs:0.02,16.4131213696608)
--(axis cs:0.02,32.0868786303392);

\path [draw=black, semithick]
(axis cs:1.02,38.1676030258087)
--(axis cs:1.02,67.8323969741913);

\path [draw=black, semithick]
(axis cs:2.02,46.14585072344)
--(axis cs:2.02,71.10414927656);

\path [draw=black, semithick]
(axis cs:3.02,61.734080553511)
--(axis cs:3.02,83.015919446489);

\path [draw=black, semithick]
(axis cs:4.02,69.0378573571857)
--(axis cs:4.02,90.4621426428143);

\addplot [semithick, black, mark=-, mark size=3, mark options={solid}, only marks, forget plot]
table {%
0.02 16.4131213696608
1.02 38.1676030258087
2.02 46.14585072344
3.02 61.734080553511
4.02 69.0378573571857
};
\addplot [semithick, black, mark=-, mark size=3, mark options={solid}, only marks, forget plot]
table {%
0.02 32.0868786303392
1.02 67.8323969741913
2.02 71.10414927656
3.02 83.015919446489
4.02 90.4621426428143
};
\path [draw=black, semithick]
(axis cs:0.27,5.030080846076)
--(axis cs:0.27,26.469919153924);

\path [draw=black, semithick]
(axis cs:1.27,60.9821731235736)
--(axis cs:1.27,71.5178268764264);

\path [draw=black, semithick]
(axis cs:2.27,70.0526074021556)
--(axis cs:2.27,80.1973925978444);

\path [draw=black, semithick]
(axis cs:3.27,79.7750628144669)
--(axis cs:3.27,89.7249371855331);

\path [draw=black, semithick]
(axis cs:4.27,81.4017237995838)
--(axis cs:4.27,93.8482762004162);

\addplot [semithick, black, mark=-, mark size=3, mark options={solid}, only marks, forget plot]
table {%
0.27 5.030080846076
1.27 60.9821731235736
2.27 70.0526074021556
3.27 79.7750628144669
4.27 81.4017237995838
};
\addplot [semithick, black, mark=-, mark size=3, mark options={solid}, only marks, forget plot]
table {%
0.27 26.469919153924
1.27 71.5178268764264
2.27 80.1973925978444
3.27 89.7249371855331
4.27 93.8482762004162
};
\path [draw=black, semithick]
(axis cs:0.52,4.37329254090163)
--(axis cs:0.52,19.6267074590984);

\path [draw=black, semithick]
(axis cs:1.52,67.0477821284619)
--(axis cs:1.52,80.9522178715381);

\path [draw=black, semithick]
(axis cs:2.52,77.366273267423)
--(axis cs:2.52,85.133726732577);

\path [draw=black, semithick]
(axis cs:3.52,84.937980797682)
--(axis cs:3.52,93.062019202318);

\path [draw=black, semithick]
(axis cs:4.52,89.5419007564522)
--(axis cs:4.52,96.9580992435478);

\addplot [semithick, black, mark=-, mark size=3, mark options={solid}, only marks, forget plot]
table {%
0.52 4.37329254090163
1.52 67.0477821284619
2.52 77.366273267423
3.52 84.937980797682
4.52 89.5419007564522
};
\addplot [semithick, black, mark=-, mark size=3, mark options={solid}, only marks, forget plot]
table {%
0.52 19.6267074590984
1.52 80.9522178715381
2.52 85.133726732577
3.52 93.062019202318
4.52 96.9580992435478
};
\input{graphics/legend_comparison_map}
\end{axis}

\end{tikzpicture}
    \end{subfigure}
    \\[-0.9cm]
    \caption{Real--world mAP \revTwo{(averaged over the four real environments, with standard deviations)} with our three selected models (GDs are based on \DG).}
    \label{fig:ap_comparison_detectors}
\end{figure*}

\begin{figure*}[!ht]
    \centering
    \begin{subfigure}[t]{\textwidth}
        \centering
        \def\scale{0.8}
\begin{tikzpicture}[scale=\scale]

\definecolor{crimson2143940}{RGB}{214,39,40}
\definecolor{darkgray176}{RGB}{176,176,176}
\definecolor{darkorange25512714}{RGB}{255,127,14}
\definecolor{forestgreen4416044}{RGB}{44,160,44}
\definecolor{gray}{RGB}{128,128,128}
\definecolor{lightgray204}{RGB}{204,204,204}

\begin{axis}[
width=\textwidth,
height=0.5\textwidth,
legend columns=4,
legend cell align={left},
legend style={
/tikz/every even column/.append style={column sep=0.3cm},
  fill opacity=0.8,
  draw opacity=1,
  text opacity=1,
  at={(0.5,0.98)},
  anchor=north,
  draw=lightgray204
},
tick align=outside,
tick pos=left,
title={Extended metric results in the real worlds},
x grid style={darkgray176},
xlabel={\% of qualification data},
xmin=-0.7, xmax=14.7,
xtick style={color=black},
xtick={0,1,2,3,4,5,6,7,8,9,10,11,12,13,14},
xticklabels={0,15,25,50,75,0,15,25,50,75,0,15,25,50,75},
y grid style={darkgray176},
ylabel={\%},
ymin=0, ymax=135,
ytick style={color=black},
ytick=\empty,
extra y ticks={0, 20, 40, 60, 80, 100}
]
\path [draw=forestgreen4416044, fill=forestgreen4416044, opacity=0.2]
(axis cs:0,22.1076078987317)
--(axis cs:0,40.3923921012683)
--(axis cs:1,73.9544511501033)
--(axis cs:2,75.8273790530888)
--(axis cs:3,85.9582242575422)
--(axis cs:4,92.3070879768479)
--(axis cs:4,76.1929120231521)
--(axis cs:4,76.1929120231521)
--(axis cs:3,70.0417757424578)
--(axis cs:2,57.1726209469112)
--(axis cs:1,52.0455488498967)
--(axis cs:0,22.1076078987317)
--cycle;

\path [draw=crimson2143940, fill=crimson2143940, opacity=0.2]
(axis cs:0,5.58578643762691)
--(axis cs:0,8.4142135623731)
--(axis cs:1,6.70801280154532)
--(axis cs:2,6.23205080756888)
--(axis cs:3,3.5)
--(axis cs:4,2.81649658092773)
--(axis cs:4,1.18350341907227)
--(axis cs:4,1.18350341907227)
--(axis cs:3,1.5)
--(axis cs:2,2.76794919243112)
--(axis cs:1,1.29198719845468)
--(axis cs:0,5.58578643762691)
--cycle;

\path [draw=darkorange25512714, fill=darkorange25512714, opacity=0.2]
(axis cs:0,12.7546308209909)
--(axis cs:0,30.7453691790091)
--(axis cs:1,49.4780910734102)
--(axis cs:2,51.0057682070089)
--(axis cs:3,30.0011260575853)
--(axis cs:4,24.1605780251386)
--(axis cs:4,9.33942197486143)
--(axis cs:4,9.33942197486143)
--(axis cs:3,11.4988739424147)
--(axis cs:2,18.4942317929911)
--(axis cs:1,20.0219089265898)
--(axis cs:0,12.7546308209909)
--cycle;

\path [draw=forestgreen4416044, fill=forestgreen4416044, opacity=0.2]
(axis cs:5,9.63981982138663)
--(axis cs:5,30.3601801786134)
--(axis cs:6,71.8304589153965)
--(axis cs:7,80.5460605656619)
--(axis cs:8,90.6062674281112)
--(axis cs:9,93.5976185412489)
--(axis cs:9,82.4023814587511)
--(axis cs:9,82.4023814587511)
--(axis cs:8,80.8937325718888)
--(axis cs:7,71.4539394343381)
--(axis cs:6,62.1695410846035)
--(axis cs:5,9.63981982138663)
--cycle;

\path [draw=crimson2143940, fill=crimson2143940, opacity=0.2]
(axis cs:5,0.156023557858696)
--(axis cs:5,7.3439764421413)
--(axis cs:6,3.00830573921179)
--(axis cs:7,2.70742710775634)
--(axis cs:8,1.07735026918963)
--(axis cs:9,0.75)
--(axis cs:9,-0.25)
--(axis cs:9,-0.25)
--(axis cs:8,-0.0773502691896257)
--(axis cs:7,0.792572892243662)
--(axis cs:6,0.491694260788208)
--(axis cs:5,0.156023557858696)
--cycle;

\path [draw=darkorange25512714, fill=darkorange25512714, opacity=0.2]
(axis cs:5,5.12004436032342)
--(axis cs:5,10.3799556396766)
--(axis cs:6,5.87995563967658)
--(axis cs:7,3.79099444873581)
--(axis cs:8,2.25)
--(axis cs:9,2.79099444873581)
--(axis cs:9,0.209005551264194)
--(axis cs:9,0.209005551264194)
--(axis cs:8,1.25)
--(axis cs:7,1.20900555126419)
--(axis cs:6,0.620044360323416)
--(axis cs:5,5.12004436032342)
--cycle;

\path [draw=forestgreen4416044, fill=forestgreen4416044, opacity=0.2]
(axis cs:10,7.24404964229087)
--(axis cs:10,24.7559503577091)
--(axis cs:11,81.5414518843274)
--(axis cs:12,86.6091263510296)
--(axis cs:13,93.7360788111948)
--(axis cs:14,97.122281323269)
--(axis cs:14,91.377718676731)
--(axis cs:14,91.377718676731)
--(axis cs:13,87.7639211888052)
--(axis cs:12,80.3908736489704)
--(axis cs:11,73.4585481156726)
--(axis cs:10,7.24404964229087)
--cycle;

\path [draw=crimson2143940, fill=crimson2143940, opacity=0.2]
(axis cs:10,2.36700683814455)
--(axis cs:10,5.63299316185545)
--(axis cs:11,4.82574185835055)
--(axis cs:12,3.25)
--(axis cs:13,2.81649658092773)
--(axis cs:14,1.25)
--(axis cs:14,0.25)
--(axis cs:14,0.25)
--(axis cs:13,1.18350341907227)
--(axis cs:12,2.25)
--(axis cs:11,1.17425814164945)
--(axis cs:10,2.36700683814455)
--cycle;

\path [draw=darkorange25512714, fill=darkorange25512714, opacity=0.2]
(axis cs:10,7.03264421739165)
--(axis cs:10,11.4673557826083)
--(axis cs:11,31.4650239645194)
--(axis cs:12,17.6429694486001)
--(axis cs:13,16.9673557826083)
--(axis cs:14,14.0092497528229)
--(axis cs:14,4.99075024717711)
--(axis cs:14,4.99075024717711)
--(axis cs:13,12.5326442173917)
--(axis cs:12,13.8570305513999)
--(axis cs:11,13.0349760354806)
--(axis cs:10,7.03264421739165)
--cycle;

\addplot [semithick, forestgreen4416044, mark=*, mark size=2.5, mark options={solid}, forget plot]
table {%
0 31.25
1 63
2 66.5
3 78
4 84.25
};
\addplot [semithick, crimson2143940, mark=triangle*, mark size=2.5, mark options={solid}, forget plot]
table {%
0 7
1 4
2 4.5
3 2.5
4 2
};
\addplot [semithick, darkorange25512714, mark=diamond*, mark size=2.5, mark options={solid}, forget plot]
table {%
0 21.75
1 34.75
2 34.75
3 20.75
4 16.75
};
\addplot [semithick, forestgreen4416044, mark=*, mark size=2.5, mark options={solid}, forget plot]
table {%
5 20
6 67
7 76
8 85.75
9 88
};
\addplot [semithick, crimson2143940, mark=triangle*, mark size=2.5, mark options={solid}, forget plot]
table {%
5 3.75
6 1.75
7 1.75
8 0.5
9 0.25
};
\addplot [semithick, darkorange25512714, mark=diamond*, mark size=2.5, mark options={solid}, forget plot]
table {%
5 7.75
6 3.25
7 2.5
8 1.75
9 1.5
};
\addplot [semithick, forestgreen4416044, mark=*, mark size=2.5, mark options={solid}, forget plot]
table {%
10 16
11 77.5
12 83.5
13 90.75
14 94.25
};
\addplot [semithick, crimson2143940, mark=triangle*, mark size=2.5, mark options={solid}, forget plot]
table {%
10 4
11 3
12 2.75
13 2
14 0.75
};
\addplot [semithick, darkorange25512714, mark=diamond*, mark size=2.5, mark options={solid}, forget plot]
table {%
10 9.25
11 22.25
12 15.75
13 14.75
14 9.5
};
\addplot [black, forget plot]
table {%
-0.7 -3.5527136788005e-15
14.7 -3.5527136788005e-15
};
\path [draw=gray, draw opacity=0.7, semithick, dash pattern=on 5.55pt off 2.4pt]
(axis cs:4.5,0)
--(axis cs:4.5,115);

\path [draw=gray, draw opacity=0.7, semithick, dash pattern=on 5.55pt off 2.4pt]
(axis cs:9.5,0)
--(axis cs:9.5,115);

\draw (axis cs:2,107) node[
  text=black,
  rotate=0.0
]{DETR~\cite{detr}};
\draw (axis cs:7,107) node[
  text=black,
  rotate=0.0
]{YOLOv5~\cite{yolov5}};
\draw (axis cs:12,107) node[
  text=black,
  rotate=0.0
]{Faster~R--CNN~\cite{fasterrcnn}};
\input{graphics/legend_comparison_extended_metric}
\end{axis}

\end{tikzpicture}
    \end{subfigure}
    \\[-0.9cm]
    \caption{Real--world performance of the operational performance indicators with our three selected models (GD is based on \DG). \revTwo{Results are averages and standard deviations across the four real environments of \Dreal.}}
    \label{fig:extended_comparison_detectors}
\end{figure*}

By observing the mAP performance shown in Figure~\ref{fig:ap_comparison_detectors} we can see that the best GD is based on DETR that, not requiring task--oriented knowledge, better addresses the sim--to--real gap (between our dataset \DG{} and the real acquisitions of \Dreal). While YOLOv5 lies in the middle, Faster R--CNN reaches the worst performance when trained in simulation (with our dataset \DG) and tested in the real world. As discussed in Section~\ref{sec:Results:GD}, Faster R--CNN, being a two--stage detector, tends to overfit the distribution of the training data acquired in simulation. This outcome is reverted by the qualification procedure. When fine--tuned for a target environment, Faster R--CNN reaches the best mAP results while DETR the worst (see the green and red bars in Figure~\ref{fig:ap_comparison_detectors}). This is caused by the Transformer that, although popular, requires huge amounts of data (hundreds of millions) to effectively learn the architecturally inherent biases of the CNN--based models (such as the translation equivariance and the locality principle~\cite{surveytransformer}). Moreover, by carefully examining the extended metric's results, we can see that the $BFD_{\%}$ of YOLOv5 is considerably lower than the other detectors (both the GD and its qualified versions). In a robotic domain where detections are translated into actions, this fact is extremely important because drastically reduces robot failures. Figure~\ref{fig:confidence_threshold} shows how the additional indicators of $QD_e^{15}$ vary according to the confidence threshold. The results demonstrate that our choice of $\rho_c = 75\%$ is a good compromise between the correct ($TP_{\%}$) and the wrong ($FP_{\%}$, $BFD_{\%}$) predictions. 

\begin{figure*}[!ht]
    \centering
    \begin{subfigure}[t]{\textwidth}
        \centering
        \def\scale{0.8}
        \include{graphics/qualified_15_confidences}
    \end{subfigure}
    \\[-0.9cm]
    \caption{Operational performance indicators w.r.t. confidence for $QD_e^{15}$ (averages \revTwo{and standard deviations} over the real environments, GDs based on \DG).}
    \label{fig:confidence_threshold}
\end{figure*}

Despite it is well--known from the literature that the two--stage detectors (like Faster R--CNN) are generally better than single--stage ones~\cite{objectdetectionsurvey}, YOLOv5 is more suitable for edge devices typically mounted in service robots. First, it is compatible with the NVIDIA Jetson TX2 mounted on our Giraff--X robotic platform~\cite{movecare, GIRAFF} (depicted in Figure~\ref{fig:giraff}) where it can run at 20 fps with the TensorRT framework. Since the other models are not compatible with the NVIDIA SDK for our specific hardware, we deploy all the architectures relying on ONNXRuntime, a less efficient inference framework able to run YOLOv5, DETR, and Faster R--CNN at 14, 6, and 0.7 fps, respectively.  In our experimental setting, YOLOv5 represents the best compromise between performance and inference time, thus appearing as the most convenient model for door detection with service robots.

\revI{\subsection{Evaluation on a Downstream Task: Topology Mapping}\label{sec:onTheFieldExp}}


The goal of equipping an autonomous mobile robot with an object detection method is to allow the robot to have an updated representation of its working environment that can be used to plan and execute the tasks assigned to the robot. In the scenario we consider, the ability to detect doors can be used by the robot for the downstream task of reconstructing the current topology of the environment, that is, to infer which are the sub-areas that are currently accessible by the robot and those that are not. In the experiments presented in this section, we consider the environment's topology to be a graph, wherein the nodes represent rooms and the edges correspond to open paths connecting them. This knowledge can be used by the robot to plan its activities~\cite{longtermnavigation} considering the constraint that only a subset of the sub-areas are accessible at the current time.

We evaluate how the door detector can be used to obtain such a knowledge. As discussed in Section \ref{sec:qualification}, we assume that the robot can rely on a 2D map acquired during its setup, but we consider a situation where the current topology of the environment has changed with respect to the one encoded in such a map, since some doors might be closed at the moment (for obvious reasons, when the 2D map is acquired all the doors are left opened). The task that the robot must face is to infer the current topology of the free space it can cover, assuming that door statuses do not change during the execution of this task. To carry it out, we consider the setting exemplified in Figure~\ref{fig:giraff}: the robot follows a trajectory spanning multiple rooms (the trajectory can be either functional to this topology-inference task or to another higher-level task the robot is performing). While doing so, it can observe, on purpose or incidentally, the status (\texttt{open} or \texttt{closed}) of multiple doors. While some doors are perceived from a frontal view, others will likely be observed only from a side angle as the robot moves in a different direction. We assume that the robot has full knowledge of all the locations of doors on the map $D = \{d_1, d_2, \ldots d_n \}$. Furthermore, we assume to have a method that,  given the image $x\in X$ where a door $\hat{y} \in \hat{Y}$ has been identified by a door detector, along with the pose from which the image was acquired, can determine the specific door instance $d \in D$ being observed at the moment. Note that multiple doors can be observed within the same image. The result of this step is that each $\hat{y} \in \hat{Y}$ that is not a $BFD$ is associated to a door instance $d$. The robot thus counts, along the whole trajectory, how many times each door $d \in D$ has been identified either as \texttt{open} or \texttt{closed}, and infers its status as the one of the majority label. This information is used to infer the current topology, which, for evaluation, we compare with the one obtained by repeating the process using true detections instead of predictions.

\begin{figure*}[!ht]
\centering
\begin{scriptsize}
\begin{tabular}{c|ccccc}
\toprule
        \textbf{Exp.} & \greensquare{} \redsquare{} $\uparrow$ & \greencross{} \redcross{} $\downarrow$ & \yellowcross{} $\downarrow$ &  \graycross{} $\downarrow$ & $RA$ $\uparrow$ \\
        \midrule
 $GD$ & 20 & 4 & 1 & 2 & 71\% \\
 $QD_e^{15}$ & 25 & 1 & 1 & 1 & 89\% \\
\bottomrule
\end{tabular}
\end{scriptsize}\\[5pt]
\subfloat[$GD$]{
  \includegraphics[width=0.47\linewidth]{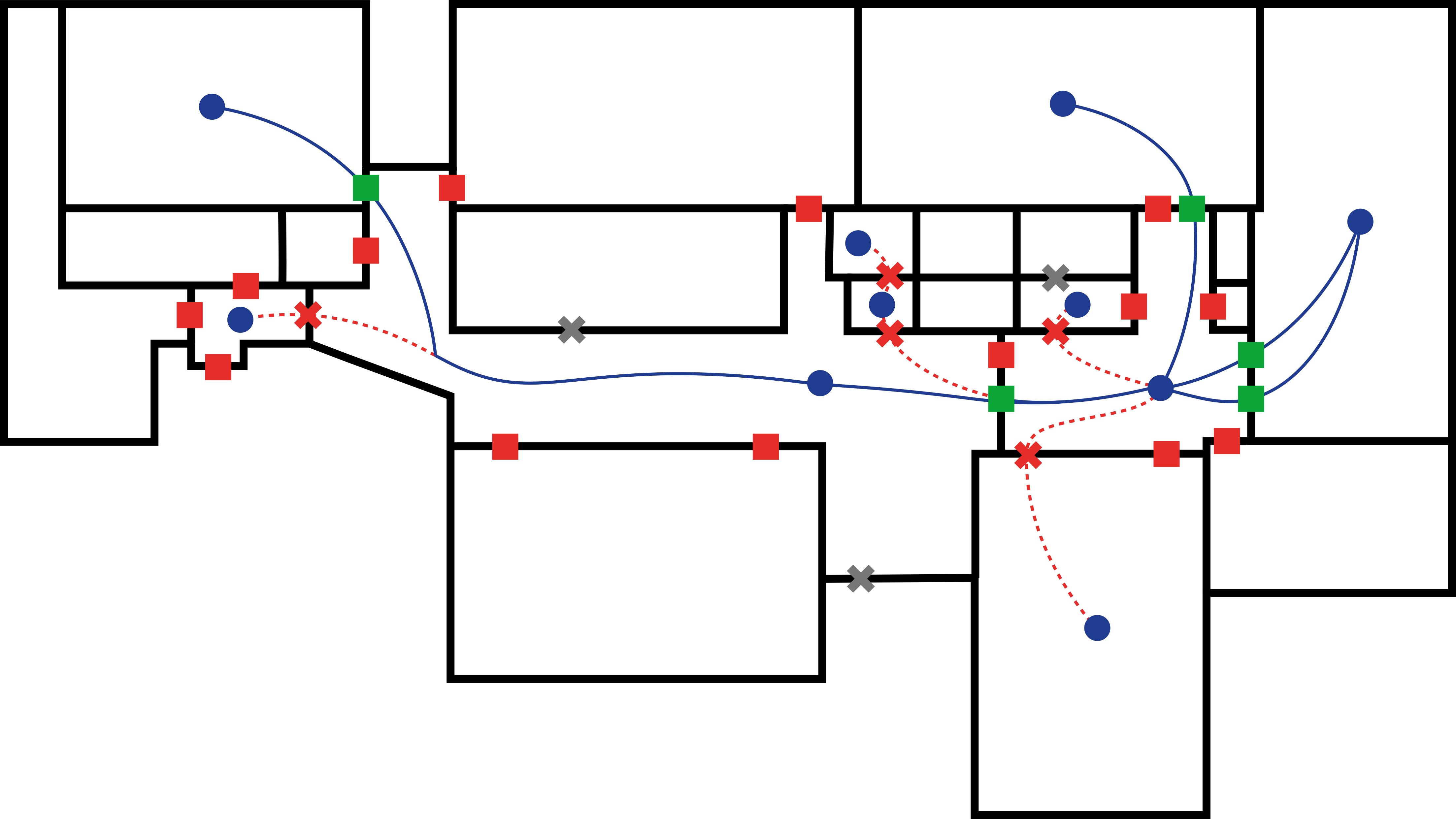}
  \label{fig:p1GD}
}
\hfill
\subfloat[$QD^{15}_e$]{
  \includegraphics[width=0.47\linewidth]{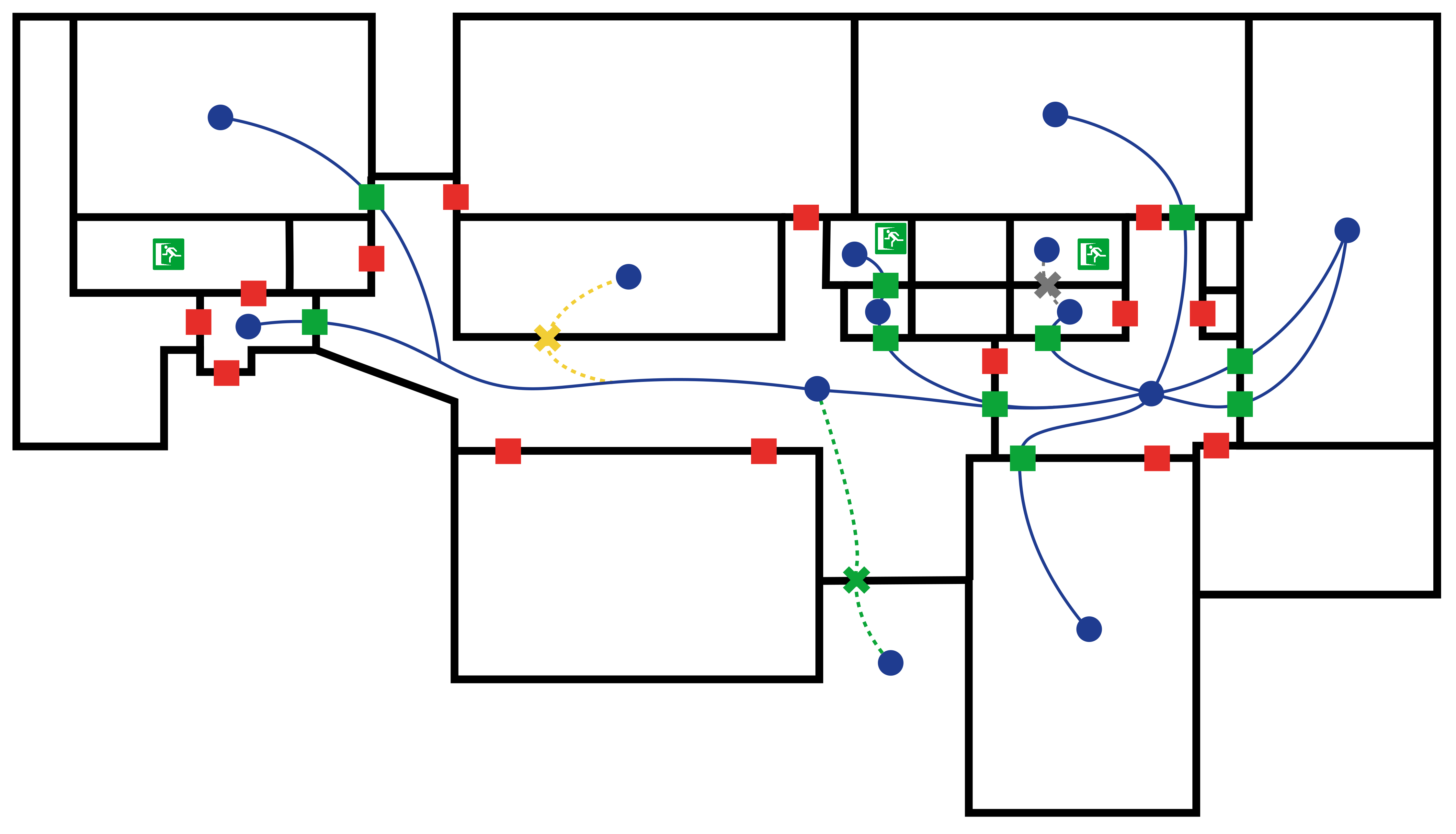}
  \label{fig:p1QD}
}
\caption{Topology of the environment for $e_1$ - \Classrooms{} during night-time as identified using $GD$ and $QD^{15}_e$ to detect the status of each door.}
\label{fig:P1}
\end{figure*}

\begin{figure*}[!ht]
\centering
\begin{scriptsize}
\begin{tabular}{c|ccccc}
\toprule
        \textbf{Exp.} & \greensquare{} \redsquare{} $\uparrow$ & \greencross{} \redcross{} $\downarrow$ & \yellowcross{} $\downarrow$ &  \graycross{} $\downarrow$ & $RA$ $\uparrow$ \\
        \midrule
 $GD$ & 26 & 9 & 4 & 3 & 62\% \\
 $QD_e^{15}$ & 40 & 1 & 1 & 0 & 95\% \\
\bottomrule
\end{tabular}
\end{scriptsize}\\[5pt]
\subfloat[$GD$]{
  \includegraphics[width=0.8\linewidth]{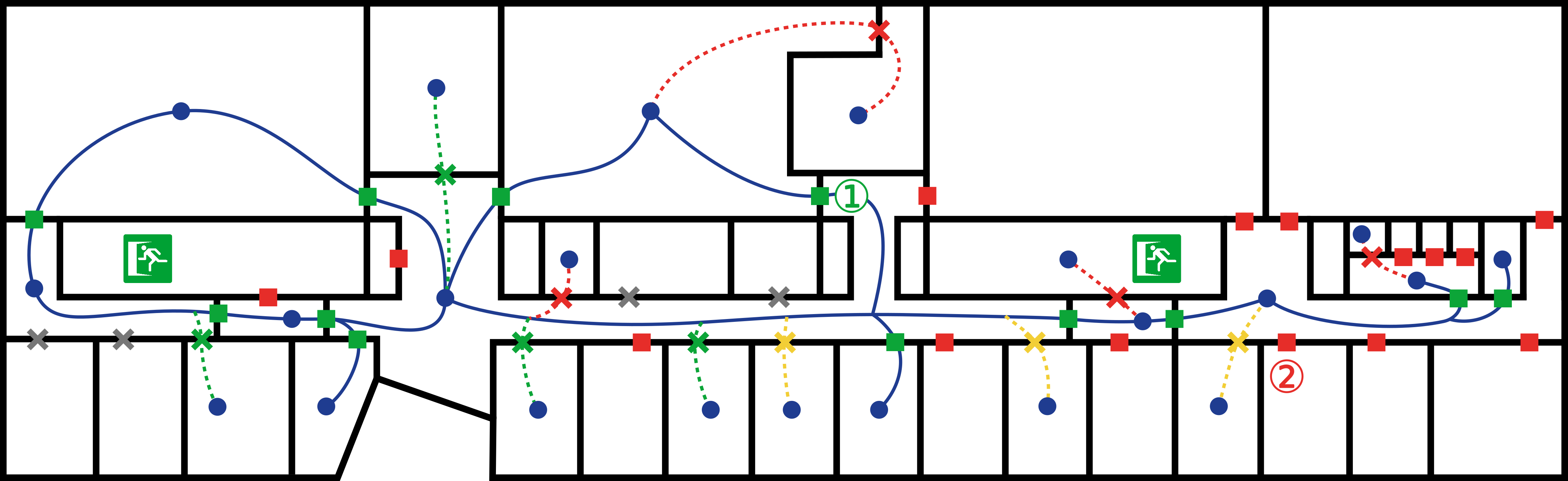}
  \label{fig:p4GD}
}
\hfill
\subfloat[$QD_e^{15}$]{
  \includegraphics[width=0.80\linewidth]{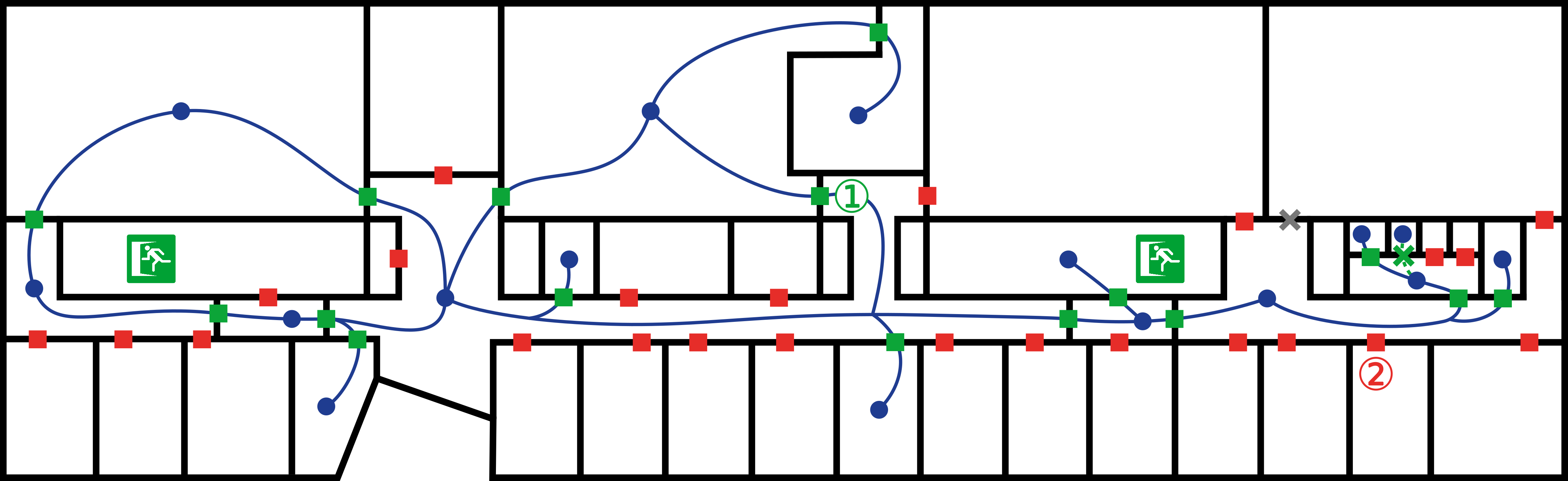}
  \label{fig:p4QD}
}
\caption{Topology of the environment for $e_2$ - \Offices{} during daytime as identified using $GD$ and $QD^{15}_e$ to detect the status of each door.}
\label{fig:P4}
\end{figure*}

We had the robot following two trajectories in a real-world experiment with the same setup described in Section~\ref{sec:ExperimentalSetting:Datasets}. The first trajectory is performed in $e_1$ - \Classrooms{} during nighttime, \revI{using the same run of Section~\ref{sec:Results:LongTerm}.} The second trajectory is performed in $e_2$ - \Offices{} with daylight. We compare the performance in inferring the topology with $GD$ against $QD^{15}_e$ (both based on YOLOv5 and \DG). 
In both cases, the $QD_e^{15}$ is trained with data collected at mapping time, with daylight. \revI{Note that the evaluation in $e_1$ is performed in challenging conditions because the qualified detector is tested under light variations (i.e., with nighttime data).}
The floor plans and the topologies of $e_1$ - \Classrooms{} and $e_2$ - \Offices{} inferred with this framework are shown in Figure~\ref{fig:P1} and~\ref{fig:P4}, respectively.  We indicate with \greensquare{} (\redsquare{}) a door correctly recognized as \texttt{open} (\texttt{closed}) and with \redcross{} (\greencross{}) a door that has been wrongfully recognized as \texttt{closed} (\texttt{open}) when its current status is \texttt{open} (\texttt{closed}). We indicate with \yellowcross{} the event where the number of detections $\hat{y}$ where a door is labeled as \texttt{open} is equal to those where it is perceived as \texttt{closed}, and thus the robot is undecided. If a door has been observed in multiple images, but the door detector was always unable to detect any door due to false negatives, we label such door with a \graycross{}. We indicate with \blueroom{} the location of a room that the robot can access, and we highlight the location of the main entrance/exits of the environment. We indicated with a solid blue line a path across two different rooms that is open for the robot, and with a dashed line a path between two rooms that has been wrongfully estimated, following the same color schema as above: a red (green) path when a passage is estimated to be \texttt{closed} (\texttt{open}) when actually it is \texttt{open} (\texttt{closed}).

To understand the impact of having a qualified detector in estimating the topology of an environment, we report the topology obtained with $GD$ and $QD^{15}_e$ in Figure \ref{fig:P1}-\ref{fig:P4}, as well as the number of doors whose status is correctly/wrongly detected, and the total \emph{recognition accuracy} $RA$, that is the percentage of doors $d$ whose status has been successfully detected during the robot run. These metrics are specific to the detection domain and are downstream OPI, following the definition of Section \ref{sec:ExperimentalSetting:Metrics}.

From Figure \ref{fig:P1}-\ref{fig:P4}, we can see how the qualified detector can correctly identify the topological status of the environment, albeit making minor errors. In both environments, the QD identifies correctly the status of most doors, with an $RA$ of around $90\%$ ($89,29\%$ in $e_1$, $95,23\%$ in $e_2$).

We noticed how, while both $GD$ and $QD^{15}_e$ could detect successfully the status of a door when it is observed from a frontal position by the robot, the $GD$ often fails when the robot is at one side of a door, when the door\revZ{'s view} is partially occluded, or when there are challenging light conditions. In all those cases, $QD^{15}_e$ does not suffer from the same limitations.
An example of this can be seen in the doors connected to the main corridor of $e_2$, shown in Figure \ref{fig:P4}. While $QD^{15}_e$ can identify how most doors are closed (\redcross ), $GD$ often fails to understand the status of those doors, thus identifying them as open (\greencross{} or \yellowcross ) or failing to identify them (\graycross ).



\begin{figure*}[!ht]
\centering
\subfloat[]{
  \includegraphics[width=0.23\linewidth]{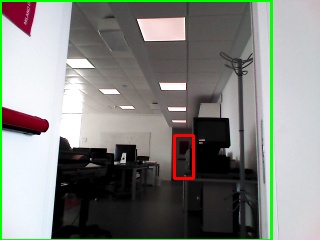}
  \label{fig:door71}
}
\subfloat[]{
  \includegraphics[width=0.23\linewidth]{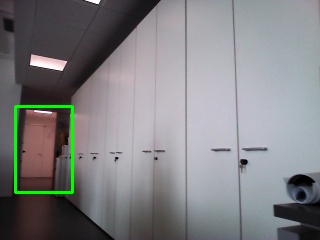}
  \label{fig:door72}
}
\subfloat[]{
  \includegraphics[width=0.23\linewidth]{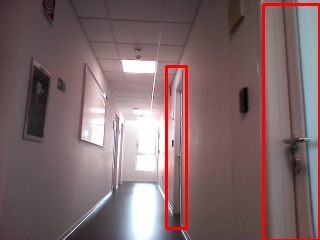}
  \label{fig:door211}
}
\subfloat[]{
  \includegraphics[width=0.23\linewidth]{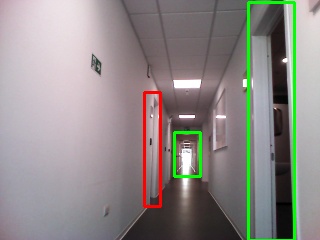}
  \label{fig:door212}
}
\caption{Two examples where the $QD$ identifies the door \textcolor{ForestGreen}{\ding{192}} in tho challenging images (a-b), and the door \textcolor{red}{\ding{193}} in other two challenging images (c-d). In all four cases, the $GD$ does not identify the two doors in these images.}
\label{fig:P4Examples}
\end{figure*}

To better highlight this event, we provide detailed results about how many times two doors, that are highlighted with \textcolor{ForestGreen}{\ding{192}} and \textcolor{red}{\ding{193}} in Figure \ref{fig:P4}, have been viewed by the robot in the two runs. Door \textcolor{ForestGreen}{\ding{192}} was observed in $60$ images by the robot. While $QD^{15}_e$ was able to correctly detect the status of the door $52$ times and was unable to detect the door on $8$, the $GD$ was able to detect the door only on $2$ occasions and was unable to detect it $58$ times. Figure \ref{fig:P4Examples}(a-b) shows two of the 60 images, with the bounding box identified by $QD^{15}_e$. At the beginning of the run, the door was closed, and was briefly perceived in that condition when the robot was outside the room; nevertheless, the robot was able to correctly label it, as in Figure \ref{fig:door71}. Later, the robot enters the room and the door status is open, as in Figure \ref{fig:door72}. In both cases, $GD$ fails to identify any bounding box from those images.
The door \textcolor{red}{\ding{193}} was observed in $40$ images as closed; $QD^{15}_e$ was able to correctly identify the status of the door $32$ times, and was unable to detect the door $8$ times.  The $GD$ is far less accurate in detecting doors in the same set of images: it correctly identified the door as \texttt{closed} $5$ times, wrongly identified the door as \texttt{open} $4$ times, and did not identify any door in $31$ perceptions. Two examples of these images are shown in Figure \ref{fig:P4Examples}(c-d); in Figure \ref{fig:door211} the door \textcolor{red}{\ding{193}} is the second one on the right, while in Figure \ref{fig:door212} it is the one on the left side of the corridor. In both images, $QD^{15}_e$ was able to identify successfully the door status and location, while $GD$ failed to identify the presence of a door. Similar examples can be made for all of the rooms that are connected to the central corridor of Figure~\ref{fig:P4}, that are seen by the robot from a similar perspective.

These results show how the general detector manages to partially reconstruct the topology of the environment due to a high number of false positives and wrong detections. On the other hand, the qualified detector obtains more stable and robust performance compared to its general version when used in its deployment environment. This demonstrates how the qualification step described in Section~\ref{sec:qualification} substantially improves (with a little cost) the performance of the detection method in a downstream task, providing more accurate domain-specific knowledge to the robot.
\revI{\revTwo{\section{Lessons Learned}\label{sec:lessons_learned}}}

\revI{
\revTwo{With our extensive experimental campaign in the real world, we assess the effectiveness of our pipeline involving simulation and qualification for the development and deployment of deep-learning based vision modules for service robots. In this section, we synthesize some key lessons that, while specific to our scenario, might apply to object detection with service robotics at large.}

\emph{Simulation pursues domain relevance, lowers costs.} Analogously to what happens in other robotic domains, simulation can be properly engineered to synthesize domain-relevant training data for object detection with service robots. In this specific scenario, domain relevance is not only influenced by photorealism. The alignment with the robot's perception model plays a major role that cannot be neglected in the development phase. Simulations offering an acceptable level of photorealism and, at the same time, allowing to generate robot-centric perceptions produce valuable training data. This simulated data is remarkably more cost-effective when compared with real-world acquisition with mobile robots.

\emph{Qualification is key, yet affordable.} Training on varied data in the attempt of generalizing across different environments will inevitably hit a performance ceiling. During its operational time, a service robot will encounter detection instances that remarkably shift away from the training distribution and that constitute hard cases peculiar of the specific environment in which the robot is deployed. For service robots, the priority is to be capable of dealing with such instances and not to generalize on each possible environment. Qualification leverages this condition and allows to break this performance limit. Its impact is remarkable since difficult detection instances are typically connected with critical steps in a robot's task. Additionally, qualification shows a diminishing-return trend in performance where the first improvement steps outperform the subsequent ones and already lead to effective and robust detectors. As a consequence, training a qualified detector incurs in affordable data preparation costs.

\emph{Model and performance assessment must be setup-driven.} In our robotic scenario, detections are meant to directly translate to decisions and actions. This is an aspect often, and rightfully, neglected in the broad field of object detection. Selecting the proper model to deploy and identify the most relevant performance metrics plays a crucial role in tailoring the robotic setup to the use case at hand.
}


\revTwo{\section{Conclusions}\label{sec:conclusions}}

\revTwo{Our work devises and evaluates a method for on-the-field object detection with service robots, focusing on the task of real-time detection of doors, intended as variable-traversability passages. We leverage state-of-the-art deep-learning techniques combined with simulation and fine-tuning to cost-effectively synthesize detectors that operate with satisfying performance, even when faced with challenging instances and conditions. We conducted an extensive experimental campaign exploiting and adapting public datasets and simulation frameworks, while also carrying out on-the-field data acquisition and experimentation in four distinct real-world settings.

We envisage future directions building upon the limitations of our method. Enhancing the photorealism in our simulation framework would allow to further close the sim-to-real gap. One interesting objective in this direction is to improve the visual quality of simulators like iGibson~\cite{igibson} in such a way as to fully exploit its high level of automation. Our method could gain a significant boost by integrating automatic scene design/generation, overcoming the limit to rely on hand-crafted scenes. This is a flourishing area of research whose recent progresses could find in our setting an intriguing use case. 
While LiDAR and depth data have well-known limits for the task of robotic vision, integrating them in the RGB pipeline with a sensor fusion approach could introduce significant advantages. \revTwo{An interesting solution is to use these technologies to confirm the status of previously identified doors when the robot is close enough to them.} 
Another undoubtedly interesting direction of research would be to conduct a large-scale experimentation in a pilot campaign with a fleet of service robots deployed in real setups.}

\printbibliography
\end{document}